\documentclass[12pt,a4paper]{book}
\usepackage[Lenny]{fncychap}

\usepackage[english]{babel}
\usepackage[letterpaper,top=2cm,bottom=2cm,left=3cm,right=3cm,marginparwidth=1.75cm]{geometry}
\usepackage{setspace}
\usepackage{csquotes}

\usepackage{microtype}
\usepackage{soul}
\usepackage{multirow}
\usepackage{rotating}
\usepackage{pifont} %
\usepackage{booktabs} %
\usepackage{tabularx}    %
\usepackage{makecell} %
\usepackage{graphicx} %
\usepackage{caption}  %
\usepackage{mathrsfs} 
\usepackage{array}
\usepackage{hyperref}
\usepackage{multicol}
\usepackage{amssymb} %
\usepackage{graphicx} %
\usepackage{float} %
\usepackage{algorithm}
\usepackage{algorithmic}
\usepackage{amsmath}  %
\usepackage{pifont}%

\usepackage[nottoc]{tocbibind} %

\usepackage[natbib=true, maxcitenames=1, maxbibnames=99]{biblatex}

\newcommand{\thesistitle}{Multimodal Group Emotion Recognition In-the-Wild Towards a Privacy-Safe Non-Individual Approach}

\newcommand{\thesisauthor}{Anderson AUGUSMA}
\newcommand{\thesisuniv}{Thèse de Doctorat de l'Université Grenoble Alpes}
\newcommand{\thesissupervisors}{Dominique Vaufreydaz, Frédérique Letué}

\newcommand{\thesisdate}{January 16, 2026}

\newcommand{\makecustomtitle}{%
\begin{titlepage}
  \thispagestyle{empty}
  \begin{center}

    \vspace*{0.5cm}
    \includegraphics[width=0.12\textwidth]{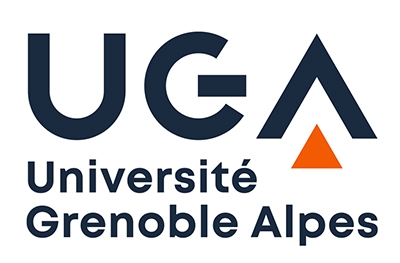}\hspace{0.8cm}%
    \includegraphics[width=0.11\textwidth]{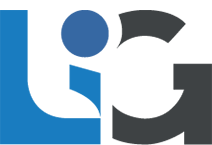}\hspace{0.8cm}%
    \includegraphics[width=0.08\textwidth]{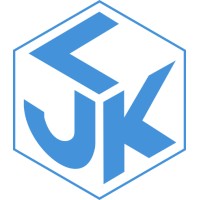}\hspace{0.8cm}%
    \includegraphics[width=0.28\textwidth]{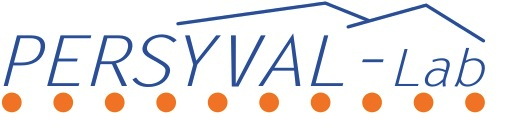}

    \vspace{1.2cm}

    {\Large\bfseries \thesistitle \par}
    \vspace{0.35cm}

    \vspace{1.0cm}
    \rule{0.9\linewidth}{0.6pt}\par
    \vspace{0.8cm}

    {\large \textbf{\thesisauthor}\par}
    \vspace{0.3cm}
    {\normalsize \emph{Supervisors:} \thesissupervisors\par}

    \vspace{0.6cm}
    {\normalsize \thesisuniv\par}

    \vspace{0.8cm}
    \rule{0.9\linewidth}{0.4pt}\par
    \vspace{0.5cm}

    {\footnotesize
    \textbf{Jury Composition}\par
    \vspace{0.3cm}

    \renewcommand{\arraystretch}{1.15}
    \begin{tabular}{@{}p{3cm}p{10cm}@{}}
      \textbf{President} & Prof. Didier Schwab — Univ. Grenoble Alpes \\
      \textbf{Reviewer}  & Prof. Alessandro Vinciarelli — Univ. of Glasgow \\
      \textbf{Reviewer}  & Dr. Antitza Dantcheva — Inria \\
      \textbf{Examiner}  & Prof. Christine Keribin — Univ. Paris-Saclay \\
      \textbf{Examiner}  & Dr. Bernd Dudzik — TU Delft \\
    \end{tabular}
    }

    \vfill
    {\thesisdate\par}

  \end{center}
\end{titlepage}
}

\begin{document}
\makecustomtitle

\section*{Abstract}
\addcontentsline{toc}{chapter}{Abstract} %

This thesis addresses the challenge of group emotion recognition (GER) in-the-wild. Traditional approaches to emotion recognition often rely on individual-level cues such as facial recognition, gaze tracking, or voice profiling. While effective in some settings, these methods raise serious concerns about privacy and surveillance. To overcome these limitations, this thesis prioritizes privacy preservation by leveraging only collective audio–visual signals, focusing on group-level rather than individual-level emotion recognition. The overall objective is to develop multimodal models that can infer group emotions while avoiding the risks associated with individual monitoring and surveillance. Two complementary frameworks are proposed to achieve this goal. The first introduces a cross-attention multimodal architecture for audio–video fusion, combined with a Frames Attention Pooling (FAP) strategy. This framework is further supported by synthetic data augmentation and validated through extensive ablation studies. These experiments demonstrate his effectiveness and robustness for GER in real-world conditions. The second, the Variational Encoder Multi-Decoder (VE-MD), introduces a shared latent space jointly optimized for emotion classification, body, and face structural representation prediction. Two structural representation decoding strategies are explored: DETR-based and heatmap-based, highlighting their respective strengths and limitations in group versus individual settings. A detailed analysis reveals how structural representation integration impacts GER differently compared to non-GER.The scientific contributions of this thesis are threefold. First, it provides new insights into the role of multimodality and structural representation-based cues for group-level affective computing, clarifying how group and individual settings diverge in their requirements and challenges. Second, it advances methodological design through the introduction of two complementary frameworks: a cross-attention fusion model with FAP for temporal aggregation, and VE-MD as a generalizable latent space for multitask learning. Third, it establishes a privacy-preserving paradigm for GER, showing that competitive or state-of-the-art performance can be achieved without relying on individual features as input data.

\section*{Résumé}

Cette thèse aborde le défi de la reconnaissance des émotions de groupe (GER) en conditions naturelles. Les approches traditionnelles de la reconnaissance des émotions s'appuient souvent sur des indices individuels tels que la reconnaissance faciale, le suivi du regard ou le profilage vocal. Bien qu'efficaces dans certains contextes, ces méthodes soulèvent de sérieuses préoccupations en matière de confidentialité et de surveillance. Pour surmonter ces limites, cette thèse donne la priorité à la préservation de la vie privée en exploitant uniquement des signaux audiovisuels collectifs, se concentrant sur la reconnaissance des émotions au niveau du groupe plutôt qu'au niveau individuel. L'objectif global est de développer des modèles multimodaux capables de déduire les émotions d'un groupe tout en évitant les risques de manipulation et de surveillance individuelle. Deux modélisations complémentaires sont proposées pour atteindre cet objectif. La première introduit une architecture multimodale à attention croisée pour la fusion audio-vidéo, combinée à une stratégie de Frames Attention Pooling (FAP). Cette modélisation est en outre soutenue par l'augmentation des données synthétiques et validée par des études d'ablation approfondies. Ces expériences démontrent son efficacité et sa robustesse pour le GER dans des conditions réelles. La seconde, le Variational Encoder Multi-Decoder (VE-MD), introduit un espace latent partagé optimisé conjointement pour la classification des émotions et la prédiction de la représentation structurelle du corps et du visage. Deux stratégies de décodage de la représentation structurelle sont explorées : celle basée sur DETR  et celle basée sur la carte thermique, mettant en évidence leurs forces et leurs limites respectives dans des contextes de groupe et hors groupe. Une analyse détaillée révèle comment l'intégration de la représentation structurelle a un impact différent sur le GER par rapport au non-GER. Les contributions scientifiques de cette thèse sont triples. Premièrement, elle apporte de nouvelles perspectives sur le rôle de la multimodalité et des indices basés sur la représentation structurelle pour la reconnaissance affective au niveau du groupe, en clarifiant comment les contextes de groupe et individuels divergent dans leurs exigences et leurs défis. Deuxièmement, elle fait progresser la conception méthodologique grâce à l'introduction de deux modélisations complémentaires : un modèle de fusion d'attention croisée avec FAP pour l'agrégation temporelle, et VE-MD comme espace latent généralisable pour l'apprentissage multitâche. Troisièmement, elle établit un paradigme de préservation de la vie privée pour le GER, montrant que des performances compétitives ou de pointe peuvent être obtenues sans s'appuyer sur des caractéristiques individuelles comme des données d'entrée.

\section*{Rezime}
Tèz sa a adrese defi rekonesans emosyon gwoup (GER) nan kondisyon natirèl. Apwòch tradisyonèl yo pou rekonesans emosyon souvan apiye sou siyal endividyèl tankou rekonesans vizaj, swivi je, oswa pwofilaj vwa. Malgre yo efikas nan kèk kontèks, metòd sa yo soulve gwo enkyetid sou vi prive ak siveyans. Pou simonte limitasyon sa yo, tèz sa a bay priyorite ak prezèvasyon vi prive lè li itilize sèlman siyal odyovizyèl kolektif yo. Li konsantre sou rekonesans emosyon gwoup moun ansanm olye chak grenn moun nan group la. Objektif jeneral la se devlope modèl miltimodal (zouti entèlijans atifisyèl) ki kapab rekonèt emosyon gwoup moun pandan y'ap evite risk manipilasyon ak siveyans endividyèl. De apwòch modelizasyon konplemantè pwopoze pou reyalize objektif sila a. Premye a prezante yon achitekti miltimodal atansyon kwaze pou fizyon odyo-videyo, konbine avèk yon estrateji Frames Attention Pooling (FAP). Modèl sa a sipòte pa ogmantasyon done sentetik epi valide pa etid ablasyon divès. Eksperyans sa yo demontre efikasite ak robistès modèl la pou GER nan kondisyon natirèl. Dezyèm nan, Variational Encoder Multi-Decoder (VE-MD), entrodwi yon espas latan pataje optimize ansanm pou klasifikasyon emosyon ak prediksyon reprezantasyon estriktirèl kò ak vizaj moun. Gen de strateji ki eksplore pou dekode reprezantasyon estriktirèl yo: yonn ki baze sou yon modèl DETR ak yonn ki baze sou kat chalè (heatmap), aksan mete sou fòs ak limit respektif yo nan kontèks gwoup ak non-gwoup. Yon analiz detaye revele kijan entegrasyon reprezantasyon estriktirèl yo gen yon enpak diferan sou GER konpare ak sa ki pa GER (non-GER). Kontribisyon syantifik tèz sa a gen twa aspè. Premyèman, li bay yon nouvo apèsi sou wòl miltimodalite ak siyal ki baze sou reprezantasyon estriktirèl pou rekonesans siyal afektif nan nivo gwoup, li klarifye kijan kontèks gwoup ak endividyèl yo divèje nan egzijans ak defi yo. Dezyèmman, li fè pwogrese konsepsyon metodolojik atravè entwodiksyon de modèl konplemantè: yon modèl fizyon atansyon kwaze ak FAP pou agregasyon tanporèl, ak VE-MD kòm yon espas latan jeneralizab pou aprantisaj milti-tach. Twazyèmman, li etabli yon apwòch ki prezève vi prive pou GER, li montre ke pèfòmans konpetitif oswa dènye kri yo ka reyalize san yo pa konte sou karakteristik endividyèl kòm done an antre.

\newpage
\section*{Acknowledgment}
\addcontentsline{toc}{chapter}{Acknowledgment} %

First, I would like to express my deepest gratitude to my supervisors, Dominique Vaufreydaz and Frédérique Letué, for accepting me as their PhD student and for their continuous support throughout this journey. Without them, this PhD would not have been possible. Coming from a background primarily in mathematics, I had to strengthen my computer science skills in many areas during this PhD. They were always present and available, and I have learned a lot from them.

 I also thank all the members of my defense committee for taking part in my defense. In particular, I thank Alessandro Vinciarelli and Antitza Dantcheva for accepting to review my thesis manuscript and for the time and care they devoted to their reviews. I would also like to thank Christine Keribin, who was present at the beginning of my PhD as an external expert and followed my progress throughout the project. I would like to thank PERSYVAL Labex (ANR
11-LABX-0025) for funding the first years of this PhD project.

I thank all the PhD students and all members of the M-PSI team for their moral and technical support. I especially thank Fiorela Albasini, former the team engineer, for her availability whenever I needed technical test.

I would like to thank the institutions that have supported me since my master’s studies in Lyon: Asosyasyon Orijinè Granplenn (AOG), Fondasyon Konesans Ak Libète (FOKAL), and Lyon Haïti Partenaria (LHP). I would also like to thank the people who accompanied me during my master’s degree: Chantal Gérard, Yves Gérard, Gaston Jean, and Pascal Naquin. Their support and advice during my master’s studies made this doctorate possible.

I would like to thank all the teachers of the Institut des Sciences Financières et d’Assurance (ISFA) of Université Claude Bernard Lyon1 for the rigorous training they provided me in computer science, IT security, data science, and advanced algorithms. This education allowed me to explore a wide range of applied topics at the intersection of mathematics and computer science, and ultimately to pursue this thesis in artificial intelligence.

I would also like to thank the École Normale Supérieure (ENS) of Université d’État d’Haïti (UEH), in particular the teachers who provided me with a high-level, rigorous, and wide-ranging education in mathematics. This training enabled me to explore both fundamental and applied mathematics, and helped me build the foundations that made this PhD possible.

\vspace{1cm}
\noindent Je remercie la famille Louis (Ruben et Wanglaise) pour la relecture de mon manuscrit. Egalement, Je remercie Richenide Decamp pour la relecture, ainsi que pour m’avoir aidé à repérer des coquilles dans le manuscrit. Je tiens à remercier Ralph Papouche Desmard pour ses commentaires et ses retours enrichissants et chaleureux sur mes réalisations au cours de cette thèse. Je remercie tous mes compatriotes haïtiens qui m’ont accompagné durant ce voyage, et qui m’invitent toujours à des activités, m’empêchant de devenir fou durant cette thèse.

\noindent Enfin, je voudrais remercier la personne la plus importante dans ma vie, celle qui a toujours cru en moi, même durant les périodes les plus difficiles: ma mère, Angeline Ultimé. Cette thèse lui est tout particulièrement dédiée. Je remercie aussi le reste de ma famille, en l’occurrence mes trois petites sœurs: Estherline Augusma, Aderline Augusma et Schelah Augusma, qui m’ont soutenu dès le début dans mon parcours d’études.

\vspace{1cm}

                                                                \begin{center}
                                                                    Tout pwòch, ak tout moun ki te kwè nan mwen mèsi anpil !
                                                                \end{center}

   \tableofcontents

\chapter{Introduction}
\label{chap:intro}
This chapter establishes the foundational context and motivation for the thesis, emphasizing the dynamic nature of teaching and learning interactions as conveyed through multimodal signals such as facial expressions, gestures, and vocal cues. It introduces the concept of Context-Aware Classrooms (CAC), highlighting how multimodal data can inform pedagogical strategies while addressing ethical and privacy concerns. Then, it defines the objective and the methodological direction of the thesis.  The chapter also introduces Group Emotion Recognition (GER) and discusses perceived emotions in-the-wild, underscoring the challenges of interpreting emotional expressions in light of contextual, cultural, and individual variability.
Finally, it outlines the main contributions of the
thesis and presents an overview of its structure, setting the stage for the detailed work
that follows.

\section{General Context}
\subsection{Motivation}

Teaching is a complex, dynamic process where educators and learners jointly construct knowledge through continuous interactions. These interactions are mediated by multimodal signals such as facial expressions, gaze patterns, postural shifts, vocal prosody, and gestural cues that reflect engagement, comprehension, and emotional states. For example, teachers may adjust their pace according to students' frowns or redirect attention by interpreting the divergence of the collective gaze. Conversely, students assess the clarity of instruction based on the teacher's gestures and tone. Yet much of this interaction remains implicit, hidden beneath layers of transient behaviors and cognitive processes that are difficult for a teacher to discern during the teaching session.  The main reason for this is that our behavior as human beings and teaching are social, and therefore multimodal~\citep{vinciarelli2009social}.

\begin{figure}[H]
    \centering
\includegraphics[width=0.7\textwidth]{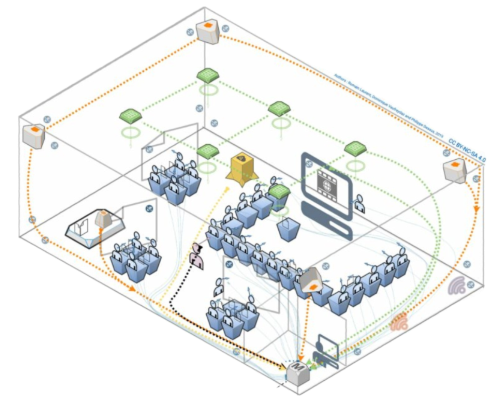}
    \caption{Smart classroom design (CAC) of the Teaching Lab project~\citep{laurent:hal-02438020}.}
    \label{fig:cac}
\end{figure}

The rise of Context-Aware Classrooms (CAC) technology-enhanced spaces equipped with ambient sensors such as cameras, microphones, and eye-trackers offers unprecedented opportunities to decode these interactions (see Figure~\ref{fig:cac}). By capturing multimodal data, such as classroom agitation, gaze scan paths, and body postures, CACs enable computational modeling of pedagogical dynamics at scale. 
This thesis is part of the interdisciplinary Teaching Lab project (\textit{MANIP: Modeling and Analysis of Instructional Processes}), which aims to improve teaching practices through the use of a CAC. Rooted in ecological psychology, MANIP conceptualizes classrooms as systems of perception and action where teachers and students engage in continuous, iterative interactions. Within this framework, educators and learners first direct their attention to resources, such as teaching materials or peers, as part of a perceptual process. Their subsequent actions then modify the environment, generating new opportunities for interaction during the action phase. Thanks to these iterative cycles, shared goals and knowledge emerge through co-construction.
While these interactions can now be captured using modern sensing technologies, such data collection raises critical questions of privacy, ethics, and responsible AI, particularly in educational contexts.

\subsection{Handling Privacy Safely}
To model instructional processes, MANIP integrates multimodal sensing such as mobile eye-trackers for teacher gaze and ambient cameras for group posture analysis with machine learning and statistical modeling. Crucially, the project adopts a privacy-safe, non-individual approach. Instead of tracking individuals, the analysis focuses on group-level features, such as global facial expression distributions and aggregated vocal patterns. Ethical data practices are prioritized, with faces masked and datasets obfuscated to comply with GDPR (General Data Protection Regulation)~\footnote{A European Union law that governs the collection, processing, and storage of personal data to ensure individuals’ privacy and data security.} and established ethical guidelines. This approach addresses two critical challenges. On the technological front, deploying deep learning models in-the-wild classroom settings is hindered by variability in lighting, occlusion, hybrid teaching setups, and sparse annotations, all of which reduce reliability. On the ethical front, the project balances innovation with privacy preservation, ensuring that context-aware classrooms enhance pedagogy without surveilling individuals.  

Importantly, the methodology aligns with the European Union’s proposed Artificial Intelligence Act (AI Act), which sets forth comprehensive regulatory frameworks for ethical usage, transparency, and robust data protection within AI systems. Educational AI systems involving biometric data, such as facial expressions and vocal patterns, were anticipated to fall under the Act’s categorization of high-risk AI. To anticipate this, the project proactively incorporates principles of privacy by design, aggregating anonymized data into group-level metrics to minimize risks associated with individual monitoring or control. To balance pedagogical insight with ethical responsibility, this thesis aims to design privacy-preserving multimodal frameworks capable of recognizing group emotions without individual tracking or providing individual features.

\subsection{Objective}

The objective of this thesis is to advance group emotion recognition as a foundation for analyzing engagement in classroom environments, with an emphasis on privacy preservation. This work lies within the broader scope of \textbf{multimodal perception}, targeting both audio and video information captured in real-world classroom settings. Two main research goals are defined:

\begin{enumerate}
    \item To develop multimodal models for group emotion recognition in-the-wild based on audio-visual signals.  
    \item To ensure privacy-preserving learning by avoiding the use of individual features in both input and output representations.  
\end{enumerate}

In this perspective, privacy preservation is defined as the exclusion of individual-specific information that could enable control, monitoring, or surveillance. For visual data, this means avoiding the use of isolated face crops, skeletons, or positional metadata as explicit inputs~\citep{dantcheva2015else}. For audio data, no individual monitoring or speaker-specific characteristics are used. Instead, models operate directly on global images, full video frames, and raw audio streams. This ensures that only collective information, such as overall facial expression distributions, ambient vocal patterns, and group-level postural trends, is used for inference.

\section{Group Emotion Recognition In The Wild}

Group emotion recognition has gained increasing attention due to its applications in education, public safety, and social interaction. In classrooms, for example, Group Emotion Recognition (GER) can provide valuable insights into collective engagement, while in public events it can help monitor crowd dynamics~\cite{dhall2017individual, huang2024survey}.  

\subsection{From Individual Emotions to Group-Level Perception}
\begin{figure}[H]
    \centering
\includegraphics[width=0.85\textwidth]{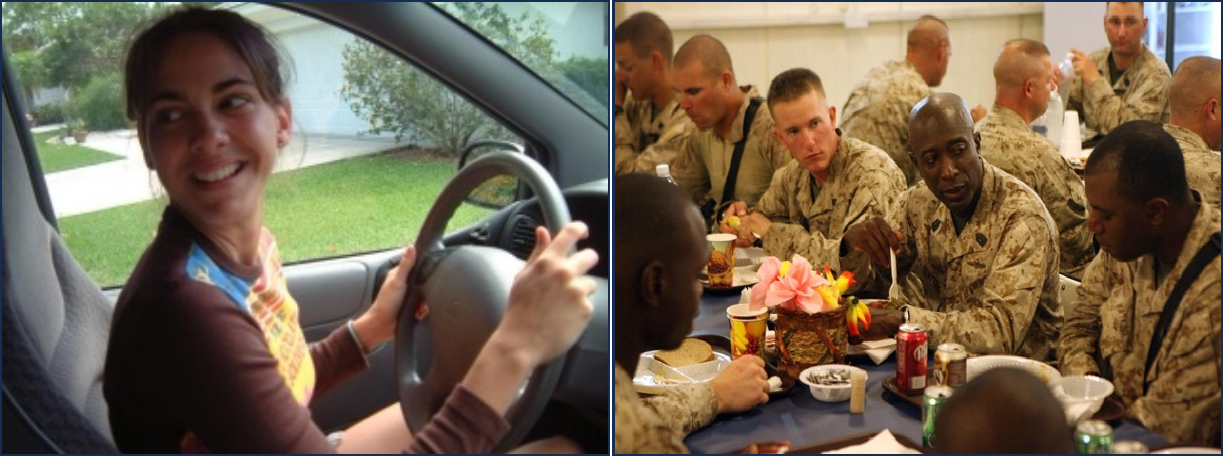}
    \caption{Example of individual and group emotions in-the-wild from EMOTIC dataset~\cite{kosti2017emotion}. At left is an individual emotion for the person in the image. On the right is a group emotion.}
    \label{fig:ind_and_ge}
\end{figure}

Human emotions can be approached from three complementary perspectives: experienced emotion, expressed emotion, and perceived emotion~\cite{Frijda01062005,gandhi2020perception}. Experienced emotion refers to the internal and subjective state felt by an individual, such as happiness or anxiety. Expressed emotion corresponds to the outward manifestation of these feelings through facial expressions, voice, posture, and gestures. Perceived emotion represents the interpretation of these signals by others, whether human observers or AI systems. In the context of GER, the focus is on perceived emotion, which must be inferred from multimodal cues expressed by several individuals simultaneously. An example of group versus individual emotion is given in Figure~\ref{fig:ind_and_ge}.

\subsection{Challenges in GER In The Wild}

Several challenges arise when moving from individual to group-level perception. First, emotional expression is inherently multimodal, encompassing not only the face but also voice, posture, gestures, and physiological cues. Reliance on a single modality, such as facial expressions, is insufficient for robust GER. Second, emotional meaning is highly context-dependent: the same expression may signal different emotions depending on the situation and cultural norms. Third, emotions often serve social and evolutionary functions beyond reflecting internal states, such as pride, embarrassment, or politeness, which complicates their interpretation. Finally, human affect extends beyond Ekman’s six basic emotions to include a wide range of complex categories such as awe, relief, or affection~\cite{keltner2016expression}, making AI-based recognition even more challenging.  

In summary, GER requires AI systems that move beyond basic facial analysis to integrate multimodal cues, contextual information, and the richness of human emotions. These challenges highlight the need for novel approaches capable of addressing the complexity of group-level affect recognition in-the-wild, motivating the multimodal and privacy-preserving frameworks proposed in this thesis.

\section{Thesis Contributions}

The contributions of this thesis are organized around two complementary frameworks for group emotion recognition in-the-wild. 

The first is a cross-attention multimodal model enhanced by synthetic data augmentation. This framework introduces a privacy-preserving architecture that avoids reliance on individual-specific features, while enabling effective audio–video fusion through a cross-attention mechanism. In addition, we adapt a Frames Attention Pooling~(FAP), inspired from Attentive Statistic Pooling (ASP) without the standard deviation modeling presented by ~\citet{okabe2018attentive}, which allows robust temporal aggregation without covariance modeling. Together, these innovations provide a contribution to multimodal affective computing by demonstrating how cross-attention and pooling strategies can improve group-level emotion recognition in the classroom and other real-world scenarios.  

The second major contribution is the Variational Encoder Multi-Decoder (VE-MD) framework, which leverages a shared latent space for multi-task learning. Unlike traditional architectures, VE-MD jointly optimizes emotion classification, body, and face structural representation prediction within a unified model. Through extensive analysis, we reveal distinct roles of structural representation integration in group versus non-group datasets, providing new insights into how group affect differ from individual-level expression. VE-MD also extends to multimodal settings, incorporating audio and text alongside visual signals, and demonstrates the privacy-preserving potential of group-level structural representations.  

The VE-MD architecture is designed with two structural representation decoding strategies (DETR-based and heatmap-based), an emotion decoder enhanced with spatio-temporal graph convolution (ST-GCN), and a multimodal classification head that supports late fusion, cross-attention, and attention-guided fusion. All these processing is done through an automatic annotation pipeline for body and face structural representations, a unified multi-task loss for joint optimization, and extensive empirical validation, for group and non-group,  across six benchmark datasets (GAF-3.0~\citep{gupta2018attention}, VGAF~\citep{dhall2020emotiw}, SAMSEMO~\citep{bujnowski2024samsemo} , MER-MULTI~\citep{lian2023mer}, DFEW~\citep{jiang2020dfew}, and EngageNet~\citep{emotiw2023}). These experiments establish new state-of-the-art results on several datasets and confirm the effectiveness of our proposed frameworks.  

In summary, this thesis provides both novel scientific insights into group-level emotion recognition and concrete technical solutions, contributing to the development of privacy-preserving, multimodal, and scalable affective computing systems.

\section{Thesis Overview}

After the introduction  , Chapter~\ref{chap:sota} consists in the state of the art in emotion recognition, with a focus on multimodal approaches in in-the-wild settings. The chapter highlights the advantages of multimodal over unimodal methods, recent advances in deep learning architectures and attention mechanisms, and the growing importance of Group Emotion Recognition (GER). Particular emphasis is placed on privacy and ethical concerns, reviewing privacy-preserving strategies such as synthetic data generation and non-individual approaches. The chapter concludes with a survey of widely used public datasets, a description of those adopted in this thesis, and a snapshot of performance benchmarks.

Chapter~\ref{chap:mger} depicts the proposed framework cross-attention multimodal model for group emotion recognition. This framework introduces a cross-attention mechanism for effective audio–video fusion and a Frames Attention Pooling (FAP) strategy, inspired by Attentive Statistic Pooling, to aggregate temporal features without the standard deviation modeling. The chapter also explores the role of synthetic data augmentation to improve model robustness. Extensive ablation studies are reported, analyzing frame count, fusion methods, pooling strategies, and training configurations. This approach \textbf{wins} the EmotiW 2023 challenge by achieving the best performance, demonstrating the effectiveness of cross-attention and privacy-preserving multimodal modeling in-the-wild.

Chapter~\ref{chap:ve-md} introduces the Variational Encoder Multi-Decoder (VE-MD) framework for group and non-group emotion recognition. This approach leverages a shared latent space jointly optimized for emotion classification, global structural representation prediction for body and face. The chapter compares two structural representation decoding strategies, DETR-based and heatmap-based, and analyzes the effect of structural representation integration on group versus non-group datasets. It further extends VE-MD to multimodal fusion with audio and text, supported by a unified multitask loss and an automatic annotation pipeline for person structural representations. Extensive experiments across six datasets demonstrate the effectiveness of the framework, with VE-MD achieving new state-of-the-art results on GAF-3.0, VGAF, and SAMSEMO, while also being competitive on MER-MULTI, EngageNet, and DFEW.

In Chapter~\ref{chap:conclusion}, we summarize the main scientific contributions of the thesis, emphasizing the development of privacy-preserving multimodal frameworks for group emotion recognition. The chapter discusses the limitations encountered, particularly in terms of privacy trade-offs, dataset annotation quality, and robustness in noisy environments. Building on these observations, it outlines future research perspectives, including strategies for enhancing privacy in latent spaces, improving annotation consistency with the support of vision–language models, and advancing audio processing through separation techniques.

\chapter{State of the Art and Literature Review}
\label{chap:sota}

\section{Introduction}
This thesis is part of a project on Context-Aware Classrooms (CAC) technology, enhanced with cameras, microphones, and opportunities to decode interactions between students and teachers. This is a shared environment with people and objects, and classroom noise. In the objective of modeling pedagogical activities, the current thesis focuses on multimodal emotion recognition since it is considered as a sub-dimension of student engagement. That means the modeling approach should handle audio and video for emotion recognition in the wild. However, working in a CAC with students reveals a lot of concern about privacy, safety, fairness, and ethical issues on individual-level tracking.

The aim of this chapter is twofold. First, it demonstrates that \textbf{multimodal approaches to Emotion Recognition (ER) in-the-wild} are more effective than unimodal ones, as they integrate richer and more diverse information relevant to emotion detection. Second, it demonstrates that \textbf{emotion recognition, being inherently focused on individual-level data, poses significant risks to personal privacy}.

To support these claims, the chapter presents a comprehensive review of the current state of the art in emotion recognition, covering both \textbf{unimodal and multimodal methodologies}. It explores recent developments in \textbf{visual, audio, and textual modalities}, with particular attention to advances in {deep learning architectures, attention mechanisms}, and \textbf{privacy-preserving techniques}. A key focus is placed on \textbf{Group Emotion Recognition (GER)}, which is the main exploration in this thesis. The chapter discusses the strengths and limitations of the proposed methodologies, alongside their recent evolutions and challenges in real-world (in-the-wild) scenarios.

Furthermore, the chapter examines the \textbf{issue of privacy in ER}, reviewing various efforts to address it at both the \textbf{individual and group levels}, in controlled environments and in the real world. Particular attention is paid to synthetic data generation and non-individual approach techniques, which offer promising directions for large-scale, privacy-aware emotion recognition. The chapter concludes with a summary of public datasets widely used in the in-the-wild GER community and a list of the datasets used in this thesis, along with a snapshot of state-of-the-art (SOTA) performance benchmarks at the beginning of the thesis.

\section{Emotion Recognition in Unimodal Settings}
Emotion recognition in unimodal settings involves analyzing a single type of data, visual, audio, or textual, to compute emotional states. The visual modality refers to either a single image or video. The audio refers to any kind of sound, like speech, music, background noise, etc.  And the textual modality refers to text information. Each modality possesses distinct strengths and challenges, and the choice of modality typically aligns with the specific context or application.~\citet{tomar2023unimodal} extensively reviews the various approaches utilized in unimodal emotion recognition, highlighting visual, audio, and textual modalities as the most commonly studied areas. This section discusses each modality in detail, emphasizing key methodologies, challenges, and recent advancements.

\subsection{Visual Modality}
\label{visual_modal}

The visual modality is extensively explored in emotion recognition, as facial expressions and body language provide rich emotional information ~\citep{cao2025deep, muradulloyeva2025importance, asatullaev2025body, baxtiyarovich2025interpreting}.  Research within this domain operates at two primary levels: \textbf{individual} and \textbf{group}. \textbf{Individual level} will refer to the fact of applying emotion recognition to a single person in an image or video. \textbf{Group-level} refers to applying emotion recognition to a group of people when there are at least two people in an image or video.

\subsubsection{Individual Level Visual Modality}

Emotion recognition at the individual level is very common in the research area. Most of the time, it is addressed via Facial Expression Recognition (FER). It is one of the dominant use cases at the individual level.~\citep{ko2018brief, cao2025deep, muradulloyeva2025importance, asatullaev2025body, baxtiyarovich2025interpreting}.  Applications of FER are widespread, including healthcare, human-computer interaction, education,  and security~\citep{argaud2018facial, huang2023study, ortmann2025unimodal, chumachenko2024mma, el2023emonext, dada2023facial, nawaz2025novel, pordoy2024multi}.

Despite significant advancements due to deep learning, persistent challenges remain, such as age bias, occlusion, and cultural specificity. ~\citet{huang2024facial} addresses age bias by proposing the Age Group Expression Feature Learning (AEFL) framework, which employs a multi-branch network structure. Each network branch captures expression features specific to distinct age groups (e.g., infants, adolescents, young adults, middle-aged adults, and the elderly). A global feature extractor gathers shared emotional cues, with an age-guided attention mechanism ensuring that distinctive features from less-represented age groups remain influential. The model is evaluated on three main datasets such as FACES, RAF-DB, and AffectNET. 

\begin{figure}
  \centering
 \includegraphics[width=0.95\textwidth]{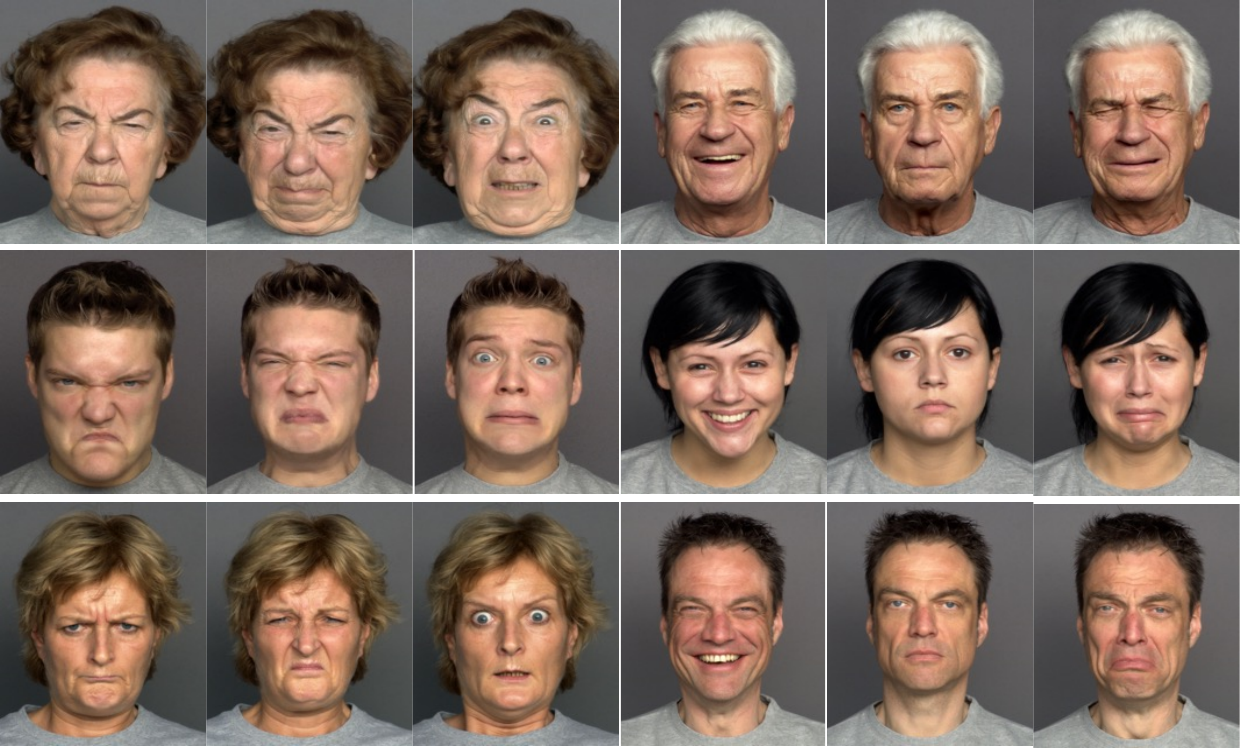}
  \caption{FACES dataset overview: From left to right, the emotion goes like: Anger, Disgust, Fear, Happy, Neutral, and Sad.}
  \label{fig:faces}
\end{figure}

\begin{figure}
  \centering
 \includegraphics[width=0.95\textwidth]{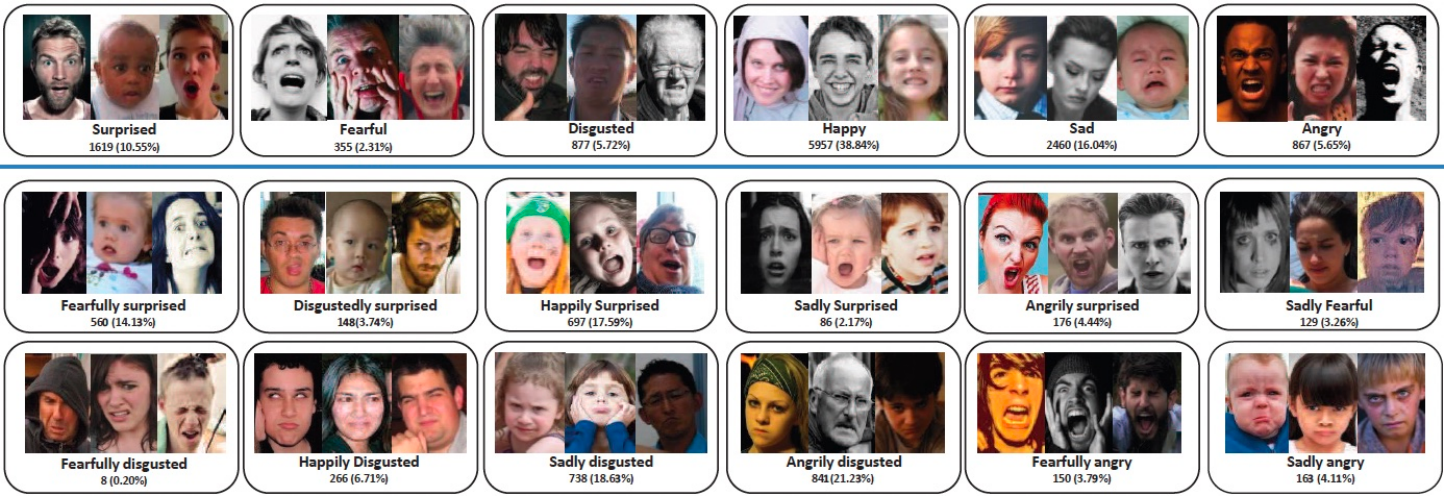}
  \caption{RAF DB overview: six-class basic emotions and twelve-class compound emotions, source:~\citet{li2017reliable}.}
  \label{fig:rafdb}
\end{figure}

The FACES database~\citep{ebner2010faces}  comprises 2,052 faces of individuals balanced for gender, each displaying six prototypical facial expressions (neutrality, sadness, disgust, fear, anger, and happiness).  RAF-DB~\citep{li2017reliable} is a dataset that comprises around 30000 in-the-wild facial images. The dataset is annotated by the basic emotions (surprise, fear, disgust, happiness, sadness, anger, and neutral).  The dataset is partitioned into a single-label subset of images with a dominant emotion and a two-label (compound) subset of samples spanning 11 compound-emotion classes. An overview of these FACES and RAF-DB is respectively given in Figure~\ref{fig:faces}, and Figure~\ref{fig:rafdb}. The AffectNet~\citep{mollahosseini2017affectnet} dataset is the largest publicly available in-the-wild facial affect corpus, comprising over one million images. A subset of 450,000 images was annotated the same as RAF-DB with seven basic emotion categories (neutral, happiness, sadness, surprise, fear, disgust, anger), but with continuous valence and arousal scores in addition.

Beyond the efficient methodology used in that work, the FER allows achieving promising results on the RAF-DB  (Real-World Affective Faces Database) and on the FACES dataset, for a performance of $90.09\%$ and $73.49\%$ with an improvement of $+2.69\%$ and $+5.27\%$  of accuracy, respectively, compared to the baseline. Their approach achieves the state of the art on the  RAF-DB ($90,09\%$ accuracy), AffectNet-7 ($65,57\% $ accuracy), and AffectNet-8 ($61,58\%$ accuracy).  But the use of multi-branch structures and attention mechanisms such as Age Guided Attention (AGA), Age Group Feature (AGF), and Global Age Guided Feature introduced additional computational complexity, while performance results compared to the state of the art are barely improved with less than 1\% gain in accuracy for all three datasets. However, \citet{balachandran2025facial}  propose a novel age-adaptive Facial Emotion Recognition (FER) model called MViT-CnG, which combines multi-scale Vision Transformers (ViT~\citep{Dosovitskiy2020}) with contrastive learning to improve the recognition of emotions across diverse age groups.  The model leverages the ViT’s ability to capture both fine-grained local and global facial features at multiple scales, while contrastive learning enhances feature discriminability by bringing similar expressions closer and pushing dissimilar ones apart in the embedding space. The authors apply an Improved Single-Shot Multi-box Detector (ImSSD,  ~\citep{kumar2020object} ) for precise facial region extraction. The model is trained on two benchmark datasets: FER-2013  ~\citep{goodfellow2013challenges}, which includes images spanning various age groups, and the CK+ dataset  ~\citep{lucey2010extended}, with posed expressions in controlled settings. Experimental results show outstanding performance, with the MViT-CnG model achieving accuracy rates of 99.6\% and 99.5\% on FER-2013 and CK+ datasets, respectively. It demonstrates superior precision, recall, and F1 scores compared to existing FER models, confirming robustness and generalizability, especially in recognizing subtle facial cues across different ages. Limitations include challenges in accurately detecting emotions in children and elderly groups due to dataset biases and the need for further diversification of training samples.

\citet{petrou2023lightweight} tackle occlusions by simulating the partial facial occlusion typically caused by virtual reality (VR) headsets or smart glasses, building on the work of ~\citet{rodrigues2022classification}. Using geometric masking and transfer learning with a mini-model of Xception~\citep{arriaga2019real}, they show that despite a moderate drop in performance, the model remains effective in occlusion scenarios. The best-performing model (finely tuned with all re-entrainable layers) achieved the highest accuracy (69\%) and F1 score (68\%) for occluded faces with only minimal performance degradation (4\%) compared to the non-occluded scenario. However, the limitation is using artificial generation of occlusion instead of using natural occluded images.  The artificial ones may not fully capture the variability of occlusion in the real world.

Recently,~\citet{li2025occlusion} also addressed the challenging problem of facial expression recognition (FER) under occlusion, where important facial features are partially blocked, leading to degraded model performance. The authors propose a Multi-Angle Feature Extraction (MAFE) framework that improves recognition accuracy by extracting and fusing global, fine-grained, and key regional features from occluded faces. The approach integrates two powerful feature extractors: Pyramid Transformer ResNet-50 for capturing multi-scale global features and Swin Transformer Encoder (SWIN-E~\citep{liu2021swin}) for fine-grained local features. Facial landmarks guide the reframing to important areas, and an attention mechanism focuses on these areas while suppressing occluded parts. A novel Regional Bias Loss (RB-Loss~\citep{wang2020region}) encourages focus on critical facial areas, and a Consistent Feature Recognition (CFR) module with \textbf{con-feature loss}~\footnote{Con-feature loss is a loss function used to combine features extracted from two networks, intending to optimize model performance. The concept of cofunction loss is based on the idea that “two networks don't make the same mistake”. Extracted features are therefore more precise and discriminating.} ensures mutual guidance between fused and global features to enhance discriminative power. Experiments on two occlusion benchmark datasets, Occlusion-RAF-DB~\citep{wang2020region}  and Occlusion-FERPlus~\citep{wang2020region}, demonstrate the model’s superiority with accuracies of 89.42\% and 86.94\%, respectively, outperforming state-of-the-art methods. The model also achieves strong performance on the original RAF-DB and FERPlus datasets, indicating robust generalization. Ablation studies confirm the effectiveness of multiscale feature fusion and the \textbf{con-feature loss} in improving recognition accuracy. Visualization of attention maps reveals that MAFE effectively reduces focus on occluded regions, highlighting its occlusion-robust design. The MAFE model provides a comprehensive and effective solution for occlusion-robust facial expression recognition, leveraging multi-angle feature extraction and attention-guided fusion to handle complex occlusion challenges in FER.

Cultural and demographic specificity are also critical considerations for FER. \citet{khajontantichaikun2023facial} develop a tailored emotion recognition system for Thai elderly individuals using the YOLOv7  ~\citep{wang2023yolov7}, Faster R-CNN  ~\citep{renNIPS15fasterrcnn}, and SSD  ~\citep{lufficc2018ssd}  architecture. Their method successfully detects six core emotions: Neutral, Anger, Joy, Sadness, Fear, and Surprise, demonstrating YOLOv7’s efficacy for real-time monitoring and intervention in mental health contexts among specific demographic groups. YOLOv7 achieved the highest mean average precision (mAP) of 95\%, surpassing Faster R-CNN (mAP 87\%) and SSD (mAP 84\%). However, while beneficial for the Thai elderly, results might not generalize well across different ethnicities or age groups without additional data.

\begin{figure}[H]
  \centering
 \includegraphics[width=0.99\textwidth]{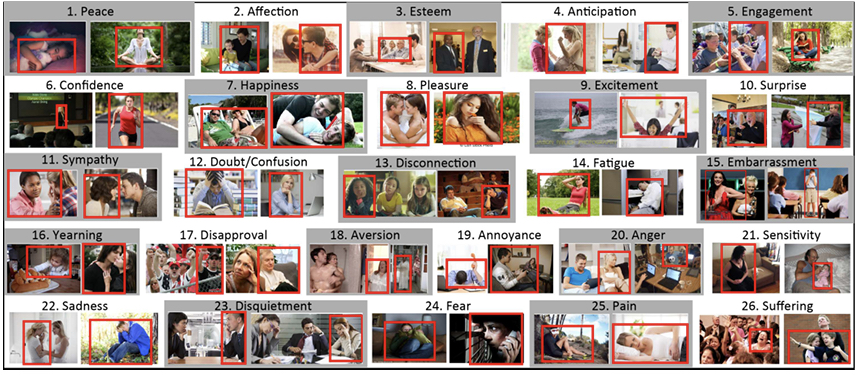}
  \caption{Examples of the 26 feeling categories of EMOTIC dataset. In each category are shown two images where the person marked with the red bounding box has been annotated with the corresponding category. source:~\citet{kosti2017emotion}.}
  \label{fig:emotic}
\end{figure}

Another way of performing emotion recognition at the individual level, in addition to FER, is emotion recognition in context, where the focus is on one person, considering the location, situation, and position that define the context.  In contrast to previous work focusing primarily on facial expressions (e.g., Ekman's six basic emotions). \citet{kosti2017emotion} has introduced EMOTIC (Emotion in Context Database) to explicitly handle the role of environmental context in affective perception. The EMOTIC dataset contains 18,316 images (23,788 annotated individuals) with a dual representation of emotions with 26 discrete categories (e.g. sad, anticipation, yearning, etc. See Figure~\ref{fig:emotic}) and continuous VAD dimensions (Valence, Arousal, Dominance). To take advantage of contextual cues (essential for 25\% of images with masked faces). The two-stream CNN they propose fuses features from body regions and global scenes, jointly trained via a combined loss function for categories and VAD. The results show that context fusion (body + scene) performs better than body alone or scene alone. Although the model allows for nuanced inferences beyond facial expressions (e.g., inferring engagement in social contexts), limitations include the subjectivity of annotation, class imbalance (low average accuracy for rare emotions such as embarrassment). A summary of results presented in this section is given in Table~\ref{tab:visual_individual}.

While much research has focused on recognizing emotions at the individual level, real-world scenarios often involve groups where collective emotional states emerge. Extending FER approaches from individuals to groups introduces new challenges and opportunities. This extension presents challenges addressed in the following section.

\begin{table}[H]
\centering
\caption{Summary of results from Visual Modality (Individual Level).}
\label{tab:visual_individual}
\resizebox{\textwidth}{!}{%
\begin{tabular}{|l|l|l|l|}
\hline
\textbf{Authors} & \textbf{Methodology} & \textbf{Dataset(s)} & \textbf{Performance/Key results} \\ \hline
\citet{huang2024facial} & Age Guided Attention & \makecell[l]{RAF-DB,\\ AffectNet(7/8),\\ FACES}   & \makecell[l]{90.09\% accuracy,\\ 65.57\% / 61.58\% accuracy,\\ 73.49\% accuracy}  \\ \hline
\citet{balachandran2025facial} & \makecell[l]{Multi-scale ViT and\\ contrastive learning} & \makecell[l]{FER-2013,\\ CK+} & \makecell[l]{99.6\% accuracy,\\ 99.5\% accuracy }\\ \hline
\citet{petrou2023lightweight} & \makecell[l]{Geometric masking and\\ transfer learning} & \makecell[l]{Occlusion scenario\\ (VR headsets)} & \makecell[l]{69\% accuracy,\\ minimal degradation (4\%)} \\ \hline
\citet{li2025occlusion} & Multi-Angle Feature Extraction & \makecell[l]{Occlusion-RAF-DB,\\ Occlusion-FERPlus} & \makecell[l]{89.42\% accuracy, \\ 86.94\% accuracy} \\ \hline
\citet{khajontantichaikun2023facial} & YOLOv7, Faster R-CNN, SSD & Thai elderly dataset & \makecell[l]{YOLOv7 achieved highest\\ mAP (95\%) }\\ \hline
\end{tabular}%
}
\end{table}

\subsubsection{Group Level Visual Modality}

Group emotion recognition extends individual FER methodologies to collective settings like classrooms, social gatherings, and public events, with education, security, and crowd management implications~\citep{tan2017group, rassadin2017group}.
To stimulate progress in this area, the EmotiW 2017 challenge introduced the Group Affect Database 2.0~\cite{dhall2017individual}, a collection of images of emotional states from diverse scenarios sourced from Google and Flickr, annotated with a global group label: positive, neutral, or negative.

\begin{figure}[H]
  \centering
 \includegraphics[width=0.75\textwidth]{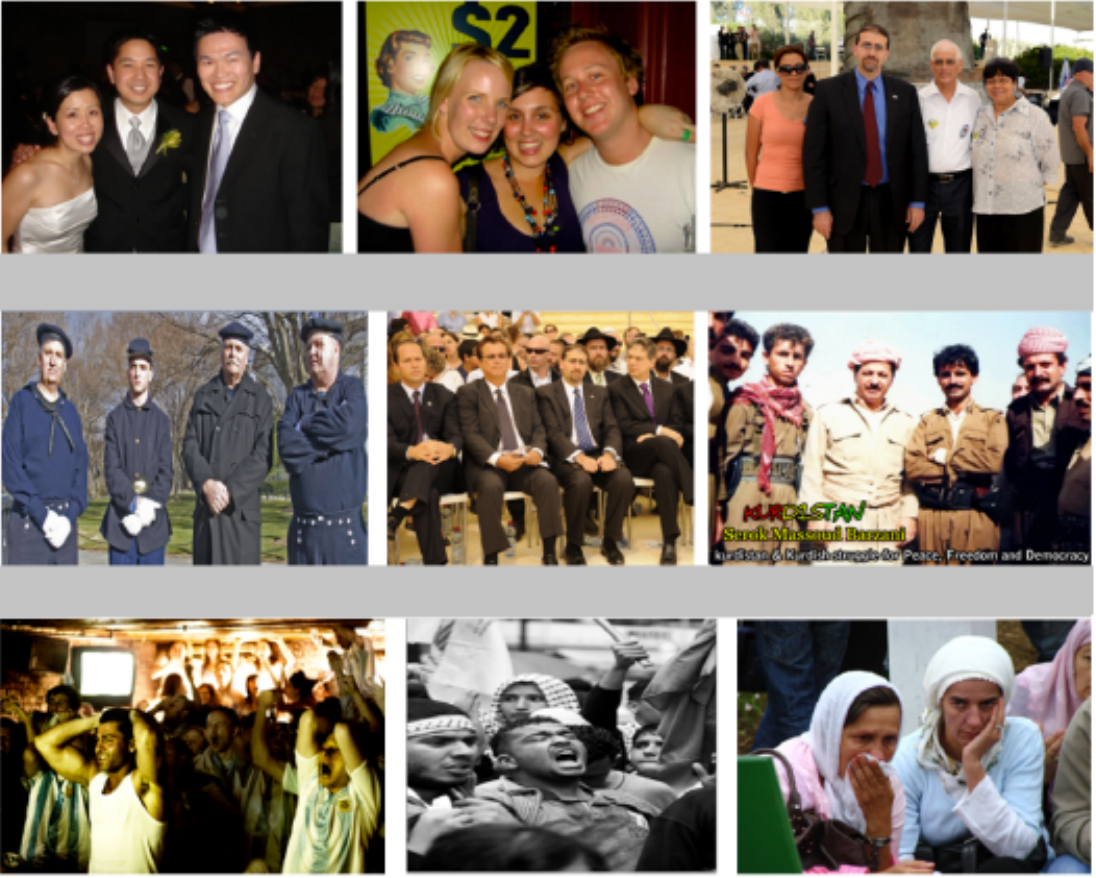}
  \caption{Group Affect Database 2.0 overview: from the top to the bottom, the emotion rows are Positive, Neutral, and Negative. source:~\citet{dhall2017individual}.}
  \label{fig:gaf2}
\end{figure}

\citet{gupta2018attention} propose a dual-branch model combining global scene analysis and detailed facial expression detection. Using DenseNet 161~\citep{huang2017densely}  for global emotion cues and SphereFace~\citep{liu2017sphereface}  for local facial analysis, their model incorporates three attention mechanisms to prioritize the most emotionally informative faces, improving recognition robustness and accuracy. The attention mechanisms perform significantly better than simple averaging, demonstrating the effectiveness of capturing key facial features. Their architecture achieved 80.90\% on the validation set and 64.83\% accuracy on the test set of the EmotiW 2018 challenge~\citep{dhall2018emotiw}. A notable gap between validation and test performance indicates potential domain change issues requiring further investigation.

\citet{Petrova20} proposed a non-individual approach to group emotion recognition in-the-wild using the VGAF dataset~\citep{dhall2020emotiw}. To preserve individual privacy, their approach uses the whole image as input instead of cropping faces or extracting bodies from the overall scene. They use synthetic images to augment the data by superimposing faces showing well-known basic emotions (from FACES dataset~\citep{ebner2010faces}) on a random background to attenuate non-specific image information. They used fine-tuning mechanisms with VGG19~\citep{simonyan2014very}  to train the model. The approach achieved an accuracy of 52.36\% on validation, a gain of 1.06\% over the baseline validation set. Their participation in the EmotiW 2020 challenge~\citep{dhall2020emotiw}  achieved 59.12\% accuracy on the test set, a gain of 11.24\% over the baseline test set. Although the model performed well on the test set, it remained weak on the validation set, which could be explained by the fact that the single-image approach chooses one image per video, and potentially not the best image within the video. In addition, the absence of temporal modeling of the video could explain the weak performance. 

\citet{gong2025hybrid} propose a hybrid fusion model aimed at improving group-level emotion recognition in complex and dynamic scenarios, such as crowded environments and diverse social interactions. Recognizing the challenges inherent in accurately capturing collective emotions, especially when individual expressions vary considerably within a group, the authors introduce a sophisticated multimodal framework that effectively integrates both facial and contextual information. The proposed hybrid model comprises three main modules: an individual emotion recognition module using facial cues extracted by a convolutional neural network (CNN), a contextual module capturing scene and environment information, and human pose detection. The fusion approach combines individual and contextual features through attention mechanisms that dynamically weight the importance of these features according to their relevance to the group's emotional state. Experimental results validate the model's effectiveness, demonstrating superior performance to reference methods on the Group Affect Database 2.0~\citep{dhall2018emotiw} with a performance accuracy of 78.99\%. 

\begin{figure}[H]
  \centering
\includegraphics[width=0.75\textwidth]{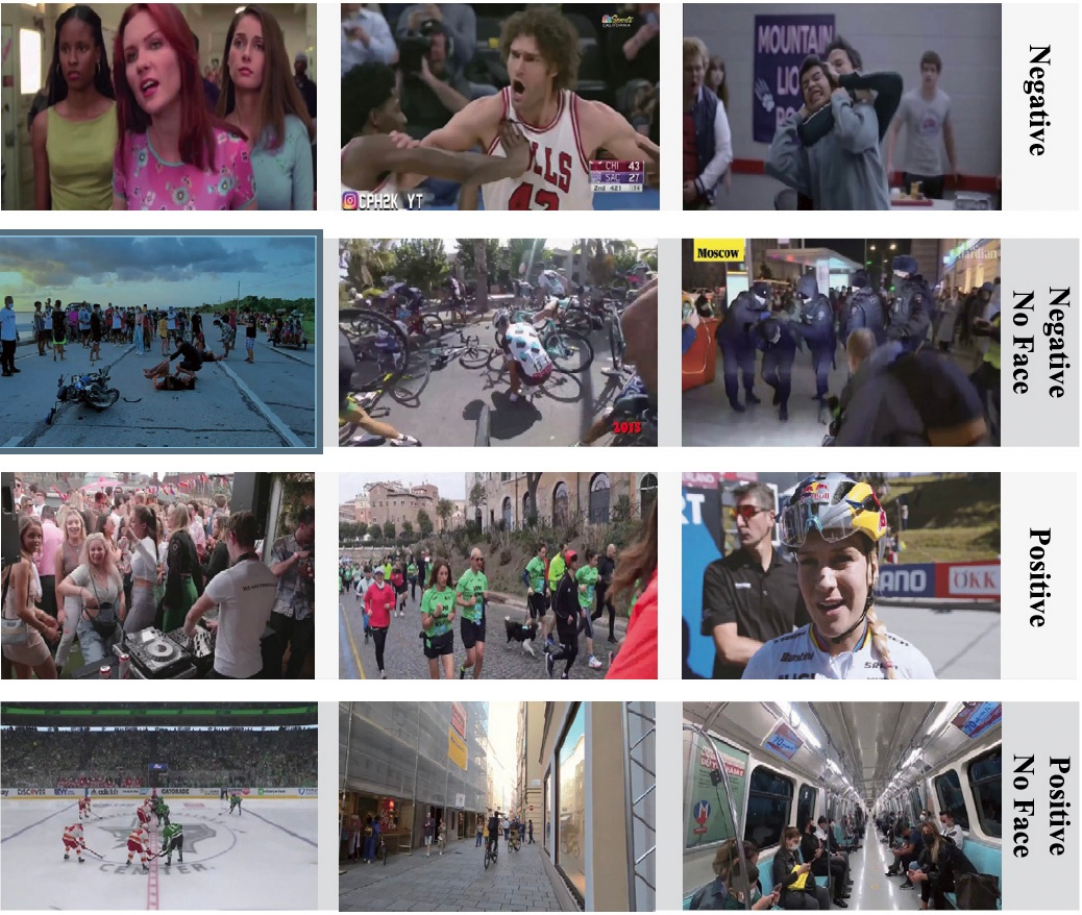}
  \caption{Group and Scene Database overview: The first two rows contain negative frames with faces and without faces. And the two last ones contain positive frames with faces and without faces, source:~\citet{gong2025hybrid}.}
  \label{fig:groupscene}
\end{figure}

Another experiment is made on a new collected dataset called: Group and Scene Emotions Dataset. It comprises 16,386 key frames randomly sampled from 128 publicly available YouTube video clips (totaling around 700 minutes) covering diverse group‐level scenarios, films, television programs, sports events, protests, concerts, queues, and more. To preserve clip‐level consistency, all frames from a given video were confined to a single split and then partitioned into training (11,471), validation (3,294), and test (1,621) sets. 10,806 frames contain clearly visible human faces, while 5,580 depict broader scene contexts without discernible faces. Each image was annotated by multiple volunteers as positive (1) or negative (0), defaulting to positive in the absence of explicit negative cues, and assigned a confidence score from 0 (highest uncertainty) to 5 (highest confidence). Example frames are shown in Figure~\ref{fig:groupscene}.

Promising performance accuracy on Group and Scene Emotion with 97.51\% on the validation set and 97.90\% on the test set. Despite these strengths, the model faces certain limitations. Its complexity requires considerable computing resources. In addition, performance falls short of the state-of-the-art on the Group Affect 2.0 database, and is 3.3\% lower than that of~\citet{ghosh2018automatic}  applying the same approach.

A graph-based deep learning framework, called Prototype Network Subgraph with Multi-Head Attention Framework (PSMF), has been presented by \citet{huangpsmf2025}. It aims to significantly improve emotion recognition at the group level. The approach addresses two major challenges: inadequate fusion of multimodal emotional features and low accuracy in recognizing infrequent emotional expressions. The proposed model relies on prototype networks combined with a graph-based reasoning approach. Specifically, it constructs a prototype network subgraph using support and query sets derived from multimodal signals (face, scene, and object features), enhanced by a multi-headed attention mechanism~\citep{Vaswani2017}. This enables disparate emotional signals to be effectively integrated and improves recognition of subtle and rare emotions by refining feature competition and interaction between different emotional categories. Experiments were conducted on widely recognized datasets, including AFEW  ~\citep{kossaifi2017afew}, GAF-2.0, GAF-3.0  ~\citep{dhall2018emotiw}, GroupEmoW  ~\citep{guo2020graph}  , and AffectNet  ~\citep{mollahosseini2017affectnet} . Performances were 62.43\%, 86.37\%, 83.58\%, 94.77\%, and 73.48\%, respectively, on the validation set. Despite its impressive results, the PSMF approach also has certain limitations. The inherent complexity of graph-based, attention-driven mechanisms introduces a considerable computational overhead. In the Table~\ref{tab:visual_group}, a summary of results for this section is given.

\begin{table}[H]
\centering
\caption{Summary of results from Visual Modality (Group Level).}
\label{tab:visual_group}
\resizebox{\textwidth}{!}{%
\begin{tabular}{|l|c|l|l|}
\hline
\textbf{Authors} & \textbf{Methodology} & \textbf{Dataset(s)} & \textbf{Performance/Key results} \\ \hline
\citet{gupta2018attention} & DenseNet 161 + SphereFace & GAF-2.0 & \makecell[l]{80.90\% (validation),\\ 64.83\% (test) accuracy} \\ \hline
\citet{Petrova20} & Synthetic augmentation, VGG19 & VGAF dataset & \makecell[l]{59.12\% accuracy,\\ 11.24\% gain over baseline} \\ \hline
\citet{gong2025hybrid} & Hybrid multimodal fusion & \makecell[l]{ GAF-É.0,\\ Group and Scene Emotions} & \makecell[l]{78.99\% accuracy,\\ 97.90\% accuracy} \\ \hline
\citet{huangpsmf2025} & \makecell[c]{Graph-based Prototype\\ Network Subgraph} & \makecell[l]{AFEW,\\ GAF-2.0,\\ GAF-3.0,\\ GroupEmoW,\\ AffectNet} & \makecell[l]{62.43\% accuracy,\\ 86.37\% accuracy,\\ 83.58\% accuracy,\\ 94.77\% accuracy,\\ 73.48\% accuracy} \\ \hline
\end{tabular}%
}
\end{table}

\subsection{Vocal and Audio Modality}  
The audio modality leverages speech and audio signals to infer emotional states, finding utility in telemedicine and human-computer interaction. Recent advancements in deep learning and self-supervised learning have enhanced the robustness of speech emotion recognition (SER) systems~\citep{sahu2018adversarial, zhao2025temporal, tang2025speech, kang2025speech}.

\subsubsection{Individual Level Audio Modality}  

\citet{davletcharova2015detection} demonstrates the importance of acoustic feature extraction (e.g., MFCC, peak to peak distances) in SER. Their work reveals that models trained on single-speaker data outperform multi-subject models, highlighting the need for personalized systems to account for individual variability. 
Hybrid models combining linguistic and emotional cues prove to be effective. The findings show joy showed the smallest peak distances, anger the next smallest, while sadness yielded the highest distances, indicating a correlation between emotional state and acoustic features. However, only three emotions (neutral, anger, joy) were used in classification, potentially limiting the model’s expressiveness for real-world applications.

 \citet{du2025speech} designed a Dual-Path Speech Emotion Recognition by introducing a biologically inspired framework combining Spiking Neural Networks (SNN)~\citep{zhang2021new}  and CNN to capture dynamic pulse features in speech signals, to capture more spectrogram cues. The approach used a novel Perceptual Neuron Encoding Layer (PNEL) to convert raw speech directly into spike trains (rather than first converting to images), preserving temporal–pulse information.  
 Weighted fusion of SNN/CNN is evaluated on IEMOCAP (Interactive Emotional Dyadic Motion Capture) dataset~\citep{Busso2008}. This dataset comprises approximately 12 hours of multimodal dyadic interaction recordings collected by USC’s Speech Analysis and Interpretation Laboratory (SAIL) from ten professional actors (five male–female pairs). Each pair performed three carefully selected emotional scripts and eight improvisational scenarios designed to elicit happiness, anger, sadness, frustration, and neutral states. The corpus is 
 segmented into 10,039 speaker turns (mean duration 4.5 s; mean 11.4 words), manually transcribed and forced-aligned to phonemes, and annotated both categorically (basic emotions plus frustration and excitement) and continuously (valence, activation, dominance) via multiple human raters. The weighted fusion of SNN/CNN achieved 65.3\% accuracy on the IEMOCAP dataset, surpassing single-path CNN (57.3\%) and SNN (42.2\%) and outperforming contemporary models. However, performance degrades under fixed-frequency noise, indicating room for improvement in noisy conditions. Another limitation is evaluation is only on IEMOCAP; cross-dataset and cross-language generalization is not tested.

\citet{radhika2025reliable} has proposed a combination of Tangent Flight (TF), Light Gradient Boosting Machine (LGBM)~\citep{nematzadeh2022tuning}, Exponential Pelican Optimization Algorithm (EPOA), called TEL (TF-EPOA + LGBM) framework classifier with Threshold-based Feature Selection (TFS).  It offers significant advantages for multi-regional SER, achieving maximum accuracy (99.27\% in Tamil, 98.42\% in Malayalam, 96.70\% in Indian English) by explicitly accounting for linguistic diversity and cultural nuances in emotional expression.  The Threshold-based Feature Selection (TFS) algorithm~\citep{subramanian2024effective}  reduces feature size by 40-50\% while increasing discriminative power (+7.48\% accuracy for MFBEE features ~\citep{subramanian2024effective}). However, limitations persist concerning scalability validation. The study covers only three Indian languages (excluding major languages such as Hindi), uses small cohorts of speakers (n=26), and relies on noise-free recordings.

\citet{sadok2023vector} proposed VQ-MAE-S framework that demonstrates significant advantages in self-supervised recognition of audio emotions (SER), leveraging vector-quantized discrete tokens from a pre-trained VQ-VAE to improve Masked Auto-Encoder (MAE)~\citep{he2022masked}  representations. This approach outperforms spectrogram-based MAEs in four benchmark datasets: RAVDESS-Speech, RAVDESS-Song, IEMOCAP, and  EMODB. 

The Ryerson Audio-Visual Database of Emotional Speech and Song (RAVDESS~\citep{livingstone2018ryerson}) is a richly annotated multimodal corpus comprising 7,356 recordings of vocal affect produced by 24 
well-balanced gender professional North American actors (12 male, 12 female) speaking and singing two neutral statements under eight emotion conditions (neutral, calm, happy, sad, angry, fearful, surprise, disgust) and two intensity levels (normal, strong) across three modalities (audio-video, video-only, audio-only). The Berlin Database of Emotional Speech (EMODB~\citep{burkhardt2005database} corpus consists of roughly 800 acted utterances produced by ten professional German speakers (5 female, 5 male), each rendering ten everyday German sentences (five short, five long) across seven discrete emotion categories (neutral, anger, fear, joy, sadness, disgust, and boredom). 

The maximum accuracy performances on these datasets are 84.1\% on RAVDESS-Speech, 85.8\% on RAVDESS-Song, 66.4\% on IEMOCAP  and 90.2\% on EMODB. Discrete tokenization, which avoids blurring of the spectrogram reconstruction, frame masking (80\% ratio), which outperforms patch-based strategies by up to 15.1\%, and the Query2Emo ~\citep{liu2021query2label}  fine-tuning module, which increases accuracy through cross-attention. However, with experiments conducted exclusively on English and German datasets, research on other datasets is still necessary. In the Table~\ref{tab:audio_individual}, a summary of results for this section is given.

\begin{table}[H]
\centering
\caption{Summary of results from Audio Modality (Individual Level).}
\label{tab:audio_individual}
\renewcommand{\arraystretch}{2} %
\resizebox{\textwidth}{!}{%
\begin{tabular}{|l|c|l|l|}
\hline
\textbf{Authors} & \textbf{Methodology} & \textbf{Dataset(s)} & \textbf{Performance/Key results} \\ \hline
\citet{davletcharova2015detection} & MFCC, peak distances & Custom dataset & \makecell[l]{ Personalized models\\ outperform multi-subject models} \\ \hline
\citet{du2025speech} & Dual-Path (SNN + CNN) & IEMOCAP & \makecell[l]{65.3\% accuracy,\\ outperformed single-path models} \\ \hline
\citet{radhika2025reliable} & \makecell[c]{TEL framework \\ (TF-EPOA + LGBM)} & \makecell[l]{Indian
regional\\ datasets} & \makecell[l]{99.27\%  accuracy (Tamil),\\ 98.42\% accuracy(Malayalam),\\96.70\% accuracy (English) } \\ \hline
\citet{sadok2023avector} & VQ-MAE-S framework &\makecell[l]{RAVDESS (speech, song),\\ IEMOCAP,\\ EMODB} & \makecell[l]{84.1\% ; 85.8\% accuracy,\\ 66.4\% accuracy, \\ 90.2\% accuracy } \\ \hline
\end{tabular}%
}
\end{table}

\subsubsection{ Group Level Audio Modality}  
Although significant progress has been made in the recognition of audio emotions at the individual level, little work has been done on the recognition of audio emotions at the group level. The complexity of overlapping voices and dynamic group interactions makes this a challenging but important area for further exploration. Most research focuses instead on emotion recognition in conversations or dialogues~\citep{zhang2017interaction, he2025dialoguemmt, alhussein2025speech}.

A Deep Spectrum framework that demonstrates significant advantages in group-level recognition of audio emotions, using the VGAF dataset, was presented in the EmotiW challenge 2020 by~\citet{ottl2020group}. Leveraging transfer learning from the pre-trained CNN ImageNet to extract robust audio representations from Mel spectrograms. Strategic fusion techniques were applied with DenseNet-121~\citep{huang2017densely}  and OpenSMILE, where a late fusion of all Deep Spectrum networks achieves a test accuracy of 62.70\%, an improvement of 14.82\% over the challenge baseline (47.88\%).

\citet{yeh2020dialogical} introduced a novel inference mechanism, Dialogical Emotion Decoder (DED), designed to enhance emotion recognition performance in multi-turn spoken dialog. Unlike conventional utterance-level speech emotion recognition (SER) systems that process each utterance independently, DED leverages the dialog structure by modeling temporal emotion flow and inter-speaker emotional influence. The proposed method decodes emotion over a conversation sequence by combining three components: a pre-trained SER classifier (IAAN~\citep{yeh2019interaction}), and an emotion shift model (which accounts for changes in a speaker's emotional state). DED treats dialog as a sequential prediction problem, integrating prior emotional context into current predictions. The approach is validated on two benchmark datasets: IEMOCAP and MELD~\citep{poria2018meld}. Results show that DED significantly improves performance over the base classifier IAAN, particularly on IEMOCAP, where DED with beam search decoding~\footnote{Beam search is a decoding algorithm used in speech recognition to efficiently find the most likely sequence of words from audio input. Speech recognition systems generate many possible word sequences, each assigned a probability by a neural network or statistical model.} achieves 70.1\% unweighted accuracy (UA), a 3.0\% absolute improvement. On MELD, a more challenging multi-party dataset with shorter dialogs and noisier conditions, DED achieves a modest 0.9\% improvement (40.3\% UA). Detailed analysis shows that modeling emotion shifts and integrating dialog-level context enhances classification, especially for ambiguous categories like "neutral." However, effectiveness on MELD was constrained by the lower performance of the base classifier and the brevity of dialog sequences, which limits the benefit of long-term context modeling. 

\citet{morgan2021classifying} present a comprehensive study on using deep learning for emotion recognition in natural group discourse. A novel dataset was collected from 14 group meetings involving 45 participants performing the Lunar Survival Task, during which participants’ speech was recorded and annotated both categorically and along three emotional dimensions: activation, dominance, and valence. The study compares the performance of four neural network architectures 2D CNN, 1D LSTM, CNN-LSTM, and a fusion network on two tasks: categorical emotion classification and 3D emotional regression. Transfer learning with the IEMOCAP dataset was applied to address the limited size of the collected dataset. Results in Table~\ref{tab:morgan_results} show that the CNN-LSTM network, especially when pretrained on IEMOCAP, performed better for both emotion classification and dimensional regression.

\begin{table}[H]
\centering
\caption{Concordance Correlation Coefficient (CCC) for arousal, dominance, valence, and Unweighted accuracy (UA,\%) and Weighted-by-category
(WA, \%) for fivefold cross-validation for each of the four neural network architectures on the lunar task dataset. Pretrained on the IEMOCAP database.}
\label{tab:morgan_results}
\begin{tabular}{|l|c|c|c|r|r|}
\hline
\textbf{Network}     &   \hspace{0.5cm} \textbf{Arousal}        & \hspace{0.8cm}\textbf{Dominance}      & \hspace{0.5cm} \textbf{Valence}        & \hspace{0.5cm} \textbf{UA}  & \hspace{0.5cm} \textbf{WA} \\
\hline
CNN          & \textbf{0.404} & 0.377          & 0.195 &34.10         & {32.20} \\
\hline
LSTM         & 0.332          & 0.248          & 0.056  &26.60  &{24.50}          \\
\hline
CNN‐LSTM     & 0.394          & \textbf{0.381} & \textbf{0.197}  &\textbf{34.20} &\textbf{32.60}         \\
\hline
Fusion       & 0.331          & 0.296          & 0.054 &31.20        & 29.40          \\
\hline
\end{tabular}
\end{table}

Pretraining significantly improved classification performance for the CNN and CNN-LSTM architectures, indicating the benefits of transfer learning from larger, more diverse emotion datasets. Emotion distributions were also analyzed for their predictive power in identifying emergent leaders and contributors within the groups. Notably, categorical emotion distributions correctly predicted 71\% of group leaders and 86\% of major contributors, revealing a strong relationship between emotional speech content and perceived group dynamics. The study highlights the limitations of using only 1D input for emotion recognition, the challenges of accurately modeling valence, and the importance of emotion balance in datasets. Despite the difficulty of recognizing nuanced emotions in natural speech, the study demonstrates that deep learning can reliably predict social roles such as leadership and contribution from speech emotion data.

\subsection{Textual Modality}

Textual emotion recognition infers a speaker’s affective state from written or transcribed text, with applications in sentiment analysis, social media monitoring, and affect‐aware interaction~\citep{yang2007building,batbaatar2019semantic,gu2025research,bacsal2025natural,babu2025enhancing}. Early approaches relied on lexicon‐spotting, syntactic rules, and knowledge bases to map words and phrases onto emotion categories.

\citet{ma2005emotion} presented one of the first hybrid systems: it used WordNet‐Affect for keyword detection, part‐of‐speech tagging, and negation handling to locate sentiment targets, and common‐sense inference from Open Mind Common Sense (OMCS) to infer one of Ekman’s six basic emotions~\citep{ekman1992there,singh2002open}. Integrated into a real‐time chat avatar platform, this client–server architecture demonstrated affect‐sensitive behavior but treated each utterance in isolation.

Multilingual extensions followed~\citet{jain2017extraction} compared monolingual classifiers, machine‐translation pipelines, and aligned multilingual lexicons across English, Italian, and Dutch. They showed that high‐quality Multilingual Text (MT) plus English‐trained models often outperformed native classifiers, though “disgust” and “surprise” remained challenging due to cultural nuances and lexical gaps~\citep{strapparava2004wordnet}.

The advent of large language models (LLMs) has revitalized text‐based emotion recognition. LLMs naturally model dialogue context and support multi‐step reasoning, enabling richer, personalized inference~\citep{brown2020language,santoso2024large,hong2025aer}. Most recently, \citet{li2025revise} proposed an R3 (Revise-Reason-Recognize) pipeline that embeds emotion‐specific prompts and ASR‐error correction directly into the inference prompt. On IEMOCAP, fine‐tuned LLaMA-2 13B~\citep{touvron2023llama} with R3 achieved 64.7 \% UA substantially above earlier baselines while also highlighting the sensitivity of LLMs to prompt design and transcription quality.

By combining traditional symbolic methods with the modern capabilities of LLM models, current systems are able to offer both linguistic accuracy and contextual depth in understanding emotions from text.

\section{Emotion Recognition in Multimodal Settings}

Emotion recognition has evolved from unimodal methods, which rely on a single data source (e.g., facial expressions, speech, or text), to multimodal approaches that integrate multiple modalities for improved accuracy and robustness~\citep{ramaswamy2024multimodal,kalateh2024systematic,hazmoune2024using,zhao2025review, chen2025enhancing}.
Multimodal emotion recognition leverages the complementary nature of different data sources, such as audio, visual, and textual cues, to better capture the complexity of human emotions. \citet{ezzameli2023emotion} provide a comprehensive review of this transition, highlighting key fusion techniques, deep learning advancements, and challenges in synchronization and computational complexity. Their work emphasizes the growing role of multimodal emotion recognition in applications like human-computer interaction, healthcare, and education.

In this section, we explore multimodal emotion recognition at both the individual and group levels, focusing on key methodologies such as feature representation for combined modalities and recent advancements.

\subsection{Multimodal Emotion Recognition at the Individual Level}
Multimodal emotion recognition at the individual level integrates multiple data sources from the individual, such as facial expressions, speech, and/or physiological signals, to infer emotional states. The multimodal approach addresses the limitations of unimodal methods such as incomplete emotion representation, by drawing on the complementary strengths of the different modalities.

\subsubsection{Feature Representation in Multimodal Learning}

\citet{ghaleb2019multimodal} demonstrate that audio-visual fusion significantly improves emotion recognition performance. Based on the McGurk effect  ~\citep{mcgurk1976hearing} , which illustrates how visual perception influences auditory processing, their work highlights the interdependence of speech and facial expressions in human communication. 
The appraoch is evaluated on RAVDESS and CREMA-D dataset. 

The Crowd-sourced Emotional Multimodal Actors Dataset (CREMA-D~\citep{cao2014crema}) consists of 7,442 short clips of 91 professional actors (48 males and 43 females; diverse ages and ethnicities) each uttering one of twelve semantically neutral sentences under six “basic” emotion prompts (happy, sad, anger, fear, disgust, neutral). Each clip was presented in three modalities (audio-only, visual-only, and audio-visual) and crowd-sourced to 2,443 raters via Survey Sampling International.

By modeling temporal dynamics and leveraging deep metric learning, their framework achieves state-of-the-art results on benchmark datasets such as CREMA-D~\citep{cao2014crema}  and RAVDESS~\citep{livingstone2018ryerson} . Central to their approach is temporal modeling, which deals with the evolution of emotions over time. Recognizing that positive and negative emotions are perceived at different speeds, the authors use long-term memory networks (LSTM) to capture emotional fluctuations incrementally. This approach contrasts with traditional late fusion methods, which assume the simultaneous expression of emotions in all modalities. 

Their final approach combines a Gating Paradigm~\citep{grosjean1996gating}  on all integration representations. On this basis, they introduce Deep Metric Learning (DML) to improve alignment between modalities. Rather than relying on conventional feature concatenation, deep metric learning maps audio and visual features into a shared latent space, minimizing divergence between modalities and maximizing mutual reinforcement. Triple loss optimizes this process by grouping emotionally similar instances while distancing dissimilar ones, thus improving recognition accuracy. The results of these strategies are presented in Table~\ref {tab:ghaleb}, showing the difference between unimodal and multimodal based on feature strategy representation. The combination of audio and video improves performance by around 9\% over the single modality for video only.

\begin{table}[H] 
  \centering
  \caption{Performance accuracy(\%) comparison for unimodal and multimodal on CREMA-D and RAVDESS of the work of ~\citep{ghaleb2019multimodal}.}
  \renewcommand{\arraystretch}{0.9} %

  \label{tab:ghaleb}
    \vspace{3mm}
  \begin{tabular}{|l|c|c|}
    \hline
    \textbf{Inputs} & \textbf{CREMA-D} & \textbf{RAVDESS} \\
    \hline
    Audio Only                          & 56.4 & 50.1 \\
    Audio Only (LSTM)                             & 50.2 & 40.1 \\
    Audio Only (Gating Paradigm)                 & 57.0 & 45.3 \\
    \hline
    Video Only                          & 63.1 & 60.2 \\
    Video Only (LSTM)                            & 66.8 & 60.5 \\
    Video Only (Gating Paradigm)                 & 65.0 & 60.1 \\
    \hline
    Audio-video                          & 69.0 & 65.7 \\
    Concatenation of Audio 
 (LSTM) and Video (LSTM) & 72.9 & 65.8 \\
    Audio-video  (Gating Paradigm)                 & \textbf{74.0} & \textbf{67.7} \\
    \hline
  \end{tabular}
\end{table}

As mentioned in the above lines, representation learning in terms of feature representation and the combination manner for all modalities are crucial. \citet{nemati2019hybrid} propose a hybrid latent space fusion method for multimodal emotion recognition, combining feature-level fusion for audiovisual data with decision-level fusion for textual integration. Their framework employs techniques like  Canonical Correlation Analysis (CCA)~\citep{correa2010canonical}, Cross-Modal Factor Analysis (CFA)~\citep{li2003multimedia}, and  Marginal Fisher Analysis (MFA)~\citep{sharma2012generalized} for latent space alignment, alongside Dempster-Shafer (DS) theory~\citep{dempster1968generalization, basiri2014sentiment} for uncertainty-aware decision fusion.  
For latent space fusion, the authors address the limitations of traditional feature concatenation, such as dimensionality explosion and redundancy by projecting audio and visual features into a shared latent space. Unlike unsupervised methods like CCA and CFA, their supervised MFA approach maximizes intra-class compactness and inter-class separability, significantly improving classification accuracy.  
Building on this, they implement decision-level fusion to incorporate textual data. Using DS theory, the framework aggregates evidence from multiple modalities while reducing uncertainty, enhancing robustness. This hybrid fusion strategy achieves state-of-the-art performance on the extended DEAP dataset  ~\citep{nemati2016incorporating, nemati2017exploiting}, demonstrating the effectiveness of combining latent space alignment with evidential reasoning. The DEAP (Database for Emotion Analysis using Physiological signals) dataset is a multimodal corpus for analysing human affective states via synchronized video signals, peripheral physiological cues, and electroencephalogram recordings, accompanied by self-reported arousal and valence ratings on nine-point scales provided by 14–16 volunteers per clip. Results displayed in Table~\ref{tab:nemati} multimodal fusion with support vector machine (SVM) classifier show a gain of 11\%, 12\%, 17\% of accuracy, for AVT-MFA compared to audio (A), video (V), and text (T) alone, respectively.

\begin{table}[H]
  \centering
  \caption{Comparison of the performance of using audio (A), video (V), textual modality (T), feature‐level fusion of audio and visual modalities (AV) with their decision‐level fusion (AVT) using CCA, CFA, and MFA fusion methods of ~\citep{nemati2019hybrid}.}
  \renewcommand{\arraystretch}{0.75} %
  \label{tab:nemati}
  \vspace{3mm}
  \begin{tabular}{|c|l|l|c|c|c|c|}
    \hline
    \textbf{Modality} & \textbf{Classifier} & \textbf{Precision} & \textbf{Recall} & \textbf{F1‐measure}  & \textbf{Accuracy (\%)} \\
    \hline
    A      & Naive Bayes & 0.48 & 0.44 & 0.46  & 0.72 \\
               & SVM         & 0.65 & 0.65 & 0.65  & 0.82 \\
    V      & Naive Bayes & 0.76 & 0.69 & 0.72  & 0.85 \\
               & SVM         & 0.62 & 0.65 & 0.63  & 0.81 \\
    T          & Naive Bayes & 0.79 & 0.53 & 0.64  & 0.70 \\
               & SVM         & 0.82 & 0.61 & 0.70  & 0.76 \\
    \hline
    AV‐CCA     & Naive Bayes & 0.82 & 0.67 & 0.74  & 0.84 \\
               & SVM         & 0.66 & 0.63 & 0.64  & 0.81 \\
    AVT‐CCA    & Naive Bayes & 0.84 & 0.77 & 0.80  & 0.88 \\
               & SVM         & 0.84 & 0.81 & 0.82  & 0.91 \\
    \hline
    AV‐CFA     & Naive Bayes & 0.75 & 0.63 & 0.68  & 0.82 \\
               & SVM         & 0.76 & 0.70 & 0.73  & 0.85 \\
    AVT‐CFA    & Naive Bayes & 0.74 & 0.66 & 0.70  & 0.84 \\
               & SVM         & 0.85 & 0.80 & 0.83  & 0.91 \\
    \hline
    AV‐MFA     & Naive Bayes & 0.83 & 0.79 & 0.81  & 0.89 \\
               & SVM         & 0.85 & 0.84 & 0.84  & 0.92 \\
    AVT‐MFA    & Naive Bayes & 0.78 & 0.77 & 0.78  & 0.88 \\
               & SVM         & \textbf{0.88 }&\textbf{ 0.86} &\textbf{ 0.87}  &\textbf{ 0.93} \\
    \hline
  \end{tabular}
\end{table}

\citet{hsu2023applying} introduce a segment-level approach to multimodal emotion recognition, addressing challenges such as modality inconsistency and cross-modal misalignment. Their framework combines a  Segment Level Attention (SLA)  mechanism with a  Bi-Modal Transformer Encoder (BMT)  to enhance temporal consistency and alignment across modalities.
Central to this framework, the  Segment Level Attention (SLA)  mechanism prioritizes segments where audio and visual emotional cues align, assigning them higher weights to improve reliability. By focusing on regions of high emotional agreement, SLA mitigates the risks of modality inconsistency, where conflicting signals from different modalities could degrade recognition accuracy. 
Building on the SLA, the  Bi-Modal Transformer Encoder (BMT)  dynamically models relationships between modalities through self-attention and cross-modal interaction. Unlike concatenation or decision-level fusion, the BMT captures temporal dependencies, ensuring emotions evolve coherently across modalities over time. This approach is evaluated on BAUM-1~\citep{zhalehpour2016baum}  and CMU-MOSEI~\citep{zadeh2018multimodal}  datasets, demonstrating its effectiveness in handling real-world emotional dynamics. 

The BAUM-1 corpus~\citep{zhalehpour2016baum}  comprises both acted and spontaneous audio-visual recordings from 31 native Turkish speakers (17 female; age 19-65). The acted subset (BAUM-1a) contains 273 clips in which subjects deliver scripted utterances to portray eight target states, six basic emotions (happiness, anger, sadness, disgust, fear, surprise) plus boredom, contempt, and confusion. The spontaneous subset (BAUM-1s) includes 1,184 segments elicited by 29 carefully selected images/videos (e.g. illusions, family conflict, horror scenes) that prompt unscripted verbal responses; these cover six basic emotions and seven non-basic mental states (boredom, contempt, unsure, thinking, concentrating, bothered, neutral) and were annotated for dominant state and intensity (0–5). The CMU Multimodal Opinion Sentiment and Emotion Intensity dataset (CMU-MOSEI~\citep{zadeh2018multimodal}) comprises 23,453 video segments drawn from 3,228 high-quality monologue videos by 1,000 distinct YouTube speakers across 250 topics, multimodal data (language, vision, and audio). Gender balanced (57\% male, 43\% female), and videos were vetted by experts for visual and audio quality, resulting in a final set of 3,228 monologue clips. Annoted with the four-point Ekman emotion scale (neutral, happiness, sadness, anger, fear, disgust, surprise).

For feature extraction, it uses pretrained wav2vec 2.0 for audio and VGGNet  ~\citep{simonyan2014very}  for visual inputs, fine-tuned on the re-annotated BAUM-1 dataset. The training strategy involves combining losses from segment and signal levels, enabling the model to learn both local and global emotional representations. Experiments on BAUM-1 and CMU-MOSEI datasets show the proposed model outperforms state-of-the-art methods, achieving 74.31\% accuracy on BAUM-1 and 76.81\% on CMU-MOSEI improvements of 3.05\% and 2.57\% over previous bests, respectively. However, the method is limited by its high annotation overhead for segment-level labeling and potential data imbalance issues, especially in negative emotions.

\citet{sadok2024multimodal} introduce the  Multimodal Dynamical Variational Autoencoder (MDVAE), an unsupervised generative model designed to learn structured latent representations from audiovisual speech data. MDVAE disentangles static information (e.g., speaker identity) from dynamic information (e.g., phonemes, lip movements) while separating modality-common features from modality-specific ones.
Central to MDVAE’s design is its hierarchical latent space, which organizes latent variables into three categories: static audiovisual (e.g., speaker identity), dynamic audiovisual (e.g, phoneme articulation), and modality-specific (e.g., acoustic or visual nuances). This hierarchical structure enables better disentanglement of emotional and speech-related features, offering a flexible and interpretable framework for analyzing multimodal emotional speech.
MDVAE learns disentangled embeddings by encouraging modality-specific encoders to capture private variations while enforcing shared latent space to encode emotion-relevant features common across modalities. Additionally, a cross-modal reconstruction loss ensures that the shared latent representation is robust and informative enough to reconstruct input from any modality. Using audio and visual modalities, the approach is trained and evaluated on the CMU-MOSEI dataset  ~\citep{zhalehpour2016baum}. Experimental results show that MDVAE outperforms several strong baselines, including multimodal Transformer and tensor fusion networks, achieving state-of-the-art performance in binary sentiment classification (average F1 of 81.3\%) and 7-class emotion recognition tasks.
The main advantages of the MDVAE framework include improved interpretability of latent features, robustness to missing modalities (due to its cross-modal capabilities), and better generalization through disentangled learning. However, limitations involve increased model complexity and missing experiments on in-the-wild datasets.

\subsubsection{Reasoning in Multimodal Learning}

Recent emotion recognition systems have begun to incorporate true multimodal reasoning, jointly interpreting audio, video, and text to handle real-world (‘in-the-wild’) signals.  This shift has been driven by the emergence of large language models (LLMs) that excel at long-range contextual reasoning in pure text~\citep{brown2020language}.  By extending these architectures to accept multiple input streams what we call multimodal LLMs (MLLMs). It is possible to perform cross-modal inference: for example, correlating a speaker’s tonal cues with the semantic content of their words and facial expressions~\citep{touvron2023llama}.
Initial MLLM-based emotion studies have shown improved robustness under noisy, real-world conditions, thanks to LLMs’ ability to dynamically weight and reason over disparate cues.  In the next section, we’ll survey key MLLM architectures and how they leverage reasoning primitives to advance in-the-wild affective computing.

\citet{cheng2024emotion}  introduce \textbf{ Emotion-LLaMA}, an advanced multimodal model designed to accurately recognize and reason about human emotions using audio, visual, and textual inputs. Recognizing the limitations of traditional single-modality approaches and existing Multimodal Large Language Models (MLLM)~\citep{alayrac2022flamingo, bai2023qwen, chen2023shikra, chiang2023vicuna, peng2023kosmos, wang2023visionllm}. \textbf{Emotion-LLaMA} integrates specialized encoders for audio (HuBERT~\citep{hsu2021hubert}) and visual data (MAE~\citep{he2022masked}, VideoMAE~\citep{tong2022videomae}, EVA~\citep{fang2023eva}) alongside text into a unified representation space using a modified LLaMA architecture~\citep{touvron2023llama}  enhanced by instruction tuning. The authors also present the Multimodal Emotion Recognition and Reasoning (MERR) dataset, which contains 28,618 coarse-grained and 4,487 fine-grained annotated samples covering a wide range of emotions, including complex and subtle expressions. The dataset helps the model generalize across various scenarios and emotional contexts. \textbf{Emotion-LLaMA} significantly outperforms other models in extensive evaluations, achieving state-of-the-art results with an F1 score of 90.36\% on the MER-SEMI 2023 dataset  ~\citep{lian2023mer} , and top performance in zero-shot evaluations on the DFEW dataset  ~\citep{jiang2020dfew}  (UAR: 45.59, WAR: 59.37). These results underscore the model's effectiveness in both recognizing subtle emotional cues and explaining emotional reasoning. Despite its notable performance,\textbf{ \textbf{ Emotion-LLaMA}} faces challenges, such as computational complexity due to multimodal fusion and reliance on high-quality annotation data.

\citet{yang2025omni} addresses critical limitations in current multimodal large language models (MLLM) for emotion analysis with \textbf{Omni-Emotion}. Notably, their difficulty to capturing subtle facial expressions and effectively integrating audio cues. To support the training and evaluation of \textbf{Omni-Emotion}, two novel datasets were constructed: a self-reviewed emotion (SRE) dataset containing 24,137 samples automatically annotated with high alignment scores, and a smaller, manually verified human-reviewed emotion (HRE) dataset with 3,500 carefully annotated samples. These datasets are meticulously curated through advanced multimodal pipelines that extract detailed facial features, audio cues, and textual contexts, which are subsequently validated for consistency using GPT-3.5~\citep{quintans2023chatgpt}. The arcbitecure is evaluated on DFEW and MAFW. 
The MAFW dataset~\citep{liu2022mafw} comprises 10,045 video–audio clips “in the wild” sourced from over 1,600 movies and TV dramas as well as 20,000 short videos from reality shows, talk shows, news, variety programs, and more, covering a broad range of themes and cultures. Each clip was independently annotated by eleven trained raters using an 11-dimensional confidence scoring scheme to assign one or more of eleven emotions: anger, disgust, fear, happiness, neutral, sadness, surprise, contempt, anxiety, helplessness, and disappointment.  Every clip is paired with bilingual (English–Chinese) descriptive captions that detail environmental context, body movements, facial action units, and other affective behaviors, enabling tasks such as emotion captioning. MAFW is inherently multi-modal, providing aligned video frames, audio tracks, and text, and includes automatic annotations of 68 frame-level facial landmarks, precise face regions, and gender estimates (58.1 \% male, 41.9 \% female).

Experimental results demonstrate \textbf{Omni-Emotion} achieved the state-of-the-art on these benchmarks with a remarkable Unweighted Average Recall (UAR) of 68.80\% on DFEW  ~\citep{jiang2020dfew}  and 53.81\% on MAFW  ~\citep{liu2022mafw}. Moreover, \textbf{Omni-Emotion} significantly improves open-vocabulary emotion recognition and emotion reasoning tasks, surpassing previous methods such as \textbf{\textbf{ Emotion-LLaMA}} and AffectGPT  ~\citep{lian2024affectgpt}  by considerable margins. %
However, \textbf{Omni-Emotion} faces certain limitations, including challenges related to mixed emotional states, substantial computational requirements, and reliance on high-quality training data, which might limit deployment in real-time or resource-constrained scenarios. Overall, \textbf{Omni-Emotion} represents a substantial advancement in multimodal emotion analysis, providing a foundation for more nuanced and contextually aware emotion recognition systems across diverse real-world applications. In the Table~\ref{tab:multimodal_individual}, a summary of results for this section is given.

\begin{table}[H]
\centering
\caption{Summary of results from Multimodal Settings (Individual Level).}
\label{tab:multimodal_individual}
\renewcommand{\arraystretch}{1.1} %
\resizebox{\textwidth}{!}{%
\begin{tabular}{|l|c|l|l|}
\hline
\textbf{Authors} & \textbf{Methodology} & \textbf{Dataset(s)} & \textbf{Performance/Key results} \\ \hline
\citet{ghaleb2019multimodal} & Audio-visual fusion & \makecell[l]{ CREMA-D,\\ RAVDESS } & \makecell[l]{74.0\% accuracy,\\ 67.7\% accuracy } \\ \hline
\citet{nemati2016incorporating} & \makecell[c]{Hybrid latent space fusion\\ (MFA, DS theory)} & Extended DEAP & \makecell[l]{Up to 93\% accuracy\\ with multimodal fusion} \\ \hline
\citet{hsu2023applying} & \makecell[c]{Segment Level Attention \\ + Bi-Modal Transformer} &\makecell[l]{BAUM-1,\\ CMU-MOSEI} & \makecell[l]{74.31\% accuracy,\\ 76.81\% accuracy} \\ \hline
\citet{sadok2024multimodal} & Multimodal Dynamical VAE & CMU-MOSEI & \makecell[l]{ 81.3\% F1 score\\ in binary sentiment} \\ \hline
\citet{cheng2024emotion} & Emotion-LLaMA (multimodal LLM) & \makecell[l]{ MER-SEMI,\\ DFEW} & \makecell[l]{90.36\% F1 score\\ 59.37\% WAR, 45.59\% UAR}\\ \hline

\citet{yang2025omni} & Omni-Emotion (multimodal LLM) & \makecell[l]{ MAFW,\\ DFEW} & \makecell[l]{53.81\% UAR\\ 68.80\% UAR}\\ \hline
\end{tabular}%
}
\end{table}

\subsection{Multimodal Emotion Recognition at the Group Level}

Building on advances in individual-level multimodal emotion recognition, researchers have begun exploring how to extend these approaches to groups, where multiple individuals’ emotions and contextual information must be integrated to infer collective affective states. \citet{liu2020group} propose a hybrid network for group-level emotion recognition, integrating facial emotion analysis, environmental context, and temporal dynamics. 
Central to their framework, the  Facial Emotion Stream employs DenseNet169~\citep{huang2017densely}  to classify emotions at the group level. This stream aggregates multiple facial features, capturing dominant, weak, and variable emotional representations to infer collective emotional states.
To incorporate environmental context, the  Environmental Object Statistics (EOS) Stream detects contextual objects (e.g., birthday cakes, weapons) using YOLOv3~\citep{yolov3}. The EOS stream enhances prediction accuracy by analyzing how these objects correlate with emotional scenarios (e.g., joy at celebrations or tension in conflict settings).
Building on these spatial features, the  Temporal Modeling component combines a Temporal Shift Module (TSM)~\citep{lin2019tsm}  with LSTM networks to capture dynamic emotion patterns in video sequences. This dual approach addresses temporal variability, ensuring the model adapts to evolving group emotions. Together, these streams demonstrate the critical role of multimodal integration in decoding complex group-level affective states.  

Using the VGAF dataset~\citep{dhall2020emotiw}  as the primary benchmark, the authors show that their fusion of multimodal data significantly outperforms baseline models, achieving a test set accuracy of 76.85\%, which is 26.89\% higher than the official baseline. This result was achieved by combining two facial streams, audio, EOS, optical flow, the fighting detector, and global features using a linear SVM. Among individual streams, facial emotion and video features performed best (around 64\% each), while others like object statistics and body pose were slightly weaker but still additive in the final fusion. Additionally, the authors generated auxiliary data and designed novel classifiers, such as the fighting detector, which enhanced recognition of negative emotions prevalent in aggressive scenes. While the approach provides a good performance, the approach also depends heavily on accurate detection (e.g., face, objects) and requires multiple pretrained models, which increases system complexity. %

\citet{Wang2020} introduce the  K-injection audiovisual network, which enhances group-level emotion recognition by integrating explicit audiovisual features with implicit knowledge representations. Their model combines cross-attention fusion and knowledge injection to improve accuracy on the VGAF dataset  ~\citep{dhall2020emotiw}, bridging the gap between data-driven and cognitively inspired emotion analysis.
Central to this framework,  cross-attention fusion dynamically aligns video and audio representations using multi-head attention~\citep{Vaswani2017}. By emphasizing emotion-relevant features (e.g. ,synchronized facial expressions and audio tone shifts), this mechanism ensures coherent multimodal alignment, mimicking how humans naturally associate visual and auditory cues during emotion perception.
Complementing this,  knowledge injection incorporates implicit contextual information from video descriptions (e.g., scene semantics). Experiments on the VGAF dataset show that the proposed model significantly outperforms a strongly the baseline (66.40\% test accuracy vs. 47.88\%), with the best performance coming from the integration of both linguistic and acoustic K-injection subnetworks. The inclusion of implicit knowledge helped improve recognition of complex group dynamics and emotions, especially in social or ambiguous contexts.
The approach mirrors human cognitive processes, where prior knowledge and environmental context inform emotional interpretation. 

Another work, on VGAF dataset,  that recently outperforms the state-of-the-art is presented by~\citep{kumar2025fusing}. The authors propose a multimodal fusion integrating multiple data streams, audio, video, pose, and frame-level features into a unified framework. Advanced feature extraction methods like TimeSformer~\citep{bertasius2021space}  for spatiotemporal video data, wav2vec2.0~\citep{baevski2020wav2vec}  for audio embeddings, and YOLOv8~\citep{hussain2023yolo}  for pose estimation significantly enhance the emotional inference capabilities of the model. Experiments conducted on the VGAF dataset demonstrate that the proposed multimodal approach significantly outperforms traditional unimodal and less comprehensive multimodal methods, achieving a notable classification accuracy of 81.98\%  on the validation set. In addition, analyses highlight modality-specific strengths, for example, audio features excel at detecting neutral emotions due to their contextual depth, while video features captured by TimeSformer effectively identify positive emotional states. Pose and frame-level features particularly aid in distinguishing negative emotions by capturing nuanced body movements and spatial-temporal changes. However, the model is computationally demanding due to multiple modality streams and requires meticulous parameter tuning.

\citet{wang2022congnn} introduced ConGNN (Context-consistent Cross-Graph Neural Network for Group Emotion Recognition in the Wild). A robust framework for recognizing group-level emotions in complex real-world scenarios. Unlike traditional facial-expression-based methods that overlook scene and contextual cues, ConGNN models both intra and inter-relations among multiple emotion cues, facial expressions, local objects, and global scenes by constructing a cross-graph neural network. The architecture is composed of three components: Multi-branch Feature Extractors (MFE) to capture distinct emotional features, a Cross-Graph Neural Network (C-GNN) that constructs graphs representing emotional relations within and across modalities, and an Emotion Context Consistent Learning (ECL) mechanism, which introduces a Bias Penalty Function (BPF) to align inconsistent emotional cues across branches. ConGNN is evaluated on two group emotion datasets: GroupEmoW~\citep{guo2020graph}  and the newly proposed SiteGroEmo, which contains 10,034 crowd images from diverse global contexts. The model achieves 85.59\% accuracy on GroupEmoW and 83.57\% on SiteGroEmo, outperforming state-of-the-art models such as CAER-Net~\citep{lee2019context}  and GNN-based methods by margins of 3.35\% and 4.32\%, respectively. Ablation studies confirm that each component MFE, C-GNN, and ECL contributes to performance gains, with the full model showing the best results. The ECL module particularly proves effective in reducing emotion bias, improving model robustness when facial, object, and scene cues exhibit conflicting sentiments. Furthermore, cross-database validation demonstrates the generalizability of ConGNN, achieving 76.24\% accuracy when trained on GroupEmoW and tested on SiteGroEmo without fine-tuning. Although ConGNN introduces some computational overhead due to its complex graph and multitask training structure.

\citet{lee2024group} propose a novel, interpretable approach to group emotion recognition (GER) by leveraging psychological theories and cinematographic composition rules through a fuzzy logic framework. Unlike traditional GER models that average individual facial expressions, this method acknowledges that humans intuitively focus on faces based on their size and centrality in an image. The model incorporates 89 fuzzy rules derived from expert intuition and cinematographic principles, assessing group emotion based on face size, position, and individual emotion rates to determine a dominant group emotion from the seven basic emotions categories.

The fuzzy system architecture includes three main components: fuzzification, fuzzy rule evaluation, and defuzzification. These components translate ambiguous, non-numerical cues into linguistic variables and inference scores. The model assigns higher emotional weight to faces that are larger and more centrally located in the image, following psychological principles about visual attention and perception. Additional “event rules” are introduced to resolve conflicting emotional cues (e.g., when both happy and sad expressions are present), allowing for refined reasoning in mixed-emotion scenarios. Experiments on a curated subset of the EMOTIC dataset~\citep{kosti2017emotion}  (augmented with web images) demonstrate that this fuzzy-rule-based method improves GER accuracy by approximately 14\% compared to the simple averaging of individual emotions. It also performs better across valence classes, especially in detecting dominant emotional trends in complex group settings. The system showed that face size and emotion rate were more influential than face position, and it provided more human-like, interpretable results than black-box deep learning alternatives. While promising in performance and interpretability, the method has limitations, including dependence on accurate facial detection, lack of standardized group emotion datasets, and inability to account for sarcasm or implicit emotional states. In the Table~\ref{tab:multimodal_group}, a summary of results for this section is given.

\begin{table}[H]
\centering
\caption{Summary of results from Multimodal Emotion Recognition at the Group Level.}
\label{tab:multimodal_group}
\resizebox{\textwidth}{!}{%
\begin{tabular}{|l|c|l|l|}
\hline
\textbf{Authors} & \textbf{Methodology} & \textbf{Dataset(s)} & \textbf{Performance/Key results} \\ \hline
\citet{liu2020group} & Hybrid (Facial Emotion, EOS, TSM, LSTM) & VGAF & \makecell[l]{76.85\% accuracy,\\ 26.89\% improvement over baseline} \\ \hline
\citet{wang2020region} & K-injection audiovisual network & VGAF & 66.40\% accuracy vs. 47.88\% baseline \\ \hline
\citet{kumar2020object} & \makecell[c]{Multimodal fusion \\ (TimeSformer, wav2vec2.0, YOLOv8)} & VGAF & 81.98\% accuracy on validation set \\ \hline
\citet{wang2022congnn} & ConGNN (cross-graph neural network) & \makecell[l]{ GroupEmoW,\\ SiteGroEmo} &  \makecell[l]{85.59\% accuracy,\\ 83.57\% accuracy} \\ \hline
\citet{lee2024group} & \makecell[c]{Fuzzy logic\\ (psychological theories and cinematographic rules)} & EMOTIC subset & \makecell[l]{Improved accuracy by approximately\\ 14\% over simple averaging} \\ \hline
\end{tabular}%
}
\end{table}

\section{Privacy in Emotion Recognition}

Emotion recognition plays a crucial role in various domains such as healthcare, education, marketing, customer service, security, and autonomous robotics perception. However, it also raises significant concerns regarding personal privacy and individual safety ~\citep{oh2016faceless}. The collection and analysis of individuals’ emotional data without explicit consent can lead to unauthorized data usage and misuse, posing serious ethical questions about personal dignity and freedom ~\citep{ortiz2023implications, van2025emotion}. 

When referring to the European Community (EU), the European Union's Artificial Intelligence Act (AI Act~\citep{cancela2024eu}) introduces comprehensive regulation for emotion recognition systems, with a particular emphasis on safeguarding personal data and fundamental rights. Adopting a risk-based framework, the AI Act classifies emotion recognition technologies as high-risk or prohibited applications, depending on their context and potential impact. Specifically, the Act bans AI-based emotion recognition used for biometric mass surveillance, predictive policing, or any scenario likely to infringe privacy and personal autonomy, except for narrowly defined exceptions related to public safety, such as locating missing persons or preventing imminent threats. Emotion recognition systems that are permitted must strictly adhere to the EU's General Data Protection Regulation (GDPR), ensuring transparency, explicit consent, data minimization, and purpose limitation. Systems deploying emotion recognition are further required to ensure human oversight to avoid purely automated decisions that could negatively impact individual rights and freedoms. Over recent years, many researchers have focused on addressing these challenges by developing methods that preserve individual privacy while maintaining the effectiveness of emotion recognition systems.

\subsection{Privacy Approaches in Visual Emotion Recognition}

This section reviews recent advances in privacy-preserving visual emotion recognition, focusing on techniques that protect user identity while maintaining high recognition accuracy.

\citet{zitouni2022privacy} tackle the challenge of affective state recognition by emphasizing non-facial visual cues such as upper body movements and background context. These cues are less intrusive than facial expressions yet retain rich emotional information. Using the K-EmoCon dataset ~\citep{park2020k}, they apply a face-masking technique to conceal identity, combined with a CNN-LSTM model to classify affective states based on arousal and valence dimensions, including quadrant classifications. The model processes video segments of naturalistic conversations and evaluates emotion recognition from both self and partner annotated perspectives. Results show that this privacy-preserving approach achieves performance comparable to raw, unmasked video data, with recognition accuracies of up to 96.82\%, 95.91\%, and 91.52\% for arousal, valence, and quadrant classifications, respectively. Partner annotations, which better reflect visual cues, yield higher accuracy than self-annotations. The authors also identify an optimal temporal window of 25–35 seconds to effectively capture affective state changes. Overall, the study demonstrates that body gestures and contextual information provide sufficient cues for emotion recognition without revealing identity, making this system suitable for privacy-conscious human-machine interaction applications. Despite these promising results, limitations remain. The approach experiences reduced performance for participants with high movement variability and may be affected by cultural biases in body language interpretation. Furthermore, it relies on controlled recording conditions, potentially requiring adaptation for more diverse or unconstrained environments. Nevertheless, this work represents a significant step toward privacy-aware affective computing by showing accurate emotion inference without facial identity disclosure.

Building on visual obfuscation techniques, \citet{pentyala2021privacy} propose a cryptographic approach that enhances privacy through Secure Multi-Party Computation (MPC) protocols. Their framework enables clients to classify video content, such as human emotion recognition, without exposing the video data or the classification model parameters to external parties. This is especially critical in sensitive applications like surveillance, healthcare, and empathy-based AI systems. The authors develop novel MPC protocols for oblivious frame selection and secure label aggregation, facilitating an end-to-end encrypted classification pipeline using convolutional neural networks (ConvNets~\citep{simonyan2014two}). Their method aggregates predictions from individual video frames to classify the entire video while ensuring all computations occur on secret-shared data, effectively preventing data leakage. Evaluated on the RAVDESS dataset~\citep{Busso2008}, the framework achieves classification accuracy comparable to state-of-the-art non-private models (around 56.8\%), while maintaining strict privacy guarantees. It operates efficiently on multi-party cloud setups with honest-majority adversary models, processing 3-5 second video clips within approximately 9 seconds on Azure F32 machines. This demonstrates practical feasibility despite the computational overhead inherent to MPC. Current limitations include higher runtimes than non-private models, which impede real-time deployment, and the use of 2D ConvNets without explicit temporal modeling, through extensions to spatiotemporal architectures are possible.

Together, these studies exemplify two complementary privacy approaches in visual emotion recognition: identity obfuscation through data masking and encryption-based secure computation. Both approaches demonstrate promising trade-offs between privacy preservation and recognition performance, highlighting active research directions in privacy-aware affective computing.

\subsection{Privacy Approaches in Audio Emotion Recognition}

Privacy concerns in audio-based emotion recognition primarily focus on anonymizing speaker identity while preserving both linguistic content and emotional states. Recent research has advanced this area through challenges and novel anonymization architectures.

\citet{chen2022system} introduced the Voice Privacy Challenge, which aims to foster the development of voice anonymization systems that conceal speaker identity without distorting the original speech or emotional cues. The challenge emphasizes utterance-level anonymization, requiring pseudo-speaker voices that vary across utterances while maintaining linguistic and emotional fidelity. Baseline results demonstrated varied trade-offs between privacy protection and utility preservation, with some submissions achieving strong anonymization but sometimes at the expense of linguistic or emotional accuracy.

Building on these foundations, \cite{yao2024npu} present the NPU-NTU speaker anonymization system, which improves identity concealment while preserving linguistic and paralinguistic information, including emotional state. The system employs a disentangled neural codec architecture with a serial disentanglement strategy that progressively separates global, time-invariant speaker identity from time-variant linguistic and paralinguistic features. Key innovations include multiple distillation methods: semantic distillation for linguistic content, supervised speaker distillation for identity, and frame-level emotion distillation to preserve emotional cues, ensuring effective disentanglement and enhanced emotion preservation. During anonymization, speaker identity is replaced by a weighted average of candidate speaker embeddings combined with a randomly generated speaker identity, enabling flexible pseudo-speaker synthesis. Experiments on the VoicePrivacy 2024 datasets LibriTTS ~\citep{zen2019libritts}, LibriSpeech ~\citep{panayotov2015librispeech}; IEMOCAP~\citep{Busso2008} show that this approach achieves strong privacy protection, measured by equal error rate (EER), while maintaining high utility, reflected in low word error rates (WER) and high unweighted average recall (UAR) for emotion recognition. This system outperforms previous baselines, offering an excellent balance between privacy and utility. However, limitations include the complexity of the disentanglement training process and reliance on accurate modeling of speaker and emotional features.

Another recent advancement is the lightweight Emotion-Preserving Prosody Anonymization (EPPA) network proposed by~\citet{he2025emotion}. EPPA extracts speaker-independent prosodic features and converts them into the style of a pseudo-speaker to anonymize speech while retaining emotional cues. Integrated with FACodec~\citep{ju2024naturalspeech}, a neural codec for timbre cloning, this dual anonymization framework synthesizes both timbre and prosody, achieving comprehensive anonymization without sacrificing emotional or linguistic fidelity. The EPPA framework utilizes a conditional variational autoencoder architecture and incorporates multiple components: a prosody encoder and decoder, a speaker prompter, a prosody discriminator, and a gradient reversal layer-based speaker classifier. These components jointly disentangle prosody from speaker identity while preserving emotional expression. A novel pseudo-speaker selection strategy, Closest Center Distance (CCD), improves naturalness by choosing pseudo-speakers whose timbre embeddings are closest to the dataset center. Evaluations on the VoicePrivacy Challenge 2024 datasets LibriTTS~\citep{zen2019libritts}, LibriSpeech~\citep{panayotov2015librispeech}; IEMOCAP~\citep{Busso2008} demonstrate that FACodec+EPPA achieves state-of-the-art performance across key metrics: unweighted average recall (UAR) for emotion preservation, word error rate (WER) for content clarity, and equal error rate (EER) for privacy protection. The model ranks among the top solutions, effectively balancing privacy and utility. Ablation studies confirm that both EPPA and the CCD strategy significantly contribute to improvements in emotion preservation and anonymization strength. Nonetheless, the pretrained FACodec and the complexity of the dual anonymization framework present integration challenges. Although EPPA converges quickly on smaller datasets, combining it with FACodec requires careful coordination.

These works illustrate the progression from foundational voice anonymization challenges to sophisticated disentanglement and prosody-preserving frameworks, reflecting the ongoing effort to safeguard privacy in audio emotion recognition without compromising data utility.

\subsection{Privacy-Preserving Approaches in Multimodal Settings}

Recent research has increasingly focused on privacy preservation in multimodal emotion recognition, which involves combining information from multiple modalities such as audio, video, and text. This section reviews key approaches that address the unique privacy challenges posed by multimodal data fusion.

\citet{jaiswal2020privacy} investigate how demographic information, particularly gender, can unintentionally leak through learned representations in multimodal emotion recognition models. Their analysis shows that textual, acoustic, and combined multimodal inputs may reveal sensitive demographic details, compromising user privacy. To address this, the authors propose an adversarial learning framework that trains models to be invariant to gender by effectively \textbf{unlearning} demographic features from latent representations. They rigorously evaluate this method across multiple datasets including IEMOCAP ~\citep{Busso2008}, MuSE ~\citep{jaiswal2019muse}, MSP-Podcast ~\citep{martinez2020msp}, and MSP-Improv ~\citep{busso2016msp} using metrics that quantify demographic leakage and membership inference risks. Results indicate that adversarial training significantly reduces demographic leakage without substantially degrading emotion recognition accuracy. Interestingly, multimodal models exhibit greater demographic leakage than unimodal ones, particularly due to audio features. Furthermore, applying adversarial components separately to individual modalities yields stronger privacy protection than joint application. The study also demonstrates that training for speaker invariance reduces membership inference risks, though stronger adversarial training does not always equate to better privacy due to potential overfitting and variability across demographic groups. 

Building on these insights, \citet{yin2024primonitor} introduce PriMonitor, an adaptive privacy-preserving framework designed to protect driver privacy in multimodal emotion detection systems within intelligent vehicles. Recognizing that fusing unprotected modalities like video, audio, and text may enable privacy breaches via correlation attacks, PriMonitor employs a unified differential privacy framework customized per modality ~\citep{wang2016using, he2023clustered}. Gaussian noise is added to video and audio features to ensure privacy, while textual data undergoes a novel Generalized Random Response (GRR ~\citep{wang2017locally}) technique that selectively replaces sensitive words probabilistically. This approach preserves semantic meaning and emotional content better than traditional perturbations. A key innovation is the privacy budget reallocation algorithm, which adaptively optimizes privacy budget distribution across modalities to maximize emotion detection accuracy under a fixed global privacy constraint. This optimization leverages Bayesian methods and a pre-aggregator that discards suboptimal budget configurations to reduce computational overhead~\citep{yin2021privacy}. Evaluations on CH-SIMS ~\citep{yu2020ch} and CMU-MOSI~\citep{zadeh2018multimodal} datasets using state-of-the-art multimodal fusion models show that while differential privacy noise initially degrades accuracy, adaptive budget reallocation significantly recovers performance, approaching non-private baselines. The system robustly mitigates membership inference attacks, with the GRR-based text privacy method outperforming conventional word perturbations. Limitations include challenges due to modality imbalance and differing noise effects on convergence during training.

Addressing privacy in multimodal sentiment analysis (MSA), \citet{wu2024hydiscgan} propose HyDiscGAN, a hybrid distributed collaborative learning framework that preserves privacy for audio and visual modalities. Unlike centralized models that aggregate all data, risking privacy breaches HyDiscGAN processes shareable textual data centrally, while audio and visual features are generated distributively using conditional Generative Adversarial Networks (cGANs ~\citep{mirza2014conditional}). The framework generates synthetic audio and visual features conditioned on de-identified text data, approximating real private modality features without exposing sensitive information
~\citep{tsai2019multimodal}. The generated fake features are fused with real text features through gated attention units for sentiment classification. Customized contrastive losses regulate generators and discriminators to enhance feature realism and separability. HyDiscGAN eliminates client-side inference computation, reducing computational load and communication overhead for resource-constrained clients. Experiments on MOSI and MOSEI benchmark datasets ~\citep{zadeh2018multimodal} demonstrate that HyDiscGAN matches state-of-the-art centralized MSA models in accuracy while preserving privacy. It outperforms distributed learning frameworks such as federated learning and split learning in client training cost and inference efficiency. Ablation studies confirm that the generated fake features carry richer sentiment information than real private features, with visualization validating effective learned representations.

Finally, \citet{xu2025privacy} address the pressing need for privacy protection in IoT-driven multimodal sentiment analysis applications like smart assistants and healthcare monitoring. They propose Differentially Private Correlated Representation Learning (DPCRL), which integrates correlated representation learning with differential privacy mechanisms~\citep{wang2016using, he2023clustered, xu2022privacy, zheng2021efficient, zhang2023local}. The correlated representation learning module captures both correlated and uncorrelated representations from heterogeneous modalities (video, audio, text), with a tunable correlation factor that balances privacy noise interference and predictive performance~\citep{eltoft2006multivariate}. DPCRL’s architecture includes LSTM-based feature extractors per modality, separate encoders for correlated and uncorrelated embeddings, a decoder for reconstructing original features to ensure embedding quality, and a differential privacy module adding Laplace noise to enforce $\epsilon$-differential privacy~\citep{eltoft2006multivariate, hu2020differential}. The final sentiment prediction uses concatenated noisy embeddings. Extensive experiments on CMU-MOSI and CMU-MOSEI datasets~\citep{zadeh2018multimodal} show that DPCRL attains comparable or better sentiment prediction accuracy than state-of-the-art models while providing strong privacy guarantees. Ablation studies highlight the importance of both correlated and uncorrelated representations in balancing privacy and utility.

Collectively, these works illustrate diverse strategies for privacy preservation in multimodal emotion recognition, spanning adversarial learning, adaptive differential privacy, distributed generative models, and correlated representation learning. They highlight ongoing efforts to reconcile privacy protection with high utility in complex multimodal settings.

\section{Privacy Challenges and Techniques in Group Emotion Recognition}

Group Emotion Recognition (GER) aims to infer the collective emotional state of a group by analyzing both individual and contextual cues. However, leveraging personal data such as facial images raises significant privacy concerns, especially in sensitive domains like healthcare and education. 

This section explores key challenges and current privacy-preserving techniques in GER, with a focus on synthetic data generation and established practices that avoid individual identification.

\subsection{Synthetic Data for Privacy in Emotion Recognition}

A major challenge in emotion recognition (ER), particularly facial emotion recognition (FER), is balancing model accuracy with ethical and legal obligations to protect individual privacy~\citep{kung2025face}. Recent progress in synthetic data generation offers a promising solution by creating realistic datasets free from identifiable personal information~\citep{boudewijn2024legal, sarmin2024synthetic, min2025can}.

\citet{huang2023auto} exemplify this approach using StarGAN, a generative adversarial network (GAN) ~\citep{goodfellow2020generative}, to synthesize facial expression images of Parkinson’s disease patients. By training on anonymized source data, StarGAN~\citep{choi2018stargan} produces photorealistic facial expressions that preserve essential emotion-related features while dissociating them from real individuals. This ensures privacy compliance without sacrificing FER accuracy. Trained on synthetic data, achieve performance comparable to those trained on real patient datasets. The study highlights synthetic data’s potential to reduce privacy risks in sensitive healthcare applications. Experimental results demonstrate outstanding diagnostic accuracy achieving 100\% Parkinson’s disease diagnosis with EfficientNet-B7 and EfficientNetV2-Small models~\citep{tan2019efficientnet}, significantly outperforming classical methods such as LeNet~\citep{lecun2002gradient}, AlexNet~\citep{krizhevsky2012imagenet}, VGG~\citep{simonyan2014very}, and ResNet~\citep{he2016deep}. Qualitative and quantitative validations confirm the high quality of the generated facial expressions and accurate emotional representation.

Despite these advantages, limitations persist. The approach relies heavily on precise facial synthesis; errors in generation or screening can negatively impact diagnostic outcomes. Moreover, \citet{min2025can} caution that synthetic data, due to its close resemblance to original data, may still pose privacy risks, including potential reidentification.

In a different vein, \citet{Petrova20} propose a privacy-safe framework for GER that entirely avoids individual tracking. Instead of analyzing faces or identities, their method extracts global scene-level features such as crowd density and collective gestures using convolutional neural networks (CNN). For instance, in classroom settings, CNNs analyze aggregated postural shifts or gaze patterns to infer group engagement without identifying individual students. This approach complies with privacy regulations like GDPR and AI Act~\citep{cancela2024eu} by focusing on environmental context rather than personal data, facilitating scalable deployment in public or educational environments where individual consent is often impractical.

Together, these methods illustrate two complementary pathways in privacy-preserving GER: synthetic data generation that maintains emotional fidelity while protecting identities, and the extraction of non-individualized global characteristics that do not take personal data into account. Both approaches contribute to advancing ethical AI by balancing technological innovation with stringent privacy requirements.

\subsection{Conventional Methods and Privacy Concerns in Group Emotion Recognition}

Traditionally, group emotion recognition has predominantly relied on individual-level data such as facial features, key points, and action units (AUs) to infer collective emotional states~\citep{huang2024survey}. These methods typically follow a \textbf{bottom-up} approach, where individual emotional cues are first estimated and subsequently aggregated to determine the overall group affect~\citep{ghosh2018automatic, gupta2018attention, gong2025hybrid}.

A seminal example is the work of \citet{tan2017group}, who combined an individual facial emotion CNN with a global image-based CNN. Their method detects and extracts individual faces from group images, assigns group emotion labels to each face, and trains two separate CNNs: one on aligned facial images and another on non-aligned ones. The final group emotion prediction is computed by averaging scores across all detected faces, underscoring the centrality of personal data in collective emotion inference.

Building on this foundation, \citet{guo2017group} developed a hybrid model integrating global scene features, skeleton data, and local facial cues. This approach independently trains a deep CNN on three modalities: facial expressions, whole-image scenes, and skeleton representations capturing body and hand keypoints, later fusing predictions via decision fusion. The fusion of individual-level emotional signals (e.g., furrowed brows, clenched fists) with contextual scene information (e.g., classroom layouts, crowd density) enhances group-level emotion classification. Further advancing the field, \citet{Guo2018} incorporated visual attention mechanisms within the bottom-up framework. Their model combines a fine-tuned facial CNN, trained on large-scale datasets, with skeleton-based pose and gesture analysis, while attention mechanisms emphasize salient regions such as expressive faces and gesturing hands. For example, in a team meeting scenario, attention weights may prioritize speakers’ expressions over passive attendees, refining the accuracy and robustness of group emotion inference. Collectively, these conventional methods highlight a historical reliance on individual-level data in GER, which raises substantial privacy concerns, especially in domains like healthcare and education.

While bottom-up approaches have proven effective, their dependence on personal data necessitates privacy-preserving alternatives. Recent advances, including privacy-safe synthetic data and global scene-level feature extraction, offer promising avenues to meet ethical and legal standards without sacrificing recognition performance. Synthetic data decouples emotional cues from identifiable information, while global features bypass individual tracking, focusing instead on environmental and collective dynamics.

Nonetheless, traditional individual-based methods remain prevalent, underscoring the need for hybrid frameworks that balance the precision of fine-grained analysis with privacy preservation. For instance, combining anonymized facial landmarks with aggregated vocal features may preserve diagnostic value while safeguarding confidentiality~\citep{leang2024exploring, kung2025face}.

\section{Datasets}
\label{sec:datasets}
In this section, we present a description of the most recent group emotion recognition (GER) datasets available in-the-wild, as well as the datasets used in this thesis. The datasets employed in this work include both GER and non-GER datasets, covering a wide range of emotional and contextual conditions. This section therefore, provides an overview of the datasets relevant to GER, encompassing both publicly available resources in the research community and those specifically selected for our experiments. The aim is twofold: first, to review the major datasets that have shaped GER research in-the-wild and highlight their diversity in modality, scale, and annotation strategy; and second, to summarize the datasets used in this thesis, emphasizing their main characteristics and state-of-the-art (SOTA) performance since their introduction to the community. These datasets span both image-based and multimodal video-based formats, enabling the evaluation of static and dynamic aspects of group emotion recognition under real-world conditions.

\subsection{Group Emotion Recognition Datasets In-the-Wild}

Recent advancements in social media have facilitated the collection of group-level emotional data from diverse settings, ethnic groups, events, and locations. Despite these advances, annotating such data remains challenging due to the complexity and variability inherent in group interactions. Group emotion recognition (GER) datasets can broadly be categorized into two main approaches: unimodal-image datasets supporting image analysis, and multimodal-video datasets enabling dynamic analysis. Annotation for image datasets typically considers overall context and scene elements, while dynamic datasets involve annotations that integrate auditory elements such as speech and environmental noise \citet{dhall2017individual, dhall2020emotiw}. An illustration of these datasets is given in the Table~\ref{tab:ger_dataset}. It is important to mention that not all datasets mentioned for GER  in Table~\ref{tab:ger_dataset} are used in the experiments of this thesis. The main reason is availability. I requested to have access to these datasets, but there is still no answer for most of them; for others, the availability link is no longer available. The list of datasets used for the experiments in this thesis is given in Table~\ref {tab:dataset_sota_2021}.

\subsubsection{Image Datasets for Unimodal Approach}

One of the initial contributions in image GER datasets was the HAPpy PEople Images (HAPPEI) dataset, introduced by \citet{dhall2013finding}. It consists of 2638 images sourced from Flickr, collected using keyword searches focused on social events such as parties and graduation ceremonies. The annotations specifically address happiness intensity at both individual and group levels, alongside annotations related to facial occlusions and poses. HAPPEI particularly emphasizes analyzing social contexts through facial features, making it suitable for methods aimed at happiness intensity estimation within group contexts. Building upon the scope of emotional variety, \citet{mou2015group} presented the MultiEmoVA dataset comprising 250 images from diverse sources. It captures a broader emotional range by annotating images across arousal and valence dimensions into five primary emotional categories. This dataset extends beyond happiness to include varied scenarios like conferences, parties, and sports events, enhancing its utility for evaluating affect recognition in groups.

Further expanding the scope of emotional contexts, the Group Affect Database introduced by \citet{dhall2015more} contains 504 images annotated across Positive, Neutral, or Negative affective categories. Unlike earlier datasets focused predominantly on happiness, this database encompasses a wide array of scenarios, including politically charged events and protests. It integrates both contextual and facial cues, supporting more comprehensive benchmarking for affect recognition in diverse, unconstrained environments. Addressing dataset scalability, \citet{dhall2017individual} developed the Group Affect Database 2.0 (GAF2.0), significantly expanding the volume and diversity with 6471 images specifically intended for the EmotiW 2017 challenge. This dataset maintained the three-class emotional annotation schema (Positive, Neutral, Negative) but provided enhanced data for model training and validation, catering explicitly to group-level emotional analysis. Continuing the trend of scale expansion, Group Affect Database 3.0 (GAF 3.0), introduced by \citet{dhall2018emotiw}, tripled the dataset size to 17,172 images, establishing a comprehensive resource for developing robust emotion recognition models. The dataset’s expanded scope improved the ability to train and validate sophisticated models capable of capturing the complexity inherent in spontaneous, unconstrained social interactions. Extending GAF 3.0 further, the GAF-Cohesion database introduced by \citet{ghosh2019predicting} integrated annotations on group cohesion alongside emotion labels. Comprising 14,175 images, it explored correlations between emotional states and perceived group cohesion, particularly highlighting nuances between different group sizes and cultural contexts.

\citet{wang2022congnn} introduced SiteGroupEmo, a dataset specifically crafted to capture group-level emotions within highly unconstrained environments. With 8,217 images sourced from internet searches, the dataset emphasized the complexity and variability of real-world group emotions, making it particularly relevant for evaluating context-aware models capable of integrating both global scene and local facial information. Similarly, the GroupEmoW dataset by \citet{guo2020graph}, consisting of 15,894 images, further enriched the resources available for GER research. It provided detailed annotations and metadata, including bounding boxes for faces, body regions, and scene-level context, enabling advanced multimodal and context-aware modeling. Its extensive and diverse dataset structure supported the development and validation of sophisticated recognition models tailored for realistic and challenging scenarios.

\subsubsection{Multimodal Video Datasets for Dynamic Approach}

Complementing the image datasets, multimodal-video datasets incorporate temporal dynamics and auditory cues, presenting new challenges and opportunities. The Video-based Group Affect in-the-Wild (VGAF) dataset, introduced by \citet{dhall2020emotiw}, comprises videos from YouTube, covering varied group interactions across diverse demographic and event types, from interviews to protests. Annotated in Positive, Neutral, or Negative categories through consensus voting, the dataset presents complexity through variable resolution, dynamic camera angles, and challenging acoustic contexts. It includes 4183 videos, divided into training (2661), validation (766), and testing (756) subsets. Furthering complexity, the Group-level Emotion on Crowd Videos (GECV) dataset introduced by \citet{quach2022non} is specifically curated for multi-level emotion recognition in crowded environments. It features 627 videos with annotations across individual, group, and video levels, thereby enabling comprehensive analysis of spatial and temporal emotion dynamics. This dataset is structured to facilitate research in real-world applications like security and social media analytics. Recently, \citet{gong2025hybrid} introduced the Group and Scene Emotions dataset, comprising 16,386 images extracted from 128 YouTube videos. It focuses explicitly on highly complex scenarios such as explosions, stampedes, and crowded events where traditional facial detection is difficult. Annotations provide binary labels (positive or negative) emphasizing challenging, dynamic contexts. The dataset includes a clear division into training, validation, and testing subsets, promoting robust model development for realistic, challenging group emotion recognition tasks.  These multimodal video datasets significantly enhance GER research by enabling analysis of dynamic interactions, acoustic elements, and complex contextual scenarios, bridging the gap towards more practical, real-world applications.

 \begin{table}[H]
  \footnotesize
    \centering
    \caption{Group emotion dataset}
   \label{tab:ger_dataset}
    \resizebox{\linewidth}{!}
    {
   \begin{tabular}{|l|c|c|c|c|c|}
    \hline
    \textbf{Dataset} &\textbf{Type}     &\textbf{Sample Size}  &\textbf{Partition}       &\textbf{Annotations}       &\textbf{Task}\\
    \hline
        HAPPEI               & Image     & 2,638         &  \makecell{train(1500)\\ 
        val(1138)\\ 
        test(496)}     & \makecell[l]{neutral (92) \\ Small smile (147) \\ Large smile (774) \\ Small laugh (1256) \\ Large laugh (331) \\ Thrilled (38)} 
  & Regression\\
\hline
MultiEmoVA & Image & 250 &\makecell{5-fold\\ Cross-Validation} &  \makecell[l]{High-positive (46)\\ Medium-positive (64)\\
High-negative (31)\\
medium-negative (27)\\
low-negative (10)\\
neutral (72)}
& Classification\\
\hline
GAF2.0 &Image &6, 467 &\makecell{train (3,630)\\
val (2,068)\\
test (772)} &\makecell[l]{positive (2,356)\\
neutral (2,092)\\
negative (2,019)}  &Classification\\
\hline
GAF3.0 &Image &17, 172 &\makecell{train (9,836)\\
val (4,346)\\
test (3011)} &\makecell[l]{positive (6, 553)\\
neutral (5, 364)\\
negative (5,256)}  &Classification\\
\hline
Group Cohesion & Image & 16, 433 &\makecell{train (9,300)\\ val (4,244)\\ test (2,899)} &[0, 3] & Regression\\
\hline
SiteGroupEmo &Image &10, 034 &\makecell{train (6,096)\\ val (1,972)\\ test (1,966)}&\makecell[l]{positive (4,660)\\ neutral (4,355)\\ negative (1,019)}  &Classification\\
\hline
GroupEmoW &Image &15,894 &\makecell{train (11,127)\\ val (3,178)\\ test (1,589)}&\makecell[l]{positive (6,636)\\ neutral (4,947)\\ negative (4,311)}  &Classification\\
\hline
GECV &Video &627 &\makecell{train (90\%)\\ test (10\%)}&\makecell[l]{positive (204)\\ neutral (221)\\ negative (202)}  &Classification\\
\hline
VGAF &Audio Video &4, 183 &\makecell{train (2,661)\\ val (766)\\ test (756)}&\makecell[l]{positive (1,104)\\ neutral (1,203)\\ negative (1,120)}  &Classification\\
\hline
\makecell[l]{Group\\ and \\Scene\\ Emotion} & video & \makecell{(128 videos)\\ 16, 386 frames} &\makecell{train(11, 471)\\ val (3, 274)\\ test (1, 621)} &\makecell[l]{frames-positive\\ (16, 386)\\ frames-negatives\\ (7, 261)} &Classification \\
\hline
 \end{tabular}
 }
 
 \end{table}

\subsection{Datasets Used in the Thesis}
\label{sec:datasets_used}

The experiments in this thesis employ several diverse, in-the-wild datasets to comprehensively develop and evaluate the proposed models. To thoroughly assess model generalizability across different scenarios, datasets have been selected from multiple domains rather than solely focusing on group emotion recognition.

Two datasets specifically addressing Group Emotion Recognition (GER) included VGAF~\citet{dhall2020emotiw} and GAF-3.0~\citep{gupta2018attention}. Datasets focusing on individual-level emotion recognition were also considered. SAMSEMO~\citep{bujnowski2024samsemo} is a recently developed corpus that includes 23,086 manually transcribed and annotated video scenes across five languages (English, German, Spanish, Polish, Korean), featuring approximately 1,400 unique speakers. Scenes were selected from various genres (lectures, interviews, parliamentary recordings, tutorials, etc.) and validated for audiovisual clarity under in-the-wild conditions (approximately 9\% of initial samples were excluded due to low quality or unintelligibility). Each scene was annotated using a three-judge majority voting scheme, covering six Ekman emotions (Happiness, Sadness, Anger, Fear, Disgust, Surprise), as well as \emph{Neutral} and \emph{Other}. The inter-annotator agreement, defined as the fraction of scenes with unanimous labeling, is 39.37\%. Metadata provided includes gender, language, and visibility of faces, supporting multimodal tasks involving text, audio, and visual data.

MER-MULTI (MER-2023) ~\citep{lian2023mer} is a robust multimodal emotion recognition dataset curated from movies and TV series, collecting 73,148 unlabeled video clips. It specifically targets human-centered emotional expressions. The labeled subset consists of 4,196 rigorously annotated samples partitioned into three specialized subsets: MER-MULTI (411 clips) predicting discrete emotions (Anger, Sadness, Worry, Neutral, Happiness, Surprise) and dimensional valences (continuous pleasure scores); MER-NOISE (412 clips) assessing model robustness against real-world perturbations such as background noise and video blur; and MER-SEMI (834 labeled clips plus the full unlabeled set) designed for semi-supervised learning tasks. Experiments in this thesis utilize MER-MULTI. %

Further datasets addressing dynamic facial expressions and engagement recognition include Dynamic Facial Expression in the Wild (DFEW) ~\citep{jiang2020dfew}, comprising 16,372 video clips sourced from over 1,500 movies capturing facial expressions under unconstrained real-world conditions. Annotated for the seven Ekman emotions, each clip includes a distribution vector of annotator judgments, providing rich labels for benchmarking deep learning models in dynamic facial expression recognition tasks. EngageNet ~\citep{emotiw2023}, introduced during the EmotiW challenge at ICMI 2023, contains around 31 hours of video data from 127 participants (83 male, 44 female), aged 18–37 years. The dataset includes over 11,300 ten-second clips, recorded under uncontrolled conditions. Each clip is annotated into one of four engagement levels: Highly-Engaged, Engaged, Barely-Engaged, and Not-Engaged. The dataset is split into training (7,983 clips from 90 participants), validation (1,071 clips from 11 participants), and test sets (2,257 clips from 26 participants).

This diverse dataset selection ensures robust evaluation and validation of the proposed models' capability to generalize across varied conditions and emotion recognition contexts. In the Table~\ref{tab:dataset_sota_2021}, a summary of the most recent performance  (including the baseline when available) results before the beginning of the thesis is provided for these datasets.

 \begin{table}[ht]
    \centering
    \caption{\small State-of-the-art methods on the validation set of the used dataset before the start of the thesis. A: Audio, V: Video, Img: Image, Acc: Accuracy.
    TCCT-Net: Tensor-Convolution and Convolution-Transformer Network; AU: Action Units; 
    ST-GCN: Spatial-Temporal Graph Convolutional Networks}
   \label{tab:dataset_sota_2021}
    \renewcommand{\arraystretch}{1.3}
    \resizebox{\linewidth}{!}
    {
   \begin{tabular}{|l|c|l|c|c|c|}
    \hline
     \textbf{Dataset} &\textbf{Classes}  &\textbf{Reference}  &\textbf{Modality}  &\textbf{Method}       &\textbf{ Perf [\%] }\\
     \hline

       &     &\citet{dhall2018emotiw} (baseline)   & Img    & \makecell[c]{Inception-V3}          &{Acc = 65.00}\\
       \cline{3-6}
       &     &\citet{khan2018group}   & Img    & \makecell[c]{ResNet, VGG}       &{Acc = 78.39}\\
       \cline{3-6}
GAF3.0 & 3    &\citet{gupta2018attention}   & Img    & \makecell[c]{SE-ResNet\\ ResNet, VGG\\ Inception, } &{Acc = 78.98}\\
\cline{3-6}
        &     &\citet{gupta2018attention}   & Img    & \makecell[c]{DensNet, SphereFace}          &{Acc = 80.98}\\
        \cline{3-6}
       &     &\citet{wang2018cascade}   & Img    & \makecell[c]{CAN, ResNet, Se-Net}          &{Acc = 86.90}\\

\hline 
 \multirow{10}{*}{VGAF}   &\multirow{10}{*}{3}      &\citet{dhall2020emotiw} (baseline)   &A, V    & \makecell[c]{Inception-V3\\ CNN-LSTM}          &{Acc = 51.30}\\
 \cline{3-6}

   &  &\citet{Petrova20}    & Img   &VGG-19            &{Acc = 52.36}\\
   \cline{3-6}
     
   &  &\citet{ottl2020group}    &A   &Deep Spectrum &Acc = 59.40\\
   \cline{3-6}

  &   &\citet{sharma2021audio}   & A, V   &\makecell[c]{Early fusion with\\ LSTM and MLP }           &{Acc = 61.61}\\
  \cline{3-6}
  &   &\citet{pinto2020audiovisual}    & A, V   &\makecell[c]{Resnet-50\\ BiLSTM\\ and fusion SVM}   &{Acc = 62.40}\\
 \cline{3-6}
  &   &\citet{Wang2020}    & A, V   &K-injection network             &{Acc = 66.19}\\
 \cline{3-6}

&  &\citet{Savchenko2021}  &V &{EfficientNET}   &{Acc = 70.33}\\
\cline{3-6}
&  &\citet{Sun2020}    & A, V  &Fusion of 14 models  & Acc =  71.93  \\
\cline{3-6}
&  &\citet{belova2022group}  &A, V & MobileNet-v1  & Acc =  71.95\\
\cline{3-6}
&  &\citet{Liu2020}   &A, V  &Hybrid Network   &Acc = 74.28\\
\hline

 &     &\citet{bujnowski2024samsemo} (baseline)   &A    & {End2End}          &{F1 = 61.10}\\
SAMSEMO & 5    &\citet{bujnowski2024samsemo} (baseline)   &T    & {End2End}          &{F1 = 63.00}\\
 &     &\citet{bujnowski2024samsemo} (baseline)   &V    & {End2End}          &{F1 = 68.20}\\
 &     &\citet{bujnowski2024samsemo} (baseline)   &A, V, T   & {End2End}          &{F1 = 69.00}\\

   \hline
 &     &\citet{singh2023have}   & V   & \makecell{Transformer, Gaze}          &{Acc = 55.45}\\

 \cline{3-6}
  &     &\citet{singh2023have}   & V   &\makecell{Transformer\\ Gaze + Head Pose}    &{Acc = 64.45}\\
 \cline{3-6}
EngageNET & 4 &\citet{singh2023have}   & V   &\makecell{Transformer\\ Gaze + Head Pose + AU}    &{Acc = 69.10}\\
 \cline{3-6}
 &      &\citet{anand2024exceda}   & V   &\makecell{GLAMOR-Net\\Facial feature}    &{Acc = 68.72}\\
  \cline{3-6}
 &     &\citet{Vedernikov_2024_CVPR}   & V   &\makecell{TCCT-Net, Head Pose}    &{Acc = 68.91}\\
  \cline{3-6}
 &     &\citet{abedi2024engagement}   & V   &\makecell{ST-GCN\\ Facial Landmarks}    &{Acc = 71.24}\\
\hline

 &     &\citet{foteinopoulou2024emoclip}   & V   &\makecell{EmoCLIP\\ CLIP-ViT-B/32 }    &{WAR = 62.12}\\
  \cline{3-6}

 &     &\citet{mai2024ous}   & V   &\makecell{OUS, CLIP}    &{WAR = 68.85}\\
  \cline{3-6}

 &     &\citet{chen2024cdgt}   & V   &\makecell{CDGT, Transformer }    &{WAR = 70.07}\\
  \cline{3-6}

 &     &\citet{wang2024joint}   & V   &\makecell{LSGT, ResNet-18 }    &{WAR = 72.34}\\
  \cline{3-6}

DFEW &  7   &\citet{mai2024all}   & V   &\makecell{UMBEnet, CLIP}     &{WAR = 73.93}\\
  \cline{3-6}

 &     &\citet{tao2024align}   & V   &\makecell{Align-DFER\\ CLIP-ViT-L/14 }    &{WAR = 74.20}\\
  \cline{3-6}

 &     &\citet{chen2024finecliper}   & V   &\makecell{FineCLIPER\\ CLIP-ViT-L/16 }    &{WAR = 76.21}\\
  \cline{3-6}

 &     &\citet{chumachenko2024mma}   & A, V   &\makecell{MMA-DFER\\ Transformer }    &{WAR = 77.51}\\
 \hline
 &     &\citet{lian2023mer} (baseline)  & A, V   &-    &{F1-0.25MSE = 56.00}\\
\cline{3-6}
MER-MULTI &   { 6/ [-5,  5]}   &\citet{wang2023hierarchical}   & A, V   &\makecell{JDEV, HUBURT}    &{F1-0.25MSE = 68.46}\\
\cline{3-6}
 &     &\citet{zong2023building}   & A, V   &\makecell{weighted blending\\ supervision signals }    &{F1-0.25MSE = 70.05}\\
\cline{3-6}
\hline

 \end{tabular}
 
 }
  
 \end{table}

 \chapter{Multimodal Group Emotion Recognition In-the-Wild}
\label{chap:mger}

\section{Introduction}
\begin{figure}[H]
\centering
\includegraphics[width=\linewidth]{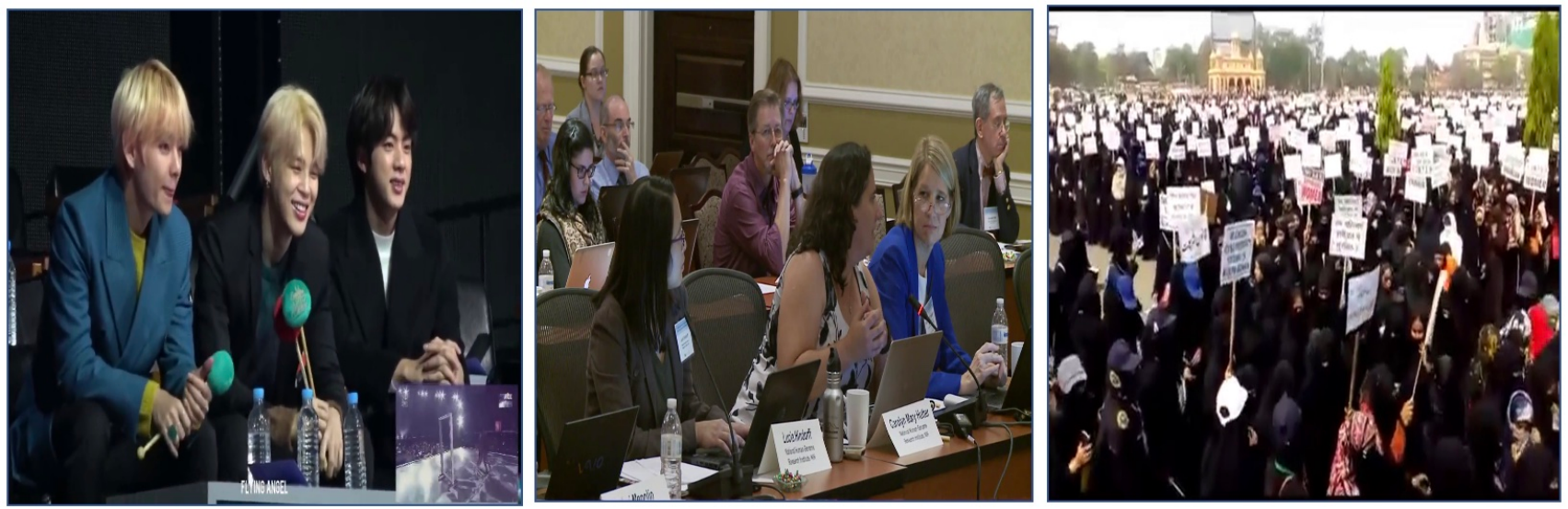}
\caption{Examples from the VGAF dataset. From left to right: Positive, Neutral, and Negative. In the last image, facial expressions are not visible; the decision depends more on context and collective behavior (e.g., protest, signs, posters).}
\label{fig:vgaf_examples}
\end{figure}

This chapter addresses group emotion recognition (GER) under strict privacy constraints in real-world conditions. We aim to infer the collective affect of a group without using individual-identifying signals (e.g., face crops, pose tracks, per-person trajectories). We develop this first research on the Video-level Group Affect (VGAF) dataset from the EmotiW challenge~\citep{Sharma2019} as our primary benchmark due to its alignment with classroom-like group dynamics, its diverse capture conditions, and robust annotations. An overview example is shown in Figure~\ref{fig:vgaf_examples}.

Our contribution is a privacy-compliant, multimodal architecture that fuses video and audio via cross-attention and a learned frame-aggregation operator (Frames Attention Pooling, FAP). Because removing identity-level cues reduces the discriminative signal, we also introduce a targeted synthetic video augmentation. Crucially, we validate this augmentation with a lightweight control model before integrating it into the main architecture. The implementation of the proposed framework is publicly available.\footnote{\url{https://github.com/augusmaa/Emotiw2023}}

\section{Proposed Multimodal Architecture}
\label{sec:proposed}
\begin{figure}[H]
\vspace{0.5cm}

  \centering  \includegraphics[width=\linewidth]{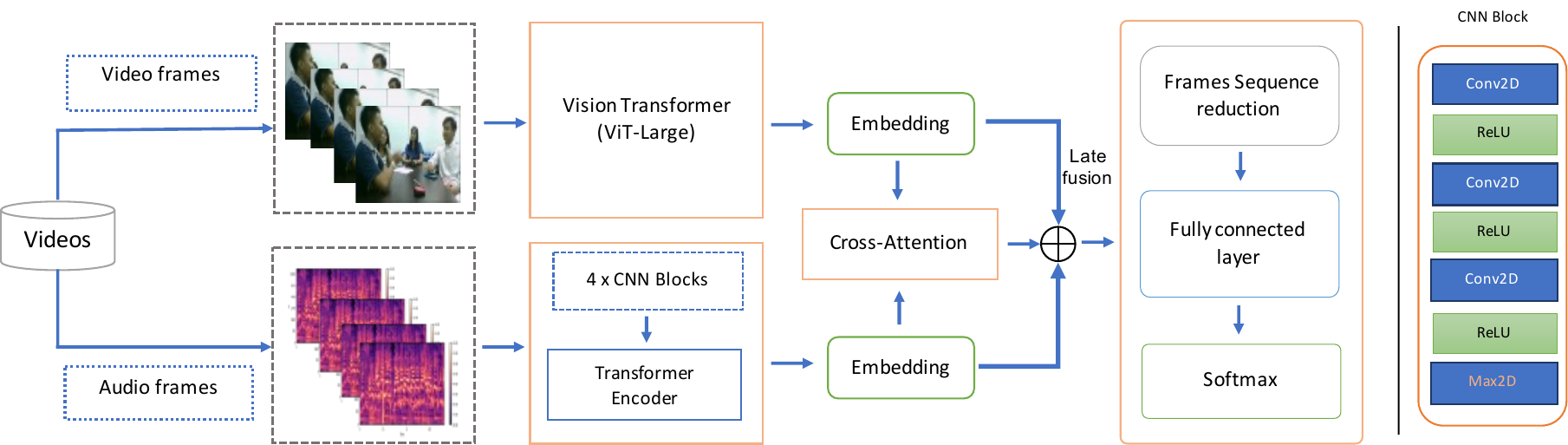}
  \caption{At left, the proposed model is a combination of two monomodal branches, a Cross-Attention and a late fusion paradigm. The video branch uses a pre-trained vision transformer (ViT) model~\cite{Dosovitskiy2020}. The audio branch encompasses 4 CNN blocks followed by a transformer encoder. At right, a description of one CNN block used in the audio branch.}
  \label{fig:model}
\end{figure}

\begin{figure}[H]
  \centering
  \includegraphics[width=0.95\textwidth]{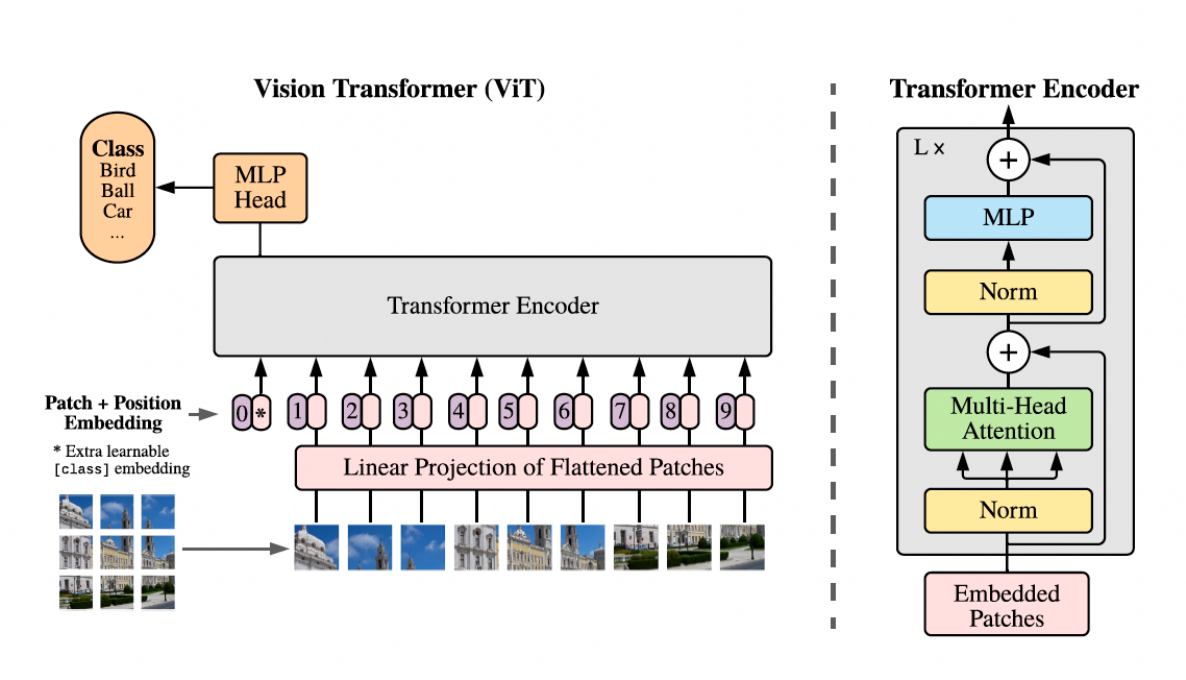}
  \caption{Vision Transformers Architecture, source~\cite{Dosovitskiy2020}.  ViT divides images into fixed-size patches, linearly embeds these patches into token sequences, and feeds them into a transformer encoder. This method leverages the self-attention mechanism to model relationships across patches, effectively capturing spatial context and global relationships.}
  \label{fig:vit}
\end{figure}

The proposed framework uses a two-branch architecture (video and audio) fused via late fusion and cross-attention (Figure~\ref{fig:model}). The video branch uses a pre-trained Vision Transformer (ViT-L/14) fine-tuned for GER. The ViT architecture is reported in Figure~\ref{fig:vit}. The audio branch converts waveforms to Mel-Spectrograms and encodes them with CNN blocks followed by a Transformer encoder. We reduce frame sequences to compact representations with Frames Attention Pooling (FAP), which learns to weight frames by relevance.

\paragraph{Video branch:} We used the 
\textit{vit\_large\_patch14\_224\_clip\_laion2b}  variant of Vision Transformer (ViT) for the task of group emotion recognition in-the-wild . This variant leverages contrastive learning from the extensive LAION-2B dataset \citep{schuhmann2021laion}, comprising approximately 2 billion text-image pairs, enabling the model to build robust semantic visual representations deeply grounded in language. Such extensive multimodal training greatly enhances the model’s ability to generalize across diverse and uncontrolled environments typically encountered in "in-the-wild" scenarios. Specifically, the chosen ViT variant uses a fine-grained patch size of 14x14 pixels, enabling it to capture subtle emotional cues and nuanced interactions among individuals in a group. Its larger architecture (ViT-Large) offers significant representational capacity, necessary for accurately modeling complex emotional interactions in groups. Moreover, the multimodal nature and inherent adaptability provided by CLIP-based~\footnote{\textit{CLIP-based} means the image encoder is initialized from CLIP pretraining (Contrastive Language–Image Pretraining) on large-scale image–text pairs (LAION-2B~\citep{schuhmann2021laion}). CLIP aligns images with captions via a contrastive objective, yielding semantically rich, robust features.} training makes this variant particularly effective for emotion recognition tasks that require nuanced semantic interpretation and robust performance in varied real-world contexts.

\paragraph{Audio branch.} Audio is resampled to 16~kHz mono and transformed into Mel-Spectrograms (128 filters). Four CNN blocks (shown in appendix~\ref{tab:cnn_audio}) extract time-frequency features; a Transformer encoder (four heads; FFN size $2\times d_{model}$) produces $d_{model}$ dimensional embeddings.

\paragraph{Fusion.} To evaluate the fusion scheme, we compare (i) concatenation of pooled audio/video embeddings and (ii) cross-attention where audio acts as queries ($Q$) over video keys/values ($K,V$), enabling audio-guided selection of visual evidence. Both branches produce frame-level tokens, which are fused before being reduced by standard averaging pooling or by Frames Attention Pooling (FAP).

 \paragraph{Frames Attention Pooling (FAP)}
\label{sec:fap}
Let $F=[f_1,\dots,f_n]\in\mathbb{R}^{p\times n}$ denote $n$ frame features, and $w\in\mathbb{R}^{p}$, $b\in\mathbb{R}$ be learnable parameters.

\begin{enumerate}

\item Compute a scalar score for each feature vector \(f_i\) via a learned linear map:
\[
s_i \;=\; w^\top f_i + b,
\qquad
i = 1,2,\dots,n.
\]

\item Turn these scores into a probability distribution over the \(n\) elements using softmax function:
\[
\alpha_i
\;=\;
\frac{\exp(s_i)}
     {\displaystyle\sum_{j=1}^n \exp(s_j)},
\qquad
\sum_{i=1}^n \alpha_i = 1.
\]

\item Compute the pooled output as the weighted sum of the original vectors:

\begin{equation}
\mathrm{FAP}(F)=\sum_{i=1}^n \alpha_i f_i
\end{equation}

\end{enumerate}
FAP is adapted from Attentive Statistic Pooling (ASP) without the standard deviation modeling presented by ~\citet{okabe2018attentive}.

\noindent Having defined the overall multimodal architecture, we next detail the preprocessing pipeline that ensures the system remains fully compliant with privacy-preserving constraints while preparing consistent multimodal inputs.

\section{Privacy-Preserving Data Processing}
\label{sec:privacy-data}
In this research, a non-individual pipeline (Figure~\ref{fig:feature_extraction}): no face cropping, pose extraction, tracking, counting, or identity-specific attributes are used as inputs. The model receives whole frames and global audio.

\subsection{Video processing}

\begin{figure}[H]
\centering
\includegraphics[width=0.9\textwidth]{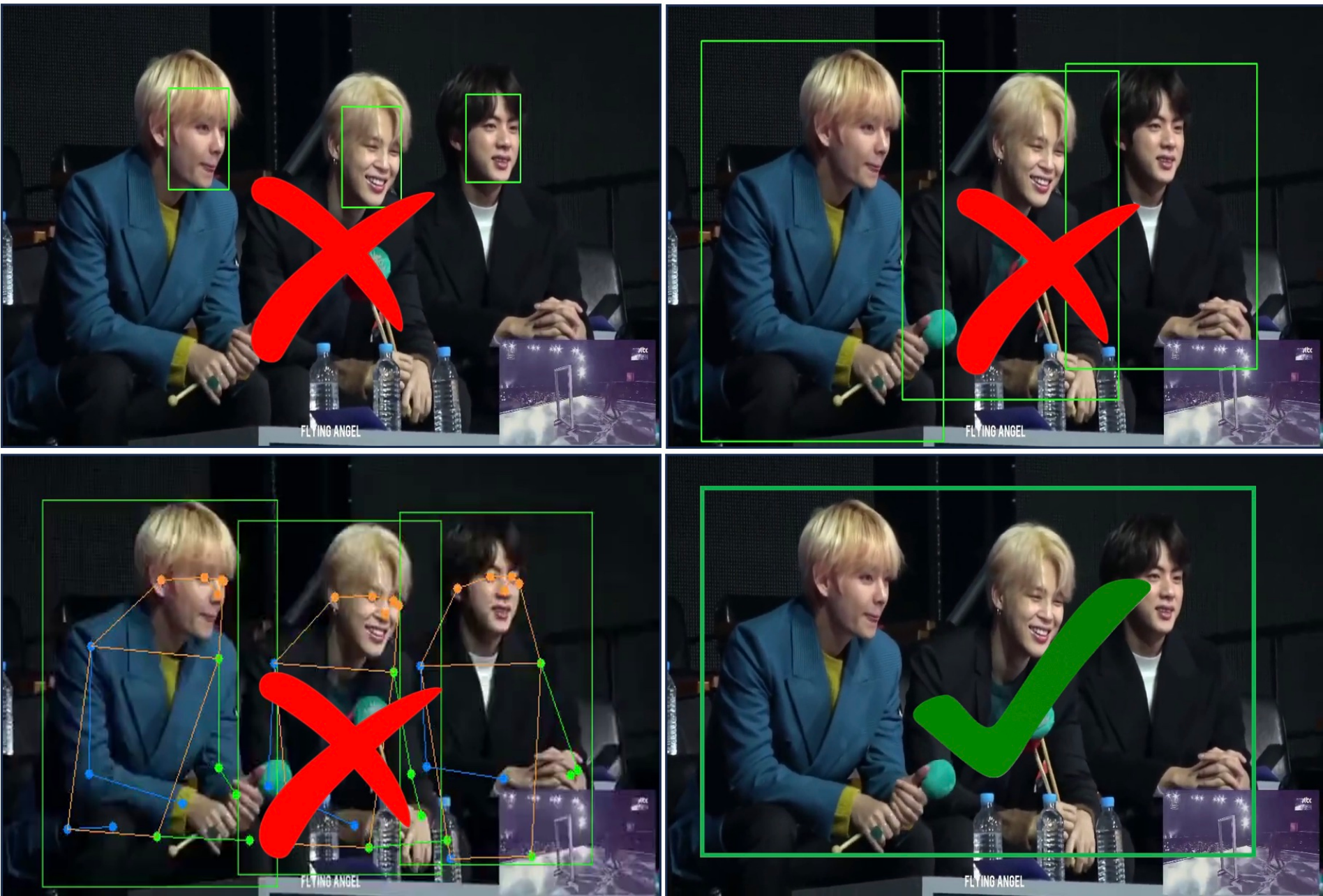}
\caption{Non-individual feature policy. The global image is the input; identity-revealing features (face crops, pose tracks) are excluded.}
\label{fig:feature_extraction}
\end{figure}

From each video, a fixed number of frames is uniformly sampled to capture temporal dynamics while controlling input length. In VGAF, the minimum available frame rate of 15 fps across all videos is considered to avoid duplication, ensuring uniform temporal coverage. Two configurations are considered: 5 frames and 75 frames per video. In the 5-frame setting, frames are extracted at 1 frame per second. For the 75-frame setting, 15 frames are extracted per second uniformly.  All extracted frames are resized to 224 × 224 pixels before being fed into the video branch. In compliance with the privacy-preserving objective (see Figure~\ref{fig:feature_extraction}), no cropping or tracking of individuals is performed, and features such as faces, body pose, or identity-specific attributes are excluded; the model receives the entire frame pixels as input.

\subsection{Audio Processing}

\begin{figure}[H]
\centering
\includegraphics[width=1\textwidth]{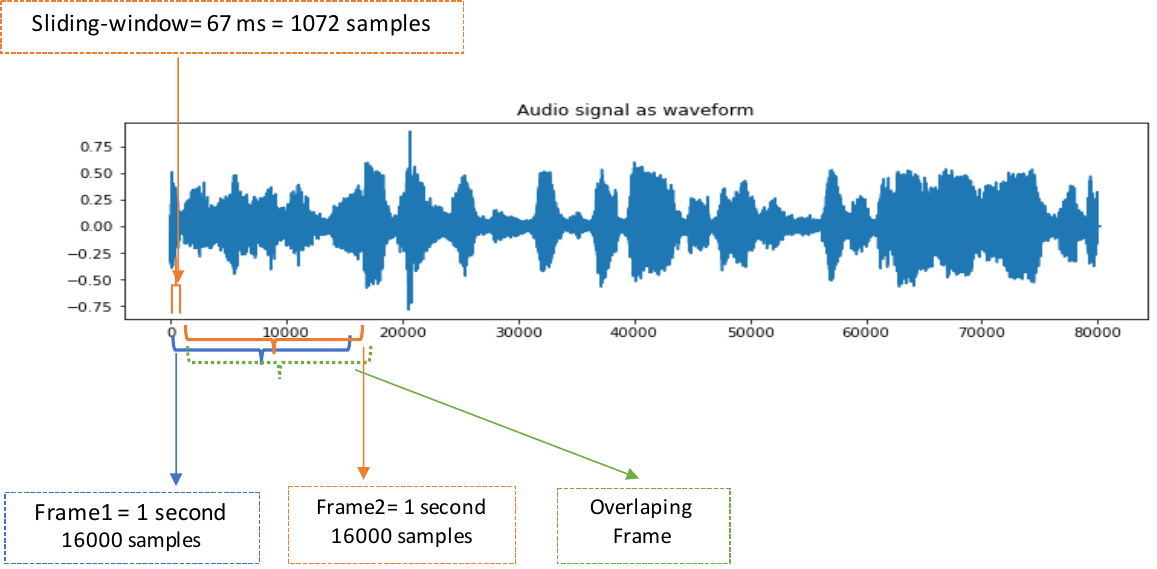}
\caption{Audio framing aligned with 5 and 75 video frames.}
\label{fig:audio_frames}
\end{figure} 

The audio of all videos is standardized by resampling it at 16 kHz and converting it to a mono channel. 5 or 75 audio frames are extracted per video in compliance with the video branch. With 5 frames, the audio frame corresponds to 1 second with no overlap. In the case of 75 frames, a sliding window is set to 67 milliseconds (1,072 samples) to get the right number of frames. An illustration of the audio frames extraction is given in Figure~\ref{fig:audio_frames}. Lastly, each audio frame is converted into Mel-Spectrograms using 128 Mel filters, to produce an input image of $128 \times 251$ adapted to the CNN blocks.

\noindent Although this non-individual processing ensures privacy, it also limits the availability of discriminative affective cues such as facial expressions and body poses. To address this trade-off, we introduce a synthetic video generation process designed to enhance the model’s ability to capture collective emotional signals without compromising privacy.

\section{Synthetic Video: Motivation, Generation, and Sanity Check}
\label{sec:synthetic}

\begin{figure}[H]
  \centering
  \includegraphics[width=0.8\textwidth]{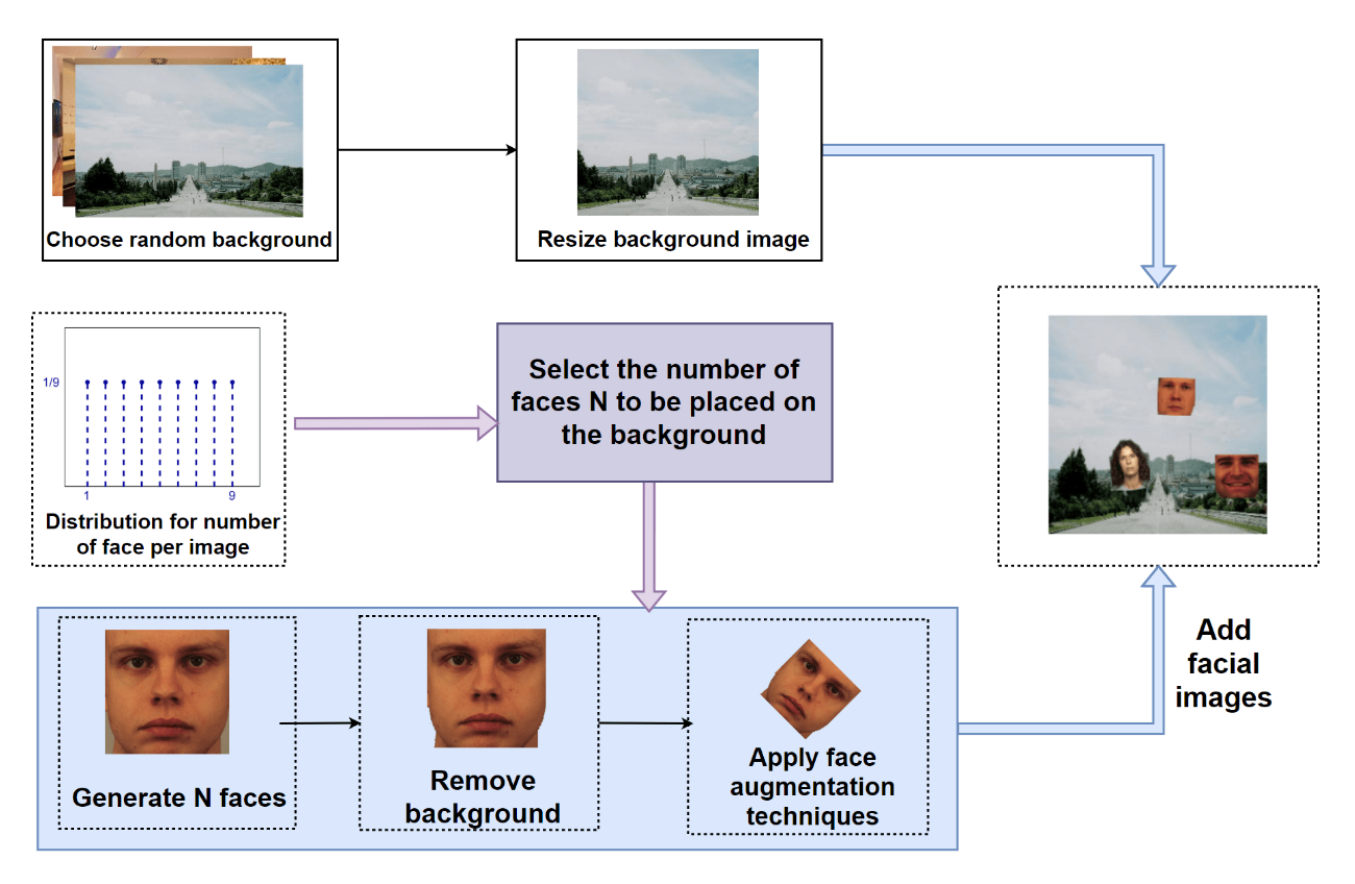}
  \caption{Synthetic image process (source~\citep{Petrova20}). }
  \label{fig:synt_process_petrova}
\end{figure}

\paragraph{Motivation.}
Because privacy-preserving processing omits facial and pose crops as inputs, the model may lack localized affective cues. To counterbalance this loss, we design a controlled synthetic data generation process. The intention is to regularize learning and improve generalization without reintroducing individual features as inputs.

\subsection{Synthetic Video Generation}
\begin{figure}[H]
  \centering
  \includegraphics[width=0.99\textwidth]{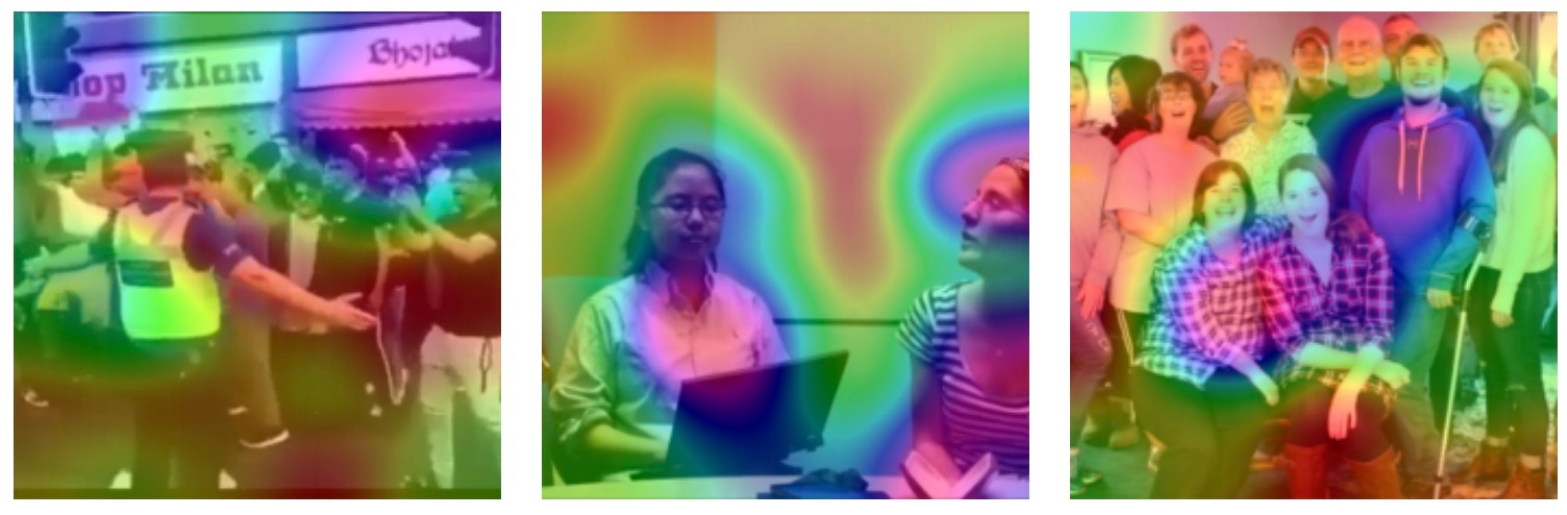}
  \caption{Negative, Neutral, Positive grad-cam visualization. (source~\citep{Petrova20}). The red areas on the heat map indicate less important pixels, while the green, blue, and purple areas indicate increasing relevance. The neutral class heatmap is the most accurate, as it focuses on both people and ignores the background.}%
  \label{fig:grad_cam_petrova}
\end{figure}

Synthetic Video Generation is grounded on prior work by our research team~\citep{Petrova20}, which composites emotion-expressive faces onto background scenes to model GER “in the wild.” The image-compositing pipeline is shown in Figure~\ref{fig:synt_process_petrova}. From that previous work, it is already shown that adding synthetic face data can help a model to focus on the faces of people by ignoring the less important parts in the environment. An example of Class Activation Map (CAM)  of three emotions is shown in Figure~\ref{fig:grad_cam_petrova}.  Here, we extend it to the video level by animating multiple faces across frames.

 The generation video process creates clips that preserve group-level, non-individual cues: real face images from FACES and KDEF~\citep{Ebner2010,calvo2018human} are composited onto LSUN backgrounds~\citep{Yu2015} to form emotion-labeled frames. The resulting face pool covers a wide range of ages and genders. Class composition is intentionally imbalanced to reflect available data: Negative (anger, disgust, fear, sadness) contributes 2{,}946 faces, Neutral 744, and Positive (happiness) 737.

\begin{figure}[H]
  \centering
  \includegraphics[width=0.99\textwidth]{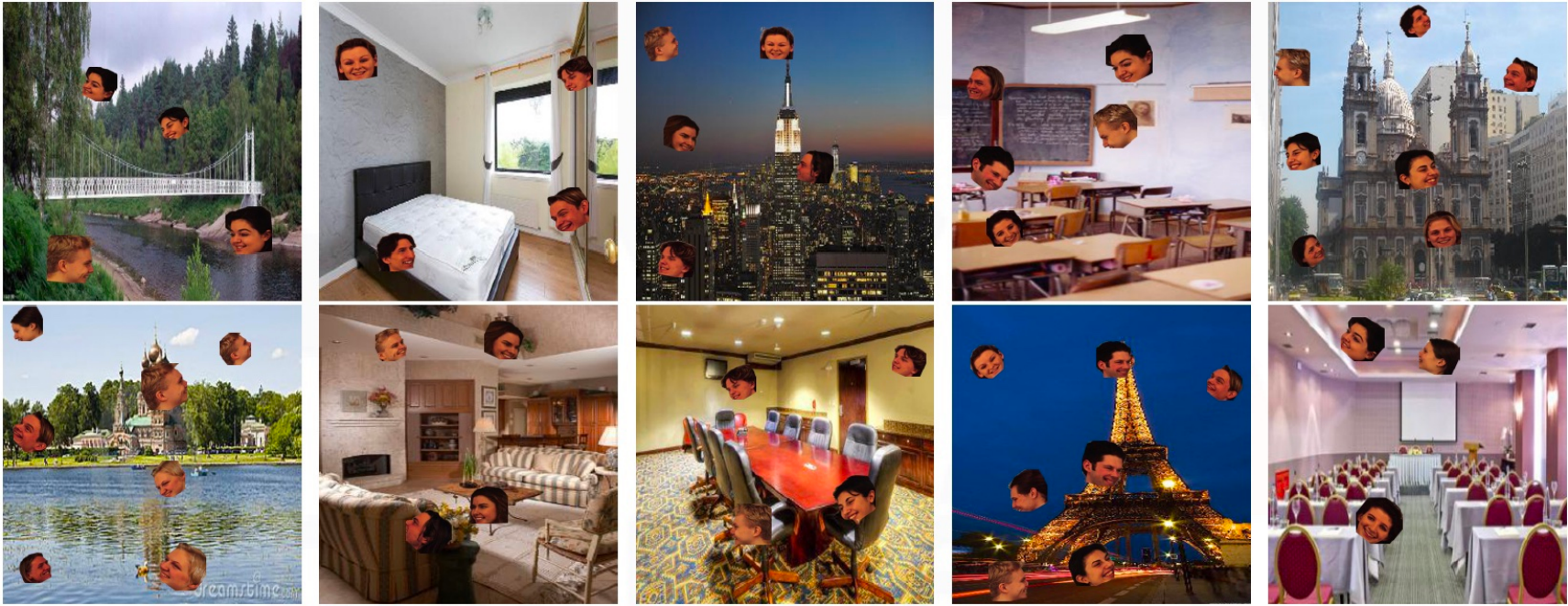}
  \caption{Ten synthetic images across environments.}
  \label{fig:synthetic_img}
\end{figure}

\begin{figure}[H]
  \centering
  \includegraphics[width=0.99\textwidth]{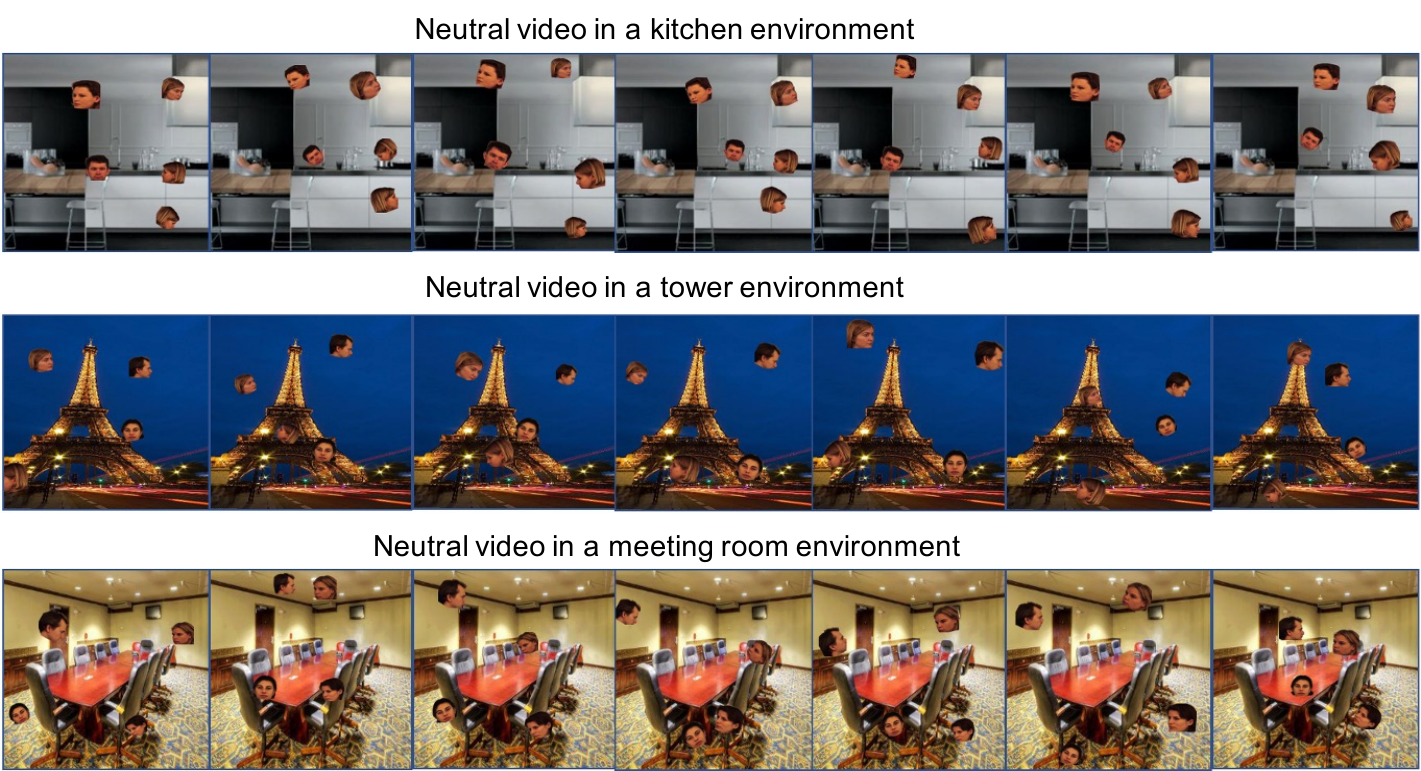}
  \caption{Example synthetic video (Neutral) composed of seven frames with animated face placements.}
  \label{fig:synthetic_video}
\end{figure}

To synthesize videos, we randomly select 3 to 9 faces from a single target class, place them on a fixed background, and animate them along random trajectories to induce mild motion and occlusion; up to 10\% of each face can be masked. Ten background environments are used: Bedroom, Bridge, Church outdoor, Classroom, Conference room, Dining room, Kitchen, Living room, Tower, and Restaurant, and for each environment, 200 distinct sample backgrounds are used. This yields 2,000 videos per class. The class label of a frame and its corresponding clip is inherited from the emotion expressed by the placed faces, not from the background. The intention is to encourage the model to attend to aggregate facial affect rather than memorize specific scenes. Figures~\ref{fig:synthetic_img} and~\ref{fig:synthetic_video} illustrate representative images of a neutral clip.

\subsection{Sanity Check with a Simple Model}
\label{sec:baseline-sanity}
Before integrating the synthetic clips into the main architecture, we first verify that they contain a meaningful label signal when combined with real VGAF data. To this end, a compact control experiment is designed to assess whether synthetic videos contribute useful emotion-related information. The sanity-check model employs VGG19 as a feature extractor, followed by a two-layer BiLSTM (hidden size 512) and an MLP classifier with softmax output. Training is performed using stochastic gradient descent (SGD) with a learning rate of $10^{-5}$. Both pretrained and frozen-backbone variants are evaluated, and the proportion of synthetic data is varied from $0\%$ to $30\%$, uniformly mixed across Positive, Neutral, and Negative classes. For readability, the full experimental configuration and per-class metrics are reported in the appendix. Table~\ref{tab:results_cnnlstm} summarizes the validation accuracy, while Figure~\ref{fig:cm_tsne} illustrates the corresponding confusion matrices and t-SNE embeddings.

On the validation set, the best accuracy (63.31\%) occurs at a 10\% synthetic ratio. As the synthetic proportion increases further, Neutral becomes the most sensitive class with a noticeable drop, whereas Negative benefits the most, which is consistent with its larger and more diverse pool of facial exemplars. The t-SNE projections suggest that modest regularization clusters become slightly more compact rather than a fundamental reshaping of the representation space. Taken together, these observations support the integration of synthetic videos into the main model, provided the synthetic ratio is controlled.

\noindent After confirming through a control experiment that synthetic clips contain usable label information, we integrate them into the full multimodal architecture. The following section describes the training strategy and the selection of optimal synthetic-to-real data ratios.

\section{Training Methodology for the Main Model}
\label{sec:training}

This section describes the training procedure applied to the main multimodal framework, including fine-tuning of the visual backbone and optimization of the audio encoder. The training process is designed to effectively combine real and synthetic data while enabling consistent multimodal learning. Specifically, the ViT-L/14 backbone is fine-tuned for 100 epochs, with all 24 transformer blocks frozen during the first 10 epochs and subsequently unfrozen for joint end-to-end optimization. The audio encoder is trained from scratch following the same schedule.

\subsection{Choosing the Synthetic Ratio}
\label{sec:synt-choice}
The impact of the synthetic-video proportion is measured using 5 frames per clip and sweep ratios from 10\% to 50\% in 10\% steps (Table~\ref{tab:syntchoice}). For this sweep, all 24 ViT blocks are kept frozen to isolate the effect of the synthetic ratio. The best validation accuracy (75.07\%) is obtained at 30\%; both lower and higher ratios underperform. We therefore choose a 30\% synthetic ratio for all subsequent experiments that combine real and synthetic data.

\begin{table}[H]
 \caption{Impact of synthetic video ratio on validation accuracy. Synt: synthetic.}
  \label{tab:syntchoice}
  \begin{tabular}{ccccc}
    \toprule
     Synt. ratio [\%] & \hspace{0.3cm} Synt. videos \hspace{0.3cm} & \hspace{0.7cm} Total train videos \hspace{0.3cm} & \hspace{0.7cm} Accuracy [\%] \hspace{0.2cm} \\
\midrule
 0   & 0     & 2661 & 70.10 \\
 10  & 297   & 2958 & 74.80 \\
 20  & 666   & 3327 & 74.93 \\
\textbf{30} & \textbf{1140} & \textbf{3801} & \textbf{75.07} \\
 40  & 1773  & 4434 & 74.41 \\
 50  & 2661  & 5322 & 74.28 \\
  \bottomrule
\end{tabular}
\end{table}

\section{Ablation Studies}
\label{sec:ablations}
\noindent Once the training configuration and data composition are established, we systematically evaluate the contribution of each design component through a series of ablation studies. These experiments help identify which elements, modality, pooling strategy, or fusion type most influence model performance.
The ablation studies cover modalities, frame pooling, sequence length, and fusion strategy to understand their contributions.

\subsection{Monomodal Audio}
\label{sec:abl-audio}
The first results are presented in Table~\ref{tab:ablation_audio}.  Audio-only performance peaks at 56.40\% with 5 frames and average pooling, and declines when using 75 frames under simple averaging, likely due to overlap-induced smoothing. Frames Attention Pooling (FAP) does not improve the audio branch.

\begin{table}[H]
\centering
\caption{Audio-only ablation comparing Average (Avg.)  pooling and Frames Attention pooling (FAP) with 5 and 75 frames.}
\label{tab:ablation_audio}
\resizebox{\linewidth}{!}{
\vspace{0.5em}
\begin{tabular}{l|cc|cc}
\toprule
  \textbf{Input Data} & \multicolumn{2}{c}{\textbf{5 Frames}} & \multicolumn{2}{c}{\textbf{75 Frames}} \\
\cmidrule{2-5}
          & Avg. Pooling & F. Att. Pooling (FAP) & Avg. Pooling & F. Att. Pooling (FAP) \\
\midrule 
 audio    & \textbf{56.40} & 53.13       & 54.96        & 54.17 \\
\bottomrule
\end{tabular}
}
\end{table}

\subsection{Monomodal Video and Synthetic Video}
\label{sec:abl-video}
For the video branch, 5 and 75 frames are evaluated with Average (Avg.) and Frames Attention pooling, and compare frozen (FW) versus released (fine-tuned) weights (RW) of ViT-L/14. Synthetic augmentation consistently improves results: the peak video-only accuracy reaches 79.24\% at 75 frames with average pooling when mixing real and synthetic videos, a gain of about 5 percentage points over real-only (74.15\%). Attention pooling further boosts performance when ViT is fine-tuned, especially with synthetic mixing (Table~\ref{tab:ablation_video}).

\begin{table}[H]
\centering
\caption{Video-only ablation: effect of frame reduction strategies on validation accuracy with 5 and 75 frames. FW: frozen weights; RW: released (fine-tuned) weights. Inputs use VGAF video and/or synthetic video (synt\_video).}
\label{tab:ablation_video}
\begin{tabular}{l|ccc|cr}
\toprule
  \textbf{Input Data} \hspace{0.7cm} & \hspace{0.3cm} \textbf{Frames Reduction} \hspace{1cm} & \multicolumn{2}{c}{\textbf{5 Frames}} & \multicolumn{2}{c}{\textbf{75 Frames}} \\
\cmidrule{2-6}
         & {-} & FW & RW & FW & RW \\
\midrule
 synt\_video          & Avg. Pooling & 55.48 & 57.44 & 55.74 & 60.05 \\
 video               & Avg. Pooling & 70.10 & 73.62 & 70.63 & 74.15 \\
 video + synt\_video & Avg. Pooling & 75.07 & \textbf{75.98} & 75.59 & \textbf{79.24} \\
\midrule
 synt\_video          & FAP &  --   & 61.88 &  --   & 59.14 \\
 video               & FAP &  --   & 76.11 &  --   & 77.72 \\
 video + synt\_video & FAP &  --   & \textbf{78.07} &  --   & \textbf{78.72} \\
\bottomrule
\end{tabular}
\end{table}

\subsection{Fusion Strategies and Frame Attention Pooling}
\label{sec:abl-fusion}
Two fusion strategies are evaluated: standard concatenation of pooled audio and video embeddings, and cross-attention where audio serves as queries over video keys/values combined with Average (Avg.) pooling and Frames Attention pooling at 5 and 75 frames. Cross-attention is applied only when ViT weights are released (fine-tuned). As summarized in Table~\ref{tab:ablation_fusion}, cross-attention consistently outperforms concatenation in the fine-tuned regime. The best multimodal validation accuracy (79.11\%) is achieved with cross-attention and \textbf{FAP} at 75 frames.  Notably, concatenating audio with synthetic video alone yields minimal improvement, underscoring the dominant role of informative visual features from real videos.

\begin{table}[H]
\centering
\caption{Ablation of fusion modalities on the VGAF dataset. We compare \textbf{Average (Avg.)} vs. (\textbf{FAP}) frame attention pooling with \textbf{Concatenation (Concat)} and \textbf{Cross Attention (CrossAtt)} fusion for 5 and 75 frames. We report results with ViT frozen (FW) or released (fine-tuned, RW). Inputs: VGAF video, VGAF audio, and synthetic video (synt\_video).}
\label{tab:ablation_fusion}
\resizebox{\linewidth}{!}{
\begin{tabular}{l|cccc|cr}
\toprule
\textbf{Input Data} & \textbf{Frames Reduction} & \textbf{Fusion} & \multicolumn{2}{c}{\textbf{5 Frames}} & \multicolumn{2}{c}{\textbf{75 Frames}} \\
\cmidrule{2-7}
 & -- & -- & FW & RW & FW & RW \\
\midrule
synt\_video + audio             & Avg. Pooling & Concat   & 60.05 & 60.96 & 52.61 & 59.40 \\
video + audio                   & Avg. Pooling & Concat   & 71.28 & 74.41 & 72.06 & 75.07 \\
video + audio + synt\_video     & Avg. Pooling & Concat   & 75.85 & \textbf{77.28} & 76.50 & \textbf{77.42} \\
\midrule
synt\_video + audio             & Avg. Pooling & CrossAtt &   --   & 62.14 &   --   & 58.75 \\
video + audio                   & Avg. Pooling & CrossAtt &   --   & 76.11 &   --   & 77.42 \\
video + audio + synt\_video     & Avg. Pooling & CrossAtt &   --   & \textbf{78.07} &   --   & \textbf{78.72} \\
\midrule
synt\_video + audio             & FAP & CrossAtt &   --   & 57.76 &   --   & 58.22 \\
video + audio                   & FAP & CrossAtt &   --   & 77.15 &   --   & 78.07 \\
video + audio + synt\_video     & FAP & CrossAtt &   --   & \textbf{77.54} &   --   & \textbf{79.11} \\
\bottomrule
\end{tabular}
}
\end{table}

\subsection{Results on VGAF and Competition Submissions}
\label{sec:results}
Five configurations were submitted to EmotiW 2023 (Table~\ref{tab:test}). Audio-only achieves about 55\% accuracy, confirming its limited standalone discriminability. The video + synthetic video model (v3) reaches 74.73\% test accuracy. The multimodal models (v4 and v5) with 5 and 75 frames both attain 75.13\% test accuracy; despite identical overall accuracy, their prediction agreement is 88\%, indicating complementary representations. Comparisons with v1 and v3 further show that audio contributes less discriminative power (agreement 53\% and 90\%, respectively), but can add complementary cues when fused. Cross-attention with \textbf{FAP} helps exploit this synergy effectively.

\begin{table}[H]
\centering
\caption{Test set accuracy of five submitted versions to EmotiW2023 challenge on the VGAF dataset.}
\label{tab:test}
\vspace{0.5em}
\resizebox{\linewidth}{!}{
\begin{tabular}{clcccc}
\hline
\textbf{Version} & \textbf{Input data} & \textbf{Frames Reduction} & \textbf{Fusion} & \textbf{Nb Frames} & \textbf{Acc. [\%]} \\
\hline
v1 & audio                        & Avg. Pooling & --        & 5   & 55.29 \\
v2 & synt\_video                  & Avg. Pooling & CrossAtt  & 75  & 54.23 \\
v3 & video + synt\_video          & Avg. Pooling & CrossAtt  & 75  & 74.73 \\
v4 & video + synt\_video + audio  & Avg. Pooling & CrossAtt  & 5   & 75.13 \\
v5 & video + synt\_video + audio  & Avg. Pooling & CrossAtt  & 75  & 75.13 \\
\hline
\end{tabular}
}
\end{table}

\subsection{Comparison with the State of the Art Until 2023}
\label{sec:sota}
Table~\ref{tab:comparison} contrasts our approach with VGAF systems before 2023 and indicates whether individual features (face crops, landmarks, pose) are used. Our privacy-preserving variant with synthetic augmentation is competitive with the strongest reported methods while avoiding identity-level inputs.

\begin{table}[H]
\caption{Comparison with SOTA systems on the VGAF dataset (2023). Columns indicate usage of individual features (Ind. Feat.), accuracy on the official validation set, and, when reported, test accuracy. A: Audio; V: Video; SV: Synthetic video.}
\label{tab:comparison}
\vspace{0.5em}
\resizebox{\linewidth}{!}{
\begin{tabular}{lccc}
\toprule
 & \textbf{Ind. feat.} & \textbf{Acc. Val. [\%]} & \textbf{Acc. Test [\%]} \\
\midrule
\textit{baseline \small(A+V)} &    & \textit{51.30} & \textit{47.88} \\
\midrule
\textit{Ours v1 \small(A, Avg. Pooling): EmotiW2023} &    & \textit{56.40} & \textit{55.29} \\
\citet{ottl2020group} \small(A) &    & 59.40 & 62.30 \\
\midrule
\citet{Petrova20} \small(V) &    & 52.36 & 59.13 \\
\textit{Ours v2 \small(SV, Avg. Pooling): EmotiW2023} &    & \textit{60.05} & \textit{54.23} \\
\citet{savchenko2022neural} \small(V) & {\checkmark} & 70.23 &    \\
\textit{Ours v3 \small(V, Avg. Pooling)} &    & \textit{74.15} & \textit{  } \\
\textit{Ours v4 \small(SV+V, Avg. Pooling): EmotiW2023} &    & \textit{\textbf{79.24}} & \textit{74.73} \\
\midrule
\citet{Evtodienko2021} \small(A+V) &    & 60.37 &    \\
\citet{sharma2021audio} \small(A+V) & {\checkmark} & 61.61 & 66.00 \\
\citet{pinto2020audiovisual} \small(A+V) &    & 65.74 &    \\
\citet{Wang2020} \small(A+V) &    & 66.19 & 66.40 \\
\citet{Sun2020} \small(A+V) & {\checkmark} & 71.93 &    \\
\citet{belova2022group} \small(A+V) &    & 71.95 &    \\
\citet{liu2020group} \small(A+V) & {\checkmark} & 74.28 & \textbf{76.85} \\
\textit{Ours v5 \small(A+SV+V, Avg. Pooling): EmotiW2023} &    & \textit{78.07} & \textit{75.13} \\
\textit{Ours v6 \small(A+SV+V, Avg. Pooling): EmotiW2023} &    & \textit{78.72} & \textit{75.13} \\
\textit{Ours v7 \small(A+SV+V, FAP)} &    & \textit{\textbf{79.11}} & \textit{  } \\
\bottomrule
\end{tabular}
}
\end{table}

\section{Discussion and Limitations}
\label{sec:discussion}

\noindent The above results position our approach among the top-performing methods on the VGAF benchmark. We now interpret these findings in greater depth, analyzing strengths, limitations, and the implications of our privacy-preserving design.

The proposed multimodal neural architecture demonstrates promising performance in recognizing group emotions while adhering strictly to privacy constraints by relying on global rather than individual-specific features. Our extensive ablation study reveals insights into the effectiveness of various modality combinations, pooling strategies, and data augmentation methods.
In monomodal experiments, the video modality, particularly when augmented with synthetic data, consistently outperformed audio. The optimal monomodal performance achieved was 79.24\% accuracy using 75 video frames and average pooling, suggesting that video features capture richer emotional information compared to audio features alone. Frames Attention Pooling (FAP) over frames further enhanced performance, highlighting its advantage in dynamically weighting frame-level features, although this effect was less pronounced in the audio branch.\\

The multimodal fusion experiments provided important insights. Although multimodal models typically benefit from complementary information across modalities, our results indicated that fusion strategies did not always surpass monomodal video performance, particularly when employing average pooling with larger frame sequences. Specifically, attention-based pooling with 75 frames achieved a maximum multimodal accuracy of 79.11\%, closely approaching but not surpassing the monomodal best performance. The relatively lower performance of the audio branch, particularly with attention pooling, likely constrained the overall effectiveness of multimodal fusion.
Comparison with existing state-of-the-art systems on the VGAF dataset underscores the strengths and limitations of the proposed approach. The audio-only model achieved lower performance (56.40\%) compared to methods employing complex audio feature extraction pipelines, such as OpenSMILE coupled with Deep Spectrum analysis (59.40\%). However, the video-only model demonstrated superior performance compared to prior approaches that also avoided individual-specific features, underscoring the value of leveraging fine-tuned vision transformers and synthetic video data augmentation.
The synthetic data augmentation strategy significantly improved video modality performance, highlighting its effectiveness in compensating for limited real-world training data. Nevertheless, our synthetic data approach is currently limited to facial information, omitting potentially valuable contextual cues from body language. Enhancements to synthetic data generation, such as incorporating full-body representations or employing generative models (e.g., GAN), may further improve performance. This strategy will be investigated in the next chapter (chapter~\ref{chap:ve-md}).

Regarding model complexity, the experiments revealed that increasing architectural parameters, such as attention heads or layers, frequently led to overfitting due to insufficient training data diversity. Therefore, a careful balance between model complexity and available training data remains essential. Exploring transfer learning from related datasets or employing more sophisticated synthetic data generation methods may help mitigate this limitation.

\begin{figure}[H]
\centering
\includegraphics[width=1\textwidth]{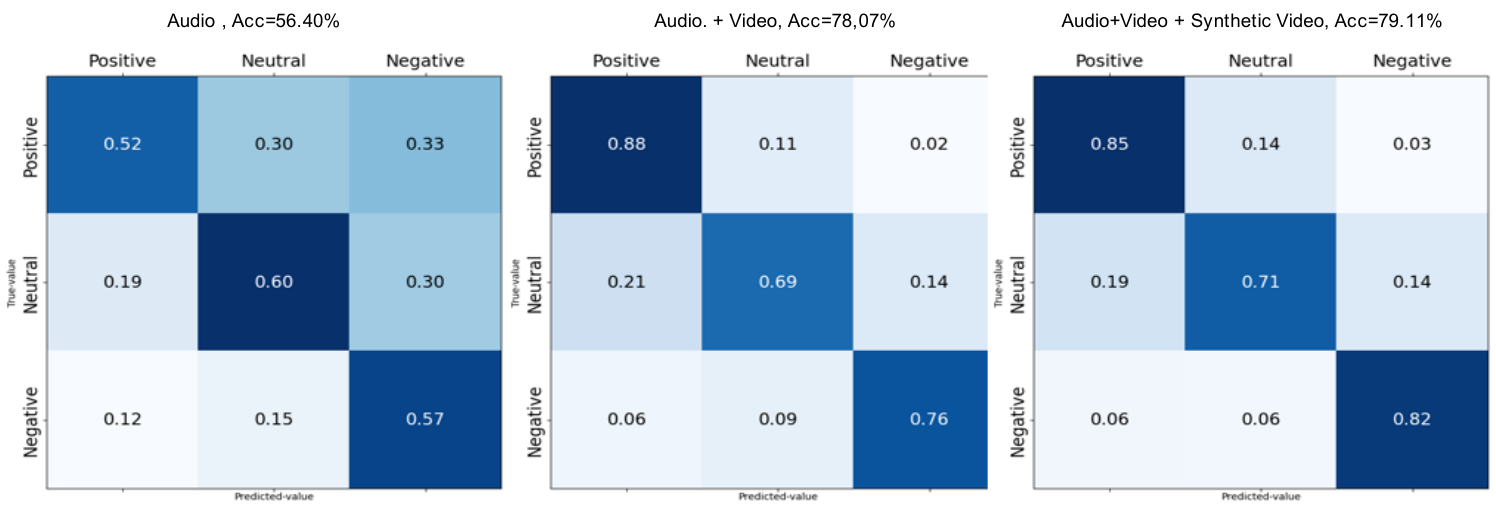}
\includegraphics[width=1\textwidth]{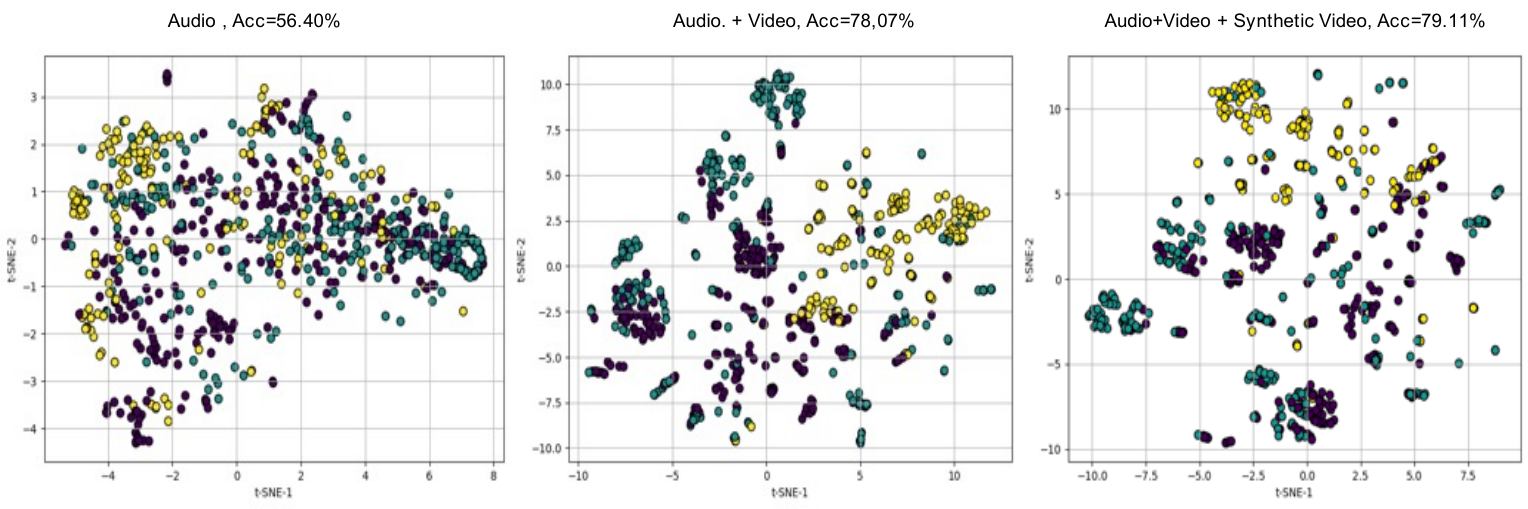}
\caption{Top: confusion matrices for the multimodal cross-attention model trained with Audio, Audio+Video, and Audio+Video+Synthetic video. Bottom: corresponding t-SNE plots of high-level features from the classification head. Yellow: Negative; Green: Neutral; Purple: Positive.}
\label{fig:cm_tsne_crossAtt}
\end{figure}

Based on the results, the audio and video modalities exhibit different behaviors in their capacity to distinguish emotion classes. Figure~\ref{fig:cm_tsne_crossAtt} shows that the Negative class becomes the most separable cluster after fusion, while Positive and Neutral remain partially overlapping.  When training the audio branch alone, the  Neutral class is predicted with the highest accuracy (60\%, as shown in the confusion matrix in Figure~\ref{fig:cm_tsne_crossAtt}); however, the corresponding t-SNE visualization reveals limited class separability. The yellow dots (negative class) are spread out almost everywhere, although there is a slight tendency to cluster toward the left. The same observation is made for the green dots (neutral class), with a small cluster toward the right. For the purple dots (positive class), there is no tendency to cluster. This suggests that, despite relatively accurate predictions, the learned audio embeddings do not form well-clustered representations across emotion classes.

In contrast, when combining audio with video inputs and further incorporating synthetic video data, the class-wise prediction behavior shifts. Similar to observations made with the sanity check, the inclusion of synthetic data improves classification performance for the  Negative class by 8\%, and for the  Neutral class by 3\%. However, this comes at a cost: performance on the  Positive class drops by 3\%. The t-SNE plot of fused embeddings indicates that  Positive and  Neutral classes are harder to separate, whereas  Negative instances form a more distinct and well-separated cluster. These observations confirm that different classes benefit unequally from synthetic augmentation and multimodal fusion. Specifically,  Negative emotions are more effectively captured through combined and augmented modalities, while  Neutral and  Positive classes exhibit overlapping distributions in the learned representation space.

\section{Conclusion}
In this chapter, a privacy-preserving multimodal transformer for GER is introduced. It fuses ViT-based video features with audio spectrogram features via cross-attention and Frames Attention Pooling (FAP). A targeted synthetic-video augmentation first validated with a sanity-check model yields consistent gains when integrated into the main architecture, with an optimal 30\% synthetic ratio. On VGAF, the approach attains 79.24\% validation accuracy and 75.13\% test accuracy, contributing to competitive results (first place)  at EmotiW 2023 \citep{emotiw2023} and supporting participation in ACII 2022 (Doctoral Consortium) \citep{augusma2022multimodal} and ICMI 2023 \citep{augusma2023multimodal}.

The study shows that video remains the dominant modality for group affect, while audio provides complementary cues best exploited through cross-attention. FAP improves frame aggregation, particularly for longer sequences over simple averaging, and ViT initialization (CLIP-based) supports robust transfer in the privacy-preserving setting.

The current limitations stem mainly from the design of synthetic data: the focus on the face with limited body cues and poor audiovisual alignment may limit multimodal gains and induce overfitting at high synthetic ratios (greater than 50\%). The next chapter addresses these gaps by enriching body-related information under the same privacy constraints, building on the foundations established here.

\chapter{Latent Space Optimization for Privacy-Preserving  Group Emotion Recognition}
\markboth{Latent Space Optimization for Privacy-Preserving GER}{} %

\label{chap:ve-md}

\section{Introduction}
\label{sec:ve-md-intro}

The previous chapter showed that targeted data augmentation using synthetic videos composed of emotion-expressive faces placed on diverse backgrounds improves performance while preserving privacy. However, this synthesis strategy omits body-related cues: the generated scenes contain faces and context but no explicit structural body information. As a result, the model becomes highly sensitive to facial evidence yet remains blind to full-body signals that often convey collective affect.

To address this limitation, this chapter proposes to learn a \textit{shared latent space} that jointly captures structural body and face representation, and scene context from full images or video frames, without relying on any individual inputs. The key idea is to optimize the latent embedding through auxiliary tasks that reconstruct human-centric structures, specifically body pose and facial landmarks, so that the latent representation encodes rich, discriminative cues relevant for emotion recognition while remaining privacy-compliant. The implementation of the proposed framework is publicly available.\footnote{\url{https://github.com/augusmaa/VE_MD_2025}}

First, a complete Variational Auto-Encoder (VAE) is impletemented with two decoders: a pixel-level reconstruction head and an emotion-classification head. Across datasets, adding the reconstruction branch consistently reduced classification accuracy, indicating that pixel-wise reconstruction encourages the model to encode background textures and other irrelevant details that compete with group-affective information. We therefore removed the reconstruction decoder and instead introduced \textit{structural decoders} that predict body and facial configurations. These structural tasks effectively guide the latent space toward person-related, semantically meaningful features that enhance stability and discriminability. The resulting model learns to (i) ingest uncropped full frames and internally infer structural body and face representations, and (ii) optimize a shared latent space jointly for emotion classification and these auxiliary structural predictions. Initially, the emotion decoder relied solely on the latent representation to ensure full privacy compliance. Experiments show that this configuration performs well on individual-level (non-GER) datasets but less effectively on group-level (GER) datasets, where relational body cues are crucial. The architecture is then extended by allowing the emotion decoder to incorporate the structural representations explicitly. This modification leads to a striking observation: the structural cues contribute much more strongly to performance on GER datasets than on non-GER datasets, suggesting that body–face geometry carries distinctive discriminative information for collective affect modeling.

The following sections provide an overview of the experimental datasets, the formulation of the Variational Encoder–Multi-Decoder (VE-MD) framework, the architectural design of the structural decoders, and the training methodology used to optimize the shared latent space.

\section{Datasets and Annotation}
\label{sec:datasets_annotation}

\begin{figure}[H]
  \centering
  \includegraphics[width=0.95\linewidth]{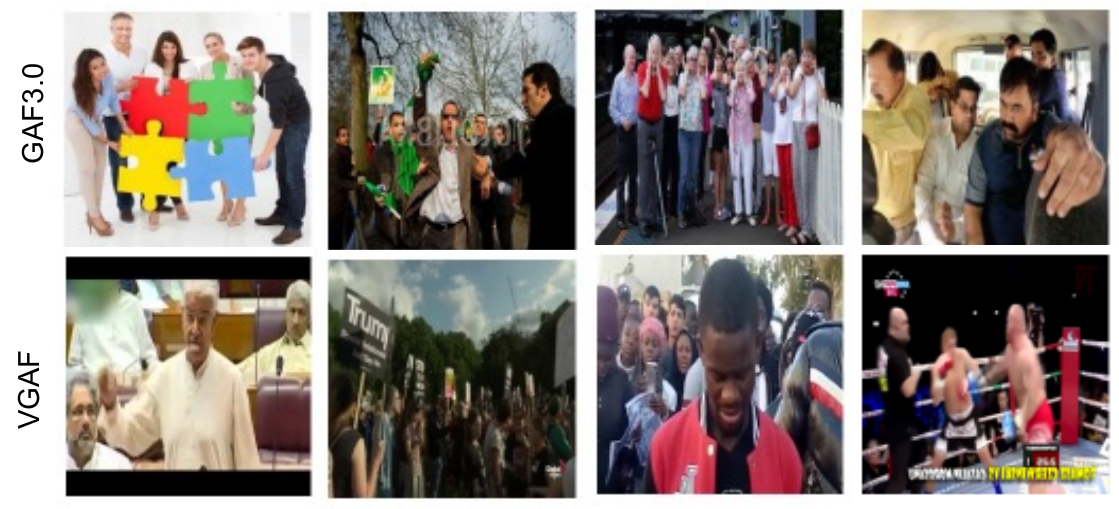}
  \includegraphics[width=0.95\linewidth]{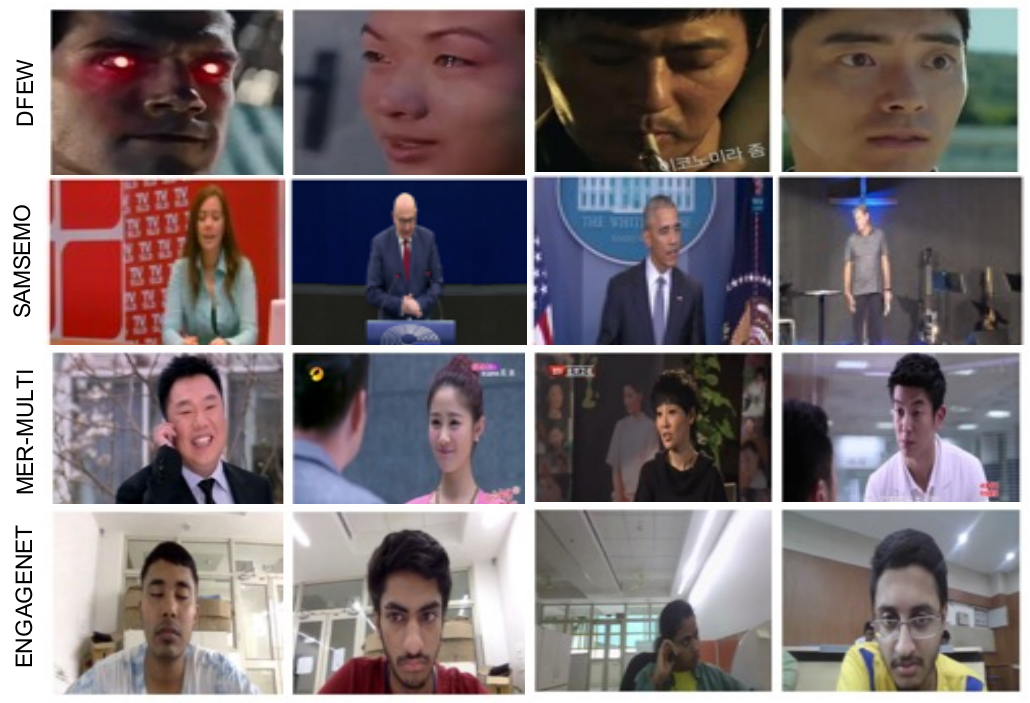}
  \caption{Overview of datasets used for experiments in this chapter. 
  The first two rows show the Group Emotion Recognition (GER) datasets (GAF-3.0 and VGAF). 
  The following rows display examples from individual-level or non-GER datasets: DFEW, SAMSEMO, MER-MULTI, and EngageNet.}
  \label{fig:datasets_used_overview}
\end{figure}

As already introduced in Section~\ref{sec:datasets_used}, this chapter employs a diverse set of \textit{in-the-wild} datasets to comprehensively evaluate the proposed models. 
Figure~\ref{fig:datasets_used_overview} illustrates how these datasets differ in modality, setting, and group composition.

\subsection{Group Emotion Recognition (GER) Datasets.}
The GAF-3.0~\citep{gupta2018attention} and VGAF~\citep{dhall2020emotiw} datasets serve as the main benchmarks for group-level affect recognition. 
GAF-3.0 contains still images labeled into three affective categories (Positive, Neutral, Negative), depicting diverse social contexts such as meetings, sports events, and protests. 
The Video-based Group Affect in-the-Wild (VGAF) dataset comprises YouTube videos capturing group interactions across varied demographics and situations, from interviews to open-air gatherings. 
Each clip is annotated by consensus voting into the same three emotion categories. 
VGAF presents particular challenges due to its variable resolution, dynamic camera motion, and complex acoustic environments.

\subsection{Individual-Level or Non-GER Datasets.}
To assess the generalization of the proposed framework beyond group emotion recognition, additional datasets focusing on individual-level or small-group emotion understanding are employed. 
SAMSEMO~\citep{bujnowski2024samsemo} is a recent multimodal corpus including manually transcribed and annotated scenes in five languages (English, German, Spanish, Polish, and Korean). 
It covers diverse domains such as lectures, interviews, and parliamentary recordings, and includes text, audio, and visual modalities annotated for six Ekman emotions plus \emph{Neutral} and \emph{Other}. 
MER-MULTI (MER-2023)~\citep{lian2023mer} consists of multimodal clips from movies and TV series, annotated with both discrete emotions (Anger, Sadness, Worry, Neutral, Happiness, Surprise) and continuous valence measures, making it a challenging benchmark for robustness to noise and blur. 
The Dynamic Facial Expression in-the-Wild (DFEW) dataset~\citep{jiang2020dfew} focuses on facial expression recognition in unconstrained conditions, providing rich annotation distributions for the seven Ekman emotions. 
Finally, EngageNet~\citep{emotiw2023}, introduced during the EmotiW 2023 challenge, targets engagement recognition in real-world educational settings, with four engagement levels: \emph{Highly Engaged}, \emph{Engaged}, \emph{Barely Engaged}, and \emph{Not Engaged}.

This diverse dataset selection ensures that the proposed framework is rigorously evaluated across a wide range of affective contexts from controlled, single-subject scenarios to dynamic, multi-person group interactions. It also enables systematic comparison between Group Emotion Recognition (GER) and Non-GER settings, which is essential for analyzing how structural representations contribute differently across these two domains. These datasets provide the experimental foundation for the following sections, where we progressively develop and evaluate our latent-space architectures.

\subsection{Structural Representation Annotation.}

The datasets used in this research originally provided annotations exclusively for emotion recognition tasks. To effectively support our multitasking approach, we performed automatic annotation for body and face structural representations. Specifically, body pose annotation was generated using ViTPose~\citep{xu2022vitpose}, while facial landmark annotation was performed using FaceAlignment~\citep{bulat2017far}.

\begin{figure}[H]
  \centering
  \includegraphics[width=0.8\textwidth]{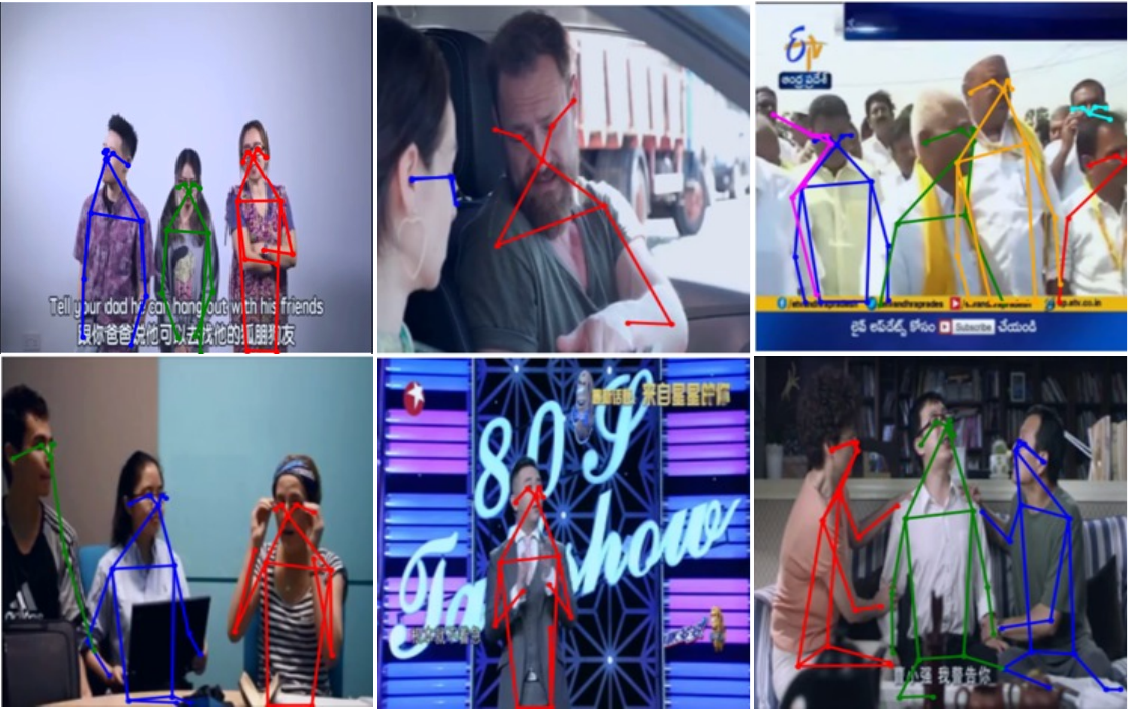}
 \caption{VitPose annotation Structural Representation for body. There are 18 limb connections based on the COCO style.}
  \label{fig:vit_annotation}
\end{figure}

\paragraph{Body Structural Representation:} For the body structural representation annotations, we followed the standard COCO format, constructing 18 limb connections from 17 keypoints. An example annotation generated by ViTPose is illustrated in Figure~\ref{fig:vit_annotation}.

\begin{figure}[H]
  \centering
  \includegraphics[width=0.8\linewidth]{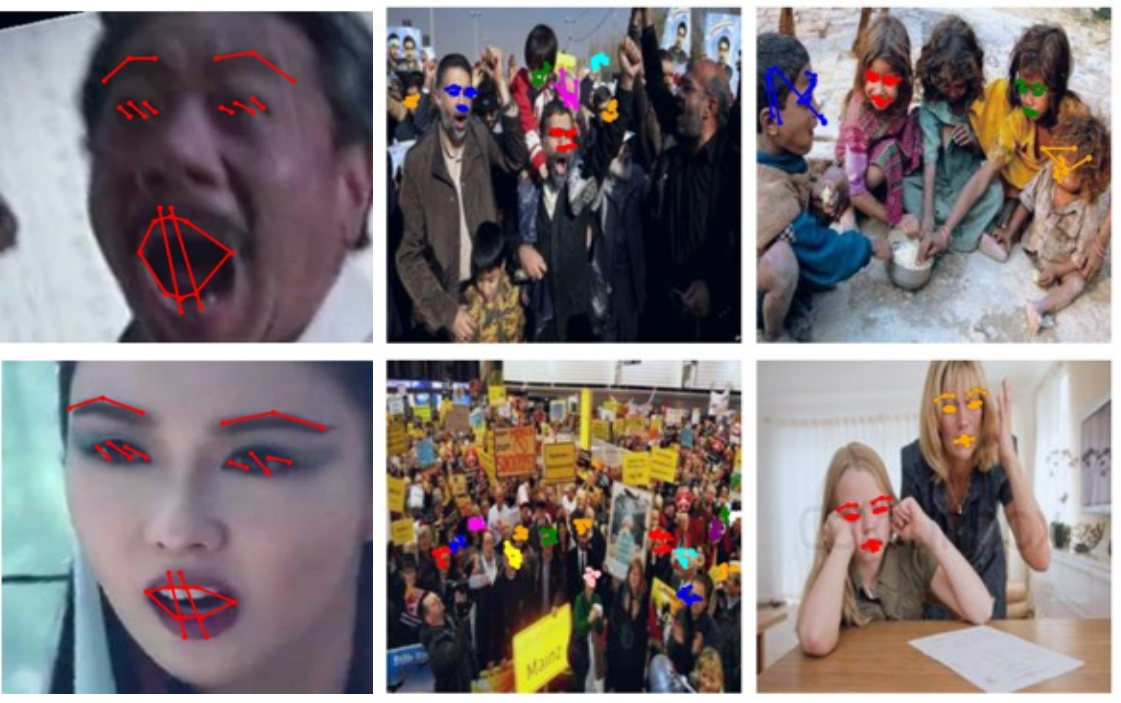}
 \caption{Custom Face annotation landmark with FaceAlignment model: They are annotated with 20 custom limb connections. }
  \label{fig:faceAlign_annotation}
\end{figure}

\paragraph{Face Structural Representation:}
Facial landmark annotations required customization to enhance their relevance to emotion recognition. Rather than employing all 68 standard facial keypoints, we created a reduced set focused on areas exhibiting significant movement, particularly those affected by speech or emotional expressions. For the DETR-based approach, we reduced the original 63 connections (limbs) to 20, specifically focusing on dynamic facial regions including the lips, eye entries, eyebrows, and mouth.
Figure~\ref{fig:faceAlign_annotation} illustrates some of the annotations produced, highlighting limitations in detection accuracy, particularly in crowded scenes.

Annotations were applied exclusively to the training sets of each dataset, since the validation of emotion recognition does not require explicit structural representation annotations. The purpose of these annotations is to enrich the model’s learning process by providing additional structural cues that enhance the representation of emotionally relevant features. However, as the datasets contain in-the-wild imagery and the annotation models are not perfect, some annotation errors and missing detections reduced the amount of usable training data. Specifically, frames in which \textit{ViTPose} or \textit{FaceAlignment} failed to detect any persons or faces were excluded from the training set. Table~\ref{tab:trainset_annotation} summarizes the final training data availability after annotation.

\begin{table}[H]
\footnotesize
    \centering
    \caption{Trainset details after Pose and Landmark annotation. }

    \label{tab:trainset_annotation}
    \begin{tabular}{|l|c|ccc|ccc|}

        \hline
        
         \textbf{Train Set} & &\multicolumn{3}{c|}{\textbf{Pose (ViTPose)}} & \multicolumn{3}{c|}{\textbf{Landmark (FaceAlignment)}}  \\ 
        \cline{3-6}
       
                        & Original Size & New Size   &AvgPers  &MaxPers   &New Size  &AvgPers &MaxPers \\
        
        \hline
        
         GAF-3.0    &9815           &9761  &6   &56      &{9558}  &6    &{104} \\ 
         VGAF       &2661           &2661  &8  &{29}     &{2636}   &5  &{71} \\ 
         SamSemo    &9822           &9821  &1    &{18}     &{9762} &1    &{14} \\ 
         MER2023    &3373           &3368  &1   &{11}     &{3272}  &1   &{16} \\ 
         DFEW       &9356           & -    &-  &{-}        &{9353}  &1   &{1} \\ 
         EngageNet  &7879           &{7879} &1   &{5}     &{7817}   &1  &{1} \\ 

        \hline
        
    \end{tabular}
\end{table}

\noindent For body-structure annotations using \textit{ViTPose}, the losses were minimal: only $0.55\%$ of images were removed from GAF-3.0, with a maximum of 56 persons detected. The VGAF dataset contained a maximum of 29 detected persons, SamSemo up to 18, and MER2023 up to 11, with a loss of only $0.15\%$ of videos. For facial-landmark annotations using \textit{FaceAlignment}, the data loss was slightly higher. In GAF-3.0, $2.61\%$ of videos were removed, with up to 104 detected faces. VGAF lost $0.95\%$ of videos (maximum 71 faces), SamSemo $0.61\%$ (maximum 14 faces), MER2023 $3.0\%$ (maximum 16 faces), and EngageNet $0.78\%$ (maximum 1 face). 

Overall, the annotation process significantly improved the suitability of the datasets for multitask learning by providing richer structural supervision. Nevertheless, limitations in detection accuracy, particularly in crowded or low-quality scenes, resulted in a slight reduction of available training data, especially for datasets with higher visual complexity.

\subsection{Video Frames Selection.}
The video frames selection is made based on the video duration of each dataset the corresponding training set. As a reminder, the duration video in the VGAF dataset is 5 seconds (fixed duration) for each video. We decided to keep 5 frames per video with one frame per second because in the previous experimentation (See chapter~\ref{chap:mger}), the alignment with audio with 5 frames audio and videos provided the best performance in the multimodal approach.  The EngageNet dataset has a fixed duration. One frame (image) per second is extracted to 10 frames (images) per video. The duration of videos in the case of MER2023 (MER-MULTI) and the SAMSEMO dataset is not fixed, so we decided to select 10 frames uniformly in each video. 16 frames (images) are used as input for the DEFEW faces dataset, as given in the dataset.

\section{Variational Auto-Encoder (VAE)}
\label{sec:ve-md-vae}

A Variational Autoencoder (VAE) is a generative model combining principles from neural networks and Bayesian inference to learn latent representations of data. Unlike traditional autoencoders, VAE assume that the latent variables follow a predefined probability distribution, typically a Gaussian. This probabilistic framework enables VAE to generate novel data points by sampling from the latent distribution. The model comprises an encoder, which maps input data into a latent distribution characterized by parameters (mean and variance), and a decoder, which reconstructs data from latent samples. The intuition behind the use of these models lies in projecting input data into a reduced dimension, followed by its reconstruction through probabilistic estimation. Mathematically, this situation can be described as follows:
Let \( X \) be an input (observed) data, \( L \) be a latent space generated from a prior distribution \( p_{\theta}(L) \), and \( X \) be generated by a conditional distribution \( p_{\theta}(X|L) \). Then the generative model learns a joint distribution $p_{\theta}(X, L)=p_{\theta}(L)p_{\theta}(X|L) $, and $\theta$  is a set of parameters to estimate as described in the study by~\citep{kingma2013auto}.\\

The VAE extends to (Dynamic Variational Auto-encoder) DVAE~\citep{girin2020dynamical, sadok2024multimodal}, for input data with time or sequence. For an input  $X_{1:T}=(x_{t_{1}}, x_{t_{1}}, ... , x_{t_{T}})$ and a latent space $L_{1:T}=(l_{t_{1}}, l_{t_{1}}, ... , l_{t_{T}})$, the generative distribution can be written as: $$ p_{\theta}(X_{1:T}, L_{1:T})= \prod_{t=1}^T p_{\theta_{L}}(L_{1:T}| \textbf{b}_{t})p_{\theta_{X}}(X_{1:T}|L_{1:T},\textbf{b}_{t}) $$ 
where $\textbf{b}_{t} = \{X_{1:T}, L_{1:T}  \} $ is a set of past observed data and latent vectors at time $t$ as explained in the study of~\citet{girin2020dynamical}.

\section{Multitask Learning for Emotion Recognition}
\label{sec:ve-md-mtl}

The approach of multitask learning has been used a lot in  Emotion Recognition (ER). \citet{foggia2023multi} explored multitask learning for facial analysis, using a shared representation to predict multiple attributes, including emotion, gender, age, and ethnicity. Their approach involved hard parameter sharing, where early network layers learned a common feature representation before branching into specialized classifiers for each task. 
\citet{hu2018deep} introduced a deep multitask learning framework for recognizing subtle facial expressions, leveraging facial landmark detection as an auxiliary task. Their model used a soft parameter sharing strategy, where layers automatically learned the degree of sharing through tensor trace norm regularization. This approach ensured an optimal balance between task-specific and shared representations, improving emotion classification. Additionally, they implemented adversarial domain alignment to mitigate dataset distribution shifts, enabling effective multitask learning across disjoint datasets. 

\citet{yin2017multi} proposed a multitask convolutional neural network for pose-invariant face recognition, where identity classification was the main task and pose, illumination, and expression estimation were auxiliary tasks (prediction with a fully connected layer in parallel). Their approach leveraged a dynamic-weighting scheme to automatically balance task contributions, ensuring effective feature disentanglement. They also introduced a pose-directed multitask CNN, which grouped faces by pose to learn pose-specific identity features, further improving robustness in face recognition.
\citet{pons2020multitask} introduced a multitask, multilabel, and multidomain learning approach for emotion recognition, leveraging facial action unit (AU) detection as an auxiliary task. Their model utilized a shared convolutional neural network (CNN) backbone, where a selective joint multitask loss (SJMT) was introduced to optimize tasks with heterogeneous labels. This approach improved emotion classification accuracy by incorporating AU knowledge, demonstrating that recognizing collective muscle movements enhances emotion recognition performance. 

\citet{ranjan2017hyperface} introduced HyperFace, a deep multitask learning framework that simultaneously performs face detection, facial landmark localization, head pose estimation, and gender recognition using a shared convolutional neural network (CNN) backbone. Their method fuses intermediate feature layers within the network, exploiting task synergy to boost performance across all four tasks. They also proposed HyperFace-ResNet, which leverages ResNet-101 to further improve accuracy. The framework uses a multi-loss optimization strategy, ensuring efficient learning of correlated facial attributes. \citet{hong2018multimodal}   introduced multitask Manifold Deep Learning (MDL) for face pose estimation, integrating multitask learning (MTL) and manifold regularization within a deep convolutional neural network (DCNN). By leveraging multi-modal data, the approach jointly learns feature mapping and pose estimation while enforcing task-related constraints to improve prediction accuracy.

Our approach here extends multitask learning by integrating a shared latent space for emotion classification, body, and face structural representation estimation. Rather than using explicit individual features like facial landmarks or body pose as direct inputs, we implicitly learn and predict them within a unified representation. This design ensures a privacy-preserving framework that avoids direct individual control or monitoring.

\section{Methodology and Model Architecture}

\begin{figure}[H]
  \centering
  \includegraphics[width=\linewidth]{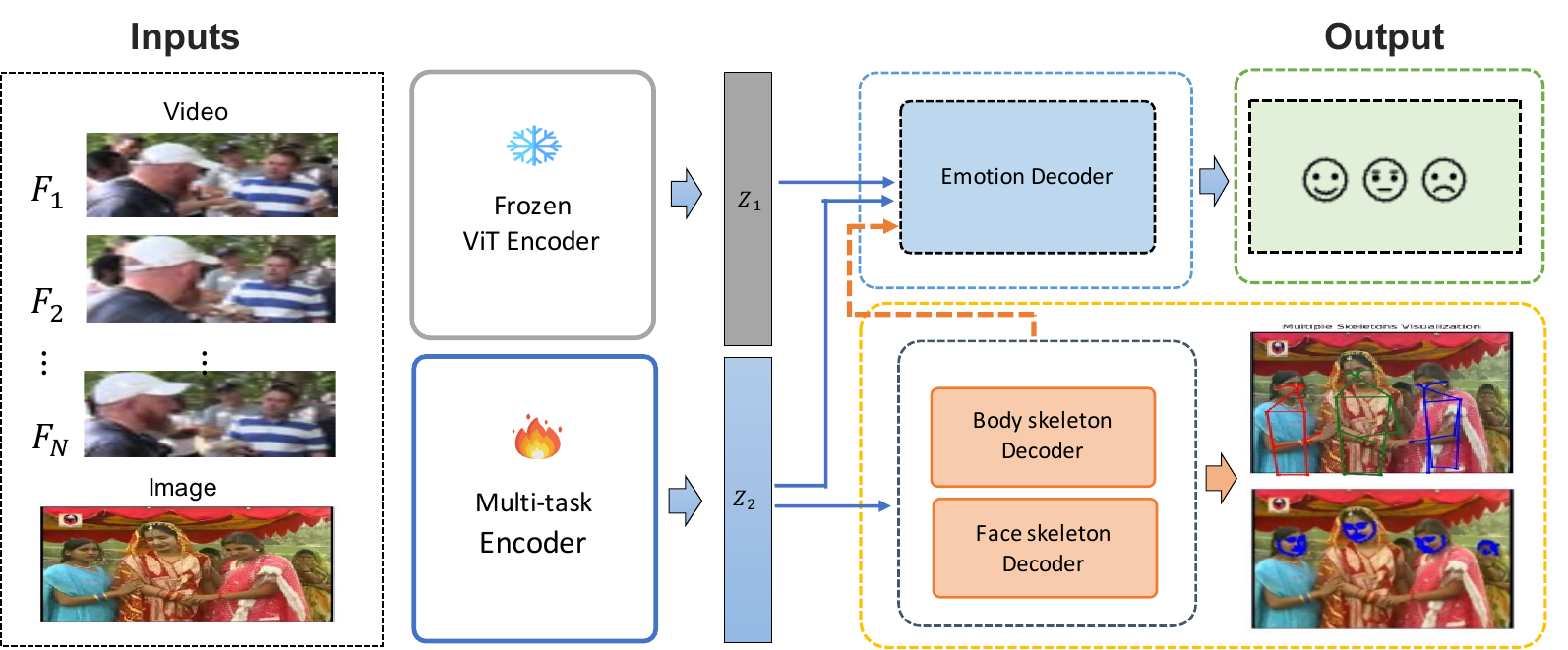}
 \caption{The proposed VE-MD 
 architecture using a multitask latent space. The left block represents the input data, which may include videos (sequences of frames) or a single image. The middle block displays encoders and corresponding latent spaces. Featuring two encoders: a multitask encoder and a frozen encoder (ViT). The right block displays the multi-decoder component includes 1) an Emotion Decoder that receives: the two latent spaces, and potentially the output structural representation from the body and face.}
  \label{fig:ve_md}
\end{figure}

\subsection{Shared Latent Space with Multi-Decoders}

The proposed VE-MD architecture (Figure~\ref{fig:ve_md}) for group emotion recognition in-the-wild is with shared latent-space multitask learning, comprising two main parts. The encoder parts where there are a frozen encoder (ViT) and a multitask encoder.  And decoder parts: Emotion decoder and Structural Representation decoder (Body and Face).

\subsection{Encoders}
\label{sec:encoders}
The proposed architecture is designed as a Variational Encoder (VE), with two encoder branches for feature extraction. 
The first branch is a  frozen ViT-Large encoder \citep{dosovitskiy2020vit}, pretrained on an emotion recognition task for the corresponding dataset. It remains frozen throughout training and serves as a specialized feature extractor for the corresponding task. The second branch is a trainable multitask encoder, responsible for learning representations for structural representation estimation (body and face) and Emotion. 

\begin{figure}[H]
  \centering
  \includegraphics[width=0.6\linewidth]{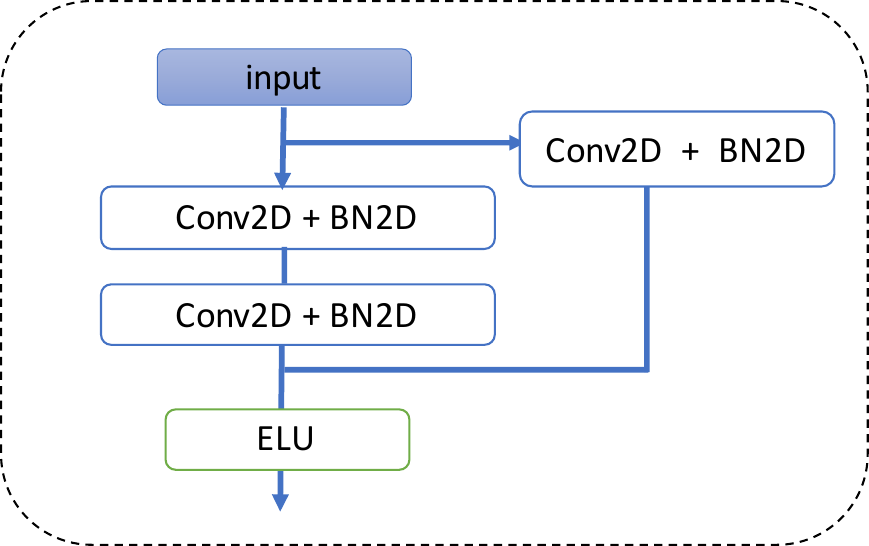}
 \caption{The CUSTom RESidual block. It performs downsampling of the input image to half its original size: Conv1, Conv2, Conv3 are 2D convolutional layers; BN1, BN2, BN3: means batch-normalization, and ELU-activation function at the end.}
  \label{fig:reblock}
\end{figure}

For the trainable multitask encoder, we use our CUSTom-RESidual encoder that has been used for the first experiments in the Variational Encoder (VE) built using residual blocks (see section~\ref{sec:vae_to_ve}), illustrated in Figure~\ref{fig:reblock}. One can employ any kind of well-known public architecture, such as ResNet50, ResNet101, VGG19,  etc. 

\subsection{From Variational Autoencoder to Variational Encoder}
\label{sec:vae_to_ve}
First, a VAE model is built to perform Emotion Recognition (ER) on video in an end-to-end manner. To do so, from the latent space, two decoders are added: One reconstruction decoder to reconstruct the video input and one classification decoder to predict emotion. For that experience, the VAE encoder is a stack of five residual blocks as defined in figure~\ref{fig:reblock}. The reconstruction decoder follows the mirror of the encoder by using Transpose convolution (ConvTranspose2D).
The latent space is fed as input. A linear projection is applied to reduce the latent space dimension to a suitable dimension before passing a transformer layer to take into account the time of video frames. A \textbf{Frames Attention Pooling} (FAP, see section~\ref{sec:fap}) is added to aggregate frames into one to pass an MLP for the classification task. 

Using that model, three kinds of experimentation were conducted. One with a reconstruction decoder (VAE), one without a reconstruction decoder by keeping the Gaussian distribution for the latent space (VE), and another one with a vanilla encoder (without reconstruction). The experimental results can be consulted in Table~\ref{tab:vae_ve_result} for five different datasets (described in section~\ref{sec:datasets_used}).  1)~VGAF: for Group Emotion Recognition in-the-wild, 2)~DFEW: for Facial Expression Recognition in-the-wild 3)~MER-MULTI (MER2023) and SAMSEMO for Emotion Recognition of single individual in-the-wild 4)~EngageNet for engagement recognition.

 \begin{table}[H]
 \centering
\caption{Comparison results between VAE,  VE, and Vanilla models. Bold indicates better accuracy (Acc.)  on the validation set.}
\label{tab:vae_ve_result}
\resizebox{\linewidth}{!}
{
   \begin{tabular}{lccc||c}
     \toprule
    Dataset  &Acc. VAE (\%) $\uparrow$ & Acc. VE (\%) $\uparrow$ & Acc. Vanilla (\%) $\uparrow$  & PSNR Recon.  $\uparrow$ \\
    \midrule
    VGAF & 48.05 & \textbf{51.54} &48.96  &14.65 \\
    MER-MULTI & 32.52 &{39.81} &\textbf{40.29} & 15.04 \\
    DFEW & 44.62 & \textbf{56.22} &53.92 &21.20 \\
    SAMSEMO & 57.19 & \textbf{62.68} &57.43 &12.78 \\
    EngageNET & 54.20 & \textbf{56.12} &55.32  &17.80 \\
   \bottomrule
 \end{tabular}
 }
 \end{table}

The results in Table~\ref{tab:vae_ve_result} show that the VE model consistently achieves higher accuracy compared to both the VAE and the Vanilla models. Except for MER-MULTI, the VE model yields at least a $3\%$ performance gain across all datasets. The \textit{PSNR Recon.} column reports the Peak Signal-to-Noise Ratio, which reflects the perceptual quality of the reconstructed images (or video frames). In the literature, a PSNR value around $20$~dB~\footnote{dB: decibels} is generally considered as a high-quality reconstruction. Based on our observations with these in-the-wild datasets, a PSNR of approximately $15$~dB already corresponds to visually acceptable reconstruction quality.

 From these results and extensive experimentation, we conclude that a multi-decoder design without the reconstruction branch while maintaining a Gaussian latent space is preferable. This configuration defines our \textit{Variational Encoder (VE)} model and shifts the learning focus from low-level pixel reconstruction toward high-level, semantically meaningful cues that are more relevant for emotion classification.

After establishing that the Variational Encoder (VE) model outperforms traditional autoencoder designs by focusing on semantic rather than pixel-level reconstruction, the next step is to extend this formulation toward a multitask configuration. To this end, we propose the \textit{Variational Encoder Multi-Decoder (VE-MD)} framework, which introduces auxiliary structural prediction tasks to enrich the latent space. The following section details the overall architecture, highlighting how multiple decoders jointly shape a shared, emotion-aware latent representation.

\subsection{Multi-Decoders}

The overall architecture consists of three main decoders: a decoder for body structural representation estimation, a decoder for facial structural representation estimation, and an emotion decoder for emotion recognition. The two structural representation decoders receive as inputs the latent space from the trainable encoder. All latent spaces are used to feed the emotion decoder as input. 

Two innovative approaches are proposed for the structural representation decoder: \textbf{Modified PETR Architecture} \citep{shi2022end}  (structural representation-based DETR) and structural representation \textbf{Based Heatmap}.  These two propositions are detailed in the next sections.

\section{VE-MD with DETR-Based Decoder}
\label{sec:ve-md-detr}

Building upon these principles, we adapt the DETR~\citep{carion2020end} framework to predict human structural representations rather than object bounding boxes. The adaptation is referred to as the \textit{Modified PETR Architecture}. The resulting design, inspired from the PETR model \citep{shi2022end}, modifies DETR’s query formulation and output heads to estimate spatial limb connections and adjacency relations between body or facial components. 

\subsection{Modified PETR Architecture (Structural Representation based on DETR)}

\begin{figure}[H]
  \centering
  \includegraphics[width=0.9\linewidth]{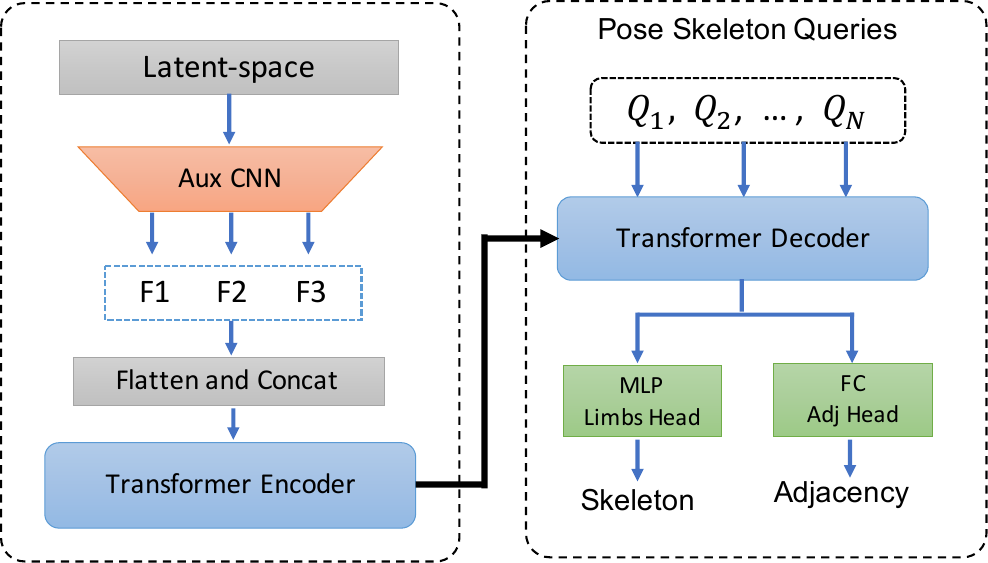}
 \caption{At left, the network takes the latent space as input. We then apply an auxiliary convolutional module that outputs three distinct scale feature tensors, \(\mathbf{F}_1, \mathbf{F}_2, \mathbf{F}_3\). Flatten and pass to the transformer encoder. At right, the transformer decoder received target queries combined with encoded features to predict the structural representation and adjacency matrix via respectively MLP head and FC head.}
  \label{fig:structural representation_decoder}
\end{figure}

To perform the structural representation decoder, the proposed architecture follows the  PETR \citep{shi2022end} style based on  DETR \citep{carion2020end} for fully end-to-end pose prediction. Instead of using the same architecture of PETR, the proposed architecture is modified and adapted to our style. As depicted in Figure~\ref{fig:structural representation_decoder}, the proposed structural representation decoder builds upon a transformer-based architecture to predict limb endpoints and the adjacency matrix among limbs simultaneously.  The network takes the latent space (feature map $\mathbf{F}$) from the above variational encoder (VE) as input. Then, it apply an auxiliary convolutional module (referred to as \(\textit{AuxiliaryConvolutions}\) which follows the residual convolution block defined in Figure~\ref{fig:reblock}) that outputs three distinct scale feature tensors, \(\mathbf{F}_1, \mathbf{F}_2, \mathbf{F}_3\). Each tensor is flattened and concatenated along the spatial dimension to form a single source sequence \(\mathbf{S}\in \mathbb{R}^{B \times S \times E}\), where \(S\) denotes the combined spatial length across scales, \(B\) is the batch size, and \(E\) is the latent embedding dimension.

\paragraph{Person-query and Transformer:} In contrast to PETR, we employ a learnable query embedding \(\mathbf{Q}\), parameterized by \(\texttt{num\_queries}\) vectors of dimension \(E\). Each query can be interpreted as a “prototype” that predicts a specific structural representation (full structural representation for one person)  or a set of joint relationships (a set of limbs). These queries, organized into a target sequence \(\mathbf{T}\in \mathbb{R}^{B \times Q \times E}\) (with \(Q = \texttt{num\_queries}\)), are passed into the Transformer decoder alongside the encoded source sequence \(\mathbf{S}\) from the transformer encoder. The number of queries should be specified for every training, and remains for the whole evaluation. Formally, the decoder refines each query representation \(\mathbf{Q}_i\) through multi-head self-attention operations:
\[
    \mathbf{T}_{\mathrm{out}} = 
    \mathrm{TransformerDecoder}(\mathbf{T}, \mathbf{S}),
\]
where \(\mathbf{T}_{\mathrm{out}}\in \mathbb{R}^{B \times Q \times E}\) are the final decoded features.

\paragraph{Heads for Limb and Adjacency Predictions:} In contrast to PETR,  the network comprises two prediction heads:
\begin{itemize}
    \item \textbf{Limb Head:} A multi-layer perceptron (MLP) ended with a sigmoid, outputs line-segment endpoints for each limb. Specifically, for \(\texttt{num\_limbs}\), we predict \(4 \times \texttt{num\_limbs}\) coordinates, denoted \([\mathrm{x}_1, \mathrm{y}_1, \mathrm{x}_2, \mathrm{y}_2, \ldots]\) for each query. This format encodes the 2D endpoints of each limb.
    \[
        \mathbf{L}_{\mathrm{pred}} = \mathrm{MLP}(\mathbf{T}_{\mathrm{out}}) \;\in\; \mathbb{R}^{Q \times 4\,\times\,\texttt{num\_limbs}}.
    \]
    \item \textbf{Adjacency Head:} A single linear layer ended with sigmoid,  outputs a flattened \(\texttt{num\_limbs}\times \texttt{num\_limbs}\) adjacency matrix for each query. This matrix specifies the connectivity scores between all pairs of limbs. Formally,
    \[
        \mathbf{A}_{\mathrm{pred}} = \mathrm{Linear}(\mathbf{T}_{\mathrm{out}}) \;\in\; \mathbb{R}^{Q \times \texttt{num\_limbs}^2}.
    \]
\end{itemize}

Hence, each query in the decoder simultaneously outputs both limb endpoints \(\mathbf{L}_{\mathrm{pred}}\) and the joint adjacency matrix \(\mathbf{A}_{\mathrm{pred}}\). The intuitive goal is for each query to specialize in decoding a consistent subset of structural representation connections or relationships. By predicting the adjacency matrix, the model can capture pairwise joint dependencies of limbs, enhancing the structural representation estimates' structure and offering additional interpretability~\citep{abedi2024engagement}.

In summary, the person's structural representation decoder takes a latent space from the proposed VE. It utilizes learnable queries to decode both (i) limb endpoint coordinates and (ii) joint-joint adjacency matrices. This unified framework allows the model to learn spatial configurations of body parts,  by hopping to lead to more robust and interpretable person structural representation.

\subsection{DETR-based Emotion Decoder}
\label{sec:emotion_decoder_detr}

The emotion decoder receives latent representations from the upstream encoders, optionally incorporates structural representation-based information. The two latent feature maps from the two encoders are linearly projected to a common latent dimension, processed by a lightweight Transformer encoder with positional encoding, summarized with Frames Attention Pooling (FAP, see section~\ref{sec:fap}), and finally classified with a multi-layer perceptron (MLP).

The emotion decoder fuses encoder features with optional body and face structural representation sequences derived from the detection transformer. structural representation dynamics are modeled using a spatio-temporal graph convolutional network (ST-GCN~\citep{abedi2024engagement}), as detailed in the next paragraph.

\paragraph{Spatio-Temporal Graph Convolution (ST-GCN).}
Let
$x \in \mathbb{R}^{T\times M\times C\times V}$ denote a sequence with $T$ frames, $M$ instances (e.g., persons), $V$ graph nodes (limbs), and $C$ feature channels per node. Let $A \in \mathbb{R}^{T\times M\times V\times V}$ denote the adjacency matrices describing the per-frame graph structure.  
For each $(t,m)$, we apply a graph convolution:
\[
\mathbf{X}^{\text{sp}}_{t,m} = \big( A_{t,m}\, \mathbf{X}_{t,m}^\top \big)^\top W
\quad\in\mathbb{R}^{V\times C_{\text{out}}},
\]
where $\mathbf{X}_{t,m}\in\mathbb{R}^{V\times C}$ stacks the node features, $A_{t,m}\in\mathbb{R}^{V\times V}$ encodes the adjacency, and $W\in\mathbb{R}^{C\times C_{\text{out}}}$ is a learned weight matrix.  

\noindent The resulting features are then processed with a temporal 3D convolution:
\[
\mathbf{X}^{\text{tmp}} = \mathrm{Conv3D}\big(\mathbf{X}^{\text{sp}}\big),
\quad \text{kernel}=(3,1,1),\ \text{padding}=(1,0,0),
\]
where $T$ is treated as temporal depth, $V$ as height, and $M$ as width. Batch normalization and ReLU activation are applied, and the output is reshaped back to $(B, T, M, C_{\text{out}}, V)$.  

\paragraph{Inputs:}  
For each clip, per-frame flattened features
\[
\mathbf{z}_1, \mathbf{z}_2 \in \mathbb{R}^{T\times (C_z \cdot h_z  \cdot w_z)}
\]
are obtained from the encoders, where $ (C_z, h_z,  w_z ) $ is the dimension of the latent space from each encoder, with $C_z$ being the latent space channel. Optionally, flattened structural representation ($\mathbf{S}^{\text{ body}}; \mathbf{S}^{\text{ face}}$)  sequences are included:
\[
\mathbf{S}^{\text{ body}} \in \mathbb{R}^{T\times (4 N_b Q)}, 
\quad
\mathbf{S}^{\text{face}} \in \mathbb{R}^{T\times (4 N_f Q)},
\]
where $N_b$ and $N_f$ denote the number of body and face limbs, respectively, and $Q$ is the number of detection queries.

\paragraph{Feature fusion:}  
Latent streams are concatenated and linearly projected:

\[
\tilde{\mathbf{z}}_t = \phi\!\left(\mathbf{C}_{\text{f}} [\mathbf{z}_{1,t}; \mathbf{z}_{2,t}\right)
\in \mathbb{R}^{C},
\]
with $C = 2C_z$, 
where $\mathbf{C}_{\text{f}} $ and $ \phi $ are respectively concatenation and linear functions, $C_z$ is the the channel dimension of the latent space ($C_z=latent\_dim$).  
If structural representations are used, two cases are considered: either the\textbf{ raw structural representations} are used or they are\textbf{ projected} via LayerNorm and Linear layers with ReLU.

\noindent If using raw structural representation, they are concatenated with $\tilde{\mathbf{z}}_t$, yielding
$$
\mathbf{x}_t = [\tilde{\mathbf{z}}_t; {\mathbf{S}}^{\text{ body}}_t\ (\text{if used}); \{\mathbf{S}^{\text{face}}_t\ (\text{if used})] \in \mathbb{R}^{D},
$$
with, 

$D=C\ (\text{no structural representation})$

$D= {C + 4QN_i}_{, i\in \{b, f\}} \ (\text{one structural representation})$

$D=C +  4Q(N_b + N_f)\ (\text{both}).$

\noindent If not using raw structural representation, they are projected as defined below:

\[
\hat{\mathbf{S}}^{\text{ body}}_t = \rho\!\left(\mathbf{S}^{\text{ body}}_t \right),\quad
 \hat{\mathbf{S}}^{\text{face}}_t = \rho\!\left(\mathbf{S}^{\text{face}}_t \right). 
\]
These are concatenated with $\tilde{\mathbf{z}}_t$, yielding

$$
\mathbf{x}_t = [\tilde{\mathbf{z}}_t; \hat{\mathbf{S}}^{\text{ body}}_t\ (\text{if used}); \hat{\mathbf{S}}^{\text{face}}_t\ (\text{if used})] \in \mathbb{R}^{D},
$$

$D=C\ (\text{no structural representation}),$

$D=C + C_S \ (\text{one structural representation}),$

$D=C + 2C_S \ (\text{both}).$

$\rho$ is the normalized linear layer projection, and $C_S$ is the projected linear dimension.

\paragraph{Temporal modeling and classification.}
To incorporate temporal order, sinusoidal positional encoding $\mathrm{PE}(t)$ is added:
\[
\mathbf{h}^{(0)}_t = \mathbf{x}_t + \mathrm{PE}(t).
\]
A Transformer encoder (1 layer, 1 head) then models temporal relations:
\[
\mathbf{H} = \mathrm{Transformer}\!\left(\{\mathbf{h}^{(0)}_t\}_{t=1}^T\right)\in \mathbb{R}^{T\times D}.
\]
Finally, Frames Attention Pooling (FAP) (see section~\ref{sec:fap}) produces a video-level embedding, which is classified by an MLP.

\begin{figure}[H]
  \centering
    \includegraphics[width=0.5\linewidth]{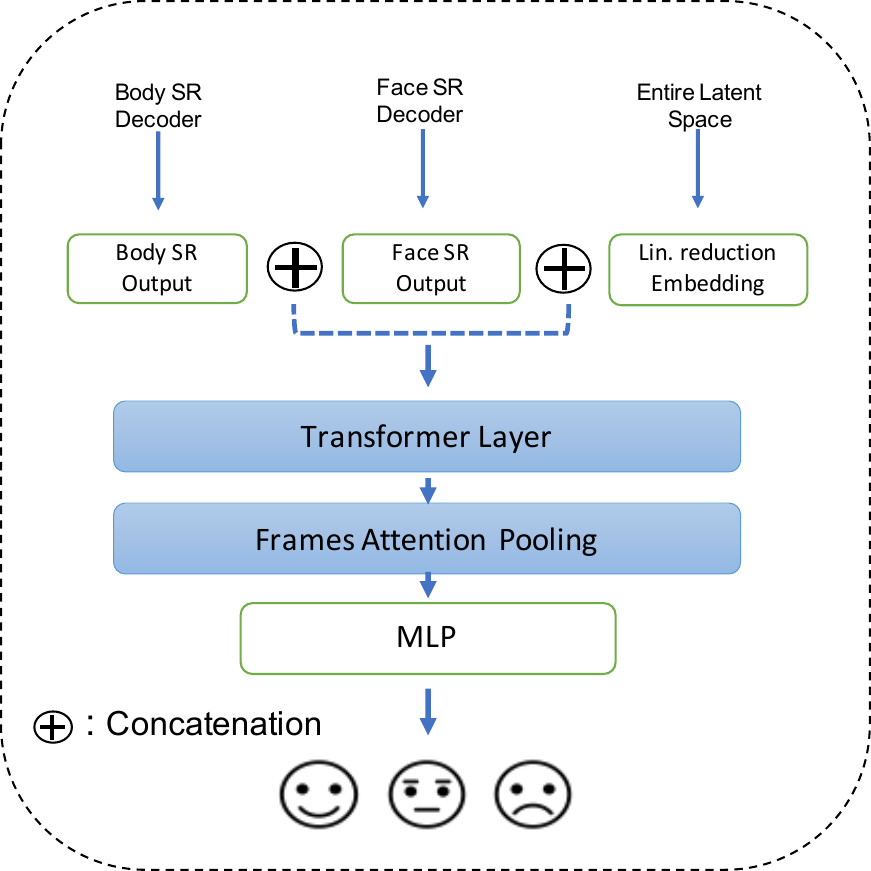}
   \caption{{Emotion Decoder. It receives the latent space as input and optionally the raw structural representation, where it is used, followed by one transformer layer over video frames. A self-learned Frames Attention Pooling (FAP) is used to reduce frames to one MLP layer for emotion classification. A detailed algorithm for the frames attention pooling is given in Section~\ref{sec:fap}.}}
  \label{fig:classification_grp}
\end{figure}

\subsection{Loss Functions}
\label{sec:ve-md-detr-loss}

The overall loss function for our model integrates multiple components, each tailored for specific tasks. (i) Emotion Classification loss, which is a standard cross-entropy loss \(\mathcal{L}_{\mathrm{cls}}\) utilized for emotion classification. (ii) Structural Representation loss, defined for body and face separately. (iii) Maximum Mean Discrepancy (MMD~\cite{dziugaite2015training}) loss which is the regularization loss defined in Appendix~\ref{sec:MMD} to ensure robust feature representation.

\noindent In the case of the DETR decoder, the Structural Representation (SR) loss combines two distinct components: a limb connection loss using Smooth L1 and an adjacency matching loss utilizing binary cross-entropy. A Hungarian matching algorithm enforces unique, optimal matching between predicted limbs and ground truth annotations. The total loss is expressed as:

\begin{equation}
\label{eq:total_loss_detr}
\mathcal{L}_{\text{total}} = \mathcal{L}_{\mathrm{cls}} + \beta_{\mathrm{p_{1}}}\mathcal{L}_{\mathrm{p_{1}}} + \beta_{\mathrm{p_{2}}}\mathcal{L}_{\mathrm{p_{2}}} + \beta_{\mathrm{mmd}}\mathcal{L}_{\mathrm{MMD}}
\end{equation}

where the  -specific losses \(\mathcal{L}_{\mathrm{p_{i}}}\), for body structural representation \((p_{1})\) and face structural representation \((p_{2})\), are defined as:

\begin{equation}
\mathcal{L}_{\mathrm{p_{i}}} = \beta_{\mathrm{limb}}\mathcal{L}_{\mathrm{limb}} + \beta_{\mathrm{adj}}\mathcal{L}_{\mathrm{adj}}, \quad i \in \{1, 2\}
\end{equation}

with \(\beta_{\mathrm{p_{i}}}\), \(\beta_{\mathrm{limb}}\), \(\beta_{\mathrm{adj}}\), and \(\beta_{\mathrm{mmd}}\) representing loss weighting factors.

\paragraph{Masked Coordinate Processing:}

To handle invalid structural representation annotations effectively, we introduce a masking mechanism~\citep{cao2019openpose}. Given a set of coordinates \(\mathbf{c} \in \mathbb{R}^{D}\) potentially containing invalid values marked explicitly (e.g., \(-1\)), we define a binary mask \(\mathbf{m} \in \{0,1\}^{D}\) and clean coordinates \(\mathbf{c}^{\text{clean}}\) as:

\begin{equation}
m_i =
\begin{cases}
1, & c_i \neq \text{mask\_value} \\
0, & c_i = \text{mask\_value}
\end{cases},
\quad c^{\text{clean}}_i = c_i \cdot m_i
\end{equation}

\paragraph{Masked Smooth L1 Loss:}
The Smooth L1 loss is computed exclusively over valid coordinates, ensuring invalid or missing keypoints do not impact model training:

\begin{equation}
\mathcal{L}_{\text{masked}} = \text{SmoothL1}(\mathbf{g} \cdot \mathbf{m}, \mathbf{p})
\end{equation}

where \(\mathbf{g}\) represents ground truth and \(\mathbf{p}\) predicted coordinates.

\subsection{Training with DETR Decoder}

The model receives as input either a sequence of video frames or a single image in the case of image-based datasets. All inputs are resized to $224 \times 224$. From the encoders, we obtain latent feature maps of size $latent\_dimension \times 7 \times 7$. As described earlier (see Section~\ref{sec:encoders}), the architecture employs two encoder branches for feature extraction: a frozen ViT encoder and a trainable multi-task encoder.

For the multi-task branch, two encoder variants are investigated: a ResNet-50 and a CUSTtom RESidual encoder. Both produce a final feature map of size $4096 \times 7 \times 7$, which is reduced to $512 \times 7 \times 7$ ($latent\_dimension = 512$) using a $2$D convolution layer. The same dimensionality reduction is applied to the ViT encoder to ensure consistent feature size. Consequently, the final latent space has dimensions $1024 \times 7 \times 7$, split into two equal parts, one from the frozen ViT branch and one from the multi-task branch. Only the multi-task branch is fed to the Structural Representation decoder, ensuring that structural prediction is learned jointly with emotion classification. During training, the ViT encoder remains frozen, as it was pre-trained for emotion recognition on the corresponding dataset. In contrast, the ResNet-50 encoder is fine-tuned jointly for emotion recognition and Structural Representation prediction, making it a true multi-task component of the architecture.

When using the DETR-based decoder, three transformer layers are used, each with eight attention heads, and employ sinusoidal positional encoding for sequence representation. The number of queries $Q_i$ determines the number of persons the model can detect, with $Q_i \in \{Q_{\text{Max}}, 50, 100\}$, where $Q_{\text{Max}}$ corresponds to the maximum number of annotated persons in the training dataset. The loss function weights are set as follows: $\beta_{\text{limb}} = 1.0$ and $\beta_{\text{adj}} = 0.5$, giving higher importance to limb detection over adjacency refinement. Since emotion classification is the primary objective, the structural decoder weights are kept lower, with $\beta_{p_i} = 0.1$ and $\beta_{\text{mmd}} = 0.1$ (see Equation~\ref{eq:total_loss_detr}). The AdamW optimizer is used with a learning rate of $10^{-7}$ and a weight decay of $10^{-4}$.

\subsubsection{Effect of Transformer-Based Encoding (ViT) on VE-MD Latent Space}
\label{sec:vit_influence}

As in our previous approach~\ref{chap:mger}, to further improve performance, a Vision Transformer (ViT) branch was integrated into the VE-MD framework. The results presented earlier in Table~\ref{tab:vae_ve_result} demonstrated the advantage of the end-to-end Variational Encoder (VE) design over the classical VAE. Building on this finding, the ViT encoder was combined with two distinct convolutional encoders, ResNet50 and our  CUSTom RESidual network, to form two hybrid configurations: \textit{ViTVE\_R50} and \textit{ViTVE\_CustRes}. The number of parameters for these two architectures is respectively $488,487,014$ and $427,910,214$. The ViT branch remained frozen during training, serving as a global semantic feature extractor, while the secondary encoder (ResNet50 or  CUSTom RESidual) was fine-tuned for emotion classification and structural representation learning.

 Table~\ref{tab:vit_ve_results} reports the validation accuracy obtained for the ViT-only model and the two ViT–VE hybrid configurations. In all cases, integrating the frozen ViT branch into the VE model yields a consistent improvement in accuracy. The combination with ResNet50 (\textit{ViTVE\_R50}) performs best on most datasets, achieving the highest results on GAF-3.0 ($83.56\%$), VGAF ($76.63\%$), and SAMSEMO ($72.26\%$). The improvement compared to ViT-only is particularly notable on MER-MULTI (+6.09\%) and EngageNet (+3.96\%).

\begin{table}[H]
\centering
\caption{Accuracy performance for VE and ViT combined with VE. \textit{ViTVE\_R50}: combination of ViT and ResNet50; \textit{ViTVE\_CustRes}: combination of ViT and the  CUSTom RESidual encoder.}
\label{tab:vit_ve_results} 
\begin{tabular}{l|ccc}
\toprule
\textbf{Dataset} & \textbf{ViT} & \textbf{ViTVE\_R50} & \textbf{ViTVE\_CustRes} \\
\midrule
GAF-3.0 & 82.27 & \textbf{83.56} & 82.41 \\
VGAF & 76.11 & \textbf{76.63} & 75.72 \\
VGAF + Synthetic data & \textit{78.07} & - & - \\
MER-MULTI & 54.74 & \textbf{60.83} & 59.22 \\
DFEW & 64.88 & 64.69 & \textbf{68.53} \\
SAMSEMO & 55.38 & \textbf{72.26} & 66.02 \\
EngageNet & 64.14 & \textbf{68.10} & 67.82 \\
\bottomrule
\end{tabular}
\end{table}

\noindent On the other hand, the combination with the  CUSTom RESidual encoder (\textit{ViTVE\_CustRes}) achieves the strongest results on DFEW ($68.53\%$), outperforming both ViT and \textit{ViTVE\_R50}. This indicates that the benefit of each encoder combination can vary depending on the nature of the dataset: while ResNet50 captures more robust global body–context interactions in group-level data, the  CUSTom RESidual encoder generalizes better to close-up facial or individual-level emotion datasets. Although the relative gains differ by dataset, both hybrid models significantly outperform the ViT baseline. The comparison with the synthetic-data-enhanced VGAF setup also highlights the complementary nature of data augmentation and hybrid feature learning.

 Integrating a frozen ViT branch within the VE-MD architecture consistently improves performance by introducing global contextual information and semantic priors into the latent space. The ViT encoder complements the convolutional branches of ResNet50 and the  CUSTom RESidual network by modeling long-range dependencies that traditional CNNs may overlook. These findings validate the design choice of maintaining a frozen ViT as a global encoder in subsequent experiments, while the multi-task VE branch focuses on refining body–face structural representations for emotion understanding. 

 Having established that the integration of a frozen ViT branch enhances the latent space by providing richer semantic and contextual features, the next step is to evaluate whether this optimized latent representation alone is sufficient for emotion recognition. In other words, we investigate whether the model can maintain strong performance when the emotion decoder relies exclusively on the latent embeddings without explicitly using the predicted structural representations (body and face). This analysis is particularly important for assessing the potential privacy-preserving of the VE-MD framework; computing internal structural parameters could limit the possibility of monitoring individual information during the inference.

\subsubsection{Emotion Decoder With No Structural Representation Inputs}
\label{sec:emotion_dec_nosr_detr}

As explained in the introduction of this chapter, the goal is to build a latent space to be optimized with a multi-task approach to capture structural body and face representation.  From the latent space produced by the previous ViTVE, we start by integrating a structural representation decoder. And the classification decoder receives only the entire optimized latent space. The idea is that the latent space, enhanced during multi-task training, might already encode sufficient information for emotion recognition, allowing the Structural Representation decoder to be discarded at inference time. In this setting, only the latent space is used for classification, and Structural Representation outputs are not explicitly reused.

\paragraph{Group Emotion datasets:}
For group emotion datasets, this approach does not yield competitive results. On GAF-3.0, performance is essentially unchanged or even slightly lower compared to ViTVE. With $Q=50$, the best accuracies are $82.25\%$ (CUSTom RESidual) and $82.37\%$ (ResNet50), while ViTVE alone achieves $82.41\%$ and $83.56\%$ respectively. A similar pattern is observed on VGAF: with $Q=50$, the model achieves $76.50\%$ ( CUSTom RESidual) and $76.63\%$ (ResNet50), close to ViTVE alone but not improved. Given this lack of benefit, further experiments on GER datasets were avoided for ecological reasons.

\paragraph{Non-group emotion datasets.}
Results on non-group emotion datasets are reported in Tables~\ref{tab:residual_detr_noSR} and ~\ref{tab:resnet_detr_noSR}. Across all evaluated datasets (MER-MULTI, SAMSEMO, DFEW, and EngageNet), the \emph{latent-only} VE\_MD configuration consistently outperforms the ViTVE baseline. On MER-MULTI, latent-only VE\_MD achieves up to {62.38\%} accuracy, improving upon ViTVE by {+3.16} points compared to the CUStom RESidual ViTVE and by {+1.55} points compared to ViTVE with a ResNet-50 backbone. For SAMSEMO, performance reaches {75.60\%}, corresponding to a {+3.34}-point gain over the best ViTVE configuration. On DFEW, latent-only VE\_MD attains {69.98\%}, yielding improvements of {+1.45} and {+5.29} points over ViTVE with ResNet-50 and the CUStom RESidual backbone, respectively.  
Finally, on EngageNet, VE\_MD achieves {68.47\%}, outperforming ViTVE by {+0.65} and {+0.37} points across the two backbone variants.

Overall, these results demonstrate that the latent-space formulation of VE\_MD provides a consistent and measurable benefit over ViTVE on non-group emotion recognition tasks even without incorporating explicit structural representations in the emotion decoder.

\begin{table}[H]
    \centering
    \caption{Accuracy comparison of VE-MD with the  CUSTom RESidual encoder when Structural Representations are not used in the emotion decoder. Results are shown for body, face, and body+face configurations across different query settings. $Q_{Max}$:~query is the maximum annotated persons in the dataset, $Q_{50}$: $query=50$,  $Q_{100}$:~$query=100$.}
    \label{tab:residual_detr_noSR}
    \begin{tabular}{@{}l|ccc|ccc|c@{}}
        \toprule
        \textbf{Dataset} & \multicolumn{3}{c|}{\textbf{Body}} & \multicolumn{3}{c|}{\textbf{Face}} & \textbf{Body+Face} \\
        \cmidrule{2-8}
           & $Q_{Max}$ & $Q_{50}$ & $Q_{100}$ & $Q_{Max}$ & $Q_{50}$ & $Q_{100}$ & Best Q \\ 
        \midrule
        MER-MULTI  & {60.92} & {62.14} & {-} & \textbf{62.38} & {60.92} & {-} & {61.17} \\
        DFEW       & { } & { } & { } & {69.46} & {69.98} & {-} & { } \\
        SAMSEMO    & {74.50} & {74.21} & {-} & {74.59} & {74.40} & {-} & \textbf{75.60} \\
        EngageNET  & {67.44} & {68.00} & {-} & {67.16} & {67.35} & {-} & {67.54} \\
        \bottomrule
    \end{tabular}
\end{table}

\begin{table}[H]
    \centering
    \caption{Accuracy comparison of VE-MD with ResNet50 encoder when Structural Representations are not used in the emotion decoder. Results are shown for body, face, and body+face configurations across different query settings.  $Q_{Max}$:~query is the maximum annotated persons in the dataset, $Q_{50}$: $query=50$,  $Q_{100}$:~$query=100$.}
    \label{tab:resnet_detr_noSR}
    \begin{tabular}{@{}l|ccc|ccc|c@{}}
        \toprule
        \textbf{Dataset} & \multicolumn{3}{c|}{\textbf{Body}} & \multicolumn{3}{c|}{\textbf{Face}} & \textbf{Body+Face} \\
        \cmidrule{2-8}
           & $Q_{Max}$ & $Q_{50}$ & $Q_{100}$ & $Q_{Max}$ & $Q_{50}$ & $Q_{100}$ & Best Q \\ 
        \midrule
        MER-MULTI  & {61.65} & {61.65} & {-} & \textbf{62.37} & {61.41} & {-} & \textbf{62.38} \\
        DFEW       & { } & { } & { } & {69.98} & {69.47} & {-} & { } \\
        SAMSEMO    & {75.31} & \textbf{75.64} & {-} & {74.26} & {75.02} & {-} & {75.41} \\
        EngageNET  & {67.63} & {66.60} & {-} & \textbf{68.47} & {68.00} & {-} & {67.88} \\
        \bottomrule
    \end{tabular}
\end{table}

\paragraph{Analysis:} The low performance observed on GER datasets can be explained by the nature of the task. In non-group emotion recognition, labels are tied to a single person or a small set of individuals, so the encoder’s latent space already captures sufficient appearance and contextual cues for classification. In contrast, group-level emotion recognition requires modeling interactions and dynamics across multiple people. Structural Representation features provide structured, per-person representations that preserve these multi-instance cues and ensure permutation invariance. When only the latent space is used, this structure is lost, and the latent space alone cannot adequately represent inter-person relationships, leading to a drop in accuracy for GER datasets.

\subsubsection{Emotion Decoder With Structural Representation Inputs}

 After finding out that using only the latent space in the emotion decoder does not work well for the GER dataset, we also explore directly feeding structural representation features into the emotion decoder to emphasize body language and facial expression cues during classification.
 
 For the sake of reducing complexity with DETR, we explore a Linear projection in the output structural representation decoder. The reason is that the size of the inputs for the emotion decoder becomes bigger and is strongly affected by the number of queries. As reminder from the section~\ref{sec:emotion_decoder_detr} the final dimension $D$ input for the emotion decoder is $D=C + 4Q(N_b + N_f)$ with $C$ the dimension channel of the entire latent space (two latent space from the two encoders), $Q$ is the number of querries, $N_b, N_f$ number of limbs connections for respectively body and face when both are used. That makes the transformer layer in the emotion decoder costly. For simplicity, the output of the structural representation decoder ($4Q(N_b + N_f)$) is projected linearly to $latent\_dim$. This is to be aligned with the latent space dimension from each encoder output.

\paragraph{Effect of Structural Representation Projection:}

We investigate two ways of integrating Structural Representations into the decoder: (i) using raw Structural Representation features, and (ii) applying a linear projection to match the latent space dimension ($latent\_dim$). A key observation is that the model behaves differently depending on whether it is trained on GER or non-GER datasets.

\begin{table}[H]
\centering
\caption{DETR accuracy comparison when using projection vs. raw structural representations in the emotion decoder. $Q_{Max}$ is used as the number of queries. GER vs. Non-GER datasets.}
\label{tab:proj_detr_results}
\resizebox{\linewidth}{!}{%
\begin{tabular}{lcccc}
\toprule
        \textbf{Dataset} &   \textbf{Encoder} &   \textbf{Structural Feature} & \textbf{Acc. (Proj.)} & \textbf{Acc. (No Proj.)}\\ 
    \midrule
     GAF-3.0    &  ResNet50   & body    & {82.65}   & \textbf{84.35} \\
     VGAF       &  ResNet50   & body    & {76.24}   & \textbf{78.46} \\
    \midrule
     MER-MULTI  &  ResNet50   & body    & \textbf{61.17}  & {56.55}  \\
     SAMSEMO    &  ResNet50   & body    & \textbf{75.29}  & {67.45}  \\
\bottomrule
\end{tabular}
}
\end{table}

 In Tables~\ref{tab:proj_detr_results}, analysis of this observation with GAF-3.0, VGAF, SAMSEMO, and MER-MULTI  reveals consistent trends. For GER datasets, projection limits performance: GAF-3.0 accuracy improves by about $+2\%$ with DETR when raw Structural Representations are used. VGAF shows a similar $+2\%$ gain without projection. This supports the view that projection acts as a bottleneck, collapsing multiple Structural Representations into a compressed representation and discarding inter-person interactions essential for group-level recognition.

For non-GER datasets such as MER-MULTI and SAMSEMO, the opposite holds. Projection improves performance substantially: for MER-MULTI, by $+5\%$ with DETR;  for SAMSEMO, by $+8\%$. In these cases, projection serves as a denoising bottleneck, removing nuisance variation in Structural Representation features and retaining salient information, which is particularly advantageous for single- or two-person emotion recognition.

\begin{figure}[H]
  \centering
  \includegraphics[width=0.99\linewidth]{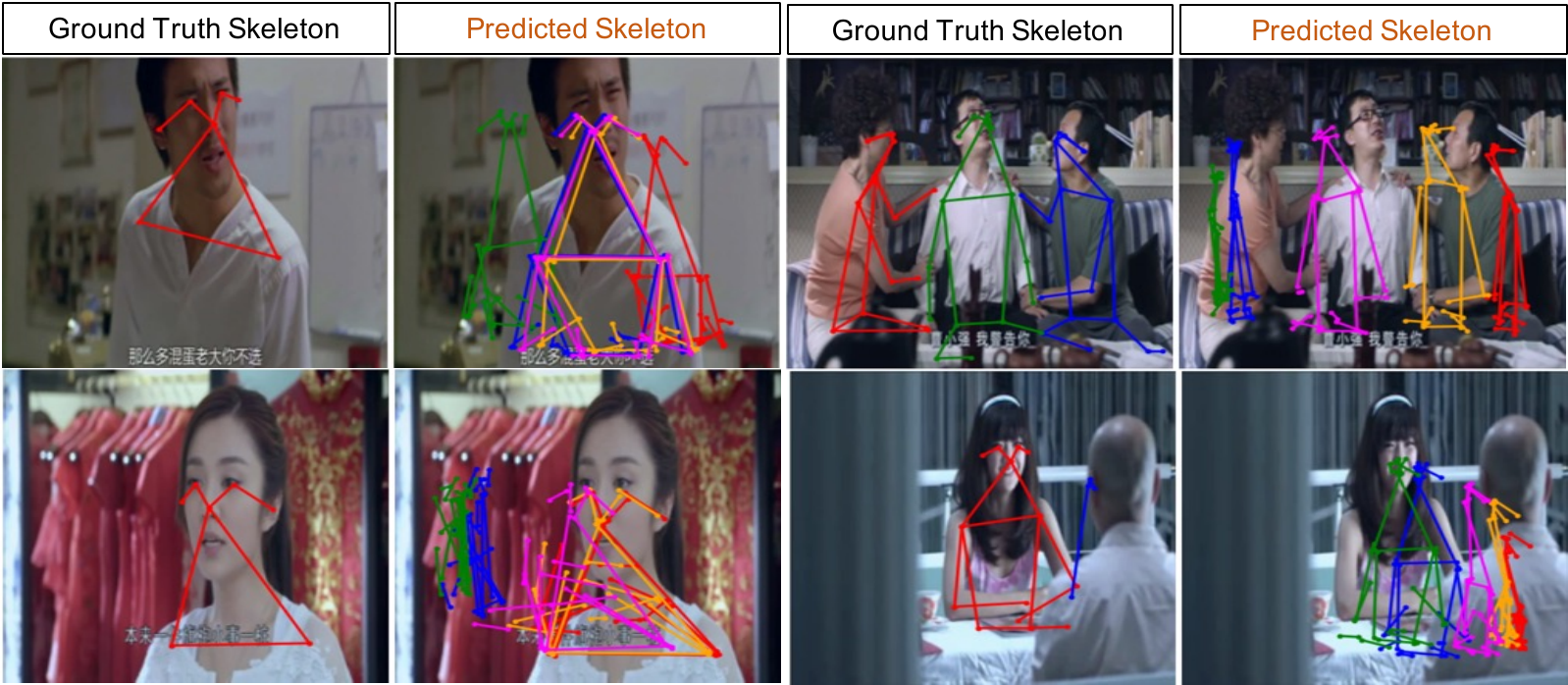}
 \caption{Predicted Structural Representation with DETR on MER-MULTI with $num\_query=6$. }
  \label{fig:pred_SR_detr}
\end{figure}

Figure~\ref{fig:pred_SR_detr} illustrates this effect on MER-MULTI with DETR ($num\_query=6$). In the first two columns, six Structural Representations are predicted, but not all align with true poses; projection suppresses these noisy Structural Representations, aiding classification. In group emotion cases (last two columns), where fewer Structural Representations are predicted, projection risks discarding useful interaction cues, thus harming performance. These results highlight the need for different emotion decoder designs in VE-MD for GER versus non-GER datasets.

 Tables~\ref{tab:residual_detr_SR} and~\ref{tab:resnet_detr_SR} present results for VE-MD with a DETR-based decoder. For GER, we evaluate VE-MD on GAF-3.0 (images) and VGAF (video), while for non-GER, we use SAMSEMO, MER-MULTI, DFEW, and EngageNET. We further conduct ablation studies varying the number of queries, Structural Representation types (body, face, or both), and the addition of ST-GCN. Two sets of experiments are reported: group emotion datasets and non-group emotion datasets. For GER, all ablation combinations are tested. For non-GER, experiments are run with $Q_{Max}$ and $Q_{50}$, and ST-GCN is applied only when improvements are observed. 

\paragraph{Ablation for Group Emotion Datasets:}  In Tables~\ref{tab:residual_detr_SR} and~\ref{tab:resnet_detr_SR}, on GAF-3.0, performance improves when adding ST-GCN, regardless of using body-only, face-only, or combined Structural Representations. The best result is $90.06\%$ with ResNet50 (body+face+ST-GCN), a +4.07\% gain over raw Structural Representations. Similarly, the  CUSTom RESidual encoder reaches $89.71\%$, a +3.87\% improvement.  
  For VGAF, the best performance with the residual encoder is $78.46\%$ (body Structural Representation, $Q=50$ with ST-GCN). Interestingly, ST-GCN does not always improve VGAF performance, suggesting dataset-specific dynamics. Using synthetic video augmentation, accuracy improves further to~$80.42\%$.

\begin{table}[H]
    \centering
\caption{Accuracy comparison of VE-MD using the  CUSTom RESidual encoder as the multi-task backbone. Results are reported for different keypoint configurations (Body, Face, and Body+Face) across query settings. $Q_{i}$ denotes the number of queries, with $Q_{Max}$ the maximum number of annotated persons in the dataset, $Q_{50}$: $query=50$,  $Q_{100}$:~$query=100$. Structural Representation (SR) outputs are used in the Emotion Decoder.}
    
    \label{tab:residual_detr_SR}
    \renewcommand{\arraystretch}{1.4} %
    \resizebox{\linewidth}{!}{%
    
    \begin{tabular}{@{}l|ccc|ccc|ccc|ccc|cc@{}}
        \toprule
         \textbf{Datasets} & \multicolumn{6}{c|}{\textbf{Body}} & \multicolumn{6}{c|}{\textbf{Face}} & \multicolumn{2}{c}{\textbf{Body+Face}} \\ 
        \cmidrule{2-15}

       & \multicolumn{3}{c|}{  SR} & \multicolumn{3}{c|}{  SR+GCN} &  \multicolumn{3}{c|}{  SR} & \multicolumn{3}{c|}{  SR+GCN}  &  \multicolumn{1}{c|}{  SR} & \multicolumn{1}{c}{  SR+GCN}  \\ 

        \cmidrule{2-15}
          &  $Q_{Max}$   & $Q_{50}  $ &$Q_{100}$ &   $Q_{Max}$ &   $Q_{50}  $ &   $Q_{100}$ &   $Q_{Max}$ &   $Q_{50}  $ &   $Q_{100}$ &   $Q_{Max}$ &   $Q_{50}  $ &   $Q_{100}$   &  Best Q &  Best Q \\ 
        
        \midrule
        GAF-3.0   &{84.13}&{84.74}&{84.27}   &{85.53}&{85.11}&{86.75}  &{85.58}&{84.92}&{85.72}  &{{88.29}}&{86.13}&{88.25}  &{85.84} &\textbf{89.71} \\ 
        VGAF     &{76.76}&{77.28}&{77.42}   &{76.50}&\textbf{{78.46}} &{76.76}  &{78.07}&{77.02}&{77.55}  &{77.42}&{76.37} &{77.28} &{77.55} &{77.15} \\ 
        \midrule
        
        MER-MULTI   &{62.14} &\textbf{62.14} &{-}  &{59.47}&{-} &{-}   &{60.68}&{60.92}&{-}   &{59.22}&{-}&{-}  &{61.65} &{-} \\ 

        DFEW   &{ }&{ }&{ }   &{ }&{ }&{ }  &\textbf{69.04}&{64.60}&{-}   &{68.88}&{64.45}&{-}  &{ } &{ } \\ 
        SamSemo  &{73.86} &{72.76} &{-}      &\textbf{{74.38}}&{72.48}&{-}  &{73.95}&{73.00}&{-} &{73.33}&{-}&{-}  &{74.05} &{74.24} \\ 
       EngageNet &{67.16}&{66.79}&{-}   &{67.16}&{67.16}&{-}  &{67.82}&{67.26}&{-} &{67.63}&{66.88}&{-}  &{67.26} &{-} \\

        \bottomrule
        
    \end{tabular}
  }
\end{table}

\begin{table}[H]
    \centering
\caption{Accuracy comparison of {VE-MD (DETR-based)} using {ResNet50} as the multi-task encoder. Results are reported for different keypoint configurations (Body, Face, and Body+Face) across query settings. $Q_{i}$ denotes the number of queries, with $Q_{Max}$ the maximum number of annotated persons in the dataset, $Q_{50}$: $query=50$,  $Q_{100}$:~$query=~100$. Structural Representation (SR) outputs are used in the Emotion Decoder.}
    \label{tab:resnet_detr_SR}
    \renewcommand{\arraystretch}{1.4} %
    \resizebox{\linewidth}{!}{%
    
    \begin{tabular}{@{}l|ccc|ccc|ccc|ccc|cc@{}}
        \toprule
         \textbf{Datasets} & \multicolumn{6}{c|}{\textbf{Body}} & \multicolumn{6}{c|}{\textbf{Face}} & \multicolumn{2}{c}{\textbf{Body+Face}} \\ 
        \cmidrule{2-15}

      & \multicolumn{3}{c|}{  SR} & \multicolumn{3}{c|}{  SR+GCN} &  \multicolumn{3}{c|}{  SR} & \multicolumn{3}{c|}{  SR+GCN}  &  \multicolumn{1}{c|}{  SR} & \multicolumn{1}{c}{  SR+GCN}  \\ 

        \cmidrule{2-15}
         &  $Q_{Max}$ &   $Q_{50}  $ &   $Q_{100}$ &   $Q_{Max}$ &   $Q_{50}  $ &   $Q_{100}$ &   $Q_{Max}$ &   $Q_{50}  $ &   $Q_{100}$ &   $Q_{Max}$ &   $Q_{50}  $ &   $Q_{100}$   &  Best Q &  Best Q \\ 
        
        \midrule
        GAF-3.0   &{84.38}&{84.22} &{84.89}   &{86.45}&{86.14}&{87.99}  &{85.72}&{85.40}&{86.11}  &{87.67}&{86.27}&{88.25} &{85.99}&\textbf{90.06} \\ 
        VGAF      &{75.98}&\textbf{78.46}&{{77.55}}  &{77.02}&{77.02}&{78.33}  &{77.28}&{77.28}&{77.55} &{77.15}&{76.11}&{76.5}  &{78.33} &{77.55} \\ 
        VGAF +  Synt-data &{-} &\textbf{80.42} &{-} &{-}  &{-} &{-} &{-} &{-} &{-} &{-}  &{-}  &{-}  &{79.37} &{-} \\ 
        \midrule

        MER-MULTI    &{61.17} &{61.41} &{-}  &\textbf{62.38}&{60.19} &{-}   &{60.19}&{61.41}&{-}   &{62.14}&{61.65}&{-}  &{60.92} &\textbf{62.38} \\ 

        DFEW  &{ }&{ }&{ }  &{ }&{ }&{ }  &{68.96}&{64.45}&{-}   &{68.66}&{-}&{-}  &{ }&{ }\\ 
       
        SamSemo  &\textbf{75.29}&{74.76}&{-}  &{74.29} &{-}&{-}  &{75.14}&{75.19} &{-}  &{74.38}&{74.63}&{-}  &{75.05} &{74.29} \\ 
       EngageNet   &{{68.53}}&{67.35}&{-}   &{67.82}&{67.98}&{-}  &{67.91}&{67.63}&{-} &{68.19}&{66.88}&{-}  &{68.38} &\textbf{68.98} \\

        \bottomrule
        
    \end{tabular}
  }
\end{table}

\paragraph{Ablation for Non-Group Emotion Datasets:} In Tables~\ref{tab:residual_detr_SR} and~\ref{tab:resnet_detr_SR}, performance trends vary by encoder. For MER-MULTI, the residual encoder achieves $62.14\%$ (body Structural Representation, $Q_{Max}$/$Q_{50}$), while ResNet50 achieves $62.38\%$ (body+face, best $Q$, with ST-GCN). For SAMSEMO, the residual encoder reaches $74.38\%$ (body Structural Representation, $Q_{Max}$ with ST-GCN), while ResNet50 achieves $75.29\%$ (body-only).  
  For EngageNET, the best accuracy is $68.98\%$ (ResNet50, body+face+ST-GCN), while for DFEW the best is $69.04\%$ (residual encoder, face-only).  

Overall, VE-MD demonstrates consistent improvements over ViT-only baselines across all datasets, with ST-GCN providing the most benefit in body+face configurations for group datasets.

It is worth noting that, for ecological reasons, only $Q_{Max}$ and $Q_{50}$ were evaluated for non-GER datasets. $Q_{100}$ was omitted after observing a drop in performance when increasing the number of queries. In addition, the ST-GCN module was sometimes excluded when no improvement was observed with its inclusion.

\subsubsection{DETR Approach Analysis}
The proposed VE-MD DETR-based decoder achieves promising results on several datasets, notably GAF-3.0 and VGAF. However, its performance is not optimal. For instance, on VGAF, combining face and body structural representations does not improve accuracy compared to using a single modality. One explanation lies in the alignment issues within the automatically generated annotations: body structural representations are derived from ViTPose and facial landmarks from FaceAlignment, both of which are not perfect in the wild. Consequently, some individuals may be annotated only for face or only for body, and in video datasets such as VGAF, annotations can be inconsistent across frames, present in one frame but missing in the next. This variability weakens the benefit of combining modalities.  

These annotation inconsistencies also explain why ST-GCN performs well on single-image datasets (e.g., GAF-3.0) but less reliably on videos. ST-GCN relies on temporal continuity to model structural representation dynamics; if a person disappears from the frame or parts of the body are occluded, the sequential relationships are disrupted, degrading performance. A similar issue arises with DETR’s fixed query mechanism: the number of persons detected can fluctuate from frame to frame, yet the model assumes a fixed number of queries, introducing further misalignment.  

Frame selection strategies also play a role. When frames are sampled densely or continuously, ST-GCN has a better chance of capturing consistent temporal dynamics. This explains why it sometimes improves results in MER-MULTI and SAMSEMO (10 frames per video), but less so in VGAF (5 frames, one per second), where temporal coherence is weaker.

\noindent The DETR-based approach proved effective in learning structured representations but remained constrained by its fixed query mechanism, which restricts the number of individuals the model can represent simultaneously. To move beyond this limitation and achieve a more flexible, end-to-end formulation, we adopt a heatmap-based strategy that directly infers structural representations through limb-connection estimation.

\section{VE-MD with Heatmap-Based Decoder}
\label{sec:ve-md-heatmap}
In the previous approach based on DETR~\citep{carion2020end}, the model’s capacity was constrained by the fixed number of queries defined during training. Each query corresponds to a potential person to detect; thus, setting the number of queries to 100, for example, allows the model to predict only 100 persons, even if more or fewer are present in the image. This fixed-query mechanism inherently restricts scalability and flexibility when dealing with in-the-wild group scenes containing a variable number of people. To overcome this limitation, we propose an end-to-end heatmap-based estimation approach for direct structural representation prediction through limb-connection estimation. In contrast to DETR, the heatmap decoder naturally adapts to any number of individuals present in an image, since person structures emerge directly from the spatial activation patterns of the predicted heatmaps.

In the literature, most structural representation or pose-estimation methods~\citep{cao2017realtime, cao2019openpose, osokin2019real, jo2022comparative, zhang2025two} rely on keypoint detection followed by a separate post-processing step to infer the body or face connections. In our approach, this two-step procedure is replaced by a single-stage prediction: rather than estimating discrete keypoints and linking them afterward, the model directly predicts limb-connection heatmaps representing the structural configurations of both body and face. This design allows structural information to be embedded directly within the latent space, enhancing its semantic richness for emotion recognition while maintaining an end-to-end differentiable training pipeline.

\subsection{Heatmap Structural Representation Decoder} 

\begin{figure}[H]
  \centering
  \includegraphics[width=0.99\linewidth]{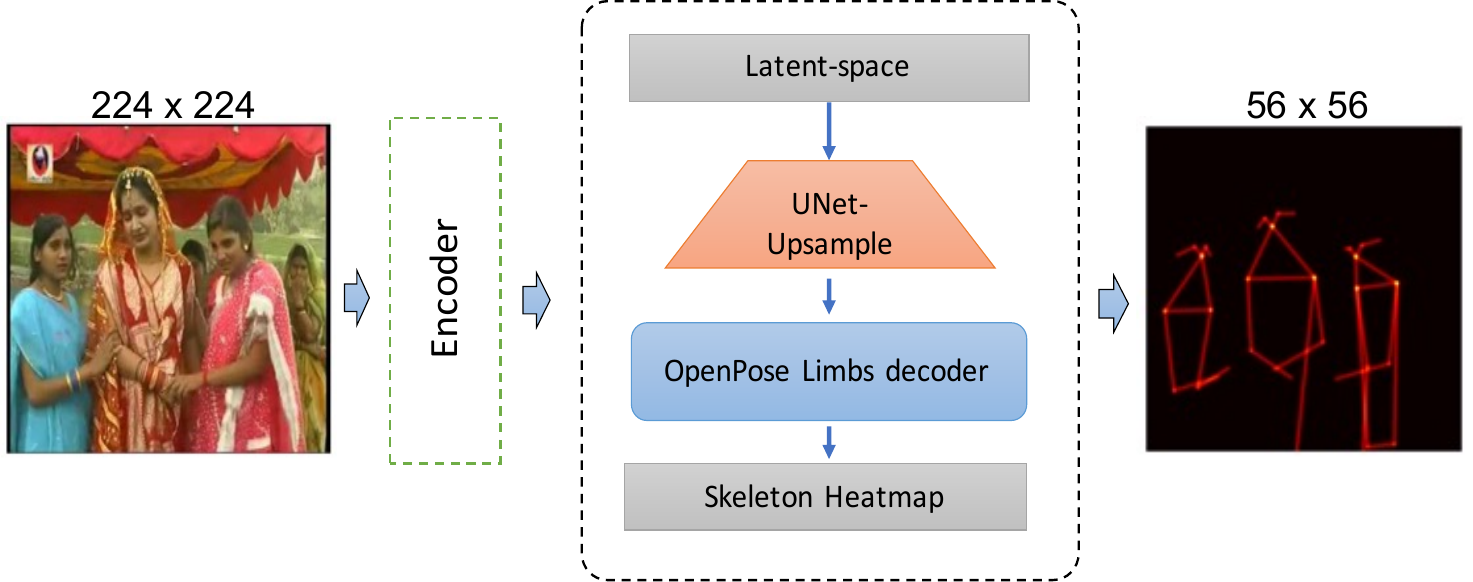}
 \caption{ At the top is the latent space input, followed by a custom UNet upsample style, then a Modified OpenPose style decoder to predict the limbs heatmap.}
  \label{fig:SR_decoder_heatmap}
\end{figure}

In Figure~\ref{fig:SR_decoder_heatmap}, it is shown that the proposed heatmap structural representation decoder is composed of two main stages: the first stage is an upsampling feature from the latent space by using a custom residual U-Net style~\citep{ronneberger2015u}.  The description of the custom U-NetUpsample feature is given in appendix ~\ref{tab:unetres}. From that output feature, a custom OpenPose \citep{cao2017realtime} is used for the second stage, with output channels assigned to the number of limbs. 

\subsection{Heatmap-based Emotion Decoder}
The emotion decoder, the same in section~\ref{sec:emotion_decoder_detr}, receives the entire latent representations from the upstream encoders and, optionally, incorporates structural representation-based information.

\paragraph{Inputs.}  
For each clip, per-frame features, optionally, structural representation heatmaps are included:
\[
\mathbf{S}^{\text{body}},\ \mathbf{S}^{\text{face}} \in \mathbb{R}^{T\times (H_S \cdot W_S)}.
\]

\paragraph{Feature fusion.} 

If using raw  structural representation, they are concatenated with $\tilde{\mathbf{z}}_t$, yielding
$$
\mathbf{x}_t = [\tilde{\mathbf{z}}_t; {\mathbf{S}}^{\text{body}}_t\ (\text{if used}); \{\mathbf{S}^{\text{face}}_t\ (\text{if used})] \in \mathbb{R}^{D},
$$

$D=C\ (\text{no  structural representation}),$ 

$D= {C + H_S \cdot W_S} \ (\text{one  structural representation}),$

$D=C + 2( H_S \cdot W_S)\ (\text{both}).$ 

\noindent If \textbf{not} using raw  structural representation, they are projected as defined below:

\[
\hat{\mathbf{S}}^{\text{body}}_t = \rho\!\left(\mathbf{S}^{\text{body}}_t \right),\quad
 \hat{\mathbf{S}}^{\text{face}}_t = \rho\!\left(\mathbf{S}^{\text{face}}_t \right). 
\]
These are concatenated with $\tilde{\mathbf{z}}_t$, yielding

$$
\mathbf{x}_t = [\tilde{\mathbf{z}}_t; \hat{\mathbf{S}}^{\text{body}}_t\ (\text{if used}); \hat{\mathbf{S}}^{\text{face}}_t\ (\text{if used})] \in \mathbb{R}^{D},
$$ 
 
$D=C \ (\text{no  structural representation}),$

$D=C+C_S \ (\text{one  structural representation}),$

$D=C+2C_S \ (\text{both}).$

$\rho$ is the normalized linear layer projection, and $C_S$ is the projected linear dimension.

The rest is the same as in DETR version~\ref{sec:emotion_decoder_detr}.

\subsection{Loss Functions}

For the heatmap decoder, we follow a multi-stage heatmap refinement strategy inspired by OpenPose~\citep{cao2019openpose}. At each stage \(s\), predictions \(\hat{H}^{(s)} \in \mathbb{R}^{C \times H \times W}\) are compared against ground truth heatmaps \(H \in \mathbb{R}^{C \times H \times W}\) using Mean Squared Error (MSE):

\begin{equation}
\text{MSE}^{(s)} = \frac{1}{CHW}\sum_{c=1}^{C}\sum_{i=1}^{H}\sum_{j=1}^{W}(\hat{H}^{(s)}_{c,i,j}- H_{c,i,j})^2
\end{equation}

where:
\begin{itemize}
    \item \(C\): Number of heatmap channels (classes)
    \item \(H, W\): Spatial dimensions (height, width)
    \item \(S\): Total number of prediction stages
\end{itemize}

\noindent The total loss for the heatmap decoder combines all components as follows:

\begin{equation}
\label{eq:total_loss_heatmap}
\mathcal{L}_{\text{total}} = \mathcal{L}_{\mathrm{cls}} + \beta_{\mathrm{p_{1}}}\mathcal{L}_{\mathrm{p_{1}}} + \beta_{\mathrm{p_{2}}}\mathcal{L}_{\mathrm{p_{2}}} + \beta_{\mathrm{mmd}}\mathcal{L}_{\mathrm{MMD}}
\end{equation}

with \(\beta_{\mathrm{p_{i}}}\) (for body or face),  and \(\beta_{\mathrm{mmd}}\) representing loss weighting factors. (See Section~\ref{sec:ve-md-detr-loss}).

\subsection{Training with Heatmap Decoder}
Almost the same experiments conducted for DETR are conducted for the Heatmap approach, even further, for comparison purposes. Sometimes experiments can be investigated further when the results differ, because of the stability of the heatmap approach. 
For the heatmap-based decoder, the latent space is also set to $512$. Features from the U-Net upsampling stage (see in appendix Table~\ref{tab:unetres}) are projected to 256 channels. These are passed to an OpenPose-style network with six stages, each consisting of five convolutional layers ($input=output=256$, $kernel=3$, $padding=1$). The final outputs are heatmaps of size $output\_limbs \times 56 \times 56$, where $output\_limbs = 18$ for body connections and $83$ for facial landmarks. It is worth noting that for the Heatmap-based approach, we augmented the original 63 connections with our additional 20 custom connections, resulting in a total of 83 connections. The reason is that the heatmap dimension is still the same for the emotion decoder; it does not increase the complexity of the classification. For the loss function, we set $\beta_{mmd}=0.1$ and $\beta_{p_i}=1.0$ (see equation~\ref{eq:total_loss_heatmap}), giving equal importance to structural representation detection and emotion classification.

\subsubsection{Emotion Decoder With No Structural Representation Inputs}
\label{sec:emotion_dec_nosr_heatmap}

We conducted the same set of experiments as described in section~\ref{sec:emotion_dec_nosr_detr} to evaluate whether the latent space, enriched during multi-task training, already contains sufficient information for emotion recognition, making the explicit use of structural representations unnecessary at inference time. In this configuration, the emotion decoder relies solely on the latent embedding, while outputs from the structural representation decoders (body or face) are not reused.

\paragraph{Group Emotion Datasets:} Similar to the DETR-based results, we observed that using only the latent space without structural inputs does not yield competitive performance for group emotion recognition (GER) datasets. Two main experiments were performed on GAF-3.0 and VGAF using a face-structure decoder with ResNet50 as the encoder. For VGAF, the model achieved an accuracy of $76.11\%$, which is lower compared to \textit{ViTVE} ($76.63\%$). For GAF-3.0, the accuracy dropped to $82.65\%$, slightly lower than the \textit{ViTVE} baseline ($83.56\%$). Given these small and inconsistent variations, further experiments on GER datasets were not pursued for ecological reasons.

\begin{table}[H]
    \centering
    \caption{Accuracy comparison of {VE-MD (Heatmap-based)} using {CUSTom RESidual} and {ResNet50} as multitask encoders. Results are reported for different structural configurations: Body, Face, and Body+Face. No structural representation is used in the emotion decoder.}
    \label{tab:results_heatmap_noskt}
    \begin{tabular}{@{}l|ccc|ccc@{}}
        \toprule
        \textbf{Dataset} & \multicolumn{3}{c|}{\textbf{CUSTom RESidual Encoder}} 
         & \multicolumn{3}{c}{\textbf{ResNet50 Encoder}} \\
       \cmidrule{2-7}
         & \textbf{Body} & \textbf{Face} & \textbf{Body+Face} 
         & \textbf{Body} & \textbf{Face} & \textbf{Body+Face} \\ 
        \midrule
        MER-MULTI & 61.89 & \textbf{62.14} & 61.89 & 60.44 & 61.65 & 60.68 \\ 
        DFEW &   & 68.86 &  &  & 68.30 &  \\ 
        SamSemo & 73.90 & 73.91 & 74.10 & 75.00 & \textbf{75.29} & 75.10 \\ 
        EngageNet & 67.04 & 67.26 & 67.07 & 67.26 & 67.63 & 68.10 \\ 
        \bottomrule
    \end{tabular}
\end{table}

\paragraph{Non-Group Emotion Datasets:}
Table~\ref{tab:results_heatmap_noskt} reports results obtained with the heatmap-based VE\_MD configuration without explicit skeleton supervision. Overall, this setting follows the same performance trends observed with the DETR-based variant and consistently improves upon the ViTVE baselines across most non-group emotion datasets. On MER-MULTI, the proposed model reaches {62.14\%} accuracy, corresponding to an improvement of {+3.24} points over \textit{ViTVE}-CustRes and {+1.41} points over \textit{ViTVE}-R50.  
For SAMSEMO, accuracy increases to {75.29\%}, yielding a {+3.03} points gain compared to the best ViTVE configuration. On DFEW, VE\_MD achieves {68.86\%}, improving upon \textit{ViTVE}-CustRes by {+0.56} points and upon \textit{ViTVE}-R50 by {+4.17} points. Finally, on EngageNet, performance reaches {68.10\%}, matching the best ViTVE result. These results confirm that the heatmap-based VE\_MD variant maintains the advantages of the latent-space formulation observed with DETR, even in the absence of explicit skeleton supervision.

The latent-only configuration demonstrates the robustness of the VE-MD latent space but also highlights its limitations for group-level affect understanding. The following subsection reintroduces structural information into the emotion decoder to analyze its explicit contribution to emotion classification.

\subsubsection{Emotion Decoder With Structural Representation Inputs}
\label{sec:emotion_dec_sr_heatmap}

Following the analysis of the latent-only configuration, we now examine the effect of reintroducing structural representation inputs into the emotion decoder for the heatmap-based VE-MD architecture. This experiment aims to determine how explicit structural cues body and facial configurations, contribute to emotion recognition when combined with the shared latent space. Consistent with observations from the DETR-based model, the impact of structural information in the decoder differs markedly between group emotion recognition (GER) and non-group emotion datasets. For GER, we evaluate VE-MD on GAF-3.0 (image-based) and VGAF (video-based) datasets, while for non-GER, we assess performance on SAMSEMO, MER-MULTI, DFEW, and EngageNet.

\paragraph{Effect of Structural Representation Projection:}
\label{sec:emotion_dec_sr_heatmap_group_nongroup}

In the heatmap-based framework, we investigate two strategies for integrating structural representations into the emotion decoder:  
(i) directly using the raw structural representation features, and  
(ii) applying a linear projection to match the latent-space dimension ($latent\_dim$).  

To analyze this effect, we conducted experiments on VGAF (GER) and SAMSEMO (non-GER) using ResNet50 as encoder and the heatmap-based structural decoder focusing on body representations. The base latent dimension was set to $latent\_dim = 512$. When no projection is applied, the final embedding dimension is computed as:
\[
Embedding\_size = 2 \times latent\_dim + 56 \times 56,
\]
representing the two encoder outputs combined with the full structural representation heatmap.  
When projection is applied, the heatmap output ($56 \times 56$) is reduced by a linear layer to $factor \times latent\_dim$, yielding:
\[
Embedding\_size = 2 \times latent\_dim + factor \times latent\_dim,
\]
with $factor \in \{0.5, 1, 2, 3, 4\}$.

\begin{table}[H]
\centering
\caption{Heatmap accuracy comparison with and without projection in the emotion decoder. Experiments use the ResNet50 encoder with body structural representations on VGAF (GER) and SAMSEMO (Non-GER). With projection: $Embedding\_size = 2 \times latent\_dim + factor \times latent\_dim$; without projection: $Embedding\_size = 2 \times latent\_dim + 56 \times 56$.}
\label{tab:proj_heatmap_test_dim}
\resizebox{\linewidth}{!}{
\begin{tabular}{lcccccc}
\toprule
\textbf{Dataset} & \textbf{latent\_dim} & \textbf{Linear Proj.} & \textbf{Factor} & \textbf{Proj. size} & \textbf{Embedding\_size} & \textbf{Acc. (\%)} \\ 
\midrule
VGAF      & 1024 & No  & {-}   & {-}   & 5184  & 78.20 \\
VGAF      & 512  & No  & {-}   & {-}   & 4160  & \textbf{79.77} \\
VGAF      & 512  & Yes & 4     & 2048  & 3072  & 77.81 \\
VGAF      & 512  & Yes & 3     & 1536  & 2560  & 78.07 \\
VGAF      & 512  & Yes & 2     & 1024  & 2048  & 77.94 \\
VGAF      & 512  & Yes & 1     & 512   & 1536  & 77.42 \\
\midrule
SAMSEMO   & 512  & No  & {-}   & {-}   & 4160  & 67.48  \\
SAMSEMO   & 512  & Yes & 4     & 2048  & 3072  & 73.05 \\
SAMSEMO   & 512  & Yes & 3     & 1536  & 2560  & 72.81 \\
SAMSEMO   & 512  & Yes & 2     & 1024  & 2048  & 73.29 \\
SAMSEMO   & 512  & Yes & 1     & 512   & 1536  & \textbf{74.81} \\
SAMSEMO   & 512  & Yes & 0.5   & 256   & 1280  & 73.57 \\
\bottomrule
\end{tabular}
}
\end{table}

 Table~\ref{tab:proj_heatmap_test_dim} summarizes the results. For VGAF, accuracy improves slightly when no projection is used, reaching $79.77\%$ at $latent\_dim = 512$. Increasing the latent size to 1024 brings no further gain. In contrast, SAMSEMO performs poorly without projection ($67.48\%$) but improves significantly when the projection is applied, peaking at $74.81\%$ for a projection size of 512. These outcomes confirm an opposite trend between GER and non-GER datasets: while projection degrades GER performance, it enhances results on non-GER datasets.

\begin{table}[H]
\centering
\caption{Heatmap accuracy comparison when using projection versus raw structural representations in the emotion decoder, across GER and non-GER datasets.}
\label{tab:proj_heatmap_results}
\resizebox{\linewidth}{!}{%
\begin{tabular}{lcccc}
\toprule
\textbf{Dataset} & \textbf{Encoder} & \textbf{Structural Features} & \textbf{Acc. (Proj.)} & \textbf{Acc. (No Proj.)} \\ 
\midrule
GAF-3.0    & ResNet50   & Face    & 82.56 & \textbf{89.60} \\
VGAF       & ResNet50   & Face    & 77.42 & \textbf{79.77} \\
\midrule
MER-MULTI  & ResNet50   & Face    & \textbf{61.17} & 54.85 \\
SAMSEMO    & ResNet50   & Face    & \textbf{75.00} & 55.34 \\
\bottomrule
\end{tabular}
}
\end{table}

Table~\ref{tab:proj_heatmap_results} summarizes extended analysis to additional datasets, namely GAF-3.0 and MER-MULTI, confirming these opposing behaviors. For GER datasets, applying projection consistently limits performance: on GAF-3.0, accuracy improves by approximately $+7\%$ when using raw (non-projected) structural representations, and VGAF exhibits a similar $+2\%$ gain. This suggests again that projection acts as a compression bottleneck, reducing the dimensional richness of the structural cues and discarding inter-person interaction patterns that are critical for group-level affect understanding. 

Conversely, for non-GER datasets such as MER-MULTI and SAMSEMO, projection substantially enhances performance ($+8\%$ and $+10\%$, respectively). In these cases, the projection layer acts as a beneficial denoising bottleneck, filtering out irrelevant variations in structural features while retaining the most salient information. This effect is particularly useful for single-person or small-group emotion recognition, where the structural signal is more stable and less affected by inter-person variability. Based on these observations, subsequent experiments were conducted using the optimal configurations, with a projection size of 512 when it is applied.

\begin{table}[H]
    \centering
\caption{Accuracy comparison of {VE-MD (Heatmap-based)} using {CUSTom RESidual} and {ResNet50} as multi-task encoders. Results are reported for different keypoint configurations (Body, Face, and Body+Face). Structural representation outputs are used in the Emotion Decoder.}
    \label{tab:heatmap_skt_results}
    \begin{tabular}{@{}l|ccc|ccc@{}}
        \toprule
        \textbf{Datasets}   & \multicolumn{3}{c|}{{CUSTom RESidual Encoder}} 
         & \multicolumn{3}{c}{{Resnet50 Encoder}} \\
       \cmidrule{2-7}

         &{\textbf{Body}} & {\textbf{Face}} &{\textbf{Body+Face}} &{\textbf{Body}} & {\textbf{Face}} &{\textbf{Body+Face}} \\ 
        
        \midrule
         GAF-3.0    &{86.91}   &{87.67} &\textbf{89.09}    &{86.72} &{89.60} &\textbf{89.92}   \\ 
         VGAF   &77.15   &\textbf{78.98} &{77.42}    &{79.24} &{79.77} &{79.37}    \\ 
        VGAF+Synt-data   &-   &{-} &{-}    &{80.68} &\textbf{80.81} &{80.55}    \\ 

         \midrule

        MER-Multi 2023    &{60.92}   &{60.68} &{60.92}   &{61.17}   &{60.19} &\textbf{61.89}   \\
        DFEW    &{ }   &{70.73} &{ }    &{ }   &{70.31} &{ }  \\ 
        SamSemo   &{73.90}   &\textbf{74.48} &{74.38}    &{74.81}   &{75.00} &\textbf{75.10}  \\ 
        EngageNet    &{67.35}   &{67.72} &{67.44}    &{67.63}   &{67.26} &{67.26}  \\ 
        \bottomrule
        
    \end{tabular}
\end{table}

\paragraph{Ablation For Group Emotion Datasets:} In Table~\ref{tab:heatmap_skt_results}, for GER datasets, the heatmap-based decoder achieves competitive and often superior performance compared to DETR. Table~\ref{tab:heatmap_skt_results} shows that on GAF-3.0, combining face and body structural representations consistently outperforms single-modality setups. With our CUSTom RESidual encoder, the model reaches $89.09\%$ accuracy, a gain of about $2\%$ over using either body or face alone. A similar trend is observed with ResNet50, achieving $89.92\%$ accuracy. Compared to DETR, the heatmap approach shows clear advantages, especially when DETR uses fewer than 100 queries ($Q<100$, see Table~\ref{tab:resnet_detr_SR}).  

On VGAF, the heatmap decoder also surpasses DETR, both with and without synthetic data augmentation. Without augmentation, the model achieves $79.24\%$ (body) and $79.77\%$ (face) accuracy using ResNet50 as encoder. With synthetic data added, performance further improves to $80.68\%$ (body) and $80.81\%$ (face). However, combining body and face structural representations does not lead to additional gains, suggesting that the alignment of modalities is less effective for this dataset.  

\paragraph{Ablation For Non-Group Emotion Datasets:} In Table~\ref{tab:heatmap_skt_results}, for non-GER datasets, the heatmap decoder also demonstrates improvements over DETR. On DFEW, accuracy rises to $70.73\%$, approximately $1.5\%$ over DETR. For SAMSEMO, the best result is $75.10\%$ when combining body and face structural representations. The key advantage of the heatmap approach, we don't need to fix in advance the number of persons to detect; it considers all individuals present in the frame, leading to more complete representations. Importantly, all results remain substantially better than ViT-only or ViTVE baselines, confirming the contribution of VE-MD to enriching the latent space.

In summary, these results reinforce the fact that the influence of structural projection in VE-MD is highly dataset-dependent. For GER datasets, preserving raw structural information is essential to capture group interactions and contextual dependencies. In contrast, for non-GER datasets, projection enhances generalization by simplifying the latent representation and suppressing noisy or redundant structural cues. This dual behavior highlights the importance of adapting latent-space dimensionality and fusion strategies to the social complexity of the target data.

\subsubsection{Heatmap Approach Analysis}
\begin{figure}[H]
  \centering
  \includegraphics[width=0.99\linewidth]{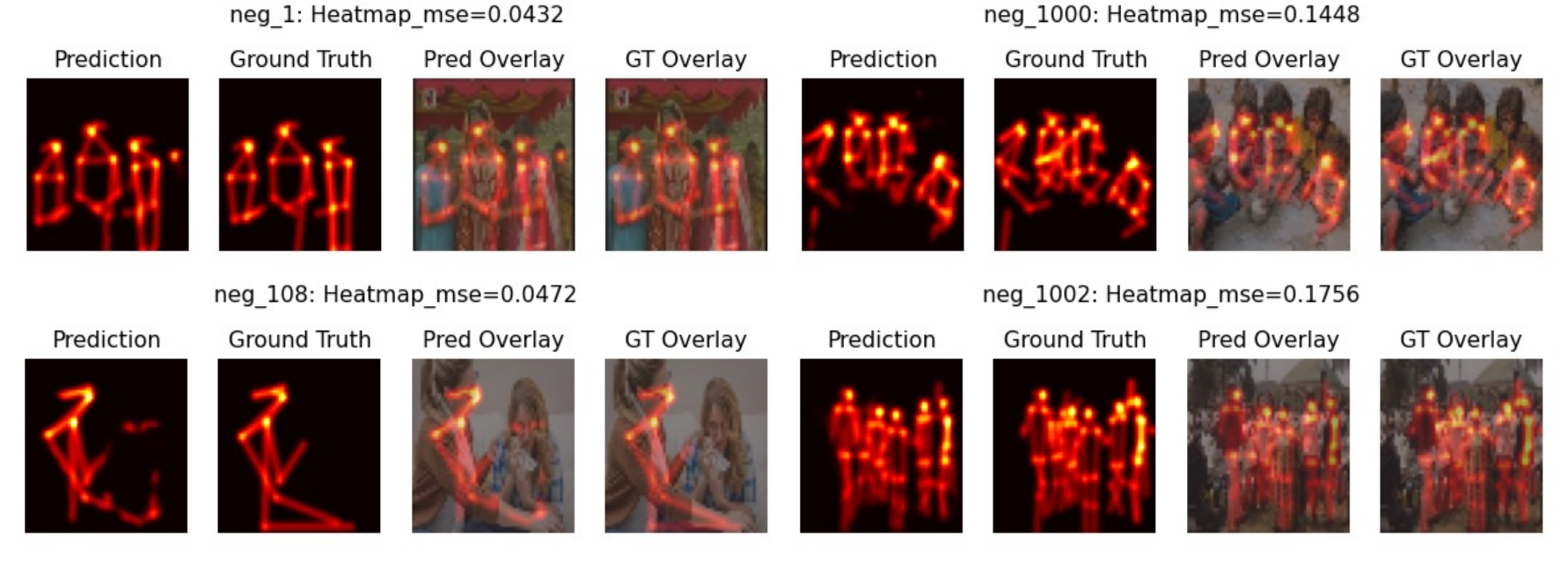}
  \includegraphics[width=0.99\linewidth]{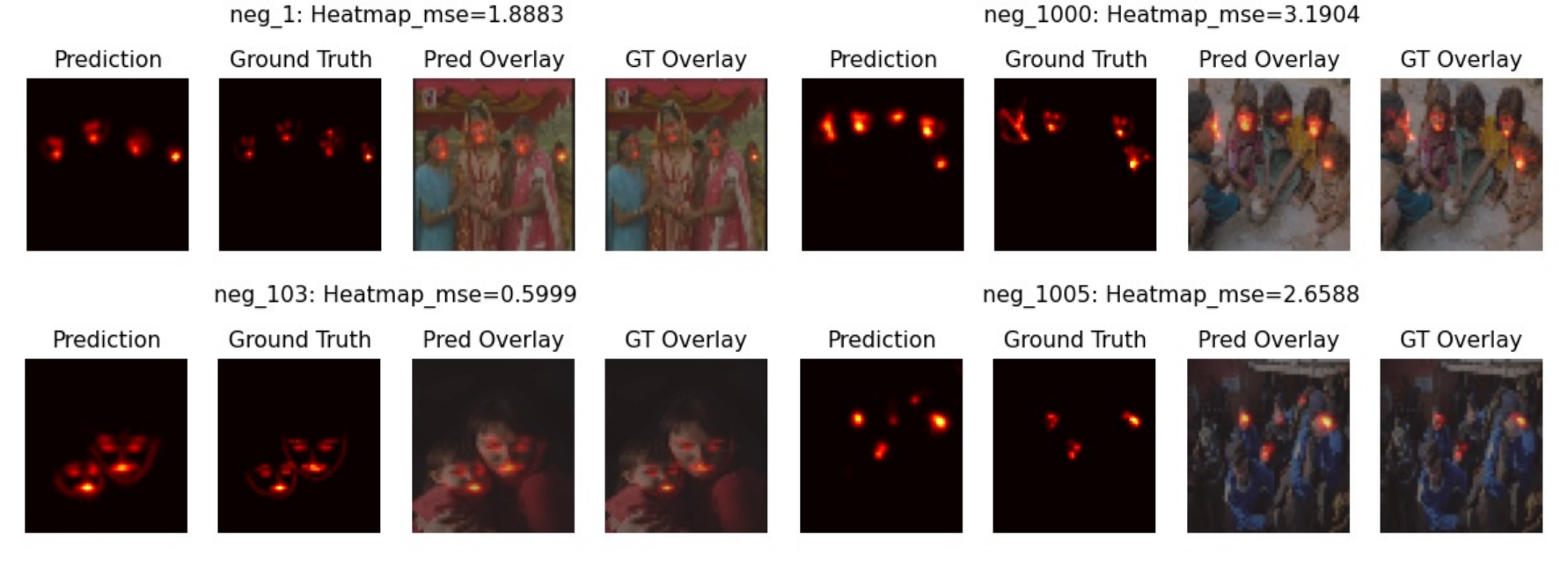}
 \caption{Predicted structural representations with Heatmap estimation on the GAF-3.0 dataset. Unlike DETR, the heatmap-based approach adapts to the number of persons in each image. Notably, the model often detects structural representations not annotated in the ground truth, reflecting higher sensitivity. For example, in image $neg\_1005$, five face structural representations are detected compared to three in the ground truth; in $neg\_1$, four body structural representations are detected compared to three labeled; and in $neg\_108$, a second person is detected despite being unlabeled. Additional examples are provided in Appendix~\ref{sec:pred_skt_examples}.}
  \label{fig:pred_skt_heatmap_gaf3}
\end{figure}

As discussed previously, the DETR-based decoder is constrained by the fixed number of queries, which limits the number of structural representations that can be predicted. To overcome this limitation, we employ a structural representation decoder based on heatmap pose estimation in the style of OpenPose. Figure~\ref{fig:pred_skt_heatmap_gaf3} shows examples of predicted structural representations using this approach. Unlike DETR, the VE-MD heatmap decoder automatically accounts for all persons present in an image, regardless of their number.

 Since the heatmap approach predicts all structural representations present in an image. This makes it particularly effective for datasets like VGAF, where the number of persons varies across frames. Nevertheless, the approach still inherits the ground-truth annotation inconsistencies described in section~\ref{sec:datasets_annotation}, which limit its stability in video datasets. On other datasets, performance is broadly comparable to DETR, but a key challenge for group-level emotion recognition remains: not all individuals in a group express the same emotion. Treating all structural representations equally may dilute the model’s ability to infer the collective group affect. This suggests that future improvements should incorporate reasoning mechanisms that weigh the contribution of individual structural representations toward the group label. One promising direction is to integrate Multimodal Large Language Models (MLLMs) with pose-based structural representation estimation reasoning, enabling the model to reason explicitly about group dynamics and predict group emotions more effectively in the wild. This observation will be part of our perspective in the conclusion of the thesis.

\subsubsection{Global Comparison for DETR-based and Heatmap-based Approaches}
\label{sec:global_conclusion_detr_heatmap}

Across both the DETR- and Heatmap-based variants of VE-MD, several consistent observations emerge. Both models confirm the central hypothesis that integrating structural representation tasks, body pose, and facial landmark prediction within a multi-decoder variational framework enriches the latent space and enhances emotion recognition performance.

\begin{figure}[H]
  \centering
  \includegraphics[width=0.95\linewidth]{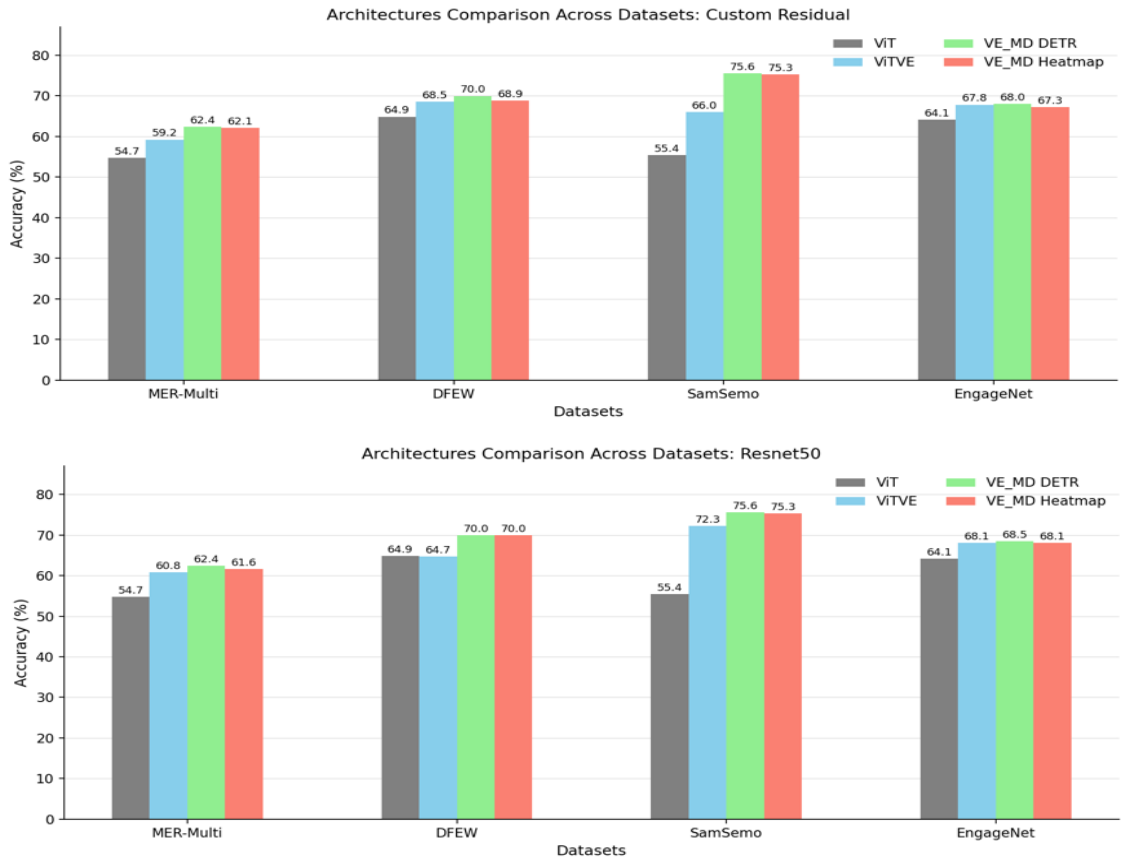}
 \caption{All comparisons are made with ViT, which is considered the baseline for the VE-MD. And all structural representation approaches are compared with the ViTVE. When not using the output structural representation in the emotion decoder, the model approach contributes significantly compared to VITVE. However, the model does not contribute to EngageNet performance prediction compared to ViTVE.}
 \label{fig:graph_comparison_noskt}
\end{figure}

\begin{figure}[H]
  \centering
  \includegraphics[width=0.95\linewidth]{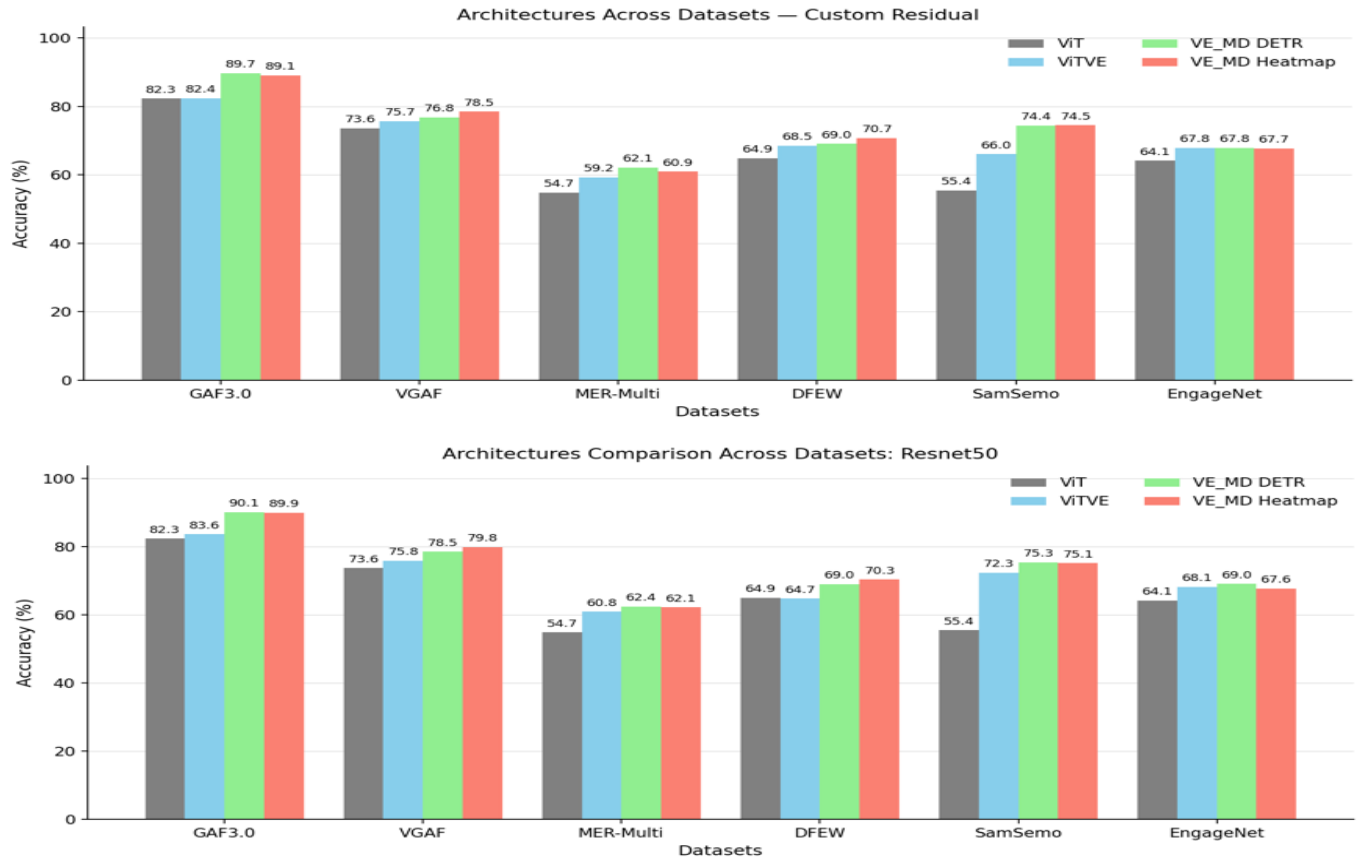}
 \caption{All comparisons are made with ViT, which is considered the baseline for the VE-MD. And all structural representation approaches are compared with the ViTVE. In all architectures, the structural representation approach contributes significantly compared to VITVE. However, in the first row (our CUSTom RESidual as encoder), the model does not contribute to EngageNet performance prediction compared to ViTVE. In the second row (Resnet50 as encoder) there is a slight difference compared to ViTVE.}
  \label{fig:graph_comparison_skt}
\end{figure}

The DETR-based decoder demonstrated the effectiveness of joint structural–affective learning but was constrained by its fixed query design, which limited its capacity to model variable group sizes in in-the-wild data. In contrast, the Heatmap-based decoder overcame this limitation by predicting all visible structural configurations through dense limb-connection maps, automatically adapting to any number of individuals per frame. This flexibility translated into higher robustness and improved accuracy on group emotion recognition (GER) datasets, notably on VGAF and competitive on GAF-3.0. It is shown in Figure~\ref{fig:graph_comparison_skt} and Figure~\ref{fig:graph_comparison_noskt} that DETR and Heatmap models achieved competitive results compared to ViT-only and ViTVE baselines, proving that the learned latent space captures rich affective cues. However, as illustrated in Figure~\ref{fig:graph_comparison_noskt}, this latent-only configuration underperformed for GER datasets, confirming that explicit structural cues remain crucial for modeling collective emotions. In contrast, for non-GER datasets, removing structural inputs often improved generalization and privacy preservation. It is worth noting that the Heatmap-based performance is not always better than the DETR-based approach performance. The conclusion is that, in all cases, the two architectures outperform the ViT-only and ViTVE as shown in the Figure~\ref{fig:graph_comparison_noskt}. and Figure~\ref{fig:graph_comparison_skt}. 

Figure~\ref{fig:graph_comparison_noskt} presents the results obtained when structural representations are \textbf{not} incorporated into the emotion decoder. As shown, both VE-MD architectures based on DETR and Heatmap structural decoders significantly outperform the ViT-only and ViTVE baselines. The largest improvement is observed on the SAMSEMO dataset, with a gain of +20.02 percentage points over ViT-only. Similar trends are observed for MER-MULTI, DFEW, and EngageNet, with respective gains of +7.70\%, +5.10\%, and~+4.40\%.

Figure~\ref{fig:graph_comparison_skt} presents the results obtained when structural representations are incorporated into the emotion decoder. As shown, both VE-MD architectures based on DETR and Heatmap structural decoders significantly outperform the ViT-only and ViTVE baselines. The largest improvement is observed on the SAMSEMO dataset, with a gain of +19.9 percentage points over ViT-only. Consistent performance gains are also observed on GAF-3.0, VGAF, MER-MULTI, DFEW, and EngageNet, with respective improvements of +8.4\%, +6.2\%, +7.7\%, +5.8\%, and +4.9\%.

The visual VE-MD framework successfully captures structural and affective patterns through both DETR- and Heatmap-based decoding strategies. Nevertheless, emotional expression in real-world interactions extends beyond visual signals alone. Auditory and linguistic information often convey complementary affective cues that are essential for a complete understanding of group emotion. Building upon the visual foundations established in this chapter, the next subsection extends VE-MD to a multimodal framework that integrates audio and, when available, text information for joint affective inference.

\section{Multimodal VE-MD with Audio and/or Text}

Building upon the visual VE-MD architecture developed in the chapter~\ref{chap:ve-md}, the framework is now extended to the multimodal domain. While the visual latent space enhanced through structural representations captures rich affective cues, emotion in natural environments also emerges from auditory and linguistic signals. To achieve a more comprehensive and realistic understanding of group affect, we integrate audio and, when available, text modalities into the VE-MD model.

\begin{figure}[H]
  \centering
  \includegraphics[width=0.95\linewidth]{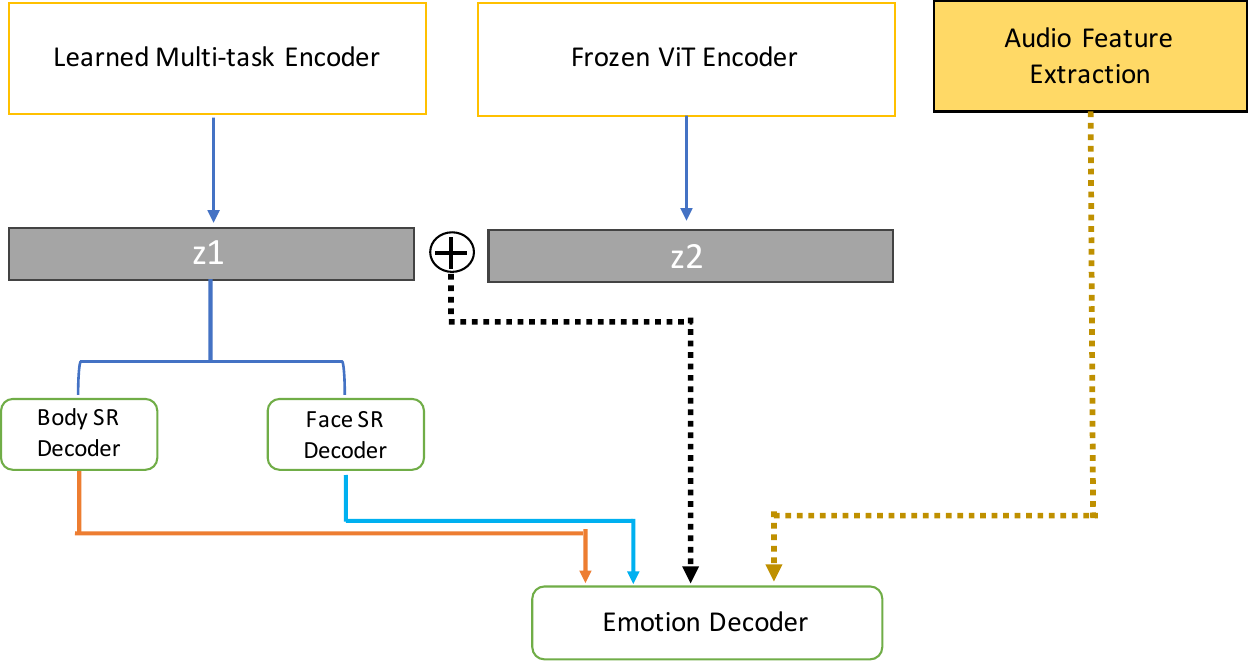}
  \caption{Multimodal combination with audio. Features from the two VE-MD latent spaces are concatenated with structural representations (SR) outputs and audio features before entering the emotion decoder.}
  \label{fig:audio_combined_vemd}
\end{figure}

 Among the datasets used in this thesis (see section~\ref{sec:datasets}), three are multimodal: VGAF (video+audio), SAMSEMO (video+audio+text), and MER-MULTI (MER-2023) (video+audio). To integrate audio (and optionally text) with VE-MD, we add a branch to the emotion decoder. As shown in Figure~\ref{fig:audio_combined_vemd}, the audio branch is an encoder that extracts features from either a pretrained foundation model or our own audio architecture.

We consider three kinds of audio models: (i) \emph{content} models (ASR-style), (ii) \emph{acoustic} representation models, and (iii) our own CNN--Transformer audio model. Intuitively, content encoders capture lexical/semantic cues that correlate with emotion, whereas acoustic encoders capture paralinguistic cues directly from the signal. Our model allows comparison with the architecture used in the previous chapter, tailored to our task rather than relying solely on off-the-shelf pretraining.

For content, we use Whisper~\citep{radford2023robust} (medium). For acoustic representations, we use WavLM~\citep{chen2022wavlm} and Wav2Vec~2.0~\citep{baevski2020wav2vec}. We also reuse our CNN--Transformer audio architecture from Chapter~\ref{chap:mger}. %

\paragraph{Audio Encoders:}
\emph{Whisper}~\citep{radford2023robust} is a transformer-based multilingual, multitask ASR system trained at scale; it provides robust features under diverse, noisy conditions and is a strong baseline for content-aware audio understanding. \emph{WavLM}~\citep{chen2022wavlm} is a transformer encoder pre-trained with masked prediction and denoising objectives to yield strong acoustic speech representations useful for ASR, speaker, and emotion tasks. \emph{Wav2Vec~2.0}~\citep{baevski2020wav2vec} learns contextualized speech features self-supervised from raw waveforms via contrastive learning and a transformer encoder, enabling strong results with limited labels on speech and paralinguistic tasks.

\subsection{Fusion Strategy}

\begin{figure}[H]
  \centering
  \includegraphics[width=0.9\linewidth]{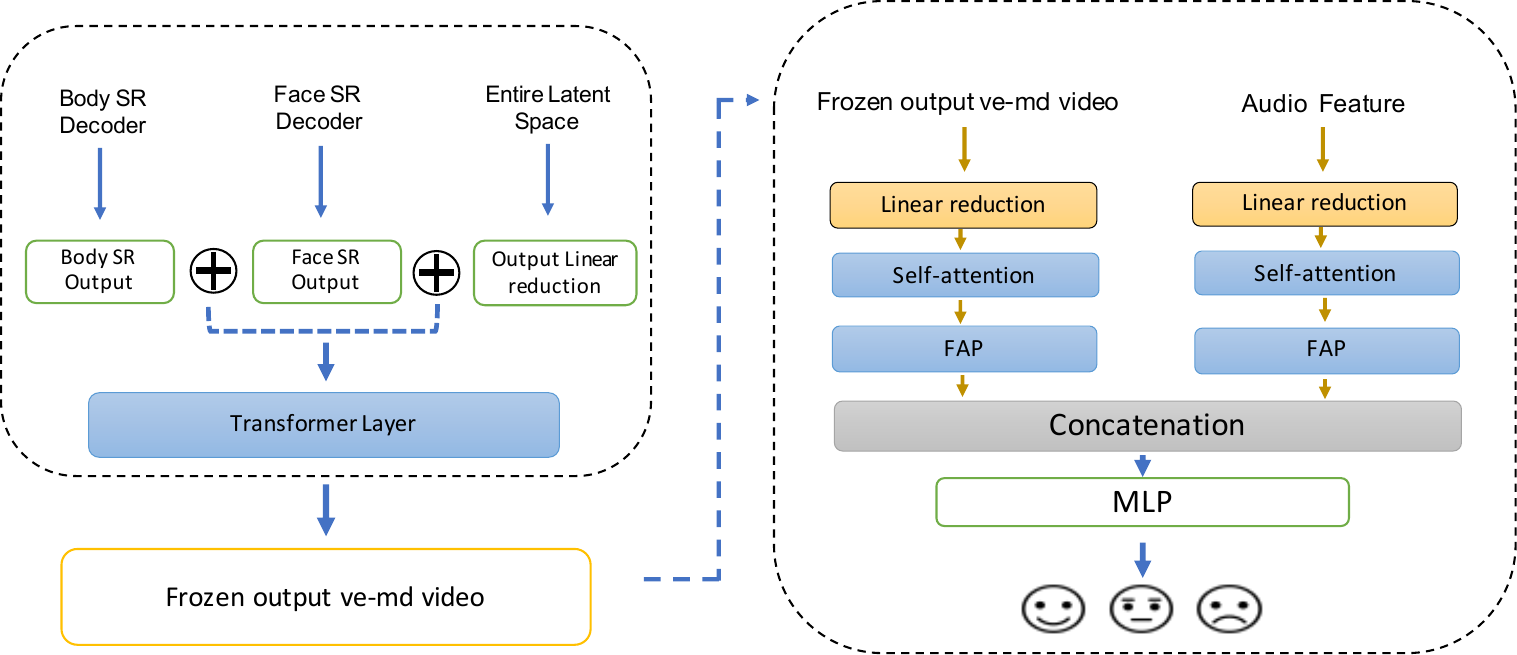}
  \caption{\textit{Multimodal classification head.} Outputs from VE\_MD (video branch) and the audio encoder pass through self-attention and learned Frames Attention Pooling (FAP), then are concatenated and projected by an MLP for emotion classification. See Section~\ref{sec:fap} for the frames attention pooling details.}
  \label{fig:classification_multimodal}
\end{figure}

To integrate the audio and VE\_MD branches, we design a multimodal classification head (Figure~\ref{fig:classification_multimodal}) supporting several fusion configurations. Our final design uses \textit{late fusion}: each branch is first linearly projected to a common dimension, followed by multi-head self-attention and a learned attention-pooling layer for sequence reduction. The fused representation is then fed to an MLP for classification.

In addition to late fusion, we evaluate \textit{cross-attention} and \textit{Attention-Guided Fusion (AFG)}~\citep{lian2019conversational}. For SAMSEMO, which includes text, we extract features using Llama~2~\citep{touvron2023llama} and Mixtral~\citep{jiang2024mixtral}, and project text embeddings to 1024 dimensions for alignment with other branches.

\paragraph{Single audio encoder.}
With one audio encoder, we test:
\begin{itemize}
  \item \textbf{Late fusion:} concatenate audio and video embeddings after sequence reduction.
  \item \textbf{Cross-attention:} let audio and video interact via
  \[
    \mathrm{Att}_{AV} = \mathrm{Attention}(A, V, V),
  \]
  where $A$ is the audio branch and $V$ is the VE\_MD video branch.
\end{itemize}
The final representation is $\mathrm{Concat}(A, \mathrm{Att}_{AV}, V)$.

\paragraph{Two audio encoders.}
With two audio encoders one \textbf{acoustic} ($A$) and one \textbf{content} ($C$) we apply bidirectional cross-attention:
\[
  \mathrm{Att}_{AC} = \mathrm{Attention}(A, C, C), \quad
  \mathrm{Att}_{CA} = \mathrm{Attention}(C, A, A).
\]
We then concatenate $A$, $C$, $\mathrm{Att}_{AC}$, and $\mathrm{Att}_{CA}$, and fuse this enriched audio representation with the video branch via late fusion.

\subsection{Training and Experimentation}

Each audio branch is first trained independently. Audio features are projected to 1024 dimensions (linear layer + normalization), then classified with an MLP. Our CNN--Transformer audio model is trained from scratch. Because audio sequences are typically longer than video, we use adaptive pooling to match the audio sequence length to the number of video frames. For audio–audio cross-attention experiments, we fix the target sequence length to 128. For SAMSEMO text features, we use \texttt{Mixtral-8x7B-Instruct-v0.1} and \texttt{Llama-2-7b-chat-hf}; both produce 4096-dimensional embeddings, which are projected to 1024 for alignment.

\begin{table}[H]
    \centering
    \caption{Audio-only accuracy (\%) for different encoders. ``Cross Attention Two Encoders'' uses Whisper (content) and Wav2Vec~2.0 (acoustic).}
    \label{tab:audio_experiments}
    \resizebox{\linewidth}{!}{%
    \begin{tabular}{@{}l|cccc|c@{}}
        \toprule
        \textbf{Datasets} & \multicolumn{4}{c|}{\textbf{One-Branch Audio Encoder / Architecture}} & \multicolumn{1}{c}{\textbf{Cross Attention Two Encoders}} \\
        \cmidrule{2-6}
        & \textbf{CNN--Transformer} & \textbf{Whisper} & \textbf{WavLM} & \textbf{Wav2Vec~2.0} & \textbf{Whisper \& Wav2Vec~2.0} \\
        \midrule
        VGAF      & 56.40 & 65.79 & 51.04 & 58.49 & \textbf{69.45} \\
        SAMSEMO   & 55.62 & 64.44 & 59.77 & 61.68 & \textbf{74.78} \\
        MER-MULTI & 33.82 & 59.61 & 33.82 & 44.52 & \textbf{63.75} \\
        \bottomrule
    \end{tabular}
    }
\end{table}

\subsection{Discussion and Analysis}

\paragraph{Audio-only:}
Table~\ref{tab:audio_experiments} shows that, with a single audio branch, Whisper yields the best accuracy on VGAF, SAMSEMO, and MER-MULTI (65.79\%, 64.44\%, and 59.61\%, respectively), followed by Wav2Vec~2.0 (58.49\%, 61.68\%, 44.52\%). WavLM and our CNN--Transformer perform lower. As expected, combining \emph{content} and \emph{acoustic} branches with cross-attention substantially improves audio-only performance across datasets (69.45\%, 74.78\%, 63.75\%).

\paragraph{Audio+video fusion:}
For fusion, VE\_MD is frozen, and we select the best VE\_MD video performance for each dataset as the video branch. We observe that the video encoder choice matters for complementarity: the video features that are strongest alone do not always combine best with a given audio encoder, likely because the two branches sometimes capture overlapping (non-complementary) cues, or interfere under certain fusion strategies.

\paragraph{VGAF:} In Table~\ref{tab:multimodal_ve_md_vgaf}, the best VE\_MD video accuracies are 80.81\% (face structural representations) and 80.68\% (body). Late fusion with Wav2Vec~2.0 reaches 81.85\%. Combining AFG with the late fusion further improves results: Wav2Vec~2.0+AFG yields the best overall 82.25\%, with WavLM +late-fusion +AFG close at 82.11\%. The two-audio-encoder setup does not surpass this, suggesting that, for VGAF, acoustic cues (and AFG) dominate the gains, consistent with the dataset’s noisy, in-the-wild audio.

\begin{table}[H]
\centering
\caption{Multimodal VGAF accuracy (\%). AFG: Attention-Guided Fusion.}
\label{tab:multimodal_ve_md_vgaf}
\resizebox{\linewidth}{!}{
\begin{tabular}{lcccc}
\toprule
\textbf{Audio Encoder} & \textbf{Fusion Strategy} & \textbf{Acc. Audio} & \textbf{Acc. Video} & \textbf{Acc. Fusion} \\
\midrule
WavLM           & late fusion         & 51.04 & 80.55 & 81.59 \\
Wav2Vec~2.0     & late fusion         & 58.49 & 80.55 & 81.85 \\
Whisper         & late fusion         & 65.79 & 80.55 & 81.20 \\
CNN--Transformer & late fusion        & 56.04 & 80.55 & 81.33 \\
\midrule
WavLM           & late fusion + AFG   & 51.04 & 80.55 & \textbf{82.11} \\
Wav2Vec~2.0     & late fusion + AFG   & 58.49 & 80.55 & \textbf{82.25} \\
Whisper         & late fusion + AFG   & 65.79 & 80.55 & 81.59 \\
CNN--Transformer & late fusion + AFG  & 56.04 & 80.55 & 81.62 \\
\midrule
WavLM           & cross-attention     & 51.04 & 80.55 & 81.33 \\
Wav2Vec~2.0     & cross-attention     & 58.49 & 80.55 & 81.59 \\
Whisper         & cross-attention     & 65.79 & 80.55 & 81.46 \\
CNN--Transformer & cross-attention    & 56.04 & 80.55 & 81.46 \\
\midrule
Wav2Vec~2.0 + Whisper & cross-attention + late fusion & 65.79 & 80.55 & 81.59 \\
\bottomrule
\end{tabular}
}
\end{table}

\begin{table}[H]
\centering
\caption{Multimodal MER-MULTI accuracy (\%).}
\label{tab:multimodal_ve_md_mer}
\resizebox{\linewidth}{!}{
\begin{tabular}{lcccc}
\toprule
\textbf{Audio Encoder} & \textbf{Fusion Strategy} & \textbf{Acc. Audio} & \textbf{Acc. Video} & \textbf{Acc. Fusion} \\
\midrule
WavLM           & late fusion         & 33.82 & 62.38 & 60.83 \\
Wav2Vec~2.0     & late fusion         & 44.52 & 62.38 & 60.84 \\
Whisper         & late fusion         & 59.61 & 62.38 & \textbf{63.02} \\
CNN--Transformer & late fusion        & 38.93 & 62.38 & 60.10 \\
\midrule
WavLM           & late fusion + AFG   & 33.82 & 62.38 & 60.34 \\
Wav2Vec~2.0     & late fusion + AFG   & 44.42 & 62.38 & 60.35 \\
Whisper         & late fusion + AFG   & 59.61 & 62.38 & 62.77 \\
\midrule
Wav2Vec~2.0     & cross-attention     & 44.52 & 62.38 & 60.10 \\
Whisper         & cross-attention     & 44.52 & 62.38 & 61.55 \\
\midrule
Wav2Vec~2.0 + Whisper & cross-attention + late fusion & 63.75 & 62.38 & \textbf{64.23} \\
\bottomrule
\end{tabular}
}
\end{table}

\paragraph{MER-MULTI:} In table~\ref{tab:multimodal_ve_md_mer},  Whisper aligns best with the video branch under late fusion ($63.02\%$), whereas AFG and cross audio–video attention do not help further. However, combining content+acoustic audio (Whisper+Wav2Vec~2.0) with cross-attention, then late fusing with video, gives the best $64.23\%$.

\paragraph{SAMSEMO:} In Table~\ref{tab:multimodal_ve_md_samsemo_1}, Whisper again pairs best with video: late fusion reaches 76.21\% and late+AFG gives 76.31\%. Using content and acoustic audio (Wav2Vec~2.0+Whisper with cross–audio attention) and late fusion with video yields 78.07\%. Table~\ref{tab:multimodal_ve_md_samsemo_2} includes five translated text variants (EN, KO, PL, ES, DE). When we add text (via Llama~2 or Mixtral features) to the audio–video fusion, performance improves further. The best results are achieved with Mixtral features, yielding 79.18\% accuracy and 77.94\% weighted F1.

\begin{table}[H]
\centering
\caption{Multimodal SAMSEMO accuracy (\%).}
\label{tab:multimodal_ve_md_samsemo_1}
\resizebox{\linewidth}{!}{
\begin{tabular}{lcccc}
\toprule
\textbf{Audio Encoder} & \textbf{Fusion Strategy} & \textbf{Acc. Audio} & \textbf{Acc. Video} & \textbf{Acc. Fusion} \\
\midrule
WavLM           & late fusion         & 59.81 & 75.31 & 75.41 \\
Wav2Vec~2.0     & late fusion         & 61.68 & 75.31 & 74.97 \\
Whisper         & late fusion         & 64.44 & 75.31 & \textbf{76.21} \\
CNN--Transformer & late fusion        & 55.62 & 75.31 & 75.26 \\
\midrule
WavLM           & late fusion + AFG   & 59.81 & 75.31 & 75.41 \\
Wav2Vec~2.0     & late fusion + AFG   & 61.68 & 75.31 & 75.07 \\
Whisper         & late fusion + AFG   & 64.44 & 75.31 & \textbf{76.31} \\
\midrule
Wav2Vec~2.0     & cross-attention     & 61.68 & 75.31 & 75.50 \\
Whisper         & cross-attention     & 64.44 & 75.31 & 75.36 \\
\midrule
Wav2Vec~2.0 + Whisper & cross-attention + late fusion & 74.78 & 75.31 & \textbf{78.07} \\
\bottomrule
\end{tabular}
}
\end{table}

\begin{table}[H]
\centering
\caption{SAMSEMO: adding text features (per-language) to audio--video fusion. Llama and Mixtral are text-only or combined with audio+video.}
\label{tab:multimodal_ve_md_samsemo_2}
\resizebox{\linewidth}{!}{%
\begin{tabular}{@{}l|cc|cc|cc|cc@{}}
\toprule
 & \multicolumn{2}{c|}{\makecell[c]{Llama Features \\ (Text Only)}} 
 & \multicolumn{2}{c|}{\makecell[c]{Mixtral Features \\ (Text Only)}}  
 & \multicolumn{2}{c|}{Llama + Audio + Video}
 & \multicolumn{2}{c}{Mixtral + Audio + Video} \\
\cmidrule{2-9}
{Language} & \textbf{Acc.} & \textbf{F1} & \textbf{Acc.} & \textbf{F1} & \textbf{Acc.} & \textbf{F1} & \textbf{Acc.} & \textbf{F1} \\
\midrule
EN & 61.92 & -- & 64.49 & -- & 79.22 & 77.76 & 79.46 & 78.15 \\
KO & 59.63 & -- & 60.49 & -- & 79.22 & 77.78 & 78.41 & 77.46 \\
PL & 58.77 & -- & 61.82 & -- & 79.16 & 77.72 & 79.22 & 77.94 \\
ES & 60.48 & -- & 63.63 & -- & 79.22 & 77.63 & \textbf{79.55} & \textbf{78.24} \\
DE & 60.10 & -- & 63.06 & -- & 79.17 & 77.68 & 79.27 & 77.91 \\
\midrule
Average & 60.18 & -- & 62.69 & -- & 79.20 & 77.71 & \textbf{79.18} & \textbf{77.94} \\
\bottomrule
\end{tabular}
}
\end{table}

Finally, the multimodal extension of VE-MD with audio and text demonstrated the framework’s scalability beyond the visual domain, confirming its potential as a unified, privacy-preserving affect recognition architecture. Having validated the VE-MD framework across visual, audio, and multimodal configurations, we now benchmark its performance against existing state-of-the-art methods across all datasets.

\section{Comparisons with State-of-the-Art}
In this section, we compare VE\_MD with baselines and state of the art (SOTA) across all datasets.

\paragraph{GAF-3.0:} In Table~\ref{tab:gaf3_sota}, VE\_MD surpasses prior work by a clear margin: our best configuration (predicted body+face structural representation) improves accuracy by \textbf{+3.16 points} over the strongest prior method (90.06\% vs.\ 86.90\%). Even using a single predicted structural representation decoder (face \emph{or} body), VE\_MD exceeds SOTA, underscoring its effectiveness for group emotion recognition in images.

It is worth noting that Zhu et al.~(2025)~\citep{zhu2025adaptive} recently proposed the Key Role Guided Hierarchical Relation Inference (KR-HRI) model to enhance group-level emotion recognition. This method is not included in the comparison presented in Table~\ref{tab:gaf3_sota}, as it reports performance using a different evaluation metric, Unweighted Average Recall (UAR) rather than accuracy, which is commonly adopted by other studies. On the GAF-3.0 dataset, KR-HRI achieved a UAR of 81.29\%. In contrast, our best VE-MD Heatmap-based model reached a UAR of \textbf{88.70\%} on the same dataset, outperforming KR-HRI by \textbf{+7.41} percentage points.

\begin{table}[H]
  \footnotesize
  \centering
  \caption{Comparison on GAF-3.0. CAN: Cascade Attention Network; SR: Structural Representation.}
  \label{tab:gaf3_sota}
  \resizebox{\textwidth}{!}{%
  \begin{tabular}{lccccc}
    \toprule
    Year & Methodology & Ind.\ Features & Features & \textbf{Acc [\%]} & \textbf{Rank} \\
    \midrule
    2018 & ResNet, VGG~\cite{khan2018group} & {\checkmark} & Face, Skeleton & 78.39 & 9 \\
    2018 & DenseNet, SphereFace~\cite{gupta2018attention} & {\checkmark} & Face, Skeleton & 80.98 & 8 \\
    2018 & CAN, ResNet, SE-Net~\cite{wang2018cascade} & {\checkmark} & Face, Skeleton, Pose & 86.90 & 6 \\
    2025 & PSMF~\cite{huangpsmf2025} & {\checkmark} & Face, Scene & 83.58 & 7 \\ 
    \midrule
    2025 & VE-MD DETR \textit{(ours)} &  & Pred.\ Body SR & 87.99 & 5 \\
    2025 & VE-MD DETR \textit{(ours)} &  & Pred.\ Face SR & 88.29 & 4 \\
    2025 & VE-MD Heatmap \textit{(ours)} &  & Pred.\ Face SR & 89.60 & 3 \\
    2025 & VE-MD Heatmap \textit{(ours)} &  & Pred.\ Body+Face SR & 89.92 & 2 \\
    2025 & VE-MD DETR \textit{(ours)} &  & Pred.\ Body+Face SR & \textbf{90.06} & \textbf{1} \\
    \bottomrule
  \end{tabular}
  }
\end{table}

\paragraph{VGAF:} In Table~\ref{tab:vgaf_sota_2025}, VE\_MD also proves effective. Our multimodal approach (VE\_MD+ Wav2Vec~2.0) achieves \textbf{82.25\%} accuracy, improving over the strongest listed prior fusion (81.98\%) by \textbf{+0.27 points}.

\begin{table}[H]
  \footnotesize
  \centering
  \caption{Comparison on VGAF (since 2023). SR: Structural Representation.}
  \label{tab:vgaf_sota_2025}
  \resizebox{\textwidth}{!}{%
  \begin{tabular}{lcccccc}
    \toprule
    Year & Modalities & Methodology & Ind.\ Features & Features & \textbf{Acc [\%]} & \textbf{Rank} \\
    \midrule
    2023 & A   & CNN+Transformer~\cite{augusma2023multimodal}\textit{(ours)} &   & Acoustic & \textit{56.40} & 13 \\
    2023 & A,V & Cross Attention~\cite{augusma2023multimodal}\textit{(ours)} &   & Global Image & 78.72 & 9 \\
    2023 & V   & ViT + Synthetic data~\cite{augusma2023multimodal}\textit{(ours)} &   & Global Image & 79.24 & 8 \\
    
    2024 & A & \makecell[c]{Wav2Vec~2.0 +1D CNN~\cite{kumar2020object}} & {} & Acoustic & 64.09 & 12 \\
    
    2024 & V & \makecell[c]{TimeSformer, YOLOv8~\cite{kumar2020object}} & {\checkmark} & Body Pose & 73.10 & 10 \\
    2024 & A,V & \makecell[c]{Multimodal fusion \\ (TimeSformer, Wav2Vec~2.0, YOLOv8)~\cite{kumar2020object}} & {\checkmark} & Body Pose & 81.98 & 3 \\
    \midrule
    2025 & A   & Wav2Vec~2.0 + Whisper \textit{(ours)} &   & Acoustic, Content & 69.45 & 11 \\
    2025 & V   & VE-MD DETR \textit{(ours)} &   & Pred.\ Body SR & 80.42 & 7 \\
    2025 & V   & VE-MD Heatmap \textit{(ours)} &   & Pred.\ Body SR & 80.68 & 5 \\
    2025 & V   & VE-MD Heatmap \textit{(ours)} &   & Pred.\ Face SR & 80.81 & 4 \\
    2025 & V & VE-MD Heatmap \textit{(ours)} &   & Pred.\ Body+Face SR & 80.55 & 6 \\ %
    2025 & A,V & VE-MD + WavLM \textit{(ours)} &   & Pred.\ Body+Face SR & 82.11 & 2 \\
    2025 & A,V & VE-MD + Wav2Vec~2.0 \textit{(ours)} &   & Pred.\ Body+Face SR & \textbf{82.25} & \textbf{1} \\
    \bottomrule
  \end{tabular}
  }
\end{table}

\begin{table}[H]
  \footnotesize
  \centering
  \caption{Comparison on SAMSEMO. SR: Structural Representation.}
  \label{tab:samsemo_sota}
  \resizebox{\textwidth}{!}{%
  \begin{tabular}{lcccccc}
    \toprule
    Year & Modalities & Methodology & Ind.\ Features & Features & \textbf{F1-Score [\%]} & \textbf{Rank} \\
    \midrule
    2024 & A     & E2E~\cite{bujnowski2024samsemo} &  & -- & 61.10 & 9 \\
    2024 & T     & E2E~\cite{bujnowski2024samsemo} &  & -- & 63.00 & 8 \\
    2024 & V     & E2E~\cite{bujnowski2024samsemo} &  & -- & 68.20 & 7 \\
    2024 & A,V,T & E2E~\cite{bujnowski2024samsemo} &  & -- & 69.00 & 6 \\
    \midrule
    2025 & A     & Wav2Vec~2.0 + Whisper \textit{(ours)} &  & Acoustic + Content & 74.78 & 5 \\
    2025 & A,V   & Whisper + VE\_MD DETR \textit{(ours)} &  & Pred.\ Body SR & 74.97 & 4 \\
    2025 & A,V   & Wav2Vec~2.0 + Whisper + VE\_MD DETR \textit{(ours)} &  & Pred.\ Body SR & 76.68 & 3 \\
    2025 & A,V,T & \makecell[c]{Wav2Vec~2.0 + Whisper + VE\_MD DETR \\ + Llama2 Feat.\ \textit{(ours)}} &  & Pred.\ Body SR & 77.71 & 2 \\
    2025 & A,V,T & \makecell[c]{Wav2Vec~2.0 + Whisper + VE\_MD DETR \\ + Mixtral Feat.\ \textit{(ours)}} &  & Pred.\ Body SR & \textbf{77.94} & \textbf{1} \\
    \bottomrule
  \end{tabular}
  }
\end{table}

\paragraph{SAMSEMO:} In Table~\ref{tab:samsemo_sota}, VE\_MD substantially outperforms the baseline for the tri-modal setting (A, V, T): our best configuration (Wav2Vec~2.0 + Whisper + VE\_MD + Mixtral features) reaches a weighted F1 of \textbf{77.94\%}, exceeding the E2E baseline’s 69.00\% by \textbf{+8.94 points} and establishes the SOTA on SAMSEMO.

\paragraph{MER-MULTI:} In Table~\ref{tab:mermulti_sota}, VE\_MD attains 63.80\% (F1-0.25MSE), which is below the current SOTA of 70.05\%. We attribute the gap primarily to heavy, multi-level data augmentation and complex multi-branch supervision used by \citep{zong2023building}, as well as hierarchical audio-video modeling and additional face streams with Attention Guided Fusion (AFG) in \citep{wang2023hierarchical}, which differ from our streamlined, shared-latent VE\_MD design.

\begin{table}[H]
  \footnotesize
  \centering
  \caption{Comparison on MER-MULTI. SR: Structural Representation.}
  \label{tab:mermulti_sota}
  \resizebox{\textwidth}{!}{%
  \begin{tabular}{lcccccc}
    \toprule
    Year & Modalities & Methodology & Ind.\ Features & Features & \textbf{F1-0.25MSE [\%]} & \textbf{Rank} \\
    \midrule
    2023 & A,V & Baseline~\cite{lian2023mer} &  & -- & 56.00 & 4 \\
    2023 & A,V & JDEV, HuBERT~\cite{wang2023hierarchical} & {\checkmark} & Face stream & 68.46 & 2 \\
    2023 & A,V & \makecell{Weighted blending of \\ supervision signals~\cite{zong2023building}} &  & Feature augmentation & \textbf{70.05} & \textbf{1} \\
    \midrule
    2025 & A,V & Wav2Vec~2.0 + Whisper + VE\_MD \textit{(ours)} &  & Pred.\ Body+Face SR & 63.80 & 3 \\
    \bottomrule
  \end{tabular}
  }
\end{table}

\begin{table}[H]
  \footnotesize
  \centering
  \caption{Comparison on EngageNet. TCCT-Net: Tensor-Convolution and Convolution-Transformer Network; AU: Action Units; ST-GCN: Spatiotemporal Graph Convolutional Networks; SR: Structural Representation.}
  \label{tab:engagenet_sota}
  \resizebox{\textwidth}{!}{%
  \begin{tabular}{lcccrc}
    \toprule
    Year & Methodology & Ind.\ Features & Features & \textbf{Acc [\%]} & \textbf{Rank} \\
    \midrule
    2023 & Transformer~\cite{singh2023have} & {\checkmark} & Gaze & 55.45 & 7 \\
    2023 & Transformer~\cite{singh2023have} & {\checkmark} & Gaze + Head Pose & 64.45 & 6 \\
    2023 & Transformer~\cite{singh2023have} & {\checkmark} & Gaze + Head Pose + AU & 69.10 & 2 \\
    2024 & GLAMOR-Net~\cite{anand2024exceda} & {\checkmark} & Facial features & 68.72 & 5 \\
    2024 & TCCT-Net~\cite{Vedernikov_2024_CVPR} & {\checkmark} & Head Pose & 68.91 & 4 \\
    2024 & ST-GCN~\cite{abedi2024engagement} & {\checkmark} & Facial landmarks & \textbf{71.24} & \textbf{1} \\
    \midrule
    2025 & VE\_MD \textit{(ours)} &  & Pred.\ Body+Face SR & 68.98 & 3 \\
    \bottomrule
  \end{tabular}
  }
\end{table}

 \paragraph{EngageNet:} In Table~\ref{tab:engagenet_sota}, VE\_MD achieves 68.98\% accuracy versus the SOTA of 71.24\%. Methods surpassing us rely on explicit individual features as inputs (e.g., facial landmarks, head pose, AUs) computed beforehand and fed directly into their models, whereas VE\_MD operates from global images and predicted structural representations without using such pre-extracted individual cues. Notably, our performance is comparable to \citet{singh2023have} (69.10\%) despite their direct use of individual features.

 \paragraph{DFEW:} In Table~\ref{tab:dfew_sota}), VE\_MD does not reach SOTA (76.21\%). Top results leverage large vision-language models (e.g., CLIP variants, MLLMs like LLaVA), specialized face encoders (e.g., FaceXFormer), and sophisticated test-time similarity calculations, which are computationally intensive. In contrast, VE\_MD maintains a simpler, shared-latent formulation with an intermediate face-structural representation decoder.

\begin{table}[H]
  \footnotesize
  \centering
  \caption{Comparison on DFEW (video); SR: Structural Representation.}
  \label{tab:dfew_sota}
  \resizebox{\textwidth}{!}{%
  \begin{tabular}{lccccc}
    \toprule
    Year & Methodology & Ind.\ Features & Features & \textbf{WAR [\%]} & \textbf{Rank} \\
    \midrule
    2024 & \makecell{EmoCLIP \\ CLIP-ViT-B/32~\cite{foteinopoulou2024emoclip}} & {\checkmark} & Face & 62.12 & 8 \\
    2024 & OUS, CLIP~\cite{mai2024ous} & {\checkmark} & Face & 68.85 & 7 \\
    2024 & LSGT, ResNet-18~\cite{wang2024joint} & {\checkmark} & Face & 72.34 & 4 \\
    2024 & \makecell{UMBEnet, CLIP~\cite{mai2024all}} & {\checkmark} & Brain, Face & 73.93 & 3 \\
    2024 & \makecell{Align-DFER, CLIP-ViT-L/14, \\ MLLM~\cite{tao2024align}} & {\checkmark} & \makecell[c]{Face analysis \& parsing, \\ Landmarks} & 74.20 & 2 \\
    2024 & \makecell{FineCLIPER, CLIP-ViT-L/16, \\ LLaVA, MLLM~\cite{chen2024finecliper}} & {\checkmark} & \makecell[c]{Face analysis \& parsing, \\ Landmarks} & \textbf{76.21} & \textbf{1} \\
    \midrule
    2025 & VE\_MD DETR \textit{(ours)} &  & Pred.\ Face SR & 69.98 & 6 \\
    2025 & VE\_MD Heatmap \textit{(ours)} &  & Pred.\ Face SR & 70.73 & 5 \\ %
    \bottomrule
  \end{tabular}
  }
\end{table}

\noindent\paragraph{MER-MULTI discussion.}
Our best result (63.80\%) trails the 70.05\% SOTA. \citet{zong2023building} employs extensive data augmentation at image, waveform, and spectrogram levels and jointly trains modalities via a weighted blend of supervision signals, improving robustness. \citet{wang2023hierarchical} uses a hierarchical audio-video design combining low/mid/high-level features and dual face encoders with attention-guided features, which differs from our simpler VE\_MD pipeline.

\noindent\paragraph{EngageNet discussion.}
The 71.24\% SOTA uses explicit individual features (landmarks, head pose, AUs) directly as inputs. VE\_MD, which relies on global images and predicted structural representations (no per-person pre-extraction), still achieves a comparable 68.98\% and is close to the 69.10\% result of \citet{singh2023have}.

\noindent\paragraph{DFEW discussion.}
Top-performing methods exploit large vision-language models (CLIP backbones, MLLMs like LLaVA), specialized face encoders (e.g., FaceXFormer), and test-time matching strategies trading compute for accuracy. VE\_MD uses a lighter shared-latent framework with an intermediate face-structural representation decoder, which is more efficient but currently underperforms these heavy pipelines.

\begin{figure}[H]
  \centering
  \includegraphics[width=0.95\linewidth]{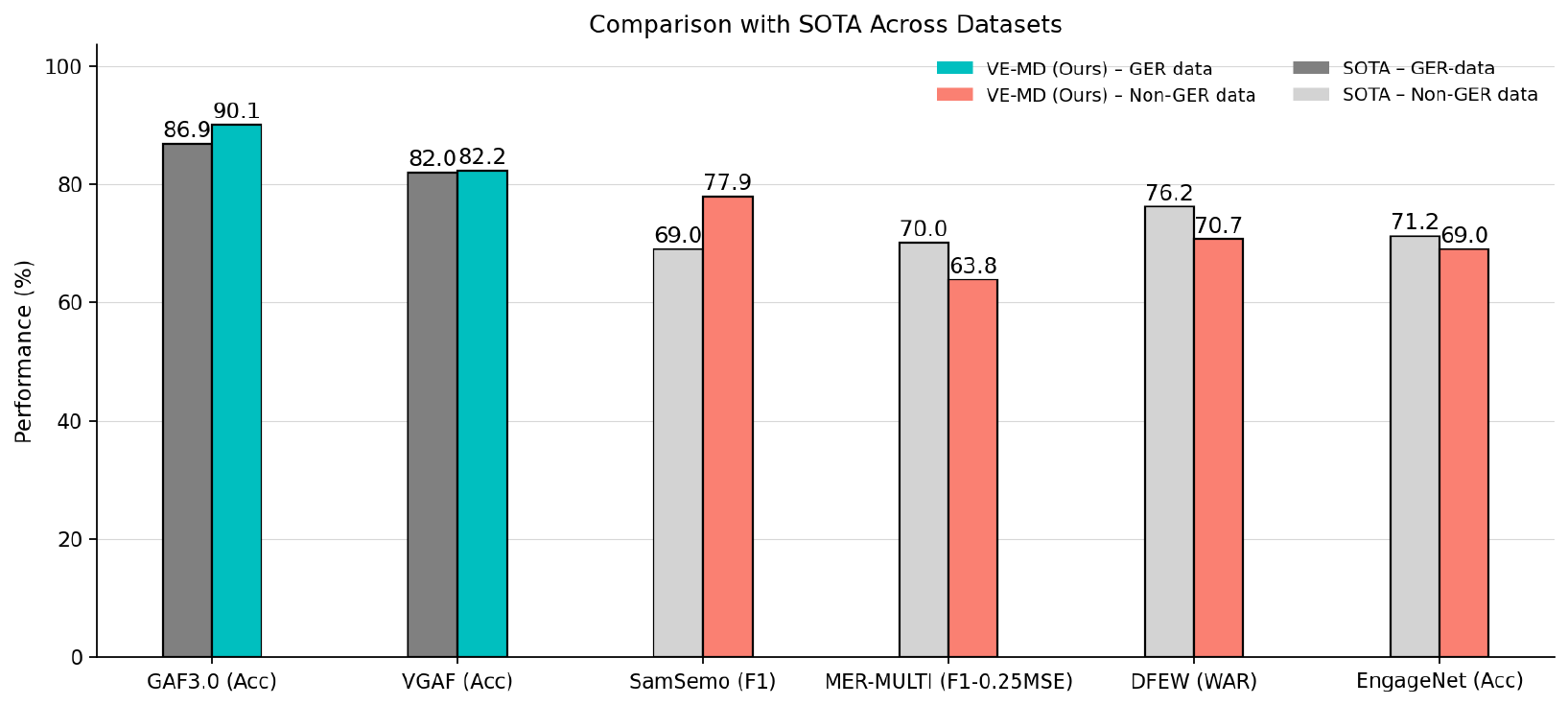}
  \caption{Summary of VE\_MD versus SOTA across datasets. Cyan/gray bars indicate GER datasets (VE\_MD vs.\ SOTA), and salmon/light-gray bars indicate non-GER datasets (VE\_MD vs.\ SOTA).}
  \label{fig:sota_graph}
\end{figure}

Overall, Figure~\ref{fig:sota_graph} summarizes the best results achieved by VE-MD in comparison with the state of the art. The proposed VE-MD framework establishes new SOTA performance on the GAF-3.0 and VGAF datasets and achieves strong results on SAMSEMO, while remaining competitive on EngageNet and DFEW. Notably, these outcomes are obtained with privacy-preserving methods and without relying on large multimodal language models (LLMs) stacks.

\section{Conclusion}

In this chapter, we introduced the Variational Encoder Multi-Decoder (VE-MD) framework for enhancing emotion recognition through multi-task learning. Motivated by the limitations of synthetic data and the need for richer contextual cues, the proposed approach leverages a shared latent space jointly optimized for emotion classification, body and face structural representation prediction. This design avoids using directly extracted (in advanced) individual features as input, thus offering a privacy-preserving solution while enriching the latent representation with complementary cues.  

From a methodological perspective, two structural representation decoder designs were investigated: a DETR-based decoder with spatio-temporal graph convolution (ST-GCN) for structured modeling, and a heatmap-based decoder for flexible in-the-wild scenarios. We also explored the role of structural representation integration within the emotion decoder, highlighting distinct behaviors between group emotion recognition (GER) and non-GER tasks: raw structural representations enhance GER performance by preserving inter-person interactions, whereas projection benefits non-GER datasets by acting as a denoising bottleneck. Extensive experiments across six datasets demonstrated the effectiveness of VE-MD. The framework achieved new state-of-the-art results on GAF-3.0, VGAF, and SAMSEMO, while remaining competitive on EngageNet, MER-MULTI, and DFEW. Ablation studies confirmed the contribution of structural representation-based features, the benefits of multimodal audio-video-text fusion, and the generality of the proposed latent-space formulation.  

We found that incorporating structural representation outputs in the emotion decoder significantly improves performance for Group Emotion Recognition (GER). When structural representation outputs are removed and only the shared latent space is used, accuracy on GER datasets decreases, indicating that structural representation cues remain critical for capturing inter-person interactions. This, however, reduces the robustness of the privacy-preserving aspect at the model output. It is important to note that the predicted structural representations represent the group rather than specific individuals, which still prevents the model from being used for individual-level monitoring or control. A key perspective for future work will be to design strategies that enable the latent space to encode sufficient interaction cues for GER classification without explicitly reusing structural representation outputs in the emotion decoder.

The contribution of this chapter establishes the value of shared latent spaces for multi-task affective computing, provides a detailed analysis of structural representation-based emotion cues, and demonstrates robust multimodal fusion strategies, achieving new state-of-the-art performance on several benchmarks. It introduces modified DETR and OpenPose-inspired decoders for end-to-end structural representation prediction, enabling the inference of intermediate group-level structural representations that enrich emotion recognition while preserving privacy.

\chapter{Conclusions and Perspectives}
\label{chap:conclusion}

This thesis contributes to the MANIP (Modeling and Analysis of Instructional Processes) project and, more broadly, to the Teaching Lab project. which aims to support teaching and learning through Context-Aware Classroom (CAC) analysis. Within this framework, we focused on developing \textbf{multimodal models for group-level emotion recognition in-the-wild}, emphasizing privacy preservation and ethical design. The objective was to infer collective emotional states from audio–visual cues without relying on explicit individual features provided in input.

Two complementary frameworks were proposed. The first integrates \textbf{cross-attention based audio–video fusion} enhanced with synthetic data augmentation to improve robustness in limited and imbalanced settings. The second, the \textbf{Variational Encoder Multi-Decoder (VE-MD)}, introduces a shared latent space jointly optimized for emotion recognition and auxiliary structural tasks (body pose and facial landmarks). Together, these frameworks advance group emotion recognition by (i) enriching multimodal feature representations, (ii) leveraging structural information in a privacy-preserving manner, and (iii) demonstrating scalability across visual, audio, and multimodal datasets.  

The following sections summarize the main contributions, discuss current limitations, and outline future research directions.

\section{Contributions}

In Chapter~\ref{chap:mger}, a privacy-preserving multimodal architecture for group emotion recognition is introduced.  
The \textbf{scientific contributions} include: (i) the design of an architecture that avoids reliance on individual-specific features in inputs, (ii) the introduction of a cross-attention mechanism for effective audio–video fusion, and (iii) the adaptation of Frames Attention Pooling (FAP), inspired by Attentive Statistic Pooling, for temporal aggregation. The proposed architecture \textbf{won} the competition at the EmotiW 2023 challenge.  

In Chapter~\ref{chap:ve-md}, we presented the Variational Encoder Multi-Decoder (VE-MD) framework for group and non-group emotion recognition.  
The {scientific contributions} include: (i) the proposal of a shared latent space jointly optimized for emotion classification, body and face structural representation prediction, (ii) a detailed analysis of structural representation integration in the emotion decoder, revealing contrasting effects for GER and non-GER datasets, and (iii) the extension of VE-MD to multimodal audio-video-text fusion. We designed a modified DETR  (with ST-GCN for temporal modeling) and an OpenPose-inspired structural representation decoders integrated into the emotion decoder.
Extensive experiments across six different datasets validated the framework. These datasets are categorized in GER and non-GER in-the-wild, across different events from street protesters to conference hosts. VE-MD achieves new state-of-the-art results on GAF-3.0, VGAF, and SAMSEMO, while remaining competitive on MER-MULTI, EngageNet, and DFEW.

\section{Limitations}

While this thesis achieved strong results and provided new insights into multimodal group emotion recognition, several limitations remain. The first concerns the balance between privacy preservation and performance in VE-MD. The shared latent space was designed to encode only intermediate body and face structural representations, ensuring that neither inputs nor outputs reveal individual-specific features. This design preserved privacy for non-GER datasets but proved less effective for complex group scenes, where inter-person relations tend to collapse in the latent space. To recover performance, explicit structural outputs were sometimes reintroduced, slightly reducing the privacy level. Although acceptable in collective scenarios, residual privacy risks remain when few individuals are present, as distinctive shapes or positions could indirectly be revealed by monitoring.

A second limitation involves the quality and consistency of dataset annotations. Body and face structural representations had to be generated using external models (ViTPose and FaceAlignment), which are imperfect and occasionally inconsistent across frames or modalities. Missing or misaligned detections between body and face annotations reduced reliability, especially in crowded or low-light conditions. Temporal inconsistencies where individuals appear in one frame but not in the next further introduced noise, hindering the modeling of inter-person dynamics. These annotation issues likely explain part of the remaining performance gap, particularly in video-based group datasets such as VGAF.

\section{Perspectives}

One key perspective is the improvement of dataset annotations. Manual annotation by humans would provide consistent face and body structural representation labels for the same individuals within an image and across video frames. Such consistency would help the model learn reliable inter-person relationships and alignments between body and face information. While manual annotation could improve consistency, it is not scalable to large datasets or real-world classroom settings due to the high cost and time requirements. For this reason, future work should explore semi-automatic annotation enhanced with reasoning from vision–language models.

\textbf{Vision-Language Models (VLMs)} could be leveraged to understand inter-person dynamics. For example, the keypoints-assisted method proposed by \citet{zhang2024keypoints} demonstrates how a vision language-driven approach can enrich pose representations and improve understanding of human interactions. Incorporating similar capabilities into VE-MD would strengthen temporal consistency and enable the model to interpret group-level emotions beyond the behavior of individual persons. This integration would help bridge the gap between visual group perception and higher-level semantic interpretation of collective affect.

An important perspective for addressing privacy limitations is to design strategies that preserve inter-person relationships and other structural cues directly within the shared latent space of VE-MD. Such approaches would allow the model to retain the discriminative power of structural representation information without explicitly reusing structural representation decoder outputs in the emotion decoder. Thus, enabling truly privacy-preserving group emotion recognition. Several technical directions could be explored. One promising approach is \textbf{knowledge distillation} with a teacher-student framework: during training, structural representation decoder outputs could be used as supervision by the teacher, while at inference time the student model would rely only on the latent space. Another direction is to \textbf{disentangle} privacy-sensitive appearance features ~\cite{zhou2023privacy, bortolato2020learning, malarkkan2025delta,jia2025latent} from group-level relational cues within the latent space. By introducing disentanglement losses, the model could separate individual characteristics (which are privacy sensitive) from collective interaction cues (which are needed). These strategies would allow VE-MD to achieve robust group emotion recognition while ensuring complete privacy at the model output.

In this thesis, we combined an acoustic audio encoder with a semantic (content-based) audio encoder to improve emotion prediction. This approach proved effective for the SAMSEMO dataset but performed less well on VGAF. The difference can be explained by the nature of the data downloaded from YouTube, which contains highly noisy scenarios such as protests, parties, large public events, or children singing while hitting chairs and furniture. In such environments, the semantic encoder aims to capture speech content, while the acoustic encoder models prosodic features such as frequency, tone, and timbre. However, when the audio is dominated by overlapping voices and broadband noise, the complementary benefits of these two encoders are reduced.  

Since group-level audio in the wild often contains uncontrolled background noise and competing sound sources, one promising future direction is the use of \textbf{audio separation techniques}. Similar to speech separation \cite{wang2018}, the idea would be to combine two encoders with latent disentanglement: one encoder focusing on the group’s speech signal, and the other on background sounds and noise. This separation could reduce interference, enhance the robustness of multimodal fusion, and improve the effectiveness of content and acoustic-based representations in noisy group environments. Comparison integration between the background noise and the clean speech could also be investigated to evaluate contribution of the noise.

In summary, this thesis proposed two complementary frameworks for multimodal group emotion recognition. While several challenges remain, particularly in privacy preservation, dataset quality, and audio robustness, the presented approaches provide a strong foundation for future work. We believe that extending VE-MD to VLM capabilities, improved annotations, and advanced audio separation techniques will open the path toward practical, privacy-aware systems for group-level affect recognition in real-world classrooms and beyond.

\appendix %

\chapter{Multimodal Group Emotion Recognition In-the-Wild}

\section{Sanity Check with a Simple Model for Synthetic Data}

In this sanity-check experiment, each video is represented by $n=5$ frames, uniformly sampled at 1~fps and resized to $224\times224$. This preprocessing is applied to both real and synthetic clips. To evaluate the effect of synthetic augmentation, we train a CNN–BiLSTM model while varying the synthetic ratio from 0\% to 30\% in 10\% steps; synthetic samples are evenly distributed across Positive, Neutral, and Negative.

Per frame, VGG19 produces a 4096 dimensional feature, which is projected to 1024 via a lightweight ConvMLP layer. The sequence of five 1024-D vectors is fed to a two-layer bidirectional LSTM (hidden size 512). The BiLSTM output is passed to a fully connected MLP and a softmax layer for 3-class prediction. We compare three backbone regimes: (i) VGG19 trained from scratch, (ii) pretrained VGG19 with unfrozen weights (fine-tuning), and (iii) pretrained VGG19 with frozen weights. In the fine-tuning setup, VGG19 is frozen for the first 10 epochs and then unfrozen; training proceeds for 100 epochs in total. Owing to the relatively small dataset, fine-tuning underperforms, likely due to overfitting during weight updates. All runs use SGD with a learning rate of $10^{-5}$. Unless otherwise stated, the VGG19 backbone remains frozen.

Validation results, including the impact of synthetic ratios, are reported in Table~\ref{tab:results_cnnlstm}.

\begin{table}[H]
  \caption{Validation accuracy of CNN–BiLSTM on VGAF with different synthetic data ratios. Synt: synthetic; Acc: accuracy.}
  \label{tab:results_cnnlstm}
  \vspace{0.5em}
  \resizebox{\linewidth}{!}{
  \begin{tabular}{lccccc}
    \toprule
    \textbf{Synt. ratio}[\%] & \textbf{Synt. Videos} & \textbf{Total Train Videos} & \textbf{Pretrained} & \textbf{Frozen Backbone} & \textbf{Acc. [\%]} \\
    \midrule
    0  & 0    & 2661 & {}          & {}            & 47.25 \\
    0  & 0    & 2661 & {\checkmark}& {}            & 62.01 \\
    0  & 0    & 2661 & {\checkmark}& {\checkmark}  & 62.79 \\
    \midrule
    \textbf{10} & \textbf{297}  & \textbf{2958} & {\checkmark} & {\checkmark} & \textbf{63.31} \\
    20 & 666   & 3327 & {\checkmark} & {\checkmark} & 63.05 \\
    30 & 1140  & 3801 & {\checkmark} & {\checkmark} & 62.80 \\
    \bottomrule
  \end{tabular}}
\end{table}

\subsection{Discussion}

Introducing synthetic clips yields a gain with the highest validation accuracy (63.31\%) at a 10\% synthetic ratio (Table~\ref{tab:results_cnnlstm}). Increasing the ratio beyond 10\% does not improve performance and slightly reduces it, suggesting a smoothing effect that becomes counterproductive when synthetic data dominates.

\begin{figure}[H]
  \centering
  \includegraphics[width=1\textwidth]{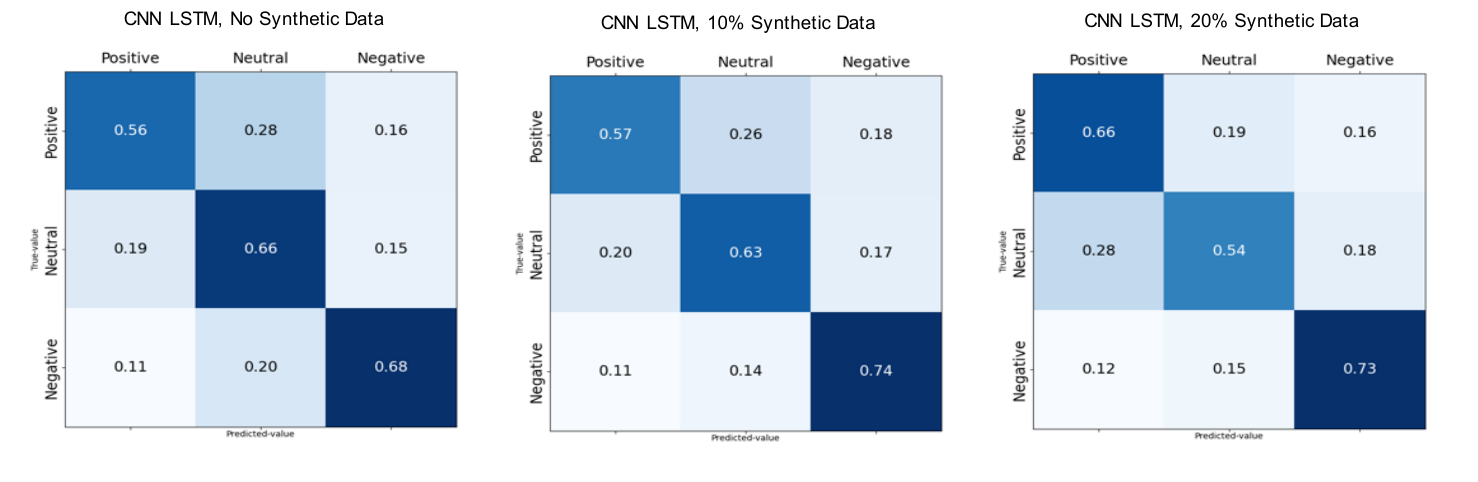}
  \includegraphics[width=1\textwidth]{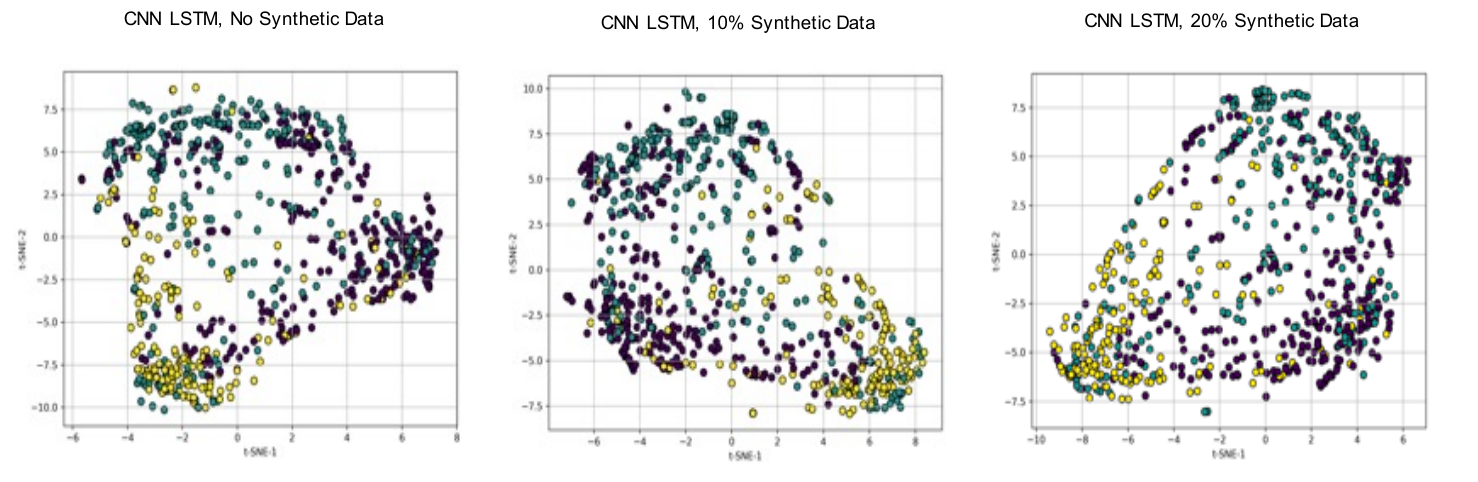}
  \caption{Top: confusion matrices for the CNN–BiLSTM trained with 0\%, 10\%, and 20\% synthetic data. Bottom: corresponding t-SNE embeddings of classification-head features. Yellow: Negative; green: Neutral; purple: Positive.}
  \label{fig:cm_tsne}
\end{figure}

To examine class-wise effects, we compare confusion matrices and t-SNE embeddings for the 0\%, 10\%, and 20\% settings in Figure~\ref{fig:cm_tsne}. In the top row of Figure~\ref{fig:cm_tsne}, the Neutral class degrades the most as synthetic proportion increases (about 3\% at 10\% and 12\% at 20\%), whereas Negative benefits (about +6\% at 10\% and +5\% at 20\%). Positive improves by about 10\% at 20\%. These trends are consistent with class composition: Negative draws from a larger and more diverse pool of 2{,}946 faces spanning four emotions, while Positive and Neutral have 737 and 744 faces, respectively, which likely yields a richer feature space for Negative and better generalization.

The t-SNE plots (Figure~\ref{fig:cm_tsne}, bottom row) indicate only modest regularization of the learned representation: Negative points (yellow) become more separable, while Positive (purple) and Neutral (green) remain partially overlapping. Overall, synthetic augmentation helps, but gains remain limited under this simple control model.

\paragraph{Note on ratio selection.}
This sanity check peaks at 10\% synthetic. In contrast, the main multimodal transformer (Chapter~\ref{chap:mger}) peaks at 30\% (Table~\ref{tab:syntchoice}), which we attribute to its higher capacity and stronger inductive biases (ViT-L/14 with attention-based frame aggregation), making it better able to exploit synthetic variation without overfitting.

\section{Audio CNN Branch}
\label{sec:cnn_audio}

\begin{table}[H]
\centering
\caption{Architecture description of the audio CNN blocks. Each layer or block is described along with its type and functionality.}
\label{tab:cnn_audio}

\resizebox{\linewidth}{!}
{
\begin{tabular}{|l|l|p{8cm}|}
\hline
\textbf{Layer} & \textbf{Type} & \textbf{Description} \\
\hline
\texttt{cnn1} & Conv2D Block & 3 convolutional layers (1$\rightarrow$32$\rightarrow$64), ReLU activations, followed by a MaxPooling layer. Extracts low-level audio features. \\
\hline
\texttt{cnn2} & Conv2D Block & 3 convolutional layers (64$\rightarrow$128), ReLU activations, followed by MaxPooling. Captures mid-level representations. \\
\hline
\texttt{cnn3} & Conv2D Block & 3 convolutional layers (128$\rightarrow$256), ReLU activations, followed by MaxPooling. Captures higher-level abstract features. \\
\hline
\texttt{cnn4} & Conv2D Block & 3 convolutional layers (256$\rightarrow$512), ReLU activations, followed by MaxPooling. Extracts deep feature representations. \\
\hline
\texttt{avgpool} & AdaptiveAvgPool2d & Reduces the spatial feature map size to a fixed shape of $(1, 2)$ regardless of the input size. \\
\hline
\end{tabular}
}
\end{table}

\chapter{Variational Encoder Multi-Decoder}

\section{Skeleton Decoder Based on Heatmaps}

\label{sec:unetupsample}

\begin{table}[H]
\centering
\caption{UNetResUpsample architecture. Here $\texttt{d}$ is the latent space dimension and $C_{\text{out}}$. BatchNorm+ReLU follow each $3{\times}3$ convolution inside the \texttt{dec} blocks. \texttt{ResUp} upsamples by $\times2$ and maps $2048{\to}512$ channels (exact internals may vary).}
\label{tab:unetres}
\resizebox{\linewidth}{!}
{
\begin{tabular}{|l|l|l|l|l|l|}
\hline
Stage & Operation & Channels (in$\to$out) & {Kernel / Stride / Pad} & {Output size} & $H_0 \times W_0$\\
\hline
\texttt{up5}  & Conv2d            & $d \to 2048$     & $1{\times}1$ / $1$ / $0$ & $(H_0,\,W_0)$          & $7{\times}7$ \\
\hline
\texttt{up4}  & ResUp ($\uparrow\times2$) & $2048 \to 512$   & —                       & $(2H_0,\,2W_0)$        & $14{\times}14$ \\
\hline
\texttt{dec4} & (Conv–BN–ReLU)$\times2$ & $512 \to 512$    & $3{\times}3$ / $1$ / $1$ & $(2H_0,\,2W_0)$        & $14{\times}14$ \\
\hline
\texttt{up3}  & ConvTranspose2d   & $512 \to 256$     & $2{\times}2$ / $2$ / $0$ & $(4H_0,\,4W_0)$        & $28{\times}28$ \\
\hline
\texttt{dec3} & (Conv–BN–ReLU)$\times2$ & $256 \to 256$    & $3{\times}3$ / $1$ / $1$ & $(4H_0,\,4W_0)$        & $28{\times}28$ \\
\hline
\texttt{up2}  & ConvTranspose2d   & $256 \to 128$     & $2{\times}2$ / $2$ / $0$ & $(8H_0,\,8W_0)$        & $56{\times}56$ \\
\hline
\texttt{dec2} & (Conv–BN–ReLU)$\times2$ & $128 \to 128$    & $3{\times}3$ / $1$ / $1$ & $(8H_0,\,8W_0)$        & $56{\times}56$ \\
\hline
\texttt{up1}  & ConvTranspose2d   & $128 \to 128$     & $1{\times}1$ / $1$ / $0$ & $(8H_0,\,8W_0)$        & $56{\times}56$ \\
\hline
\texttt{dec1} & (Conv–BN–ReLU)$\times2$ & $128 \to 128$    & $3{\times}3$ / $1$ / $1$ & $(8H_0,\,8W_0)$        & $56{\times}56$ \\
\hline
\texttt{final}& Conv2d            & $128 \to C_{\text{out}}$ & $1{\times}1$ / $1$ / $0$ & $(8H_0,\,8W_0)$        & $56{\times}56$ \\
\hline
\end{tabular}
}
\end{table}

\section{Maximum Mean Discrepancy (MMD)}  
\label{sec:MMD}
\footnotesize

\[
\mathrm{MMD}^2(P, Q) = \mathbb{E}_{x,x'}[k(x, x')] + \mathbb{E}_{y,y'}[k(y, y')] - 2\mathbb{E}_{x,y}[k(x, y)]
\]

Where:
\begin{itemize}
    \item \( k \) is a kernel function (e.g., the RBF kernel).
    \item \( x \) and \( x' \) are samples from distribution \( P \).
    \item \( y \) and \( y' \) are samples from distribution \( Q \).
\end{itemize}

For the Radial Basis Function (RBF) kernel, the kernel function \( k \) is defined as:

\[
k(x, y) = \exp \left( -\frac{\|x - y\|^2}{2\sigma^2} \right)
\]

 Where \( \sigma \) is the bandwidth parameter of the RBF kernel.

\section{Predicted Skeleton Based Heatmap Estimation}
\label{sec:pred_skt_examples}

\begin{figure}[H]
  \centering
  \includegraphics[width=0.99\linewidth]{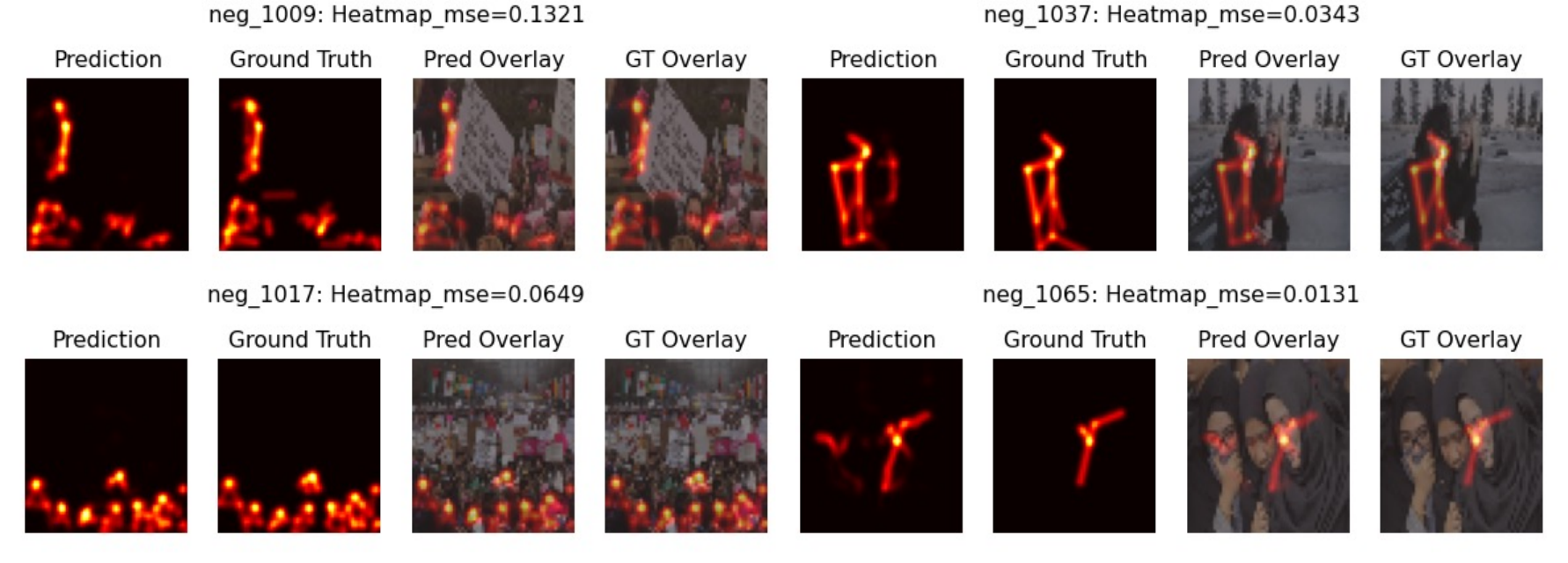}
  \includegraphics[width=0.99\linewidth]{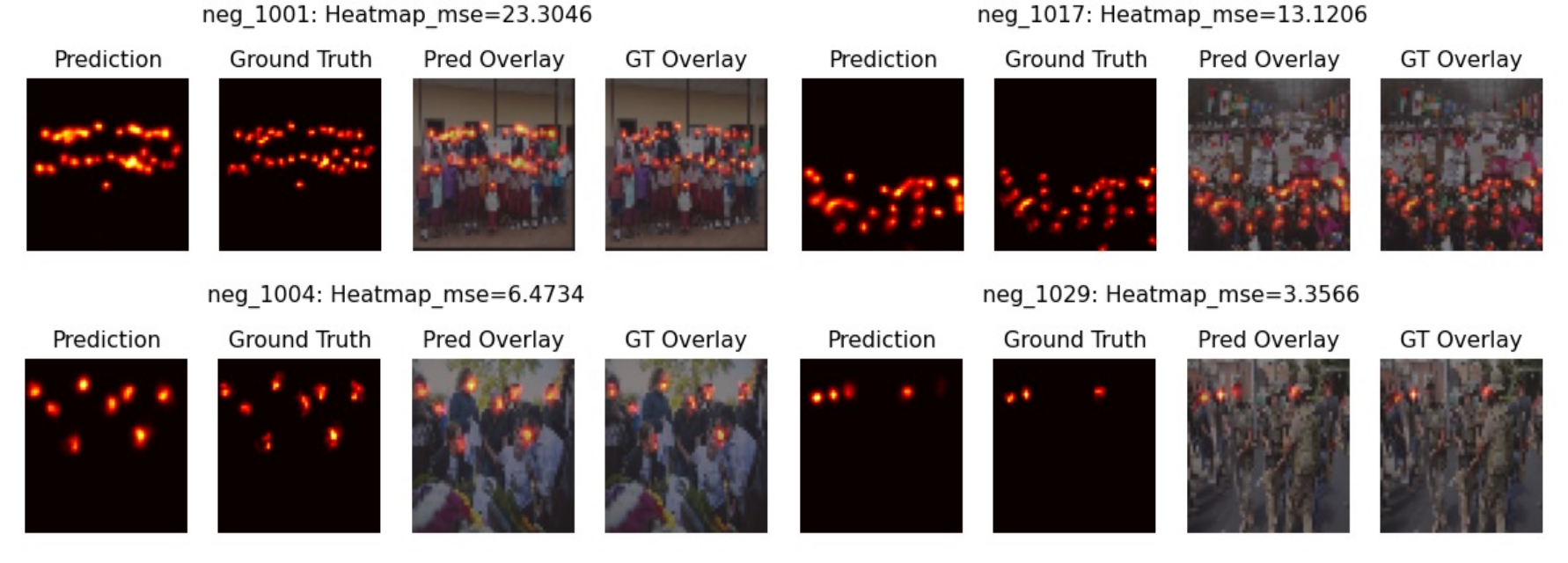}
    \includegraphics[width=0.99\linewidth]{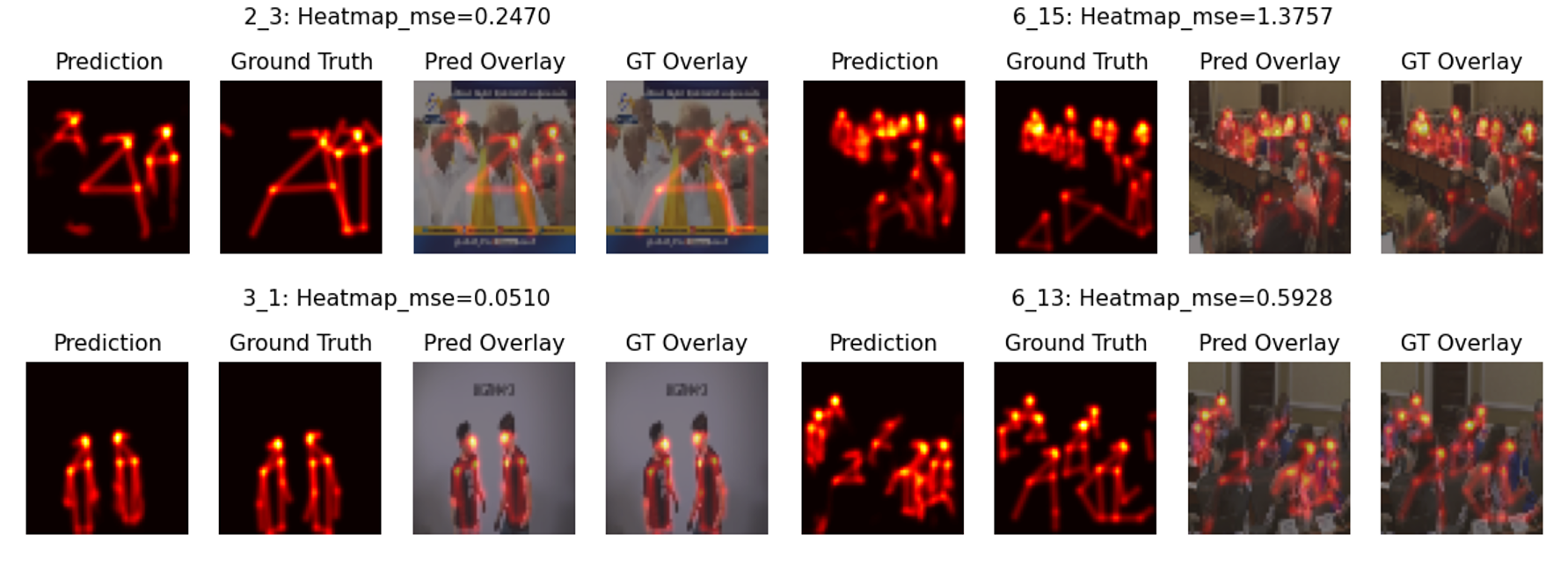}
  \includegraphics[width=0.99\linewidth]{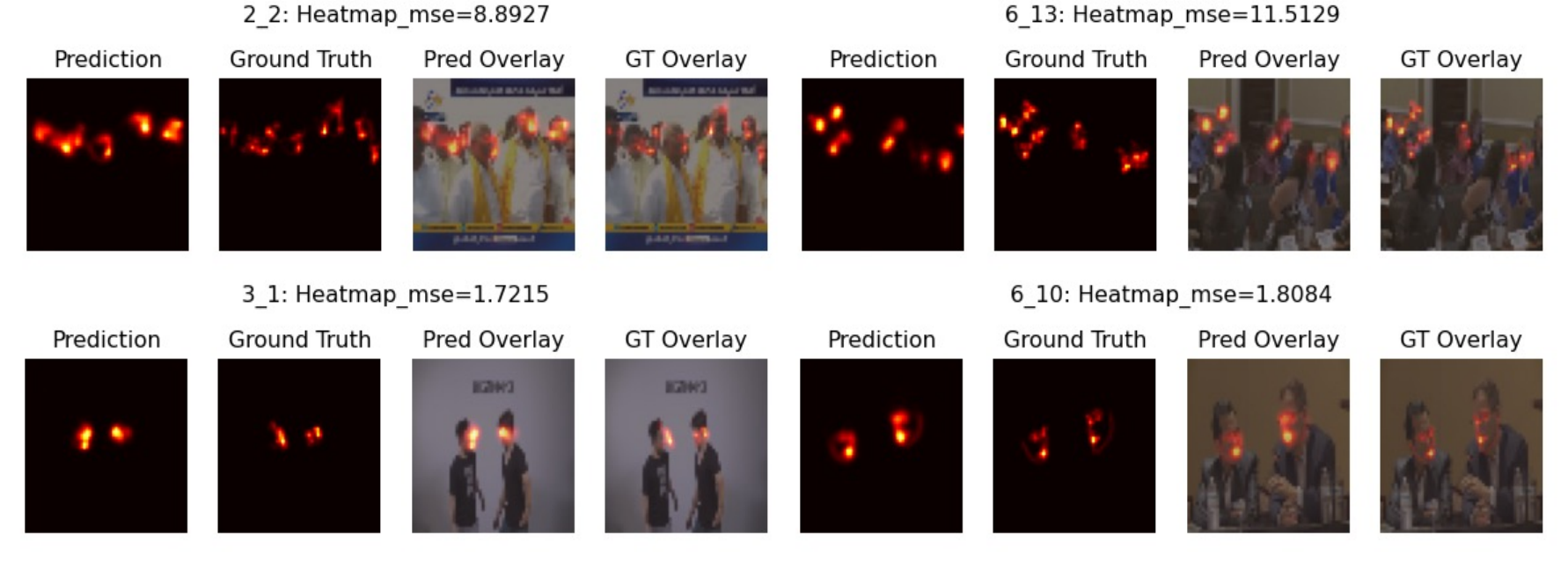}
 \caption{Examples of predicted body and face skeleton based heatmap estimation on VGAF and GAF-3.0 dataset.}
  \label{fig:pred_skt_heatmap_gaf3_app1}
\end{figure}

\begin{figure}[H]
  \centering
    \includegraphics[width=0.99\linewidth]{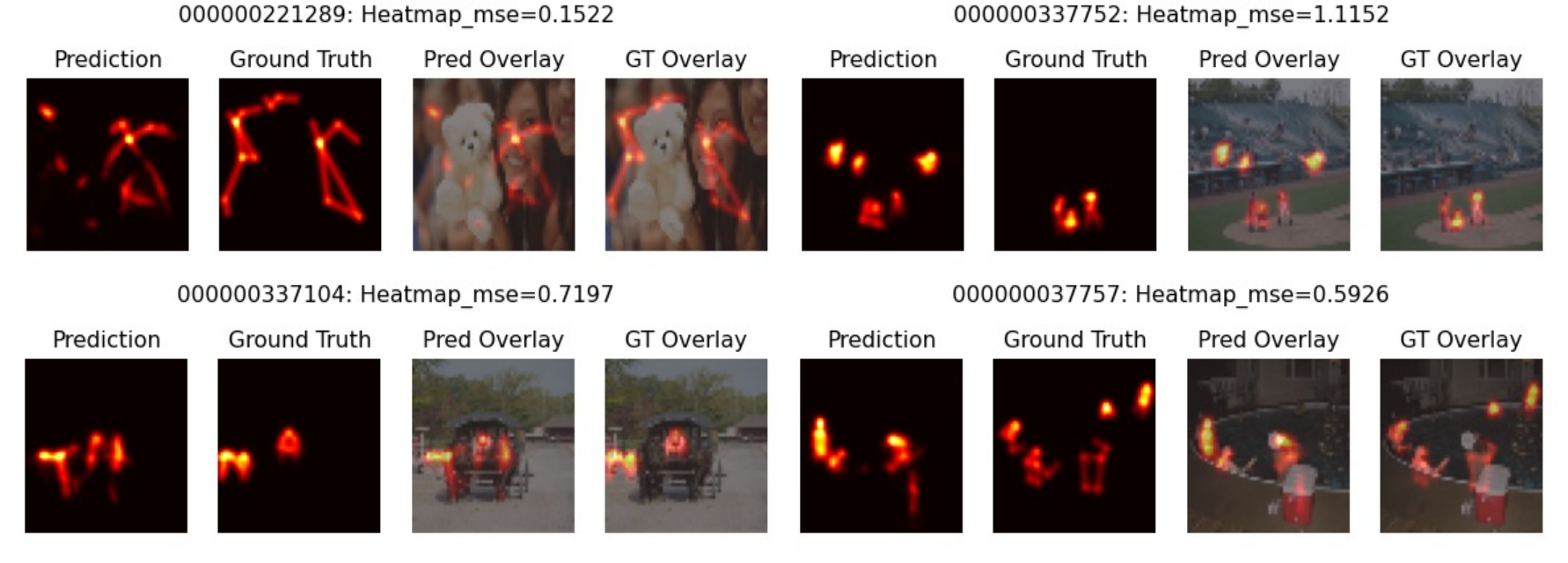}
    \includegraphics[width=0.99\linewidth]{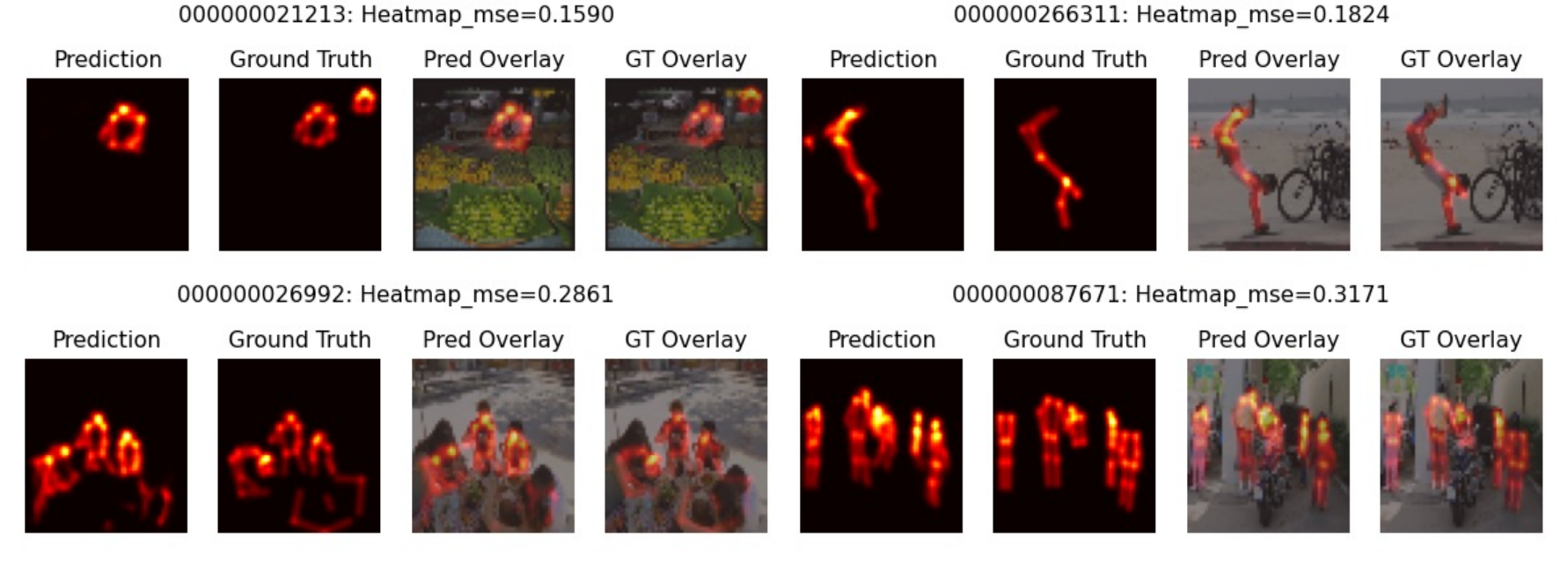}

 \caption{Examples of predicted body skeleton based heatmap estimation on COCO dataset.}
  \label{fig:pred_skt_heatmap_coco_app}
\end{figure}

\printbibliography

@inproceedings{Petrova20,
  TITLE = {{Group-Level Emotion Recognition Using a Unimodal Privacy-Safe Non-Individual Approach}},
  AUTHOR = {Petrova, Anastasia and Vaufreydaz, Dominique and Dessus, Philippe},
  URL = {https://inria.hal.science/hal-02937871},
  BOOKTITLE = {{EmotiW2020 Challenge at the 22nd ACM International Conference on Multimodal Interaction (ICMI2020)}},
  ADDRESS = {Utrecht, Netherlands},
  YEAR = {2020},
  MONTH = Oct,
  HAL_ID = {hal-02937871},
  HAL_VERSION = {v1},
}

@article{Sun2020,
  
   author = {Mo Sun and Jian Li and Hui Feng and Wei Gou and Haifeng Shen and Jian Tang and Yi Yang and Jieping Ye},
   doi = {10.1145/3382507.3417971},
   isbn = {9781450375818},
   journal = {Proceedings of the 2020 International Conference on Multimodal Interaction (ICMI 2020)},
   month = {10},
   pages = {835-840},
   publisher = {Association for Computing Machinery, Inc},
   title = {Multi-modal Fusion Using Spatio-temporal and Static Features for Group Emotion Recognition},
   year = {2020},
}

@article{Guo2018,
   author = {Xin Guo and Bin Zhu and Luisa F Polanía and Charles Boncelet and Kenneth E Barner},
   doi = {10.1145/3242969},
   isbn = {9781450356923},
   journal = {Proceedings of the International Conference on Multimodal Interaction (ICMI 2018)},
   publisher = {ACM},
   title = {Group-Level Emotion Recognition using Hybrid Deep Models based on Faces, Scenes, Skeletons and Visual Attentions},
   url = {https://doi.org/10.1145/3242969.3264990},
   year = {2018},
}

@article{Wang2018,
   author = {Kai Wang and Debin Meng and Xiaoxing Zeng and Kaipeng Zhang and Yu Qiao and Jianfei Yang and Xiaojiang Peng},
   doi = {10.1145/3242969.3264991},
   isbn = {9781450356923},
   journal = {Proceedings of the 2018 International Conference on Multimodal Interaction (ICMI 2018)},
   month = {10},
   pages = {640-645},
   publisher = {Association for Computing Machinery, Inc},
   title = {Cascade attention networks for group emotion recognition with face, body and image cues},
   year = {2018},
}

@article{Sharma2019,
   author = {Garima Sharma and Shreya Ghosh and Abhinav Dhall},
   doi = {10.1109/ACIIW.2019.8925231},
   isbn = {9781728138916},
   journal = {8th International Conference on Affective Computing and Intelligent Interaction Workshops and Demos, ACIIW 2019},
   month = {9},
   pages = {161-167},
   publisher = {Institute of Electrical and Electronics Engineers Inc.},
   title = {Automatic Group Level Affect and Cohesion Prediction in Videos},
   year = {2019},
}

@article{Vaswani2017,
  title={Attention is all you need},
  author={Vaswani, Ashish and Shazeer, Noam and Parmar, Niki and Uszkoreit, Jakob and Jones, Llion and Gomez, Aidan N and Kaiser, {\L}ukasz and Polosukhin, Illia},
  journal={Advances in neural information processing systems},
  volume={30},
  year={2017}
}

@inproceedings{Liu2020,
   author = {Chuanhe Liu and Wenqiang Jiang and Minghao Wang and Tianhao Tang},
   doi = {10.1145/3382507.3417968},
   isbn = {9781450375818},
   journal = {Proceedings of the 2020 International Conference on Multimodal Interaction (ICMI 2020)},
   booktitle={Proceedings of the 2020 International Conference on Multimodal Interaction (ICMI 2020)},
   month = {10},
   pages = {807-812},
   publisher = {Association for Computing Machinery, Inc},
   title = {Group Level Audio-Video Emotion Recognition Using Hybrid Networks},
   year = {2020},
}

@article{Ebner2010,
   author = {Natalie C. Ebner and Michaela Riediger and Ulman Lindenberger},
   doi = {10.3758/BRM.42.1.351},
   issn = {1554351X},
   issue = {1},
   journal = {Behavior Research Methods},
   month = {2},
   pages = {351-362},
   pmid = {20160315},
   title = {FACES-a database of facial expressions in young, middle-aged, and older women and men: Development and validation},
   volume = {42},
   year = {2010},
}

@article{Yu2015,
  author       = {Fisher Yu and
                  Yinda Zhang and
                  Shuran Song and
                  Ari Seff and
                  Jianxiong Xiao},
  title        = {{LSUN:} Construction of a Large-scale Image Dataset using Deep Learning
                  with Humans in the Loop},
  journal      = {CoRR},
  volume       = {abs/1506.03365},
  year         = {2015},
  url          = {http://arxiv.org/abs/1506.03365},
  eprinttype    = {arXiv},
  eprint       = {1506.03365},
  bibsource    = {dblp computer science bibliography, https://dblp.org}
}

@article{Dosovitskiy2020,
  author       = {Alexey Dosovitskiy and
                  Lucas Beyer and
                  Alexander Kolesnikov and
                  Dirk Weissenborn and
                  Xiaohua Zhai and
                  Thomas Unterthiner and
                  Mostafa Dehghani and
                  Matthias Minderer and
                  Georg Heigold and
                  Sylvain Gelly and
                  Jakob Uszkoreit and
                  Neil Houlsby},
  title        = {An Image is Worth 16x16 Words: Transformers for Image Recognition
                  at Scale},
  journal      = {CoRR},
  volume       = {abs/2010.11929},
  year         = {2020},
  url          = {https://arxiv.org/abs/2010.11929},
  eprinttype    = {arXiv},
  eprint       = {2010.11929},
  bibsource    = {dblp computer science bibliography, https://dblp.org}
}

@inproceedings{Wang2020,
   author = {Yanan Wang and Jianming Wu and Panikos Heracleous and Shinya Wada and Rui Kimura and Satoshi Kurihara},
   doi = {10.1145/3382507.3417960},
   isbn = {9781450375818},
   journal = {Proceedings of the International Conference on Multimodal Interaction (ICMI 2020)},
   booktitle={Proceedings of the International Conference on Multimodal Interaction (ICMI 2020)},
   month = {10},
   pages = {827-834},
   publisher = {Association for Computing Machinery, Inc},
   title = {Implicit Knowledge Injectable Cross Attention Audiovisual Model for Group Emotion Recognition},
   year = {2020},
}

@inproceedings{belova2022group,
  title={Group-Level Affect Recognition in Video Using Deviation of Frame Features},
  author={Belova, Natalya S},
  booktitle={Analysis of Images, Social Networks and Texts: 10th International Conference, AIST 2021, Tbilisi, Georgia, December 16--18, 2021, Revised Selected Papers},
  volume={13217},
  pages={199},
  year={2022},
  organization={Springer Nature}
}

@article{sharma2021audio,
  title={Audio-visual automatic group affect analysis},
  author={Sharma, Garima and Dhall, Abhinav and Cai, Jianfei},
  journal={IEEE Transactions on Affective Computing},
  year={2021},
  publisher={IEEE}
}

@inproceedings{pinto2020audiovisual,
  title={Audiovisual classification of group emotion valence using activity recognition networks},
  author={Pinto, Jo{\~a}o Ribeiro and Gon{\c{c}}alves, Tiago and Pinto, Carolina and Sanhudo, Lu{\'\i}s and Fonseca, Joaquim and Gon{\c{c}}alves, Filipe and Carvalho, Pedro and Cardoso, Jaime S},
  booktitle={IEEE 4th International Conference on Image Processing, Applications and Systems (IPAS 2020)},
  pages={114--119},
  year={2020},
  organization={IEEE}
}

@article{Evtodienko2021,
  author       = {Lev Evtodienko},
  title        = {Multimodal End-to-End Group Emotion Recognition using Cross-Modal
                  Attention},
  journal      = {CoRR},
  volume       = {abs/2111.05890},
  year         = {2021},
  url          = {https://arxiv.org/abs/2111.05890},
  eprinttype    = {arXiv},
  eprint       = {2111.05890},
  bibsource    = {dblp computer science bibliography, https://dblp.org}
}

@article{savchenko2022neural,
  title={Neural network model for video-based analysis of student’s emotions in E-learning},
  author={Savchenko, Andrey V and Makarov, IA},
  journal={Optical Memory and Neural Networks},
  volume={31},
  number={3},
  pages={237--244},
  year={2022},
  publisher={Springer}
}

@inproceedings{augusma2022multimodal,
  title={Multimodal Perception and Statistical Modeling of Pedagogical Classroom Events Using a Privacy-safe Non-individual Approach},
  author={Augusma, Anderson},
  booktitle={2022 10th International Conference on Affective Computing and Intelligent Interaction Workshops and Demos (ACIIW)},
  pages={1--5},
  year={2022},
  organization={IEEE}
}

@inproceedings{ottl2020group,
  title={Group-level speech emotion recognition utilising deep spectrum features},
  author={Ottl, Sandra and Amiriparian, Shahin and Gerczuk, Maurice and Karas, Vincent and Schuller, Bj{\"o}rn},
  booktitle={Proceedings of the 2020 International Conference on Multimodal Interaction},
  pages={821--826},
  year={2020}
}

@inproceedings{emotiw2023,
  title={EmotiW 2023: Emotion Recognition in the Wild Challenge},
  author={Dhall, Abhinav and Singh, Monisha and Goecke, Roland and Gedeon, Tom and Zeng, Donghuo and Wang, Yanan and Ikeda, Kazushi},
  booktitle={Proceedings of the 25th International Conference on Multimodal Interaction (ICMI 2023)},
  year={2023},
}

@inproceedings{augusma2023multimodal,
  title={Multimodal Group Emotion Recognition In-the-wild Using Privacy-Compliant Features},
  author={Augusma, Anderson and Vaufreydaz, Dominique and Letu{\'e}, Fr{\'e}d{\'e}rique},
  booktitle={Proceedings of the 25th International Conference on Multimodal Interaction},
  pages={750--754},
  year={2023}
}

@article{kingma2013auto,
  title={Auto-encoding variational bayes},
  author={Kingma, Diederik P and Welling, Max},
  journal={arXiv preprint arXiv:1312.6114},
  year={2013}
}

@article{girin2020dynamical,
  title={Dynamical variational autoencoders: A comprehensive review},
  author={Girin, Laurent and Leglaive, Simon and Bie, Xiaoyu and Diard, Julien and Hueber, Thomas and Alameda-Pineda, Xavier},
  journal={arXiv preprint arXiv:2008.12595},
  year={2020}
}

@article{sadok2024multimodal,
  title={A multimodal dynamical variational autoencoder for audiovisual speech representation learning},
  author={Sadok, Samir and Leglaive, Simon and Girin, Laurent and Alameda-Pineda, Xavier and S{\'e}guier, Renaud},
  journal={Neural Networks},
  volume={172},
  pages={106120},
  year={2024},
  publisher={Elsevier}
}

@inproceedings{singh2023have,
  title={Do i have your attention: A large scale engagement prediction dataset and baselines},
  author={Singh, Monisha and Hoque, Ximi and Zeng, Donghuo and Wang, Yanan and Ikeda, Kazushi and Dhall, Abhinav},
  booktitle={Proceedings of the 25th International Conference on Multimodal Interaction},
  pages={174--182},
  year={2023}
}

@article{dosovitskiy2020vit,
  title={An Image is Worth 16x16 Words: Transformers for Image Recognition at Scale},
  author={Dosovitskiy, Alexey and Beyer, Lucas and Kolesnikov, Alexander and Weissenborn, Dirk and Zhai, Xiaohua and Unterthiner, Thomas and  Dehghani, Mostafa and Minderer, Matthias and Heigold, Georg and Gelly, Sylvain and Uszkoreit, Jakob and Houlsby, Neil},
  journal={ICLR},
  year={2021}
}

@article{sadok2023avector,
  title={Avector QUANTIZED MASKED AUTOENCODER FOR AUDIOVISUAL SPEECH EMOTION RECOGNITION},
  author={Sadok, Samir and Leglaive, Simon and S{\'e}guier, Renaud},
  journal={arXiv preprint arXiv:2305.03568},
  year={2023}
}

@InProceedings{Vedernikov_2024_CVPR,
    author    = {Vedernikov, Alexander and Kumar, Puneet and Chen, Haoyu and Sepp\"anen, Tapio and Li, Xiaobai},
    title     = {TCCT-Net: Two-Stream Network Architecture for Fast and Efficient Engagement Estimation via Behavioral Feature Signals},
    booktitle = {Proceedings of the IEEE/CVF Conference on Computer Vision and Pattern Recognition (CVPR) Workshops},
    month     = {June},
    year      = {2024},
    pages     = {4723-4732}
}

@article{abedi2024engagement,
  title={Engagement Measurement Based on Facial Landmarks and Spatial-Temporal Graph Convolutional Networks},
  author={Abedi, Ali and Khan, Shehroz S},
  journal={arXiv e-prints},
  pages={arXiv--2403},
  year={2024}
}

@inproceedings{anand2024exceda,
  title={ExCEDA: Unlocking Attention Paradigms in Extended Duration E-Classrooms by Leveraging Attention-Mechanism Models},
  author={Anand, Avinash and Mittal, Avni and Dhawan, Laavanaya and Krishnamurthy, Juhi and Ramesh, Mahisha and Lal, Naman and Verma, Astha and Bhuyan, Pijush and Shah, Raijv Ratn and Zimmermann, Roger and others},
  booktitle={2024 IEEE 7th International Conference on Multimedia Information Processing and Retrieval (MIPR)},
  pages={301--307},
  year={2024},
  organization={IEEE}
}

@inproceedings{bujnowski2024samsemo,
  title={SAMSEMO: New dataset for multilingual and multimodal emotion recognition},
  author={Bujnowski, Pawe{\l} and Ku{\'z}ma, Bart{\l}omiej and Paziewski, Bart{\l}omiej and Rutkowski, Jacek and Marhula, Joanna and Bordzicka, Zuzanna and Andruszkiewicz, Piotr},
  booktitle={Interspeech},
  year={2024}
}

@inproceedings{khan2018group,
  title={Group-level emotion recognition using deep models with a four-stream hybrid network},
  author={Khan, Ahmed Shehab and Li, Zhiyuan and Cai, Jie and Meng, Zibo and O'Reilly, James and Tong, Yan},
  booktitle={Proceedings of the 20th ACM international conference on multimodal interaction},
  pages={623--629},
  year={2018}
}

@inproceedings{gupta2018attention,
  title={An attention model for group-level emotion recognition},
  author={Gupta, Aarush and Agrawal, Dakshit and Chauhan, Hardik and Dolz, Jose and Pedersoli, Marco},
  booktitle={Proceedings of the 20th ACM International Conference on Multimodal Interaction},
  pages={611--615},
  year={2018}
}

@inproceedings{wang2018cascade,
  title={Cascade attention networks for group emotion recognition with face, body and image cues},
  author={Wang, Kai and Zeng, Xiaoxing and Yang, Jianfei and Meng, Debin and Zhang, Kaipeng and Peng, Xiaojiang and Qiao, Yu},
  booktitle={Proceedings of the 20th ACM international conference on multimodal interaction},
  pages={640--645},
  year={2018}
}

@article{huang2023auto,
  title={Auto diagnosis of Parkinson's disease via a deep learning model based on mixed emotional facial expressions},
  author={Huang, Wei and Xu, Wenqiang and Wan, Renjie and Zhang, Peng and Zha, Yufei and Pang, Meng},
  journal={IEEE journal of biomedical and health informatics},
  year={2023},
  publisher={IEEE}
}

@article{Frijda01062005,
author = {Nico Frijda},
title = {Emotion experience},
journal = {Cognition and Emotion},
volume = {19},
number = {4},
pages = {473--497},
year = {2005},
publisher = {Routledge},
doi = {10.1080/02699930441000346},
}

@article{gandhi2020perception,
  title={Perception of expressed emotion among persons with mental illness},
  author={Gandhi, Sailaxmi and Padmavathi, Narayanasamy and Raveendran, Rajil and Jadhav, Prabhu and Sahu, Maya and Gurusamy, Jothimani and Muliyala, Krishna Prasad},
  journal={Journal of Psychosocial Rehabilitation and Mental Health},
  volume={7},
  pages={121--130},
  year={2020},
  publisher={Springer}
}

@article{ekman1992there,
  title={Are there basic emotions?},
  author={Ekman, Paul},
  year={1992},
  publisher={American Psychological Association}
}

@article{keltner2016expression,
  title={Expression of emotion},
  author={Keltner, Dacher and Tracy, Jessica and Sauter, Disa A and Cordaro, Daniel C and McNeil, Galen},
  journal={Handbook of emotions},
  volume={4},
  pages={467--482},
  year={2016},
  publisher={Guilford Press New York, NY}
}

@article{ezzameli2023emotion,
  title={Emotion recognition from unimodal to multimodal analysis: A review},
  author={Ezzameli, Kaouther and Mahersia, Hela},
  journal={Information Fusion},
  volume={99},
  pages={101847},
  year={2023},
  publisher={Elsevier}
}

@inproceedings{huang2024facial,
  title={Facial Expression Recognition with Age-Group Expression Feature Learning},
  author={Huang, Yansong and Peng, Junjie and Cai, Zesu and Guo, Jiatao and Chen, Gan and Tan, Shuhua},
  booktitle={2024 International Joint Conference on Neural Networks (IJCNN)},
  pages={1--8},
  year={2024},
  organization={IEEE}
}

@inproceedings{khajontantichaikun2023facial,
  title={Facial Emotion Detection for Thai Elderly People using YOLOv7},
  author={Khajontantichaikun, Thanapong and Jaiyen, Saichon and Yamsaengsung, Siam and Mongkolnam, Pornchai and Chirapornchai, Thanitsorn},
  booktitle={2023 15th International Conference on Knowledge and Smart Technology (KST)},
  pages={1--4},
  year={2023},
  organization={IEEE}
}

@inproceedings{petrou2023lightweight,
  title={Lightweight mood estimation algorithm for faces under partial occlusion},
  author={Petrou, Nikolas and Christodoulou, Georgia and Avgerinakis, Konstantinos and Kosmides, Pavlos},
  booktitle={Proceedings of the 16th International Conference on PErvasive Technologies Related to Assistive Environments},
  pages={402--407},
  year={2023}
}

@article{davletcharova2015detection,
  title={Detection and analysis of emotion from speech signals},
  author={Davletcharova, Assel and Sugathan, Sherin and Abraham, Bibia and James, Alex Pappachen},
  journal={Procedia Computer Science},
  volume={58},
  pages={91--96},
  year={2015},
  publisher={Elsevier}
}

@article{sahu2018adversarial,
  title={Adversarial auto-encoders for speech based emotion recognition},
  author={Sahu, Saurabh and Gupta, Rahul and Sivaraman, Ganesh and AbdAlmageed, Wael and Espy-Wilson, Carol},
  journal={arXiv preprint arXiv:1806.02146},
  year={2018}
}

@inproceedings{dhall2017individual,
  title={From individual to group-level emotion recognition: Emotiw 5.0},
  author={Dhall, Abhinav and Goecke, Roland and Ghosh, Shreya and Joshi, Jyoti and Hoey, Jesse and Gedeon, Tom},
  booktitle={Proceedings of the 19th ACM international conference on multimodal interaction},
  pages={524--528},
  year={2017}
}

@inproceedings{Savchenko2021,
  author={Savchenko, Andrey V.},
  booktitle={2021 IEEE 19th International Symposium on Intelligent Systems and Informatics (SISY)}, 
  title={Facial expression and attributes recognition based on multi-task learning of lightweight neural networks}, 
  year={2021},
  volume={},
  number={},
  pages={119-124},
  doi={10.1109/SISY52375.2021.9582508}}

@inproceedings{sadok2023vector,
  title={A vector quantized masked autoencoder for speech emotion recognition},
  author={Sadok, Samir and Leglaive, Simon and S{\'e}guier, Renaud},
  booktitle={2023 IEEE International conference on acoustics, speech, and signal processing workshops (ICASSPW)},
  pages={1--5},
  year={2023},
  organization={IEEE}
}

@inproceedings{yang2007building,
  title={Building emotion lexicon from weblog corpora},
  author={Yang, Changhua and Lin, Kevin Hsin-Yih and Chen, Hsin-Hsi},
  booktitle={Proceedings of the 45th Annual Meeting of the Association for Computational Linguistics Companion Volume Proceedings of the Demo and Poster Sessions},
  pages={133--136},
  year={2007}
}

@article{batbaatar2019semantic,
  title={Semantic-emotion neural network for emotion recognition from text},
  author={Batbaatar, Erdenebileg and Li, Meijing and Ryu, Keun Ho},
  journal={IEEE access},
  volume={7},
  pages={111866--111878},
  year={2019},
  publisher={IEEE}
}

@inproceedings{ma2005emotion,
  title={Emotion estimation and reasoning based on affective textual interaction},
  author={Ma, Chunling and Prendinger, Helmut and Ishizuka, Mitsuru},
  booktitle={Affective Computing and Intelligent Interaction: First International Conference, ACII 2005, Beijing, China, October 22-24, 2005. Proceedings 1},
  pages={622--628},
  year={2005},
  organization={Springer}
}

@article{jain2017extraction,
  title={Extraction of emotions from multilingual text using intelligent text processing and computational linguistics},
  author={Jain, Vinay Kumar and Kumar, Shishir and Fernandes, Steven Lawrence},
  journal={Journal of computational science},
  volume={21},
  pages={316--326},
  year={2017},
  publisher={Elsevier}
}

@inproceedings{ghaleb2019multimodal,
  title={Multimodal and temporal perception of audio-visual cues for emotion recognition},
  author={Ghaleb, Esam and Popa, Mirela and Asteriadis, Stylianos},
  booktitle={2019 8th International Conference on Affective Computing and Intelligent Interaction (ACII)},
  pages={552--558},
  year={2019},
  organization={IEEE}
}

@article{nemati2019hybrid,
  title={A hybrid latent space data fusion method for multimodal emotion recognition},
  author={Nemati, Shahla and Rohani, Reza and Basiri, Mohammad Ehsan and Abdar, Moloud and Yen, Neil Y and Makarenkov, Vladimir},
  journal={IEEE Access},
  volume={7},
  pages={172948--172964},
  year={2019},
  publisher={IEEE}
}

@article{hsu2023applying,
  title={Applying segment-level attention on bi-modal transformer encoder for audio-visual emotion recognition},
  author={Hsu, Jia-Hao and Wu, Chung-Hsien},
  journal={IEEE Transactions on Affective Computing},
  volume={14},
  number={4},
  pages={3231--3243},
  year={2023},
  publisher={IEEE}
}

@inproceedings{zadeh2018multimodal,
  title={Multimodal language analysis in the wild: Cmu-mosei dataset and interpretable dynamic fusion graph},
  author={Zadeh, AmirAli Bagher and Liang, Paul Pu and Poria, Soujanya and Cambria, Erik and Morency, Louis-Philippe},
  booktitle={Proceedings of the 56th Annual Meeting of the Association for Computational Linguistics (Volume 1: Long Papers)},
  pages={2236--2246},
  year={2018}
}

@article{yang2025omni,
  title={Omni-Emotion: Extending Video MLLM with Detailed Face and Audio Modeling for Multimodal Emotion Analysis},
  author={Yang, Qize and Bai, Detao and Peng, Yi-Xing and Wei, Xihan},
  journal={arXiv preprint arXiv:2501.09502},
  year={2025}
}

@inproceedings{liu2020group,
  title={Group level audio-video emotion recognition using hybrid networks},
  author={Liu, Chuanhe and Jiang, Wenqiang and Wang, Minghao and Tang, Tianhao},
  booktitle={Proceedings of the 2020 International Conference on Multimodal Interaction},
  pages={807--812},
  year={2020}
}

@article{tomar2023unimodal,
  title={Unimodal approaches for emotion recognition: A systematic review},
  author={Tomar, Pragya Singh and Mathur, Kirti and Suman, Ugrasen},
  journal={Cognitive Systems Research},
  volume={77},
  pages={94--109},
  year={2023},
  publisher={Elsevier}
}

@inproceedings{guo2017group,
  title={Group-level emotion recognition using deep models on image scene, faces, and skeletons},
  author={Guo, Xin and Polan{\'\i}a, Luisa F and Barner, Kenneth E},
  booktitle={Proceedings of the 19th ACM International Conference on Multimodal Interaction},
  pages={603--608},
  year={2017}
}

@inproceedings{tan2017group,
  title={Group emotion recognition with individual facial emotion CNNs and global image based CNNs},
  author={Tan, Lianzhi and Zhang, Kaipeng and Wang, Kai and Zeng, Xiaoxing and Peng, Xiaojiang and Qiao, Yu},
  booktitle={Proceedings of the 19th ACM international conference on multimodal interaction},
  pages={549--552},
  year={2017}
}

@inproceedings{dhall2020emotiw,
  title={Emotiw 2020: Driver gaze, group emotion, student engagement and physiological signal based challenges},
  author={Dhall, Abhinav and Sharma, Garima and Goecke, Roland and Gedeon, Tom},
  booktitle={Proceedings of the 2020 International Conference on Multimodal Interaction},
  pages={784--789},
  year={2020}
}

@article{simonyan2014very,
  title={Very deep convolutional networks for large-scale image recognition},
  author={Simonyan, Karen and Zisserman, Andrew},
  journal={arXiv preprint arXiv:1409.1556},
  year={2014}
}

@article{calvo2018human,
  title={Human observers and automated assessment of dynamic emotional facial expressions: KDEF-dyn database validation},
  author={Calvo, Manuel G and Fern{\'a}ndez-Mart{\'\i}n, Andr{\'e}s and Recio, Guillermo and Lundqvist, Daniel},
  journal={Frontiers in psychology},
  volume={9},
  pages={2052},
  year={2018},
  publisher={Frontiers Media SA}
}

@inproceedings{lian2023mer,
  title={Mer 2023: Multi-label learning, modality robustness, and semi-supervised learning},
  author={Lian, Zheng and Sun, Haiyang and Sun, Licai and Chen, Kang and Xu, Mngyu and Wang, Kexin and Xu, Ke and He, Yu and Li, Ying and Zhao, Jinming and others},
  booktitle={Proceedings of the 31st ACM International Conference on Multimedia},
  pages={9610--9614},
  year={2023}
}

@inproceedings{jiang2020dfew,
title={DFEW: A Large-Scale Database for Recognizing Dynamic Facial Expressions in the Wild},
author={Jiang, Xingxun and Zong, Yuan and Zheng, Wenming and Tang, Chuangao and Xia, Wanchuang and Lu, Cheng and Liu, Jiateng},
booktitle={Proceedings of the 28th ACM International Conference on Multimedia},
pages={2881--2889},
year={2020}}

@inproceedings{guo2020graph,
  title={Graph neural networks for image understanding based on multiple cues: Group emotion recognition and event recognition as use cases},
  author={Guo, Xin and Polania, Luisa and Zhu, Bin and Boncelet, Charles and Barner, Kenneth},
  booktitle={Proceedings of the IEEE/CVF Winter Conference on Applications of Computer Vision},
  pages={2921--2930},
  year={2020}
}

@inproceedings{shi2022end,
  title={End-to-end multi-person pose estimation with transformers},
  author={Shi, Dahu and Wei, Xing and Li, Liangqi and Ren, Ye and Tan, Wenming},
  booktitle={Proceedings of the IEEE/CVF Conference on Computer Vision and Pattern Recognition},
  pages={11069--11078},
  year={2022}
}

@inproceedings{kumar2025fusing,
  title={Fusing Multimodal Streams for Improved Group Emotion Recognition in Videos},
  author={Kumar, Deepak and Dhamdhere, Piyush and Raman, Balasubramanian},
  booktitle={International Conference on Pattern Recognition},
  pages={403--418},
  year={2025},
  organization={Springer}
}

@article{foggia2023multi,
  title={Multi-task learning on the edge for effective gender, age, ethnicity and emotion recognition},
  author={Foggia, Pasquale and Greco, Antonio and Saggese, Alessia and Vento, Mario},
  journal={Engineering Applications of Artificial Intelligence},
  volume={118},
  pages={105651},
  year={2023},
  publisher={Elsevier}
}

@inproceedings{hu2018deep,
  title={Deep multi-task learning to recognise subtle facial expressions of mental states},
  author={Hu, Guosheng and Liu, Li and Yuan, Yang and Yu, Zehao and Hua, Yang and Zhang, Zhihong and Shen, Fumin and Shao, Ling and Hospedales, Timothy and Robertson, Neil and others},
  booktitle={Proceedings of the European conference on computer vision (ECCV)},
  pages={103--119},
  year={2018}
}

@article{yin2017multi,
  title={Multi-task convolutional neural network for pose-invariant face recognition},
  author={Yin, Xi and Liu, Xiaoming},
  journal={IEEE Transactions on Image Processing},
  volume={27},
  number={2},
  pages={964--975},
  year={2017},
  publisher={IEEE}
}

@article{pons2020multitask,
  title={Multitask, multilabel, and multidomain learning with convolutional networks for emotion recognition},
  author={Pons, Gerard and Masip, David},
  journal={IEEE Transactions on Cybernetics},
  volume={52},
  number={6},
  pages={4764--4771},
  year={2020},
  publisher={IEEE}
}

@article{ranjan2017hyperface,
  title={Hyperface: A deep multi-task learning framework for face detection, landmark localization, pose estimation, and gender recognition},
  author={Ranjan, Rajeev and Patel, Vishal M and Chellappa, Rama},
  journal={IEEE transactions on pattern analysis and machine intelligence},
  volume={41},
  number={1},
  pages={121--135},
  year={2017},
  publisher={IEEE}
}

@article{hong2018multimodal,
  title={Multimodal face-pose estimation with multitask manifold deep learning},
  author={Hong, Chaoqun and Yu, Jun and Zhang, Jian and Jin, Xiongnan and Lee, Kyong-Ho},
  journal={IEEE transactions on industrial informatics},
  volume={15},
  number={7},
  pages={3952--3961},
  year={2018},
  publisher={IEEE}
}

@article{huang2023study,
  title={A study on computer vision for facial emotion recognition},
  author={Huang, Zi-Yu and Chiang, Chia-Chin and Chen, Jian-Hao and Chen, Yi-Chian and Chung, Hsin-Lung and Cai, Yu-Ping and Hsu, Hsiu-Chuan},
  journal={Scientific reports},
  volume={13},
  number={1},
  pages={8425},
  year={2023},
  publisher={Nature Publishing Group UK London}
}

@inproceedings{ortmann2025unimodal,
  title={Unimodal and Multimodal Static Facial Expression Recognition for Virtual Reality Users with EmoHeVRDB},
  author={Ortmann, Thorben and Wang, Qi and Putzar, Larissa},
  booktitle={2025 IEEE International Conference on Artificial Intelligence and eXtended and Virtual Reality (AIxVR)},
  pages={252--256},
  year={2025},
  organization={IEEE}
}

@inproceedings{chumachenko2024mma,
  title={MMA-DFER: MultiModal Adaptation of unimodal models for Dynamic Facial Expression Recognition in-the-wild},
  author={Chumachenko, Kateryna and Iosifidis, Alexandros and Gabbouj, Moncef},
  booktitle={Proceedings of the IEEE/CVF Conference on Computer Vision and Pattern Recognition},
  pages={4673--4682},
  year={2024}
}

@inproceedings{el2023emonext,
  title={Emonext: an adapted convnext for facial emotion recognition},
  author={El Boudouri, Yassine and Bohi, Amine},
  booktitle={2023 IEEE 25th International Workshop on Multimedia Signal Processing (MMSP)},
  pages={1--6},
  year={2023},
  organization={IEEE}
}

@article{dada2023facial,
  title={Facial Emotion Recognition and Classification Using the Convolutional Neural Network-10 (CNN-10)},
  author={Dada, Emmanuel Gbenga and Oyewola, David Opeoluwa and Joseph, Stephen Bassi and Emebo, Onyeka and Oluwagbemi, Olugbenga Oluseun},
  journal={Applied Computational Intelligence and Soft Computing},
  volume={2023},
  number={1},
  pages={2457898},
  year={2023},
  publisher={Wiley Online Library}
}

@article{nawaz2025novel,
  title={A Novel Transformer-based approach for adult’s facial emotion recognition},
  author={Nawaz, Uzma and Saeed, Zubair and Atif, Kamran},
  journal={IEEE Access},
  year={2025},
  publisher={IEEE}
}

@article{pordoy2024multi,
  title={Multi-Frame Transfer Learning Framework for Facial Emotion Recognition in e-Learning Contexts},
  author={Pordoy, Jamie and Farman, Haleem and Dicheva, Nevena and Anwar, Aamir and Nasralla, Moustafa M and Khilji, Nasrullah and Rehman, Ikram Ur},
  journal={IEEE Access},
  year={2024},
  publisher={IEEE}
}

@inproceedings{rodrigues2022classification,
  title={Classification of facial expressions under partial occlusion for VR games},
  author={Rodrigues, Ana Sofia Figueiredo and Lopes, J{\'u}lio Castro and Lopes, Rui Pedro and Teixeira, Lu{\'\i}s F},
  booktitle={International Conference on Optimization, Learning Algorithms and Applications},
  pages={804--819},
  year={2022},
  organization={Springer}
}

@article{ko2018brief,
  title={A brief review of facial emotion recognition based on visual information},
  author={Ko, Byoung Chul},
  journal={sensors},
  volume={18},
  number={2},
  pages={401},
  year={2018},
  publisher={MDPI}
}

@inproceedings{wang2023yolov7,
  title={{YOLOv7}: Trainable bag-of-freebies sets new state-of-the-art for real-time object detectors},
  author={Wang, Chien-Yao and Bochkovskiy, Alexey and Liao, Hong-Yuan Mark},
  booktitle={Proceedings of the IEEE/CVF Conference on Computer Vision and Pattern Recognition (CVPR)},
  year={2023}
}

@inproceedings{renNIPS15fasterrcnn,
    Author = {Shaoqing Ren and Kaiming He and Ross Girshick and Jian Sun},
    Title = {Faster {R-CNN}: Towards Real-Time Object Detection
             with Region Proposal Networks},
    Booktitle = {Advances in Neural Information Processing Systems ({NIPS})},
    Year = {2015}
}

@misc{lufficc2018ssd,
    author = {Congcong Li},
    title = {{High quality, fast, modular reference implementation of SSD in PyTorch}},
    year = {2018},
    howpublished = {\url{https://github.com/lufficc/SSD}}
}

@inproceedings{rassadin2017group,
  title={Group-level emotion recognition using transfer learning from face identification},
  author={Rassadin, Alexandr and Gruzdev, Alexey and Savchenko, Andrey},
  booktitle={Proceedings of the 19th ACM international conference on multimodal interaction},
  pages={544--548},
  year={2017}
}

@inproceedings{kosti2017emotion,
  title={Emotion recognition in context},
  author={Kosti, Ronak and Alvarez, Jose M and Recasens, Adria and Lapedriza, Agata},
  booktitle={Proceedings of the IEEE conference on computer vision and pattern recognition},
  pages={1667--1675},
  year={2017}
}

@article{argaud2018facial,
  title={Facial emotion recognition in Parkinson's disease: a review and new hypotheses},
  author={Argaud, Soizic and V{\'e}rin, Marc and Sauleau, Paul and Grandjean, Didier},
  journal={Movement disorders},
  volume={33},
  number={4},
  pages={554--567},
  year={2018},
  publisher={Wiley Online Library}
}

@inproceedings{huang2017densely,
  title={Densely connected convolutional networks},
  author={Huang, Gao and Liu, Zhuang and Van Der Maaten, Laurens and Weinberger, Kilian Q},
  booktitle={Proceedings of the IEEE conference on computer vision and pattern recognition},
  pages={4700--4708},
  year={2017}
}

@inproceedings{liu2017sphereface,
  title={Sphereface: Deep hypersphere embedding for face recognition},
  author={Liu, Weiyang and Wen, Yandong and Yu, Zhiding and Li, Ming and Raj, Bhiksha and Song, Le},
  booktitle={Proceedings of the IEEE conference on computer vision and pattern recognition},
  pages={212--220},
  year={2017}
}

@inproceedings{dhall2018emotiw,
  title={Emotiw 2018: Audio-video, student engagement and group-level affect prediction},
  author={Dhall, Abhinav and Kaur, Amanjot and Goecke, Roland and Gedeon, Tom},
  booktitle={Proceedings of the 20th ACM International Conference on Multimodal Interaction},
  pages={653--656},
  year={2018}
}

@article{du2025speech,
  title={Speech emotion recognition based on spiking neural network and convolutional neural network},
  author={Du, Chengyan and Liu, Fu and Kang, Bing and Hou, Tao},
  journal={Engineering Applications of Artificial Intelligence},
  volume={147},
  pages={110314},
  year={2025},
  publisher={Elsevier}
}

@inproceedings{zhao2025temporal,
  title={Temporal-frequency state space duality: An efficient paradigm for speech emotion recognition},
  author={Zhao, Jiaqi and Wang, Fei and Li, Kun and Wei, Yanyan and Tang, Shengeng and Zhao, Shu and Sun, Xiao},
  booktitle={ICASSP 2025-2025 IEEE International Conference on Acoustics, Speech and Signal Processing (ICASSP)},
  pages={1--5},
  year={2025},
  organization={IEEE}
}

@article{tang2025speech,
  title={Speech emotion recognition via cnn-transformer and multidimensional attention mechanism},
  author={Tang, Xiaoyu and Huang, Jiazheng and Lin, Yixin and Dang, Ting and Cheng, Jintao},
  journal={Speech Communication},
  pages={103242},
  year={2025},
  publisher={Elsevier}
}

@article{kang2025speech,
  title={Speech emotion recognition algorithm of intelligent robot based on ACO-SVM},
  author={Kang, Xueliang},
  journal={International Journal of Cognitive Computing in Engineering},
  volume={6},
  pages={131--142},
  year={2025},
  publisher={Elsevier}
}

@article{Busso2008,
  author    = {Carlos Busso and Murtaza Bulut and Chi-Chun Lee and Abe Kazemzadeh and Emily Mower and Samuel Kim and Jeannette N. Chang and Sungbok Lee and Shrikanth S. Narayanan},
  title     = {{IEMOCAP}: Interactive Emotional Dyadic Motion Capture Database},
  journal   = {Journal of Language Resources and Evaluation},
  year      = {2008},
  volume    = {42},
  number    = {4},
  pages     = {335--359},
  doi       = {10.1007/s10579-008-9076-6}
}

@article{radhika2025reliable,
  title={A Reliable speech emotion recognition framework for multi-regional languages using optimized light gradient boosting machine classifier},
  author={Radhika, Subramanian and Prasanth, Aruchamy and Sowndarya, KK Devi},
  journal={Biomedical Signal Processing and Control},
  volume={105},
  pages={107636},
  year={2025},
  publisher={Elsevier}
}

@article{livingstone2018ryerson,
  title={The Ryerson Audio-Visual Database of Emotional Speech and Song (RAVDESS): A dynamic, multimodal set of facial and vocal expressions in North American English},
  author={Livingstone, Steven R and Russo, Frank A},
  journal={PloS one},
  volume={13},
  number={5},
  pages={e0196391},
  year={2018},
  publisher={Public Library of Science San Francisco, CA USA}
}

@inproceedings{burkhardt2005database,
  title={A database of German emotional speech.},
  author={Burkhardt, Felix and Paeschke, Astrid and Rolfes, Miriam and Sendlmeier, Walter F and Weiss, Benjamin and others},
  booktitle={Interspeech},
  volume={5},
  pages={1517--1520},
  year={2005}
}

@inproceedings{he2025dialoguemmt,
  title={DialogueMMT: Dialogue Scenes Understanding Enhanced Multi-modal Multi-task Tuning for Emotion Recognition in Conversations},
  author={He, ChenYuan and Zhu, Senbin and Liu, Hongde and Gao, Fei and Jia, Yuxiang and Zan, Hongying and Peng, Min},
  booktitle={Proceedings of the 31st International Conference on Computational Linguistics},
  pages={2497--2512},
  year={2025}
}

@inproceedings{zhang2017interaction,
  title={Interaction and Transition Model for Speech Emotion Recognition in Dialogue.},
  author={Zhang, Ruo and Ando, Atsushi and Kobashikawa, Satoshi and Aono, Yushi},
  booktitle={INTERSPEECH},
  pages={1094--1097},
  year={2017}
}

@inproceedings{yeh2020dialogical,
  title={A dialogical emotion decoder for speech emotion recognition in spoken dialog},
  author={Yeh, Sung-Lin and Lin, Yun-Shao and Lee, Chi-Chun},
  booktitle={ICASSP 2020-2020 IEEE International conference on acoustics, speech and signal processing (ICASSP)},
  pages={6479--6483},
  year={2020},
  organization={IEEE}
}

@inproceedings{yeh2019interaction,
  title={An interaction-aware attention network for speech emotion recognition in spoken dialogs},
  author={Yeh, Sung-Lin and Lin, Yun-Shao and Lee, Chi-Chun},
  booktitle={ICASSP 2019-2019 IEEE International conference on acoustics, speech and signal processing (ICASSP)},
  pages={6685--6689},
  year={2019},
  organization={IEEE}
}

@article{poria2018meld,
  title={Meld: A multimodal multi-party dataset for emotion recognition in conversations},
  author={Poria, Soujanya and Hazarika, Devamanyu and Majumder, Navonil and Naik, Gautam and Cambria, Erik and Mihalcea, Rada},
  journal={arXiv preprint arXiv:1810.02508},
  year={2018}
}

@article{alhussein2025speech,
  title={Speech emotion recognition in conversations using artificial intelligence: a systematic review and meta-analysis},
  author={Alhussein, Ghada and Ziogas, Ioannis and Saleem, Shiza and Hadjileontiadis, Leontios J},
  journal={Artificial Intelligence Review},
  volume={58},
  number={7},
  pages={198},
  year={2025},
  publisher={Springer}
}

@article{morgan2021classifying,
  title={Classifying the emotional speech content of participants in group meetings using convolutional long short-term memory network},
  author={Morgan, Mallory M and Bhattacharya, Indrani and Radke, Richard J and Braasch, Jonas},
  journal={The Journal of the Acoustical Society of America},
  volume={149},
  number={2},
  pages={885--894},
  year={2021},
  publisher={AIP Publishing}
}

@article{gu2025research,
  title={Research on Multi-class Sentiment Analysis of Social Media Texts Based on the ERNIE Model},
  author={Gu, Xiaomei and Li, Busheng},
  journal={Journal of Computer Science and Artificial Intelligence},
  volume={2},
  number={1},
  pages={1--6},
  year={2025}
}

@article{bacsal2025natural,
  title={Natural Language Processing for Sentiment Analysis in Social Media Marketing},
  author={Ba{\c{s}}al, Murat},
  journal={Economics},
  volume={12},
  number={1},
  pages={39--51},
  year={2025}
}

@article{babu2025enhancing,
  title={Enhancing Sentiment Analysis With Emotion And Sarcasm Detection: A Transformer-Based Approach},
  author={Babu, Mr Suryavamshi Sandeep and Suryanarayana, SV and Sruthi, M and Lakshmi, P Bhagya and Sravanthi, T and Spandana, M},
  journal={Metallurgical and Materials Engineering},
  pages={794--803},
  year={2025}
}

@article{ramaswamy2024multimodal,
  title={Multimodal emotion recognition: A comprehensive review, trends, and challenges},
  author={Ramaswamy, Manju Priya Arthanarisamy and Palaniswamy, Suja},
  journal={Wiley Interdisciplinary Reviews: Data Mining and Knowledge Discovery},
  volume={14},
  number={6},
  pages={e1563},
  year={2024},
  publisher={Wiley Online Library}
}

@article{kalateh2024systematic,
  title={A systematic review on multimodal emotion recognition: building blocks, current state, applications, and challenges},
  author={Kalateh, Sepideh and Estrada-Jimenez, Luis A and Hojjati, Sanaz Nikghadam and Barata, Jose},
  journal={IEEE Access},
  year={2024},
  publisher={IEEE}
}

@article{hazmoune2024using,
  title={Using transformers for multimodal emotion recognition: Taxonomies and state of the art review},
  author={Hazmoune, Samira and Bougamouza, Fateh},
  journal={Engineering Applications of Artificial Intelligence},
  volume={133},
  pages={108339},
  year={2024},
  publisher={Elsevier}
}

@article{zhao2025review,
  title={A review of the emotion recognition model of robots},
  author={Zhao, Mingyi and Gong, Linrui and Din, Abdul Sattar},
  journal={Applied Intelligence},
  volume={55},
  number={6},
  pages={1--33},
  year={2025},
  publisher={Springer}
}

@article{chen2025enhancing,
  title={Enhancing robustness against adversarial attacks in multimodal emotion recognition with spiking transformers},
  author={Chen, Guoming and Qian, Zhuoxian and Zhang, Dong and Qiu, Shuang and Zhou, Ruqi},
  journal={IEEE Access},
  year={2025},
  publisher={IEEE}
}

@article{cao2014crema,
  title={Crema-d: Crowd-sourced emotional multimodal actors dataset},
  author={Cao, Houwei and Cooper, David G and Keutmann, Michael K and Gur, Ruben C and Nenkova, Ani and Verma, Ragini},
  journal={IEEE transactions on affective computing},
  volume={5},
  number={4},
  pages={377--390},
  year={2014},
  publisher={IEEE}
}

@article{mcgurk1976hearing,
  title={Hearing lips and seeing voices},
  author={McGurk, Harry and MacDonald, John},
  journal={Nature},
  volume={264},
  number={5588},
  pages={746--748},
  year={1976},
  publisher={Nature Publishing Group UK London}
}

@article{correa2010canonical,
  title={Canonical correlation analysis for data fusion and group inferences},
  author={Correa, Nicolle M and Adali, Tulay and Li, Yi-Ou and Calhoun, Vince D},
  journal={IEEE signal processing magazine},
  volume={27},
  number={4},
  pages={39--50},
  year={2010},
  publisher={IEEE}
}

@inproceedings{li2003multimedia,
  title={Multimedia content processing through cross-modal association},
  author={Li, Dongge and Dimitrova, Nevenka and Li, Mingkun and Sethi, Ishwar K},
  booktitle={Proceedings of the eleventh ACM international conference on Multimedia},
  pages={604--611},
  year={2003}
}

@inproceedings{sharma2012generalized,
  title={Generalized multiview analysis: A discriminative latent space},
  author={Sharma, Abhishek and Kumar, Abhishek and Daume, Hal and Jacobs, David W},
  booktitle={2012 IEEE conference on computer vision and pattern recognition},
  pages={2160--2167},
  year={2012},
  organization={IEEE}
}

@article{dempster1968generalization,
  title={A generalization of Bayesian inference},
  author={Dempster, Arthur P},
  journal={Journal of the Royal Statistical Society: Series B (Methodological)},
  volume={30},
  number={2},
  pages={205--232},
  year={1968},
  publisher={Wiley Online Library}
}

@article{basiri2014sentiment,
  title={Sentiment prediction based on dempster-shafer theory of evidence},
  author={Basiri, Mohammad Ehsan and Naghsh-Nilchi, Ahmad Reza and Ghasem-Aghaee, Nasser},
  journal={Mathematical Problems in Engineering},
  volume={2014},
  number={1},
  pages={361201},
  year={2014},
  publisher={Wiley Online Library}
}

@article{grosjean1996gating,
  title={Gating},
  author={Grosjean, Fran{\c{c}}ois},
  journal={Language and cognitive processes},
  volume={11},
  number={6},
  pages={597--604},
  year={1996},
  publisher={Taylor \& Francis}
}

@article{nemati2016incorporating,
  title={Incorporating social media comments in affective video retrieval},
  author={Nemati, Shahla and Naghsh-Nilchi, Ahmad Reza},
  journal={Journal of Information Science},
  volume={42},
  number={4},
  pages={524--538},
  year={2016},
  publisher={SAGE Publications Sage UK: London, England}
}

@inproceedings{nemati2017exploiting,
  title={Exploiting evidential theory in the fusion of textual, audio, and visual modalities for affective music video retrieval},
  author={Nemati, Shahla and Naghsh-Nilchi, Ahmad Reza},
  booktitle={2017 3rd international conference on pattern recognition and image analysis (ipria)},
  pages={222--228},
  year={2017},
  organization={IEEE}
}

@inproceedings{li2017reliable,
  title={Reliable crowdsourcing and deep locality-preserving learning for expression recognition in the wild},
  author={Li, Shan and Deng, Weihong and Du, JunPing},
  booktitle={Proceedings of the IEEE conference on computer vision and pattern recognition},
  pages={2852--2861},
  year={2017}
}

@article{ebner2010faces,
  title={FACES—A database of facial expressions in young, middle-aged, and older women and men: Development and validation},
  author={Ebner, Natalie C and Riediger, Michaela and Lindenberger, Ulman},
  journal={Behavior research methods},
  volume={42},
  pages={351--362},
  year={2010},
  publisher={Springer}
}

@article{mollahosseini2017affectnet,
  title={Affectnet: A database for facial expression, valence, and arousal computing in the wild},
  author={Mollahosseini, Ali and Hasani, Behzad and Mahoor, Mohammad H},
  journal={IEEE Transactions on Affective Computing},
  volume={10},
  number={1},
  pages={18--31},
  year={2017},
  publisher={IEEE}
}

@inproceedings{arriaga2019real,
  title={Real-time Convolutional Neural Networks for emotion and gender classification},
  author={Arriaga, Octavio and Valdenegro-Toro, Matias and Pl{\"o}ger, Paul},
  booktitle={27th European Symposium on Artificial Neural Networks, ESANN 2019, Bruges, Belgium, April 24-26, 2019},
  pages={221--226},
  year={2019}
}

@article{zhang2021new,
  title={A new recursive least squares-based learning algorithm for spiking neurons},
  author={Zhang, Yun and Qu, Hong and Luo, Xiaoling and Chen, Yi and Wang, Yuchen and Zhang, Malu and Li, Zefang},
  journal={Neural Networks},
  volume={138},
  pages={110--125},
  year={2021},
  publisher={Elsevier}
}

@article{nematzadeh2022tuning,
  title={Tuning hyperparameters of machine learning algorithms and deep neural networks using metaheuristics: A bioinformatics study on biomedical and biological cases},
  author={Nematzadeh, Sajjad and Kiani, Farzad and Torkamanian-Afshar, Mahsa and Aydin, Nizamettin},
  journal={Computational biology and chemistry},
  volume={97},
  pages={107619},
  year={2022},
  publisher={Elsevier}
}

@article{subramanian2024effective,
  title={An effective speech emotion recognition model for multi-regional languages using threshold-based feature selection algorithm},
  author={Subramanian, Radhika and Aruchamy, Prasanth},
  journal={Circuits, Systems, and Signal Processing},
  volume={43},
  number={4},
  pages={2477--2506},
  year={2024},
  publisher={Springer}
}

@inproceedings{he2022masked,
  title={Masked autoencoders are scalable vision learners},
  author={He, Kaiming and Chen, Xinlei and Xie, Saining and Li, Yanghao and Doll{\'a}r, Piotr and Girshick, Ross},
  booktitle={Proceedings of the IEEE/CVF conference on computer vision and pattern recognition},
  pages={16000--16009},
  year={2022}
}

@inproceedings{singh2002open,
  title={Open mind common sense: Knowledge acquisition from the general public},
  author={Singh, Push and Lin, Thomas and Mueller, Erik T and Lim, Grace and Perkins, Travell and Li Zhu, Wan},
  booktitle={On the Move to Meaningful Internet Systems 2002: CoopIS, DOA, and ODBASE: Confederated International Conferences CoopIS, DOA, and ODBASE 2002 Proceedings},
  pages={1223--1237},
  year={2002},
  organization={Springer}
}

@inproceedings{strapparava2004wordnet,
  title={Wordnet affect: an affective extension of wordnet.},
  author={Strapparava, Carlo and Valitutti, Alessandro and others},
  booktitle={Lrec},
  volume={4},
  number={1083-1086},
  pages={40},
  year={2004},
  organization={Lisbon, Portugal}
}

@article{zhalehpour2016baum,
  title={BAUM-1: A spontaneous audio-visual face database of affective and mental states},
  author={Zhalehpour, Sara and Onder, Onur and Akhtar, Zahid and Erdem, Cigdem Eroglu},
  journal={IEEE Transactions on Affective Computing},
  volume={8},
  number={3},
  pages={300--313},
  year={2016},
  publisher={IEEE}
}

@inproceedings{lin2019tsm,
  title={Tsm: Temporal shift module for efficient video understanding},
  author={Lin, Ji and Gan, Chuang and Han, Song},
  booktitle={Proceedings of the IEEE/CVF international conference on computer vision},
  pages={7083--7093},
  year={2019}
}

@article{yolov3,
  title={YOLOv3: An Incremental Improvement},
  author={Redmon, Joseph and Farhadi, Ali},
  journal = {arXiv},
  year={2018}
}

@article{wang2022congnn,
  title={Congnn: Context-consistent cross-graph neural network for group emotion recognition in the wild},
  author={Wang, Yu and Zhou, Shunping and Liu, Yuanyuan and Wang, Kunpeng and Fang, Fang and Qian, Haoyue},
  journal={Information Sciences},
  volume={610},
  pages={707--724},
  year={2022},
  publisher={Elsevier}
}

@inproceedings{lee2019context,
  title={Context-aware emotion recognition networks},
  author={Lee, Jiyoung and Kim, Seungryong and Kim, Sunok and Park, Jungin and Sohn, Kwanghoon},
  booktitle={Proceedings of the IEEE/CVF international conference on computer vision},
  pages={10143--10152},
  year={2019}
}

@article{lee2024group,
  title={Group emotion recognition based on psychological principles using a fuzzy system},
  author={Lee, Kyuhong and Kim, Taeyong},
  journal={The Visual Computer},
  volume={40},
  number={5},
  pages={3503--3514},
  year={2024},
  publisher={Springer}
}

@article{cheng2024emotion,
  title={Emotion-llama: Multimodal emotion recognition and reasoning with instruction tuning},
  author={Cheng, Zebang and Cheng, Zhi-Qi and He, Jun-Yan and Wang, Kai and Lin, Yuxiang and Lian, Zheng and Peng, Xiaojiang and Hauptmann, Alexander},
  journal={Advances in Neural Information Processing Systems},
  volume={37},
  pages={110805--110853},
  year={2024}
}

@article{alayrac2022flamingo,
  title={Flamingo: a visual language model for few-shot learning},
  author={Alayrac, Jean-Baptiste and Donahue, Jeff and Luc, Pauline and Miech, Antoine and Barr, Iain and Hasson, Yana and Lenc, Karel and Mensch, Arthur and Millican, Katherine and Reynolds, Malcolm and others},
  journal={Advances in neural information processing systems},
  volume={35},
  pages={23716--23736},
  year={2022}
}

@article{bai2023qwen,
  title={Qwen technical report},
  author={Bai, Jinze and Bai, Shuai and Chu, Yunfei and Cui, Zeyu and Dang, Kai and Deng, Xiaodong and Fan, Yang and Ge, Wenbin and Han, Yu and Huang, Fei and others},
  journal={arXiv preprint arXiv:2309.16609},
  year={2023}
}

@article{chen2023shikra,
  title={Shikra: Unleashing multimodal llm's referential dialogue magic},
  author={Chen, Keqin and Zhang, Zhao and Zeng, Weili and Zhang, Richong and Zhu, Feng and Zhao, Rui},
  journal={arXiv preprint arXiv:2306.15195},
  year={2023}
}

@article{chiang2023vicuna,
  title={Vicuna: An open-source chatbot impressing gpt-4 with 90\%* chatgpt quality, March 2023},
  author={Chiang, Wei-Lin and Li, Zhuohan and Lin, Zi and Sheng, Ying and Wu, Zhanghao and Zhang, Hao and Zheng, Lianmin and Zhuang, Siyuan and Zhuang, Yonghao and Gonzalez, Joseph E and others},
  journal={URL https://lmsys. org/blog/2023-03-30-vicuna},
  volume={3},
  number={5},
  year={2023}
}

@article{peng2023kosmos,
  title={Kosmos-2: Grounding multimodal large language models to the world},
  author={Peng, Zhiliang and Wang, Wenhui and Dong, Li and Hao, Yaru and Huang, Shaohan and Ma, Shuming and Wei, Furu},
  journal={arXiv preprint arXiv:2306.14824},
  year={2023}
}

@article{wang2023visionllm,
  title={Visionllm: Large language model is also an open-ended decoder for vision-centric tasks},
  author={Wang, Wenhai and Chen, Zhe and Chen, Xiaokang and Wu, Jiannan and Zhu, Xizhou and Zeng, Gang and Luo, Ping and Lu, Tong and Zhou, Jie and Qiao, Yu and others},
  journal={Advances in Neural Information Processing Systems},
  volume={36},
  pages={61501--61513},
  year={2023}
}

@article{hsu2021hubert,
  title={Hubert: Self-supervised speech representation learning by masked prediction of hidden units},
  author={Hsu, Wei-Ning and Bolte, Benjamin and Tsai, Yao-Hung Hubert and Lakhotia, Kushal and Salakhutdinov, Ruslan and Mohamed, Abdelrahman},
  journal={IEEE/ACM transactions on audio, speech, and language processing},
  volume={29},
  pages={3451--3460},
  year={2021},
  publisher={IEEE}
}

@article{tong2022videomae,
  title={Videomae: Masked autoencoders are data-efficient learners for self-supervised video pre-training},
  author={Tong, Zhan and Song, Yibing and Wang, Jue and Wang, Limin},
  journal={Advances in neural information processing systems},
  volume={35},
  pages={10078--10093},
  year={2022}
}

@inproceedings{fang2023eva,
  title={Eva: Exploring the limits of masked visual representation learning at scale},
  author={Fang, Yuxin and Wang, Wen and Xie, Binhui and Sun, Quan and Wu, Ledell and Wang, Xinggang and Huang, Tiejun and Wang, Xinlong and Cao, Yue},
  booktitle={Proceedings of the IEEE/CVF conference on computer vision and pattern recognition},
  pages={19358--19369},
  year={2023}
}

@article{touvron2023llama,
  title={Llama 2: Open foundation and fine-tuned chat models},
  author={Touvron, Hugo and Martin, Louis and Stone, Kevin and Albert, Peter and Almahairi, Amjad and Babaei, Yasmine and Bashlykov, Nikolay and Batra, Soumya and Bhargava, Prajjwal and Bhosale, Shruti and others},
  journal={arXiv preprint arXiv:2307.09288},
  year={2023}
}

@article{jiang2024mixtral,
  title={Mixtral of experts},
  author={Jiang, Albert Q and Sablayrolles, Alexandre and Roux, Antoine and Mensch, Arthur and Savary, Blanche and Bamford, Chris and Chaplot, Devendra Singh and Casas, Diego de las and Hanna, Emma Bou and Bressand, Florian and others},
  journal={arXiv preprint arXiv:2401.04088},
  year={2024}
}

@inproceedings{lian2019conversational,
  title={Conversational Emotion Analysis via Attention Mechanisms},
  author={Lian, Zheng and Tao, Jianhua and Liu, Bin and Huang, Jian},
  booktitle={Proc. Interspeech 2019},
  pages={1936--1940},
  year={2019}
}

@misc{quintans2023chatgpt,
  title={ChatGPT: the new panacea of the academic world},
  author={Quintans-J{\'u}nior, Lucindo Jos{\'e} and Gurgel, Ricardo Queiroz and Ara{\'u}jo, Adriano Antunes de Souza and Correia, Dalmo and Martins-Filho, Paulo Ricardo},
  journal={Revista da Sociedade Brasileira de Medicina Tropical},
  volume={56},
  pages={e0060--2023},
  year={2023},
  publisher={SciELO Brasil}
}

@inproceedings{liu2022mafw,
  title={Mafw: A large-scale, multi-modal, compound affective database for dynamic facial expression recognition in the wild},
  author={Liu, Yuanyuan and Dai, Wei and Feng, Chuanxu and Wang, Wenbin and Yin, Guanghao and Zeng, Jiabei and Shan, Shiguang},
  booktitle={Proceedings of the 30th ACM international conference on multimedia},
  pages={24--32},
  year={2022}
}

@article{lian2024affectgpt,
  title={AffectGPT: Dataset and framework for explainable multimodal emotion recognition},
  author={Lian, Zheng and Sun, Haiyang and Sun, Licai and Yi, Jiangyan and Liu, Bin and Tao, Jianhua},
  journal={arXiv preprint arXiv:2407.07653},
  year={2024}
}

@inproceedings{tan2019efficientnet,
  title={Efficientnet: Rethinking model scaling for convolutional neural networks},
  author={Tan, Mingxing and Le, Quoc},
  booktitle={International conference on machine learning},
  pages={6105--6114},
  year={2019},
  organization={PMLR}
}

@article{lecun2002gradient,
  title={Gradient-based learning applied to document recognition},
  author={LeCun, Yann and Bottou, L{\'e}on and Bengio, Yoshua and Haffner, Patrick},
  journal={Proceedings of the IEEE},
  volume={86},
  number={11},
  pages={2278--2324},
  year={2002},
  publisher={Ieee}
}

@article{krizhevsky2012imagenet,
  title={Imagenet classification with deep convolutional neural networks},
  author={Krizhevsky, Alex and Sutskever, Ilya and Hinton, Geoffrey E},
  journal={Advances in neural information processing systems},
  volume={25},
  year={2012}
}

@inproceedings{he2016deep,
  title={Deep residual learning for image recognition},
  author={He, Kaiming and Zhang, Xiangyu and Ren, Shaoqing and Sun, Jian},
  booktitle={Proceedings of the IEEE conference on computer vision and pattern recognition},
  pages={770--778},
  year={2016}
}

@article{gong2025hybrid,
  title={A Hybrid Fusion Model for Group-Level Emotion Recognition in Complex Scenarios},
  author={Gong, Wenjuan and Wang, Yifan and Wu, Yikai and Gao, Shuaipeng and Vasilakos, Athanasios V and Zhang, Peiying},
  journal={Information Sciences},
  pages={121968},
  year={2025},
  publisher={Elsevier}
}

@inproceedings{ghosh2018automatic,
  title={Automatic group affect analysis in images via visual attribute and feature networks},
  author={Ghosh, Shreya and Dhall, Abhinav and Sebe, Nicu},
  booktitle={2018 25th IEEE International Conference on Image Processing (ICIP)},
  pages={1967--1971},
  year={2018},
  organization={IEEE}
}

@article{huangpsmf2025,
  title={Psmf: Prototype Network Subgraph with Multi-Head Attention Framework for Group Emotion Recognition},
  author={Huang, Wenti and Long, Jun and others},
  journal={Not published yet (Rewiew)},
  pages={121969},
  year={2025},
  publisher={Elsevier}
}

@article{kossaifi2017afew,
  title={AFEW-VA database for valence and arousal estimation in-the-wild},
  author={Kossaifi, Jean and Tzimiropoulos, Georgios and Todorovic, Sinisa and Pantic, Maja},
  journal={Image and Vision Computing},
  volume={65},
  pages={23--36},
  year={2017},
  publisher={Elsevier}
}

@article{huang2024survey,
  title={A Survey of Deep Learning for Group-level Emotion Recognition},
  author={Huang, Xiaohua and Xu, Jinke and Zheng, Wenming and Mao, Qirong and Dhall, Abhinav},
  journal={CoRR},
  year={2024}
}

@inproceedings{bertasius2021space,
  title={Is space-time attention all you need for video understanding?},
  author={Bertasius, Gedas and Wang, Heng and Torresani, Lorenzo},
  booktitle={ICML},
  volume={2},
  number={3},
  pages={4},
  year={2021}
}

@article{baevski2020wav2vec,
  title={wav2vec 2.0: A framework for self-supervised learning of speech representations},
  author={Baevski, Alexei and Zhou, Yuhao and Mohamed, Abdelrahman and Auli, Michael},
  journal={Advances in neural information processing systems},
  volume={33},
  pages={12449--12460},
  year={2020}
}

@article{hussain2023yolo,
  title={YOLO-v1 to YOLO-v8, the rise of YOLO and its complementary nature toward digital manufacturing and industrial defect detection},
  author={Hussain, Muhammad},
  journal={Machines},
  volume={11},
  number={7},
  pages={677},
  year={2023},
  publisher={MDPI}
}

@inproceedings{leang2024exploring,
  title={Exploring VQ-VAE with Prosody Parameters for Speaker Anonymization},
  author={Leang, Sotheara and Augusma, Anderson and Castelli, Eric and Letu{\'e}, Fr{\'e}d{\'e}rique and Sam, Sethserey and Vaufreydaz, Dominique},
  booktitle={Voice Privacy Challenge 2024 at INTERSPEECH 2024},
  year={2024}
}

@inproceedings{kung2025face,
  title={Face anonymization made simple},
  author={Kung, Han-Wei and Varanka, Tuomas and Saha, Sanjay and Sim, Terence and Sebe, Nicu},
  booktitle={2025 IEEE/CVF Winter Conference on Applications of Computer Vision (WACV)},
  pages={1040--1050},
  year={2025},
  organization={IEEE}
}

@inproceedings{choi2018stargan,
  title={Stargan: Unified generative adversarial networks for multi-domain image-to-image translation},
  author={Choi, Yunjey and Choi, Minje and Kim, Munyoung and Ha, Jung-Woo and Kim, Sunghun and Choo, Jaegul},
  booktitle={Proceedings of the IEEE conference on computer vision and pattern recognition},
  pages={8789--8797},
  year={2018}
}

@article{boudewijn2024legal,
  title={Legal and Regulatory Perspectives on Synthetic Data as an Anonymization Strategy},
  author={Boudewijn, Alexander and Ferraris, Andrea F},
  journal={J. Pers. Data Prot. L.},
  pages={17},
  year={2024},
  publisher={HeinOnline}
}

@article{sarmin2024synthetic,
  title={Synthetic Data: Revisiting the Privacy-Utility Trade-off},
  author={Sarmin, Fatima Jahan and Sarkar, Atiquer Rahman and Wang, Yang and Mohammed, Noman},
  journal={arXiv preprint arXiv:2407.07926},
  year={2024}
}

@article{goodfellow2020generative,
  title={Generative adversarial networks},
  author={Goodfellow, Ian and Pouget-Abadie, Jean and Mirza, Mehdi and Xu, Bing and Warde-Farley, David and Ozair, Sherjil and Courville, Aaron and Bengio, Yoshua},
  journal={Communications of the ACM},
  volume={63},
  number={11},
  pages={139--144},
  year={2020},
  publisher={ACM New York, NY, USA}
}

@article{min2025can,
  title={Can Synthetic Data Protect Privacy?},
  author={Min, Gidan and Oh, Junhyoung},
  journal={IEEE Access},
  year={2025},
  publisher={IEEE}
}

@inproceedings{mou2015group,
  title={Group-level arousal and valence recognition in static images: Face, body and context},
  author={Mou, Wenxuan and Celiktutan, Oya and Gunes, Hatice},
  booktitle={2015 11th IEEE International Conference and Workshops on Automatic Face and Gesture Recognition (FG)},
  volume={5},
  pages={1--6},
  year={2015},
  organization={IEEE}
}

@inproceedings{dhall2013finding,
  title={Finding happiest moments in a social context},
  author={Dhall, Abhinav and Joshi, Jyoti and Radwan, Ibrahim and Goecke, Roland},
  booktitle={Computer Vision--ACCV 2012: 11th Asian Conference on Computer Vision, Daejeon, Korea, November 5-9, 2012, Revised Selected Papers, Part II 11},
  pages={613--626},
  year={2013},
  organization={Springer}
}

@inproceedings{dhall2015more,
  title={The more the merrier: Analysing the affect of a group of people in images},
  author={Dhall, Abhinav and Joshi, Jyoti and Sikka, Karan and Goecke, Roland and Sebe, Nicu},
  booktitle={2015 11th IEEE international conference and workshops on automatic face and gesture recognition (FG)},
  volume={1},
  pages={1--8},
  year={2015},
  organization={IEEE}
}

@article{quach2022non,
  title={Non-volume preserving-based fusion to group-level emotion recognition on crowd videos},
  author={Quach, Kha Gia and Le, Ngan and Duong, Chi Nhan and Jalata, Ibsa and Roy, Kaushik and Luu, Khoa},
  journal={Pattern Recognition},
  volume={128},
  pages={108646},
  year={2022},
  publisher={Elsevier}
}

@inproceedings{ghosh2019predicting,
  title={Predicting group cohesiveness in images},
  author={Ghosh, Shreya and Dhall, Abhinav and Sebe, Nicu and Gedeon, Tom},
  booktitle={2019 International Joint Conference on Neural Networks (IJCNN)},
  pages={1--8},
  year={2019},
  organization={IEEE}
}

@inproceedings{foteinopoulou2024emoclip,
  title={Emoclip: A vision-language method for zero-shot video facial expression recognition},
  author={Foteinopoulou, Niki Maria and Patras, Ioannis},
  booktitle={2024 IEEE 18th International Conference on Automatic Face and Gesture Recognition (FG)},
  pages={1--10},
  year={2024},
  organization={IEEE}
}

@article{tao2024align,
  title={Align-DFER: Pioneering Comprehensive Dynamic Affective Alignment for Dynamic Facial Expression Recognition with CLIP},
  author={Tao, Zeng and Wang, Yan and Lin, Junxiong and Wang, Haoran and Mai, Xinji and Yu, Jiawen and Tong, Xuan and Zhou, Ziheng and Yan, Shaoqi and Zhao, Qing and others},
  journal={CoRR},
  year={2024}
}

@inproceedings{mai2024all,
  title={All rivers run into the sea: Unified modality brain-inspired emotional central mechanism},
  author={Mai, Xinji and Lin, Junxiong and Wang, Haoran and Tao, Zeng and Wang, Yan and Yan, Shaoqi and Tong, Xuan and Yu, Jiawen and Wang, Boyang and Zhou, Ziheng and others},
  booktitle={Proceedings of the 32nd ACM International Conference on Multimedia},
  pages={632--641},
  year={2024}
}

@inproceedings{chen2024finecliper,
  title={Finecliper: Multi-modal fine-grained clip for dynamic facial expression recognition with adapters},
  author={Chen, Haodong and Huang, Haojian and Dong, Junhao and Zheng, Mingzhe and Shao, Dian},
  booktitle={Proceedings of the 32nd ACM International Conference on Multimedia},
  pages={2301--2310},
  year={2024}
}

@article{chen2024cdgt,
  title={CDGT: Constructing diverse graph transformers for emotion recognition from facial videos},
  author={Chen, Dongliang and Wen, Guihua and Li, Huihui and Yang, Pei and Chen, Chuyun and Wang, Bao},
  journal={Neural Networks},
  volume={179},
  pages={106573},
  year={2024},
  publisher={Elsevier}
}

@article{wang2024joint,
  title={A joint local spatial and global temporal CNN-Transformer for dynamic facial expression recognition},
  author={Wang, Linhuang and Kang, Xin and Ding, Fei and Nakagawa, Satoshi and Ren, Fuji},
  journal={Applied Soft Computing},
  volume={161},
  pages={111680},
  year={2024},
  publisher={Elsevier}
}

@article{mai2024ous,
  title={OUS: Scene-Guided Dynamic Facial Expression Recognition},
  author={Mai, Xinji and Wang, Haoran and Tao, Zeng and Lin, Junxiong and Yan, Shaoqi and Wang, Yan and Liu, Jing and Yu, Jiawen and Tong, Xuan and Li, Yating and others},
  journal={CoRR},
  year={2024}
}

@inproceedings{zong2023building,
  title={Building robust multimodal sentiment recognition via a simple yet effective multimodal transformer},
  author={Zong, Daoming and Ding, Chaoyue and Li, Baoxiang and Zhou, Dinghao and Li, Jiakui and Zheng, Ken and Zhou, Qunyan},
  booktitle={Proceedings of the 31st ACM International Conference on Multimedia},
  pages={9596--9600},
  year={2023}
}

@inproceedings{wang2023hierarchical,
  title={Hierarchical audio-visual information fusion with multi-label joint decoding for mer 2023},
  author={Wang, Haotian and Xi, Yuxuan and Chen, Hang and Du, Jun and Song, Yan and Wang, Qing and Zhou, Hengshun and Wang, Chenxi and Ma, Jiefeng and Hu, Pengfei and others},
  booktitle={Proceedings of the 31st ACM International Conference on Multimedia},
  pages={9531--9535},
  year={2023}
}

@article{muradulloyeva2025importance,
  title={THE IMPORTANCE OF BODY LANGUAGE IN COMMUNICATION},
  author={Muradulloyeva, Sevinch},
  journal={EDUCATION AND RESEARCH IN THE ERA OF DIGITAL TRANSFORMATION},
  volume={1},
  number={1},
  pages={2248--2256},
  year={2025}
}

@article{asatullaev2025body,
  title={BODY LANGUAGE INTERPRETATION: PSYCHOPHYSIOLOGICAL AND COGNITIVE ASPECTS},
  author={Asatullaev, Rustamjon and Muxamedjonova, Diyora},
  journal={Journal of Applied Science and Social Science},
  volume={1},
  number={1},
  pages={456--458},
  year={2025}
}

@article{baxtiyarovich2025interpreting,
  title={INTERPRETING BODY LANGUAGE: A SCIENTIFIC PERSPECTIVE},
  author={Baxtiyarovich, Asatullayev Rustamjon and Bahodirovich, Boboqulovc Behruz},
  journal={YANGI O ‘ZBEKISTON, YANGI TADQIQOTLAR JURNALI},
  volume={2},
  number={5},
  pages={143--146},
  year={2025}
}

@article{cao2025deep,
  title={Deep learning-based depression recognition through facial expression: A systematic review},
  author={Cao, Xiaoming and Zhai, Lingling and Zhai, Pengpeng and Li, Fangfei and He, Tao and He, Lang},
  journal={Neurocomputing},
  pages={129605},
  year={2025},
  publisher={Elsevier}
}

@article{balachandran2025facial,
  title={Facial expression-based emotion recognition across diverse age groups: a multi-scale vision transformer with contrastive learning approach},
  author={Balachandran, G and Ranjith, S and Chenthil, TR and Jagan, GC},
  journal={Journal of Combinatorial Optimization},
  volume={49},
  number={1},
  pages={1--39},
  year={2025},
  publisher={Springer}
}

@article{kumar2020object,
  title={Object detection in real time based on improved single shot multi-box detector algorithm},
  author={Kumar, Ashwani and Zhang, Zuopeng Justin and Lyu, Hongbo},
  journal={EURASIP Journal on Wireless Communications and Networking},
  volume={2020},
  number={1},
  pages={204},
  year={2020},
  publisher={Springer}
}

@inproceedings{lucey2010extended,
  title={The extended cohn-kanade dataset (ck+): A complete dataset for action unit and emotion-specified expression},
  author={Lucey, Patrick and Cohn, Jeffrey F and Kanade, Takeo and Saragih, Jason and Ambadar, Zara and Matthews, Iain},
  booktitle={2010 ieee computer society conference on computer vision and pattern recognition-workshops},
  pages={94--101},
  year={2010},
  organization={IEEE}
}

@inproceedings{goodfellow2013challenges,
  title={Challenges in representation learning: A report on three machine learning contests},
  author={Goodfellow, Ian J and Erhan, Dumitru and Carrier, Pierre Luc and Courville, Aaron and Mirza, Mehdi and Hamner, Ben and Cukierski, Will and Tang, Yichuan and Thaler, David and Lee, Dong-Hyun and others},
  booktitle={Neural information processing: 20th international conference, ICONIP 2013, daegu, korea, november 3-7, 2013. Proceedings, Part III 20},
  pages={117--124},
  year={2013},
  organization={Springer}
}

@article{li2025occlusion,
  title={Occlusion-Robust Facial Expression Recognition Based on Multi-Angle Feature Extraction},
  author={Li, Yunfei and Liu, Hao and Liang, Jiuzhen and Jiang, Daihong},
  journal={Applied Sciences},
  volume={15},
  number={9},
  pages={5139},
  year={2025},
  publisher={MDPI}
}

@inproceedings{liu2021swin,
  title={Swin transformer: Hierarchical vision transformer using shifted windows},
  author={Liu, Ze and Lin, Yutong and Cao, Yue and Hu, Han and Wei, Yixuan and Zhang, Zheng and Lin, Stephen and Guo, Baining},
  booktitle={Proceedings of the IEEE/CVF international conference on computer vision},
  pages={10012--10022},
  year={2021}
}

@article{wang2020region,
  title={Region attention networks for pose and occlusion robust facial expression recognition},
  author={Wang, Kai and Peng, Xiaojiang and Yang, Jianfei and Meng, Debin and Qiao, Yu},
  journal={IEEE Transactions on Image Processing},
  volume={29},
  pages={4057--4069},
  year={2020},
  publisher={IEEE}
}

@article{ortiz2023implications,
  title={Implications of emotion recognition technologies: balancing privacy and public safety},
  author={Ortiz-Clavijo, Luis Felipe and Gallego-Duque, Carlos Juli{\'a}n and David-Diaz, Juan Camilo and Ortiz-Zamora, Andr{\'e}s Felipe},
  journal={IEEE Technology and Society Magazine},
  volume={42},
  number={3},
  pages={69--75},
  year={2023},
  publisher={IEEE}
}

@inproceedings{jaiswal2020privacy,
  title={Privacy enhanced multimodal neural representations for emotion recognition},
  author={Jaiswal, Mimansa and Provost, Emily Mower},
  booktitle={Proceedings of the AAAI Conference on Artificial Intelligence},
  volume={34},
  number={05},
  pages={7985--7993},
  year={2020}
}

@inproceedings{jaiswal2019muse,
  title={Muse-ing on the impact of utterance ordering on crowdsourced emotion annotations},
  author={Jaiswal, Mimansa and Aldeneh, Zakaria and Bara, Cristian-Paul and Luo, Yuanhang and Burzo, Mihai and Mihalcea, Rada and Provost, Emily Mower},
  booktitle={ICASSP 2019-2019 IEEE International Conference on Acoustics, Speech and Signal Processing (ICASSP)},
  pages={7415--7419},
  year={2019},
  organization={IEEE}
}

@article{busso2016msp,
  title={MSP-IMPROV: An acted corpus of dyadic interactions to study emotion perception},
  author={Busso, Carlos and Parthasarathy, Srinivas and Burmania, Alec and AbdelWahab, Mohammed and Sadoughi, Najmeh and Provost, Emily Mower},
  journal={IEEE Transactions on Affective Computing},
  volume={8},
  number={1},
  pages={67--80},
  year={2016},
  publisher={IEEE}
}

@article{martinez2020msp,
  title={The MSP-conversation corpus},
  author={Martinez-Lucas, Luz and Abdelwahab, Mohammed and Busso, Carlos},
  journal={Interspeech 2020},
  year={2020}
}

@inproceedings{chen2022system,
  title={System description for voice privacy challenge 2022},
  author={Chen, Xiaojiao and Li, Guangxing and Huang, Hao and Zhou, Wangjin and Li, Sheng and Cao, Yang and Zhao, Yi},
  booktitle={Proc. 2nd Symposium on Security and Privacy in Speech Communication},
  year={2022}
}

@article{yao2024npu,
  title={NPU-NTU System for Voice Privacy 2024 Challenge},
  author={Yao, Jixun and Kuzmin, Nikita and Wang, Qing and Guo, Pengcheng and Ning, Ziqian and Guo, Dake and Lee, Kong Aik and Chng, Eng-Siong and Xie, Lei},
  journal={arXiv preprint arXiv:2409.04173},
  year={2024}
}

@inproceedings{he2025emotion,
  title={Emotion-Preserving Prosody Anonymization Network for Voice Privacy Protection},
  author={He, Jiabei and Zhao, Shiwan and Zhou, Jiaming and Sun, Haoqin and Wang, Hui and Qin, Yong},
  booktitle={ICASSP 2025-2025 IEEE International Conference on Acoustics, Speech and Signal Processing (ICASSP)},
  pages={1--5},
  year={2025},
  organization={IEEE}
}

@article{ju2024naturalspeech,
  title={Naturalspeech 3: Zero-shot speech synthesis with factorized codec and diffusion models},
  author={Ju, Zeqian and Wang, Yuancheng and Shen, Kai and Tan, Xu and Xin, Detai and Yang, Dongchao and Liu, Yanqing and Leng, Yichong and Song, Kaitao and Tang, Siliang and others},
  journal={arXiv preprint arXiv:2403.03100},
  year={2024}
}

@inproceedings{oh2016faceless,
  title={Faceless person recognition: Privacy implications in social media},
  author={Oh, Seong Joon and Benenson, Rodrigo and Fritz, Mario and Schiele, Bernt},
  booktitle={Computer Vision--ECCV 2016: 14th European Conference, Amsterdam, The Netherlands, October 11-14, 2016, Proceedings, Part III 14},
  pages={19--35},
  year={2016},
  organization={Springer}
}

@article{van2025emotion,
  title={Emotion Recognition: Benefits and Human Rights in VR Environments},
  author={van Noordenne, Marise},
  journal={Code and Conscience: Exploring Technology, Human Rights, and Ethics in Multidisciplinary AI Education},
  volume={14400},
  pages={17},
  year={2025},
  publisher={Springer Nature}
}

@article{zitouni2022privacy,
  title={Privacy aware affective state recognition from visual data},
  author={Zitouni, M Sami and Lee, Peter and Lee, Uichin and Hadjileontiadis, Leontios J and Khandoker, Ahsan},
  journal={IEEE Access},
  volume={10},
  pages={40620--40628},
  year={2022},
  publisher={IEEE}
}

@article{park2020k,
  title={K-EmoCon, a multimodal sensor dataset for continuous emotion recognition in naturalistic conversations},
  author={Park, Cheul Young and Cha, Narae and Kang, Soowon and Kim, Auk and Khandoker, Ahsan Habib and Hadjileontiadis, Leontios and Oh, Alice and Jeong, Yong and Lee, Uichin},
  journal={Scientific Data},
  volume={7},
  number={1},
  pages={293},
  year={2020},
  publisher={Nature Publishing Group UK London}
}

@inproceedings{pentyala2021privacy,
  title={Privacy-preserving video classification with convolutional neural networks},
  author={Pentyala, Sikha and Dowsley, Rafael and De Cock, Martine},
  booktitle={International conference on machine learning},
  pages={8487--8499},
  year={2021},
  organization={PMLR}
}

@article{simonyan2014two,
  title={Two-stream convolutional networks for action recognition in videos},
  author={Simonyan, Karen and Zisserman, Andrew},
  journal={Advances in neural information processing systems},
  volume={27},
  year={2014},
}

@article{xu2025privacy,
  title={Privacy-Preserving Multimodal Sentiment Analysis},
  author={Xu, Honghui and Li, Wei and Takabi, Daniel and Seo, Daehee and Cai, Zhipeng},
  journal={IEEE Internet of Things Journal},
  year={2025},
  publisher={IEEE}
}

@inproceedings{wang2016using,
  title={Using randomized response for differential privacy preserving data collection.},
  author={Wang, Yue and Wu, Xintao and Hu, Donghui},
  booktitle={EDBT/ICDT Workshops},
  volume={1558},
  pages={0090--6778},
  year={2016}
}

@article{he2023clustered,
  title={Clustered federated learning with adaptive local differential privacy on heterogeneous iot data},
  author={He, Zaobo and Wang, Lintao and Cai, Zhipeng},
  journal={IEEE Internet of Things Journal},
  volume={11},
  number={1},
  pages={137--146},
  year={2023},
  publisher={IEEE}
}

@article{xu2022privacy,
  title={Privacy-preserving mechanisms for multi-label image recognition},
  author={Xu, Honghui and Cai, Zhipeng and Li, Wei},
  journal={ACM Transactions on Knowledge Discovery from Data (TKDD)},
  volume={16},
  number={4},
  pages={1--21},
  year={2022},
  publisher={ACM New York, NY}
}

@article{zheng2021efficient,
  title={Efficient publication of distributed and overlapping graph data under differential privacy},
  author={Zheng, Xu and Zhang, Lizong and Li, Kaiyang and Zeng, Xi},
  journal={Tsinghua Science and Technology},
  volume={27},
  number={2},
  pages={235--243},
  year={2021},
  publisher={TUP}
}

@article{zhang2023local,
  title={A local differential privacy trajectory protection method based on temporal and spatial restrictions for staying detection},
  author={Zhang, Weiqi and Xie, Zhenzhen and Sai, Akshita Maradapu Vera Venkata and Zia, Qasim and He, Zaobo and Yin, Guisheng},
  journal={Tsinghua Science and Technology},
  volume={29},
  number={2},
  pages={617--633},
  year={2023},
  publisher={TUP}
}

@article{eltoft2006multivariate,
  title={On the multivariate Laplace distribution},
  author={Eltoft, Torbj{\o}rn and Kim, Taesu and Lee, Te-Won},
  journal={IEEE Signal Processing Letters},
  volume={13},
  number={5},
  pages={300--303},
  year={2006},
  publisher={IEEE}
}

@article{hu2020differential,
  title={Differential privacy protection method based on published trajectory cross-correlation constraint},
  author={Hu, Zhaowei and Yang, Jing},
  journal={Plos one},
  volume={15},
  number={8},
  pages={e0237158},
  year={2020},
  publisher={Public Library of Science San Francisco, CA USA}
}

@article{yin2024primonitor,
  title={PriMonitor: an adaptive tuning privacy-preserving approach for multimodal emotion detection},
  author={Yin, Lihua and Lin, Sixin and Sun, Zhe and Wang, Simin and Li, Ran and He, Yuanyuan},
  journal={World Wide Web},
  volume={27},
  number={2},
  pages={9},
  year={2024},
  publisher={Springer}
}

@inproceedings{wang2017locally,
  title={Locally differentially private protocols for frequency estimation},
  author={Wang, Tianhao and Blocki, Jeremiah and Li, Ninghui and Jha, Somesh},
  booktitle={26th USENIX Security Symposium (USENIX Security 17)},
  pages={729--745},
  year={2017}
}

@article{yin2021privacy,
  title={A privacy-preserving federated learning for multiparty data sharing in social IoTs},
  author={Yin, Lihua and Feng, Jiyuan and Xun, Hao and Sun, Zhe and Cheng, Xiaochun},
  journal={IEEE Transactions on Network Science and Engineering},
  volume={8},
  number={3},
  pages={2706--2718},
  year={2021},
  publisher={IEEE}
}

@inproceedings{yu2020ch,
  title={Ch-sims: A chinese multimodal sentiment analysis dataset with fine-grained annotation of modality},
  author={Yu, Wenmeng and Xu, Hua and Meng, Fanyang and Zhu, Yilin and Ma, Yixiao and Wu, Jiele and Zou, Jiyun and Yang, Kaicheng},
  booktitle={Proceedings of the 58th annual meeting of the association for computational linguistics},
  pages={3718--3727},
  year={2020}
}

@inproceedings{wu2024hydiscgan,
  title={HyDiscGAN: a hybrid distributed cGAN for audio-visual privacy preservation in multimodal sentiment analysis},
  author={Wu, Zhuojia and Zhang, Qi and Miao, Duoqian and Yi, Kun and Fan, Wei and Hu, Liang},
  booktitle={Proceedings of the Thirty-Third International Joint Conference on Artificial Intelligence},
  pages={6550--6558},
  year={2024}
}

@article{mirza2014conditional,
  title={Conditional generative adversarial nets},
  author={Mirza, Mehdi and Osindero, Simon},
  journal={arXiv preprint arXiv:1411.1784},
  year={2014}
}

@inproceedings{tsai2019multimodal,
  title={Multimodal transformer for unaligned multimodal language sequences},
  author={Tsai, Yao-Hung Hubert and Bai, Shaojie and Liang, Paul Pu and Kolter, J Zico and Morency, Louis-Philippe and Salakhutdinov, Ruslan},
  booktitle={Proceedings of the conference. Association for computational linguistics. Meeting},
  volume={2019},
  pages={6558},
  year={2019}
}

@inproceedings{panayotov2015librispeech,
  title={Librispeech: an asr corpus based on public domain audio books},
  author={Panayotov, Vassil and Chen, Guoguo and Povey, Daniel and Khudanpur, Sanjeev},
  booktitle={2015 IEEE international conference on acoustics, speech and signal processing (ICASSP)},
  pages={5206--5210},
  year={2015},
  organization={IEEE}
}

@article{zen2019libritts,
  title={Libritts: A corpus derived from librispeech for text-to-speech},
  author={Zen, Heiga and Dang, Viet and Clark, Rob and Zhang, Yu and Weiss, Ron J and Jia, Ye and Chen, Zhifeng and Wu, Yonghui},
  journal={arXiv preprint arXiv:1904.02882},
  year={2019}
}

@article{liu2021query2label,
  title={Query2label: A simple transformer way to multi-label classification},
  author={Liu, Shilong and Zhang, Lei and Yang, Xiao and Su, Hang and Zhu, Jun},
  journal={arXiv preprint arXiv:2107.10834},
  year={2021}
}

@article{schuhmann2021laion,
  title={Laion-400m: Open dataset of clip-filtered 400 million image-text pairs},
  author={Schuhmann, Christoph and Vencu, Richard and Beaumont, Romain and Kaczmarczyk, Robert and Mullis, Clayton and Katta, Aarush and Coombes, Theo and Jitsev, Jenia and Komatsuzaki, Aran},
  journal={arXiv preprint arXiv:2111.02114},
  year={2021}
}

@article{cao2019openpose,
  title={Openpose: Realtime multi-person 2d pose estimation using part affinity fields},
  author={Cao, Zhe and Hidalgo, Gines and Simon, Tomas and Wei, Shih-En and Sheikh, Yaser},
  journal={IEEE transactions on pattern analysis and machine intelligence},
  volume={43},
  number={1},
  pages={172--186},
  year={2019},
  publisher={IEEE}
}

@inproceedings{osokin2019real,
  title={Real-time 2D multi-person pose estimation on CPU: Lightweight OpenPose},
  author={Osokin, D},
  booktitle={ICPRAM 2019-Proceedings of the 8th International Conference on Pattern Recognition Applications and Methods},
  pages={744--748},
  year={2019}
}

@article{jo2022comparative,
  title={Comparative analysis of OpenPose, PoseNet, and MoveNet models for pose estimation in mobile devices},
  author={Jo, BeomJun and Kim, SeongKi},
  journal={Traitement du Signal},
  volume={39},
  number={1},
  pages={119},
  year={2022},
  publisher={International Information and Engineering Technology Association (IIETA)}
}

@inproceedings{cao2017realtime,
  title={Realtime multi-person 2d pose estimation using part affinity fields},
  author={Cao, Zhe and Simon, Tomas and Wei, Shih-En and Sheikh, Yaser},
  booktitle={Proceedings of the IEEE conference on computer vision and pattern recognition},
  pages={7291--7299},
  year={2017}
}

@article{zhang2025two,
  title={Two-Dimensional Human Pose Estimation with Deep Learning: A Review},
  author={Zhang, Zheyu and Shin, Seong-Yoon},
  journal={Applied Sciences},
  volume={15},
  number={13},
  pages={7344},
  year={2025},
  publisher={MDPI}
}

@article{xu2022vitpose,
  title={Vitpose: Simple vision transformer baselines for human pose estimation},
  author={Xu, Yufei and Zhang, Jing and Zhang, Qiming and Tao, Dacheng},
  journal={Advances in neural information processing systems},
  volume={35},
  pages={38571--38584},
  year={2022}
}

@inproceedings{bulat2017far,
  title={How far are we from solving the 2d \& 3d face alignment problem?(and a dataset of 230,000 3d facial landmarks)},
  author={Bulat, Adrian and Tzimiropoulos, Georgios},
  booktitle={Proceedings of the IEEE international conference on computer vision},
  pages={1021--1030},
  year={2017}
}

@inproceedings{ronneberger2015u,
  title={U-net: Convolutional networks for biomedical image segmentation},
  author={Ronneberger, Olaf and Fischer, Philipp and Brox, Thomas},
  booktitle={International Conference on Medical image computing and computer-assisted intervention},
  pages={234--241},
  year={2015},
  organization={Springer}
}

@inproceedings{dziugaite2015training,
  title={Training generative neural networks via maximum mean discrepancy optimization},
  author={Dziugaite, Gintare Karolina and Roy, Daniel M and Ghahramani, Zoubin},
  booktitle={Proceedings of the Thirty-First Conference on Uncertainty in Artificial Intelligence},
  pages={258--267},
  year={2015}
}

@inproceedings{santoso2024large,
  title={Large language model-based emotional speech annotation using context and acoustic feature for speech emotion recognition},
  author={Santoso, Jennifer and Ishizuka, Kenkichi and Hashimoto, Taiichi},
  booktitle={ICASSP 2024-2024 IEEE International Conference on Acoustics, Speech and Signal Processing (ICASSP)},
  pages={11026--11030},
  year={2024},
  organization={IEEE}
}

@article{brown2020language,
  title={Language models are few-shot learners},
  author={Brown, Tom and Mann, Benjamin and Ryder, Nick and Subbiah, Melanie and Kaplan, Jared D and Dhariwal, Prafulla and Neelakantan, Arvind and Shyam, Pranav and Sastry, Girish and Askell, Amanda and others},
  journal={Advances in neural information processing systems},
  volume={33},
  pages={1877--1901},
  year={2020}
}

@inproceedings{hong2025aer,
  title={AER-LLM: Ambiguity-aware emotion recognition leveraging large language models},
  author={Hong, Xin and Gong, Yuan and Sethu, Vidhyasaharan and Dang, Ting},
  booktitle={ICASSP 2025-2025 IEEE International Conference on Acoustics, Speech and Signal Processing (ICASSP)},
  pages={1--5},
  year={2025},
  organization={IEEE}
}

@inproceedings{li2025revise,
  title={Revise, reason, and recognize: Llm-based emotion recognition via emotion-specific prompts and asr error correction},
  author={Li, Yuanchao and Gong, Yuan and Yang, Chao-Han Huck and Bell, Peter and Lai, Catherine},
  booktitle={ICASSP 2025-2025 IEEE International Conference on Acoustics, Speech and Signal Processing (ICASSP)},
  pages={1--5},
  year={2025},
  organization={IEEE}
}

@article{cancela2024eu,
  title={The EU's AI act: A framework for collaborative governance},
  author={Cancela-Outeda, Celso},
  journal={Internet of Things},
  volume={27},
  pages={101291},
  year={2024},
  publisher={Elsevier}
}

@inproceedings{radford2023robust,
  title={Robust speech recognition via large-scale weak supervision},
  author={Radford, Alec and Kim, Jong Wook and Xu, Tao and Brockman, Greg and McLeavey, Christine and Sutskever, Ilya},
  booktitle={International conference on machine learning},
  pages={28492--28518},
  year={2023},
  organization={PMLR}
}

@article{chen2022wavlm,
  title={Wavlm: Large-scale self-supervised pre-training for full stack speech processing},
  author={Chen, Sanyuan and Wang, Chengyi and Chen, Zhengyang and Wu, Yu and Liu, Shujie and Chen, Zhuo and Li, Jinyu and Kanda, Naoyuki and Yoshioka, Takuya and Xiao, Xiong and others},
  journal={IEEE Journal of Selected Topics in Signal Processing},
  volume={16},
  number={6},
  pages={1505--1518},
  year={2022},
  publisher={IEEE}
}

@article{zhang2024keypoints,
  title={Keypoints-Integrated Instruction-Following Data Generation for Enhanced Human Pose Understanding in Multimodal Models},
  author={Zhang, Dewen and An, Wangpeng and Shouno, Hayaru},
  journal={arXiv e-prints},
  pages={arXiv--2409},
  year={2024}
}

@inproceedings{zhou2023privacy,
  title={Privacy-preserving federated learning via disentanglement},
  author={Zhou, Wenjie and Li, Piji and Han, Zhaoyang and Lu, Xiaozhen and Li, Juan and Ren, Zhaochun and Liu, Zhe},
  booktitle={Proceedings of the 32nd ACM International Conference on Information and Knowledge Management},
  pages={3606--3615},
  year={2023}
}

@inproceedings{bortolato2020learning,
  title={Learning privacy-enhancing face representations through feature disentanglement},
  author={Bortolato, Bla{\v{z}} and Ivanovska, Marija and Rot, Peter and Kri{\v{z}}aj, Janez and Terh{\"o}rst, Philipp and Damer, Naser and Peer, Peter and {\v{S}}truc, Vitomir},
  booktitle={2020 15th IEEE International Conference on Automatic Face and Gesture Recognition (FG 2020)},
  pages={495--502},
  year={2020},
  organization={IEEE}
}

@article{malarkkan2025delta,
  title={DELTA: Variational Disentangled Learning for Privacy-Preserving Data Reprogramming},
  author={Malarkkan, Arun Vignesh and Bai, Haoyue and Kaushik, Anjali and Fu, Yanjie},
  journal={arXiv preprint arXiv:2509.00693},
  year={2025}
}

@article{jia2025latent,
  title={Latent space disentangling for StyleGAN: a linear approach based on higher-dimensional geometry},
  author={Jia, Yunfei and Li, Xinyu and Hou, Fei},
  journal={Signal, Image and Video Processing},
  volume={19},
  number={7},
  pages={536},
  year={2025},
  publisher={Springer}
}

@inproceedings{carion2020end,
  title={End-to-end object detection with transformers},
  author={Carion, Nicolas and Massa, Francisco and Synnaeve, Gabriel and Usunier, Nicolas and Kirillov, Alexander and Zagoruyko, Sergey},
  booktitle={European conference on computer vision},
  pages={213--229},
  year={2020},
  organization={Springer}
}

@article{vinciarelli2009social,
  title={Social signal processing: Survey of an emerging domain},
  author={Vinciarelli, Alessandro and Pantic, Maja and Bourlard, Herv{\'e}},
  journal={Image and vision computing},
  volume={27},
  number={12},
  pages={1743--1759},
  year={2009},
  publisher={Elsevier}
}

@article{laurent:hal-02438020,
TITLE = {{Ethical Teaching Analytics in a Context-Aware Classroom: A Manifesto}}, 
AUTHOR = {Laurent, Romain and Vaufreydaz, Dominique and Dessus, Philippe},
RL = {https://hal.archives-ouvertes.fr/hal-02438020}, JOURNAL = {{ERCIM News}},
PUBLISHER = {{ERCIM}}, 
NUMBER = {120},
AGES = {39--40}, YEAR = {2020}, 
MONTH = Jan,
HAL_ID = {hal-02438020},
HAL_VERSION = {v1}, }

@article{dantcheva2015else,
  title={What else does your biometric data reveal? A survey on soft biometrics},
  author={Dantcheva, Antitza and Elia, Petros and Ross, Arun},
  journal={IEEE Transactions on Information Forensics and Security},
  volume={11},
  number={3},
  pages={441--467},
  year={2015},
  publisher={IEEE}
}

@article{zhu2025adaptive,
  title={Adaptive Key Role Guided Hierarchical Relation Inference for Enhanced Group-level Emotion Recognition},
  author={Zhu, Qing and Mao, Qirong and Dong, Wenlong and Shao, Xiuyan and Huang, Xiaohua and Zheng, Wenming},
  journal={IEEE Transactions on Affective Computing},
  year={2025},
  publisher={IEEE}
}

@article{okabe2018attentive,
  title={Attentive statistics pooling for deep speaker embedding},
  author={Okabe, Koji and Koshinaka, Takafumi and Shinoda, Koichi},
  journal={arXiv preprint arXiv:1803.10963},
  year={2018}
}

\end{document}